\RequirePackage{tikz}
\documentclass[sn-nature]{sn-jnl}

\usepackage[]{graphicx}
\usepackage[]{xcolor}
\makeatletter
\def\maxwidth{ %
\ifdim\Gin@nat@width>\linewidth
\linewidth
\else
\Gin@nat@width
\fi
}
\makeatother

\definecolor{fgcolor}{rgb}{0.345, 0.345, 0.345}

\usepackage{framed}
\makeatletter
{\par\unskip\endMakeFramed%
\at@end@of@kframe}
\makeatother

\definecolor{shadecolor}{rgb}{.97, .97, .97}
\definecolor{messagecolor}{rgb}{0, 0, 0}
\definecolor{warningcolor}{rgb}{1, 0, 1}
\definecolor{errorcolor}{rgb}{1, 0, 0}

\usepackage{graphicx}%
\usepackage{multirow}%
\usepackage{amsmath,amssymb,amsfonts}%
\usepackage{amsthm}%
\usepackage{mathrsfs}%
\usepackage[title]{appendix}%
\usepackage{xcolor}%
\usepackage{textcomp}%
\usepackage{manyfoot}%
\usepackage{booktabs}%
\usepackage{algorithm}%
\usepackage{algorithmicx}%
\usepackage{algpseudocode}%
\usepackage{listings}%
\usepackage{pgfplots}
\usepackage{hyperref}
\usepackage{rotating}
\usepackage{tabularx,ragged2e}
\usepackage{enumerate}
\usepackage[T1]{fontenc}
\usepackage{upquote}

\theoremstyle{thmstyleone}%
\theoremstyle{thmstyletwo}%

\theoremstyle{thmstylethree}%

\IfFileExists{upquote.sty}{\usepackage{upquote}}{}
\begin{document}

\title[Artificial Intelligence in the Service of Entrepreneurial Finance]{Artificial Intelligence in the Service of Entrepreneurial Finance: Knowledge Structure and the Foundational Algorithmic Paradigm}


\author*[1]{\fnm{Robert} \sur{Kudeli\'{c}}}\email{robert.kudelic@foi.unizg.hr}

\author[1]{\fnm{Tamara} \sur{\v{S}maguc}}

\author[2]{\fnm{Sherry} \sur{Robinson}}


\affil[1]{
	\orgname{University of Zagreb Faculty of Organization and Informatics}, \orgaddress{
		\country{Republic of Croatia}}}

\affil[2]{
	\orgname{Penn State University}, \orgaddress{
		\country{United States of America}}}



\abstract{While the application of Artificial Intelligence in Finance has a long tradition, its potential in Entrepreneurship has been intensively explored only recently. In this context, Entrepreneurial Finance is a particularly fertile ground for future Artificial Intelligence proliferation. To support the latter, the study provides a bibliometric review of Artificial Intelligence applications in (1) entrepreneurial finance literature, and (2) corporate finance literature with implications for Entrepreneurship. Rigorous search and screening procedures of the scientific database Web of Science Core Collection resulted in the identification of 1890 relevant journal articles subjected to analysis. The bibliometric analysis gives a rich insight into the knowledge field's conceptual, intellectual, and social structure, indicating nascent and underdeveloped research directions. As far as we were able to identify, this is the first study to map and bibliometrically analyze the academic field concerning the relationship between Artificial Intelligence, Entrepreneurship, and Finance, and the first review that deals with Artificial Intelligence methods in Entrepreneurship. According to the results, Artificial Neural Network, Deep Neural Network and Support Vector Machine are highly represented in almost all identified topic niches. At the same time, applying Topic Modeling, Fuzzy Neural Network and Growing Hierarchical Self-organizing Map is quite rare. As an element of the research, and before final remarks, the article deals as well with a discussion of certain gaps in the relationship between Computer Science and Economics. These gaps do represent problems in the application of Artificial Intelligence in Economic Science. As a way to at least in part remedy this situation, the foundational paradigm and the bespoke demonstration of the Monte Carlo randomized algorithm are presented.}




\maketitle

\section{Introduction}\label{sec:intro}

The fascinating accumulation of computational knowledge and capacity constantly gives rise to new methods, especially those that are nature/data-driven \cite{Tapeh2022} and capable of effectively replacing or improving old solutions based on traditional mathematical and statistical structures \cite{Nazareth2023}. This particularly stems from the limitations of conventional methods in solving non-linear, time-variant, and behaviorally uncertain problems, inherent in many real-life phenomena \cite{Bahrammirzaee2010}. Artificial Intelligence (AI) has been proven fruitful for such problems in lots of different areas \cite{Tapeh2022}, from Medicine \cite{Hamet2017}, Transport \cite{Kouziokas2017}, Environmental Sciences \cite{Ye2020}, and Manufacturing \cite{Arinez2020}, to Economics \cite{Dirican2015} and Finance \cite{Bahrammirzaee2010}. The latter area, although apparently saturated, has recently experienced a kind of explosion in AI-related publications \cite{Goodell2021}, and the same applies to Entrepreneurship in which the application of AI is particularly interesting given its still infancy stage \cite{Obschonka2019}.

Along with its popularity in the scientific community, the propulsiveness of AI in Finance and related disciplines has recently been recognized by practice. According to the "Hired's 2023 State of Software Engineers" survey  \cite{Perry2023}, the AI industry has risen to the top of the list of booming technology jobs in 2023. This is expected considering the AI's recent popularity achieved by the public release of Dall-E 2 and ChatGPT. However, the second most common choice of technology professionals was the Financial Technology Industry (FinTech), which overtook sectors such as Healthtech and Cybersecurity.

As per the literature search and review, this is the first study to map and bibliometrically analyze the academic field concerning the relationship between AI, entrepreneurship, and finance, and at the same time, the first review that deals with AI methods in entrepreneurship. It aims to explore and review the scientific knowledge about AI methods applicable in the entrepreneurial finance domain. The study provides a quantitative bibliometric review of applying AI in (1) entrepreneurial finance literature, and (2) corporate finance literature with implications for entrepreneurship. In addition to standard bibliometric indicators, rigorous, comprehensive, and temporal data analysis identifies various AI methods in the subject literature, showing a chronological aspect of the subject field and suggesting future application possibilities. Rich insights into the research area produce implications for different target groups dealing with AI in entrepreneurial finance (from the scientific community and computer experts to entrepreneurs and investors in entrepreneurship).

In Section~\ref{sec1} we position the study within the existing scientific opus and elaborate the scope and objectives of the research. Section~\ref{sec2} details the bibliometric methodology discussing the data search and screening procedures and describing the applied bibliometric tools. A clarification of the research methodology is followed by Section~\ref{sec3}, which presents and interprets the bibliometric results and opens horizons for discussion and research implications elaborated in Section~\ref{sec4}. Section~\ref{sec:mc} is devoted to presenting the foundational paradigm of AI and methods connected to it, with an emphasis on the practical aspect, so as to make an effort to resolve issues of AI results expounded in \cite{Biju2023}. Finally, there are conclusions of the study in Section~\ref{sec4}.

\section{Background of the Study}\label{sec1}

Here, we provide an overview of the existing literature in the emerging sphere of "AI in entrepreneurship" (Subsection~\ref{subsec2}), and the relatively saturated field of "AI in finance" (Subsection~\ref{subsec3}). The Section presents the general progress of the two domains and elaborates on key research topics. The development of the application of the following important groups of AI approaches in finance is elaborated: Expert Systems, Artificial Neural Networks, Hybrid Intelligent Systems, Support Vector Machines, and Natural Language Processing. The discussion of AI methods is followed by a detailed insight into previous literature reviews in domains of interest (Subsection~\ref{subsec4}). An overview of related bibliometric work identifies many research gaps addressed by this study. The Section ends by defining the scope and objectives of the study arising from the identified research gaps.

\subsection{Overview of AI in Entrepreneurship}\label{subsec2}

The era of AI in Entrepreneurship has begun recently \cite{Obschonka2019}. Although pioneering scientific papers in the field appeared in the 1980s\footnote{The first article in the Web of Science Core Collection database in the domain is a publication by Gordon and Key (1987) \cite{Gordon1987} on the topic of the role of AI in the information needs of small businesses \cite{Li2022}. An insight into Scopus shows that the first article in that database was published in 1986 by Friedenberg and Hensler (1986) \cite{Friedenberg1986}, who provide a kind of guide for entrepreneurs for the application of AI in various aspects of a business.}, the overall output published in the first 20 years is small, with only five papers in the Web of Science Core Collection by 2003. After a period of slow growth between 2003-2016, from 2017, the domain is experiencing a kind of publication explosion\footnote{A search of the Scopus shows that $ 83.9\% $ of scientific papers and books in the AI-entrepreneurship domain were published between 2017 and 2023. The largest number of papers and books was recorded in 2022 (as much as $ 24.9\% $). The search was conducted on June 6, 2023, using the query: "artificial intelligence" AND ("entrepreneur*" OR "small business*").} \cite{Li2022} -- not only quantitatively but also in terms of interests and topics arising from "the reciprocity of the co-evolving fields of entrepreneurship research and practice" \cite{Obschonka2019} (p. 529). As influential scholars observe \cite{Obschonka2019,Chalmers2020,Nambisan2017}, AI and technologies in general, enrich and transform entrepreneurship as a field of research but also change real-world entrepreneurial activity. Nambisan (2017) \cite{Nambisan2017} recognizes the dual transforming reflection of the proliferation of new technologies in shaping entrepreneurial pursuits. First, technological progress expands the boundaries of entrepreneurial processes and outcomes, making them more fluid and porous (e.g., advances in Financial Technology (FinTech) enrich entrepreneurial finance sources available without spatial and temporal boundaries in a specific entrepreneurial ecosystem) \cite{Nambisan2017}. Second, technology leads to a shift in the focus of an entrepreneurial agency, creating dynamic sets of agents with different characteristics, aspirations, and goals. An example is the development of new infrastructure such as crowdfunding systems that stimulated the birth of more collective forms of entrepreneurial initiative \cite{Nambisan2017,Iurchenko2023}. Such disruptive changes and novelties "on the ground" are reciprocally reflected in the agenda of entrepreneurship research -- it is not only boosted by new AI research tools and methods but also gets completely new targets that are studied with these methods \cite{Obschonka2019}.

Regarding decision-making support and business performance improvement, AI in Entrepreneurial Finance is one of the empirically fruitful research directions \cite{Giuggioli2022,Obschonka2019}. For example, AI methods have recently been applied in the codification of the communication behavior of entrepreneurs and the analysis of crowdfunding presentation campaigns \cite{Kaminski2019,Oo2023,Giuggioli2022}. Hence, AI capabilities have great potential for improving the communication strategies of entrepreneurs and rationalizing the decisions of investors \cite{Giuggioli2022}. The application of AI is also evident in Entrepreneurial Finance Management. AI-blockchain hybrid platforms support new ways of managing the financial accounting of an entrepreneurial venture \cite{Chalmers2020} and change audit processes by reducing the need for traditional audit procedures such as sampling and confirmations \cite{Giuggioli2022,Zemankova2019}. In fact, in many aspects of business management AI automation tools bring a completely new paradigm of business scaling \cite{Chalmers2020}. Predicting an entrepreneurial venture's success (failure) is another domain with AI applications \cite{Krishna2016,Koumbarakis2022,Zhang2022}. Due to prediction accuracy, handling non-linear effects in data, and ambiguity detection, AI techniques are promising compared to traditional prediction methods \cite{Koumbarakis2022}. The same applies to the segment of business planning of an entrepreneur, especially in activities such as sales forecasting, product pricing \cite{Giuggioli2022,Syam2018}, and predicting the reaction of customers to price changes \cite{Chalmers2020}.

Despite the increasingly frequent application of AI in entrepreneurship research and practice, and numerous fresh scientific topics, the research focus on the types of AI methods applicable in the field and the possibilities of these methods is relatively weak. Given the immaturity and newness of the area, this is not surprising. The AI-entrepreneurship intersection is currently mostly dealt with by entrepreneurship scholars, who have yet to seek partnerships with researchers who are experts in AI \cite{Levesque2020}. However, the time for multidisciplinary collaboration that would produce more technical scientific insights on AI in entrepreneurship is right in front of us.

\subsection{Overview of AI in Finance}\label{subsec3}

In contrast to AI research in Entrepreneurship \cite{Obschonka2019}, AI in Finance is a relatively old scientific domain \cite{Goodell2021}. The inspection of significant scientific databases (Google Scholar, Scopus, Web of Science) shows that the first relevant journal articles on the AI-finance intersection appeared in the 1970s (Google Scholar) and 1980s (Scopus), and were mostly related to the application of AI in banking and securities investment problems\footnote{The first reviewed journal article in this area appeared in 1977 and was related to the application of expert systems in Finance. \cite{Wong1995}}. Some of the early covered topics were credit card application assessment \cite{Tamai1989}, predicting the firm's financial health \cite{Elmer1988}, credit evaluation \cite{Shaw1988,Klein1989,Srinivasan1988,Chhikara1989}, stock portfolio selection \cite{Yamaguchi1989,Lee1989}, stock market behavior prediction \cite{Loo1989,Braun1987}, and assigning ratings to corporate bonds \cite{Dutta1988}. Since the mid-1980s a small group of authors has outlined the application of expert systems in accounting and auditing \cite{Shim1988,Avi1985,DUNGAN1985,VasarhelyiM.1989,Akers1986,DILLARD1987}, solving problems such as auditor's assessment of uncollectible accounts \cite{DUNGAN1985} and assessment of company solvency \cite{DILLARD1987}. The development of the domain continued in the 1990s and 2000s when a larger number of relevant applications of AI methods appeared in corporate bankruptcy prediction \cite{Zhang1999,Pacheco1996,Jo1996,Lacher1995,Altman1994,Wilson1994,Huang2008,Lee1996,Varetto1998} and financial fraud detection (accounting fraud \cite{Cerullo1999}, fraud in credit approval process \cite{Wheeler2000,McKee2009}, and credit card fraud \cite{Kim2002,Kou2004,Quah2008,Yu2009}). 

Additionally, the field of financial forecasting based on sentiment analysis began to develop, with a rise following the notable publication of Das and Chen in 2007 \cite{Das2007}. A few years later, the highly cited work of Pan (2012) paved the way for further advances in financial distress models \cite{Pan2012}. Driven by the fourth industrial revolution, AI-finance research has experienced a strong proliferation that continues to this day\footnote{The analysis of Scopus search results on April 29, 2023, indicates that $ 67.0\% $ of indexed scientific papers and books on AI-Finance were published after 2015. The most fruitful year was 2022 with $ 16.3\% $ of all scientific papers and books (the search was conducted using the query: "artificial intelligence" AND "financ*").} \cite{Goodell2021}. Recently, new topic niches are emerging, such as AI in the context of FinTech innovations (cryptocurrencies and blockchain, crowdfunding, peer-to-peer lending, financial robo-advising, and mobile payment services) \cite{Belanche2019,Goodell2021,Palmie2020,Mosteanu2020,Ashta2021,Hendershott2021}, and predicting financing success using the new, FinTech funding sources \cite{Yuan2016,Kaminski2019,Yeh2020,Wang2021}.

Expert Systems (ES) are the first form of AI in finance, with initial application in 1977 \cite{Wong1995,Bahrammirzaee2010}. Despite the hardware limitations of the time, by the mid-1990s they were pioneered in fields such as finance, investment, taxation, accounting, and administration \cite{Wilson1987} (as cited in \cite{Bahrammirzaee2010}). A more notable work from that time was published by Shaw and Gentry in 1988 \cite{Shaw1988}, developing the MARBLE system intended to assess the riskiness of business loan applicants. Although ES proved to be more practical compared to conventional statistical techniques, they failed in front of other AI methods such as Artificial Neural Networks (ANN) and Hybrid Intelligent Systems (HIS) -- they were only capable of prescription, but not of prediction and improvement of the result by experience, and were not useful for identifying the non-linear relationships \cite{Bahrammirzaee2010}. 

These deficiencies are eliminated by ANN, with the beginning of its application in the bond ranking in the 1980s \cite{Dutta1988a,Surkan1990}. ANNs are "non-parametric" methods that are data-driven, self-adaptive, and compared to parametric methods, are less sensitive to model misspecification. They are suitable for models without a priori assumptions about the data, which can be non-linear, and discontinuous \cite{Kumar2021,Tay2001}. These features have proven to be a huge strength over sophisticated statistical techniques in problems such as bankruptcy prediction and stock market prediction \cite{Tay2001,Kumar2021,Bahrammirzaee2010}, characterized by a complex set of highly correlated, nonlinear, unclearly related variables \cite{Wong1998}.

Despite the advantages, some shortcomings of ANNs have also emerged. The most popular in Finance, Back-Propagation Neural Network (BPNN) needs a large number of control parameters, hardly gives a stable solution, and suffers from potential overfitting, leading to poor generalization ability to the out-of-sample data \cite{Tay2001}. This is the reason why it is often combined with classical statistical techniques or other intelligent methods such as ES, Fuzzy Logic, Genetic Algorithms (GAs), and Robotics \cite{Wong1998}. Hybrid Intelligent Systems (HIS) aim to use the advantages of complementary methods and minimize their shortcomings, and are capable of achieving multi-functionality, technical enhancement, and multiplicity of application tasks. Although their performances are very sensitive to the right choice of integration methods and the problem of parameterization, they have generally proven to be more powerful in solving numerous problems in credit evaluation, portfolio management, and financial forecasting and planning -- especially various neuro-fuzzy systems and combinations of NNs, Fuzzy Logic, and GAs \cite{Bahrammirzaee2010}.

In addition to hybridized methods, improved generalization performance came with the Support Vector Machine (SVM) in 1998. Compared to BPNN, SVM mainly\footnote{There are also some exceptions to the majority of findings. For example, in the bankruptcy prediction problem, Ecer (2013) \cite{Ecer2013} finds a slightly better performance of ANN compared to SVM, while Tsai (2008) \cite{Tsai2008} notes the overperformance of SVM in only one of four datasets.} shows significantly or at least slightly better results in financial time series forecasting \cite{Huang2005,Tay2001,Cao2001,Kim2003,PrasadDas2012}, credit rating analysis \cite{Huang2004,Li2006}, and financial distress evaluation \cite{Hui2006}. The advantage of SVMs is the implementation of the structural risk minimization principle, minimizing an upper bound of the generalization error, in contrast to previous ANN algorithms based on the empirical risk minimization principle \cite{Tay2001,Kim2003}. "Another merit of SVMs is that the training of SVMs is equivalent to solving a linearly constrained quadratic programming" -- resulting in a unique solution, optimal solution, without the problem of converging to a local minimum which may be a drawback of BPNN \cite{Tay2001,Osuna,Bahrammirzaee2010,Li2006}. All of this has made SVM one of the most common AI methods for solving a range of (especially predictive) problems in Finance \cite{Goodell2021}, whether it is used as a single method or a component of HIS \cite{Bahrammirzaee2010}.

In the last fourteen years, there has been a growing popularity of Natural Language Processing (NLP) in Finance \cite{Xing2017,Fisher2016}. Proponents of the methods argue that a lot of data can hardly be expressed numerically, without losing the holistic meaning, endless variety and nuances, and unstructured text documents usually contain more timely information than quantitative financial sets. Moreover, text from financial news, social networks, or auditor's reports includes opinions, connections, and emotions, and all of this can be useful in a series of financial classification and prediction problems \cite{Das2014,Fisher2016}. According to Fisher et al. (2016), the most commonly used AI tools for NLP-based research in Finance are SVMs, followed by Naive Bayes (NB), hierarchical clustering, statistical methods, and Term Frequency -- Inverse Document Frequency (TF-IDF) weighting. In addition to generating and validating prototype taxonomies and thesauri, NLP has shown promising results in corporate reports readability studies, and especially in topics such as financial fraud detection, and recognizing stock price movement \cite{Fisher2016}. However, even nine years ago the domain was still in its infancy \cite{Purda2014}. Some of the identified research questions at that moment were to what extent in accounting taxonomies and thesauri problems NLP can survive independently, without the need for manual interventions, and how to overcome the problem of small data samples, distributed location of text documents and the changing nature of the accounting vocabulary \cite{Fisher2016}. More recent, future developments have been recognized in the topic analysis of accounting disclosures and the proliferation of deep learning research in, for example, quantifying the diversity of a firm's operations and locations, and labeling different types of corporate risks \cite{Bochkay2022}.

\subsection{Research Gaps and Objectives of the Study}\label{subsec4}

When it comes to the literature review opus, the AI-entrepreneurship domain has been developing recently, with papers published by Li et al. (2022) \cite{Li2022}, Giuggiol and Pellegrin (2023) \cite{Giuggioli2022}, Blanco-González-Tejero et al. (2023) \cite{GonzalezTejero2023}, and Gupta et al. (2023) \cite{Gupta2023}. In the AI-finance sphere notable review contributions are made by a plethora of authors (e.g. Wong and Selvi (1998) \cite{Wong1998}, Bahrammirzaee (2010) \cite{Bahrammirzaee2010}, Fethi and Pasiouras (2010) \cite{Fethi2010}, Omoteso (2012) \cite{Omoteso2012}, Das (2014) \cite{Das2014}, de Prado et al. (2016) \cite{Prado2016}, Alaka et al. (2018) \cite{Alaka2018}, Shi and Li (2019) \cite{Shi2019}, Königstorfer and Thalmann (2020) \cite{Koenigstorfer2020}, Kumar et al. (2021) \cite{Kumar2021}, Thakkar and Chaudhari (2021) \cite{Thakkar2020}, and Goodell et al. (2021) \cite{Goodell2021}). More recent, review works in this area are also produced by Ahmed et al. (2022) \cite{Ahmed2022}, Gómez et al. (2022) \cite{Gomez2022}, Nazareth and Reddy (2023) \cite{Nazareth2023}, Chaklader et al. (2023) \cite{Chaklader2023}, and Chen et al. (2023) \cite{Chen2023}.

Overall, most occurring review papers are systematic literature reviews, and there are several bibliometric analyses \cite{Goodell2021,Ahmed2022,Prado2016,Shi2019,Li2022,Gupta2023,GonzalezTejero2023,Chaklader2023,Chen2023}. A large part of the reviews deals with a narrower niche topic, such as AI in banking \cite{Fethi2010,Koenigstorfer2020}, AI in bankruptcy prediction \cite{Prado2016,Alaka2018,Shi2019}, or AI in sustainable entrepreneurship \cite{Gupta2023}, while two bibliometric papers seek to give a holistic outlook of the AI-finance field \cite{Goodell2021,Ahmed2022}, but with methodological limitations and without or with an incomplete review of AI methods. The same applies to the bibliometric papers from the AI-entrepreneurship domain \cite{Li2022,GonzalezTejero2023}. In terms of what we were able to gather from the literature, a review focused on the intersection of AI and entrepreneurial and/or corporate finance has not been conducted so far. In order to demonstrate all identified relevant gaps in existing knowledge, below are given deeper insights into the study-related bibliometrics. First, bibliometric studies from the AI-finance intersection are elaborated, followed by a discussion of bibliometric research from the AI-entrepreneurship domain.

Bibliometric studies on the AI-finance intersection are conducted by Goodell et al. (2021) \cite{Goodell2021}, Ahmed et al. (2022) \cite{Ahmed2022}, Chen et al. (2023) \cite{Chen2023}, Shi and Li (2019) \cite{Shi2019}, do Prado et al. (2016) \cite{Prado2016}, and Chaklader et al. (2023) \cite{Chaklader2023}. With the aim of reviewing the entire AI-finance domain, Goodell et al. (2021) \cite{Goodell2021} carried out co-citation and bibliometric-coupling analyses of 283 papers from Scopus published in the period 1986-April 2021. In addition to the generally small sample size\footnote{According to Donthu et al. (2021) \cite{Donthu2021}, bibliometric analysis is justified to be performed on sufficiently large datasets, 500 papers or more, while for a smaller volume of material (e.g. 200-300 papers or less) suitable methods are meta-analysis or a systematic literature review.} \cite{Donthu2021}, they consider only material from the subject areas "Business, management and accounting", "Economics, econometrics and finance", "Social sciences", and "Arts and humanities", with a huge number of target papers published in "Computer Science" missing \cite{Shi2019,Nazareth2023}. Although they use a rich array of search terms related to AI methods, the number and variety of used search terms from the field of finance are limited (that part of the search query included only the following: "finance" OR "financ* manag*"). The biggest contribution of the study is a thematic overview of the chronological development of the area and the identification of eight thematic clusters and three broad thematic areas: "1) portfolio construction, valuation, and investor behavior; 2) financial fraud and distress; and 3) sentiment inference, forecasting, and planning". Although the authors provide a general review of used AI methods in the analyzed finance research, the focus on methods is narrow, and the given method categorization is not adequate (for example, the general category "AI methods" is treated as a separate group of methods, in relation to machine learning methods or deep learning methods, and no insight is given as to which methods are hidden under the general category).

The entire AI-finance domain is also bibliometrically examined by Ahmed et al. (2022) \cite{Ahmed2022}. The analysis was carried out on a sample of 348 papers from Scopus, considering material only from journals categorized in the first or second quartile (Q1 and Q2 journals as per the Scopus ranking 2021). The target papers are additionally limited to the time of publication between 2011 and 2021, and just as in the previously discussed study, the material belonging to the area of "Computer Science" is not included in the analysis. Moreover, only journals in the finance field were considered. The main results refer to the identification of relevant publications in six topic research streams: (1) bankruptcy prediction and credit-risk assessment, (2) stock price prediction, portfolio management, volatility, and liquidity, (3) prediction of the prices of oil, gold, and agriculture products, (4) anti-money laundering, anti-fraud detection, and risk management, (5) behavioral finance, and (6) big data analytics, blockchain, and data mining. The AI methods used in the analyzed papers were not identified or considered in any form, which the authors themselves state as a study limitation, suggesting future research endeavors.

A recent study by Chen et al. (2023) \cite{Chen2023} narrows the AI domain, considering only Explainable Artificial Intelligence (XAI) in finance. The bibliometric dataset was taken from the Web of Science Core Collection (WoSCC) and covers the period from 2013 to 2023.  The results identified two main groups of research: (1) application-oriented research by XAI in finance, and (2) innovation-oriented studies with a focus on technology development. The contribution was made by the identification of some topic research trends and prospects. Despite the large sample of documents (N=2733), the study has methodological limitations (the analysis was performed on all search results after applying search filters, without performing any data screening and data cleaning procedures). Additional shortcomings of the study are similar to the previously mentioned one (the dataset includes only papers from finance and economics; the results are thematically focused without giving any insight into AI methods in finance).

Shi and Li (2019) \cite{Shi2019} start from a broader view of methods and narrower topic niches, trying to examine the application of various intelligent techniques (statistical, operational research and AI methods) in just one specific financial problem: corporate bankruptcy prediction. The bibliometric study was conducted on a sample of 413 publications from the WoSCC database for the period from 1968 to 2018. The results indicate the propulsiveness of the field after the financial crisis of 2007-2008 and demonstrate relatively weak cooperation among the authors. An important conclusion is that there was an approximately equal representation of papers in computer science journals and those from management and finance, supporting the seriousness of the discussed limitation of studies by Goodell et al. (2021) \cite{Goodell2021}, Ahmed et al. (2022) \cite{Ahmed2022}, and Chen et al. (2023) \cite{Chen2023}. Moreover, the highest representation of papers was found in Expert Systems with Applications (ESA) which is a computer science and engineering journal. When it comes to intelligent techniques, the study contributes in terms of identifying the most common methods in bankruptcy prediction problem (Neural Network, Multivariate Discriminant Analysis) and those less represented (Fuzzy, Rough Set, Data Mining, Adaboost, K-Nearest Neighbors, Bayesian Network).

A similar topic is considered by do Prado et al. (2016) \cite{Prado2016}. Their bibliometric analysis aimed to identify the application of multivariate data analysis techniques to credit risk and bankruptcy prediction problems. Therefore, the focus of the analysis was not on AI methods, but on 17 techniques of multivariate analysis, among which only ANNs are from the AI domain. The subject of the analysis was 393 scientific articles from the Web of Science (main collection) published between 1968 and 2014. The main conclusions are similar to those of Shi and Li (2019) \cite{Shi2019} pointing to the papers' proliferation since 2008 and the multidisciplinarity of the field covering not only Business and Economics, but Computer Science, Statistics, Mathematics, Engineering, and so on. Additionally, the study notes the widespread use of advanced AI techniques (primarily ANNs) since the 1990s and the increasing popularity of combining ANNs with traditional statistical techniques (Logistic Regression and Discriminant Analysis).

Chaklader et al. (2023) \cite{Chaklader2023} deals with AI in the context of the progress of Financial Technology (FinTech) companies. The sample includes 302 Scopus indexed papers from the period 2014 to 2022. Based on keyword analysis, the authors identify several trending topics and future research directions. The paper in no way elaborates on the applied AI methods in the FinTech topic niche.

Finally, it is important to highlight a study on machine learning (ML) in finance by Nazareth and Reddy (2023) \cite{Nazareth2023}. Although it is primarily a systematic literature review on a small sample of papers (N=126), the authors also provide bibliometric data on the field, including certain aspects of bibliometric analysis. The contribution of the study is reflected in the compilation of progress in ML in six different financial domains (stock markets, portfolio management, forex markets, bankruptcy and insolvency, financial crisis, and cryptocurrency). The study reviews more than ten ML models, with implications for their applicability in specific financial fields. However, it is not primarily bibliometric research and does not cover the entire AI domain and its intersection with entrepreneurship. Besides, the focus is on the recent literature (published in 2015 and later), without the intention of providing insight into the chronological development of the field.

In the AI-entrepreneurship domain, the following bibliometric studies were conducted: Li et al. (2022) \cite{Li2022}, Blanco-González-Tejero et al. (2023) \cite{GonzalezTejero2023}, and Gupta et al. (2023) \cite{Gupta2023}. Li et al. (2022) \cite{Li2022} carried out research on the cross-field of AI and entrepreneurial management. The analysis included only 123 papers from the Web of Science Core Collection published between 1987 and February 2021. The main results refer to the identification of thematic clusters from which ten research hotspots emerged (e.g. the impact of digitalization on different industries, the impact of AI on the development trend of enterprises, enterprise business intelligence (BI) construction and application, and others). Importantly, the study does not focus on the domain of entrepreneurial finance and does not provide any overview of AI methods in entrepreneurship.

Blanco-González-Tejero et al. (2023) \cite{GonzalezTejero2023} analyzes 520 scientific papers from the Dimensions.ai database published until July 2022. Their research does not refer to AI methods in any form and deals with the role of AI in entrepreneurship in general.

Gupta et al. (2023) \cite{Gupta2023} deals with the literature on the role of AI in sustainable entrepreneurship. The bibliometric analysis includes 482 articles from Scopus published between 1994 and 2022. The authors identify trending research topics in the field of AI-sustainable development. A deeper description of topic areas, as well as a review of AI methods, is not provided. The main conclusion of the paper is that sustainable development is a trendy scientific topic with a growing number of articles and citations.

In summary, no bibliometric study, as far as we were able to find, provides an overview of the entire spectrum of AI methods in finance or entrepreneurship. An exception is the study by Goodell et al. (2022) which takes a rough look at the groups of AI methods in finance, with certain technical shortcomings in the grouping strategy. Almost all conducted bibliometric studies are based on a relatively small sample of documents \cite{Donthu2021}, and some of them have other methodological limitations (e.g., the exclusion of Computer Science literature from the dataset, limited application of available bibliometric tools, not performing data screening procedures before analysis, etc.). Crucially, to our knowledge, none of the existing studies cover the intersection of AI, entrepreneurship, and finance to give implications for entrepreneurial finance.

Considering research gaps, the present study aims to explore and review the conceptual, intellectual and social structure of scientific knowledge \cite{Aria2017} on the intersections of AI and two economics fields: entrepreneurship and finance. Briefly speaking, the focus of the research is (1) AI-entrepreneurial finance literature, and (2) AI-corporate finance literature with implications for entrepreneurship. In the context of the study, entrepreneurial finance is defined as "the art and science of investing and financing entrepreneurial venture"\footnote{A pioneering study related to the topic of entrepreneurial finance is the work of Gorman and Sahlman (1989) \cite{Gorman1989}, and the development of the field has been taking place since the late 1990s \cite{Alemany2018}.} (p. 9) \cite{Alemany2018}. According to the definition, there are two fundamental aspects of entrepreneurial finance: investing, i.e. choosing the direction of an entrepreneur's investment (purchase of physical assets, entering a new market, etc.), and financing, or securing money for the realization of the investment plan \cite{Alemany2018}. Some of the important topics within the domain are: sources of funding for entrepreneurs (bank lending, equity capital, crowdfunding, business angels, venture capital, etc.), investor-entrepreneur negotiation strategies, business planning (including financial planning and forecasting), understanding and analyzing financial statements, and new venture and small business valuation \cite{Abor2016}. The concept is not limited to startups but also covers intrapreneurship, acquisitions of existing businesses, and new entrepreneurial ventures within corporations or family firms \cite{Alemany2018}. Moreover, some authors extend the concept to the financial and investment activities of all small and medium-sized enterprises (SMEs), distancing it from corporate finance as a "financial decision-making of large corporate organizations" (p. 4) \cite{Abor2016}. 

In accordance with the protocol of previous studies \cite{Kumar2022,Kumar2021a,Shi2019,Kumar2022a} and the research subject, the specific objectives of the study are defined as follows:
\begin{enumerate}[1.]
	\item To determine the publication productivity and evolution of scientific knowledge on the intersection of AI-entrepreneurship-finance.
	\item To identify the most influential articles, and the most prolific academic scholars, journals, institutions, and countries on the intersection of AI-entrepreneurship-finance, and to determine the degree of academic cooperation and multidisciplinarity in the knowledge field.
	\item To determine and interpret prominent topics on the intersection of AI-entrepreneurship-finance, and to identify the chronological development of prominent topics.
	\item To determine AI methods (method, algorithm, technique) used in the study of certain topics at the intersection of AI-entrepreneurship-finance, so as to determine the current state and project future possibilities.
	\item To reach a more profound insight into the research field, and to reflect on the emerging research directions and the promising AI methods for future applications in entrepreneurial finance, with implications for the scientific community, computer experts, entrepreneurs, and investors in entrepreneurship.
	\item To give recommendations for future improvement of bibliometric methodology.
\end{enumerate}

In Tables \ref{tab:VS1} and \ref{tab:VS2} we compare the characteristics of the present study against related bibliometric work. As can be seen, there are several research gaps addressed by our study: the absence of bibliometric research on AI methods with implications for entrepreneurial finance, the small sample of documents in previous studies, limitations related to the exclusion of "Computer Science" from the sample, and insufficient focus on AI methods.

\begin{table}[h]	
	\caption{Comparison of the present study against related bibliometric work -- Part I\textsuperscript{1}}\label{tab:VS1}%
	\begin{tabularx}{\textwidth}{@{}>{\RaggedRight}p{2cm}>{\RaggedRight}p{2cm}>{\RaggedRight}p{1.5cm}>{\RaggedRight}p{2cm}>{\RaggedRight}p{1cm}>{\RaggedRight}X@{}}
		\toprule
		\multicolumn{6}{@{}c@{}}{Intersection AI-Finance} \\
		\cmidrule{1-6}
		Topic Scope & Methods & Database & Subject Area & Sample\textsuperscript{2} & Review\textsuperscript{3} \\
		\midrule		
		\multicolumn{6}{@{}l@{}}{do Prado et al. (2016) \cite{Prado2016}} \\
		\midrule
		Credit Risk and Bankruptcy Prediction & Multivariate Data Analysis Techniques & WoSCC & No limit & $ 393 $ & Credit Risk and Bankruptcy Prediction multivariate techniques list and timeline \\
		\multicolumn{6}{@{}l@{}}{Shi and Li (2019) \cite{Shi2019}} \\
		\midrule
		Corporate Bankruptcy Prediction & Intelligent techniques & WoSCC & No limit & $ 413 $ & Only Bankruptcy Prediction intelligent methods list \\
		&&&&&\\
		\multicolumn{6}{@{}l@{}}{Goodell et al. (2021) \cite{Goodell2021}} \\
		\midrule
		Finance & Artificial Intelligence & Scopus & Only Social Sciences, no Computer Science & $ 283 $ & Only AI methods list, in Finance
		(mentioning some limitations) \\
		\multicolumn{6}{@{}l@{}}{Ahmed et al. (2022) \cite{Ahmed2022}} \\
		\midrule
		Finance & Artificial Intelligence & Scopus & Only Economics and Finance, no Computer Science & $ 348 $ & None \\
		\multicolumn{6}{@{}l@{}}{Chaklader et al. (2023) \cite{Chaklader2023}} \\
		\midrule
		FinTech companies & Artificial Intelligence & Scopus & No limit & $ 302 $ & None \\
		&&&&&\\
		\multicolumn{6}{@{}l@{}}{Nazareth and Reddy (2023) \cite{Nazareth2023}} \\
		\midrule
		Finance & Machine Learning & ScienceDirect & No limit & $ 126 $ & Machine Learning in Finance \\
		&&&&&\\
		\multicolumn{6}{@{}l@{}}{Chen et al. (2023) \cite{Chen2023}} \\
		\midrule
		Finance & Explainable Artificial Intelligence & WoSCC & Only Business, Business Finance and Economics, no Computer Science & $ 2733 $ & None \\
		\botrule
	\end{tabularx}
	\footnotetext[]{Data has been grouped by publication year, in ascending order}
	\footnotetext[1]{Continuation of the analysis presented in the table can be seen in Table \ref{tab:VS2}}
	\footnotetext[2]{No. of documents}
	\footnotetext[3]{What kind of AI method review provided}	
\end{table}

\begin{table}[h]	
	\caption{Comparison of the present study against related bibliometric work -- Part II\textsuperscript{1}}\label{tab:VS2}%
	\begin{tabularx}{\textwidth}{@{}>{\RaggedRight}p{2.5cm}>{\RaggedRight}p{2cm}>{\RaggedRight}p{1.5cm}>{\RaggedRight}p{2cm}>{\RaggedRight}p{1cm}>{\RaggedRight}X@{}}
		\toprule
		\multicolumn{6}{@{}c@{}}{Intersection AI-Entrepreneurship} \\
		\cmidrule{1-6}
		Topic Scope & Methods & Database & Subject Area & Sample\textsuperscript{2} & Review\textsuperscript{3} \\
		\midrule		
		\multicolumn{6}{@{}l@{}}{Li et al. (2022) \cite{Li2022}} \\
		\midrule
		Entrepreneurial management & Artificial Intelligence & WoSCC & No limit & $ 123 $ & None \\
		&&&&&\\
		\multicolumn{6}{@{}l@{}}{Blanco-González-Tejero et al. (2023) \cite{GonzalezTejero2023}} \\
		\midrule
		Entrepreneurship & Artificial Intelligence & Dimensions.ai & No data & $ 520 $ & None \\
		&&&&&\\
		\multicolumn{6}{@{}l@{}}{Gupta et al. (2023) \cite{Gupta2023}} \\
		\midrule
		Sustainable entrepreneurship & Artificial Intelligence & Scopus & No limit & $ 482 $ & None \\
		\midrule
		\multicolumn{6}{@{}c@{}}{\textbf{Intersection AI-Entrepreneurship-Finance}} \\
		\multicolumn{6}{@{}c@{}}{\textsc{The Present Study}} \\
		\midrule
		Entrepreneurial finance and corporate finance with implications for entrepreneurship & Artificial Intelligence & WoSCC & No limit & $ 1890 $ & Artificial Intelligence in entrepreneurial and corporate finance \\
		\botrule
	\end{tabularx}
	\footnotetext[]{Data has been grouped by publication year, in ascending order}
	\footnotetext[1]{First part of the analysis presented in the table can be seen in Table \ref{tab:VS1}}
	\footnotetext[2]{No. of documents}
	\footnotetext[3]{What kind of AI method review provided}
\end{table}

\section{Methodology}\label{sec2}

As introductory matters have been presented, now we will present research methodology, including a short review of Bibliometrics. Broad methodology themes are as follows: a) preliminary search and screening of the research field, b) data acquisition and preparation for bibliometric analysis, c) bibliometric data analysis; after which we will proceed to research results, produced as per the described methodology.

\subsection{Bibliometrics as a Research Method}\label{subsec5}

Despite its recent popularity, Bibliometrics is not a new research method \cite{Donthu2021}. The first documented attempt to use the method is related to statistical research on subject scattering in publications by Campbell in 1896. The application of the method was also recorded in 1917 when Cole and Eales statistically analyzed the growth of literature in comparative anatomy \cite{OSAREH1996}. Afterward, in 1922 Hulme used the term \textit{statistical bibliography} to describe "the illumination of the processes of science and technology by means of counting documents" (p. 348) \cite{Pritchard1969}. Considering statistical bibliography as an unsatisfactory and insufficiently accepted term, Pritchard (1969) \cite{Pritchard1969} proposed \textit{bibliometrics} as a new name for the subject, marking it as "the application of mathematics and statistical methods to books and other media of communication" (p. 348). Since then, the term bibliometrics has been widely accepted, and numerous definitions of it have appeared \cite{OSAREH1996,Broadus1987}. Common to almost all definitions is that it is a quantitative methodology based on statistical and mathematical techniques used to measure various constituents of certain forms of written communication (e.g., authors, locations, institutions, topics, etc.) \cite{OSAREH1996,Donthu2021}. With such a definition, bibliometrics is distanced from \textit{scientometrics}, which is a similar but broader concept covering "all quantitative aspects of the science of science" (p. 377) \cite{Broadus1987}.

According to Aria and Cuccurullo (2017), bibliometrics for scientific mapping enables insight into a specific research field with regard to its: (1) intellectual structure (identification of the knowledge base and influence of certain works and authors in the scientific community), (2) conceptual structure (finding major themes and trends), and (3) social structure (diagnosing interactions among researchers, institutions, and countries). The method generally gives a static picture of the research field at some point in time, and the inclusion of temporal analysis in data processing can provide insight into the chronological evolution of the field \cite{Aria2017}. Data for bibliometric analysis are suitable if they are objective in nature (e.g., number of papers, number of citations, etc.) and massive in quantity (the sample of documents must be greater than 500, and it is best if it is over 1000) \cite{Donthu2021}. The subject of bibliometric analysis can be different categories of materials, from books, articles, theses, and patents to the so-called "grey" literature \cite{Ellegaard2015}. Through a reliable, objective, and repeatable review of a large body of information, it provides the "big picture" of a scientific corpus \cite{Aria2017}. These are some of the main features by which bibliometric analysis differs from a systematic literature review based on a qualitative, manual analysis of a smaller set of publications. Bibliometrics should also be distinguished from meta-analysis, which is a quantitative review method, but with different goals (to summarize evidence of relationships between variables in a research field) \cite{Donthu2021}. 

\begin{figure}[h] 
	\centering
	\includegraphics[width=1\textwidth]{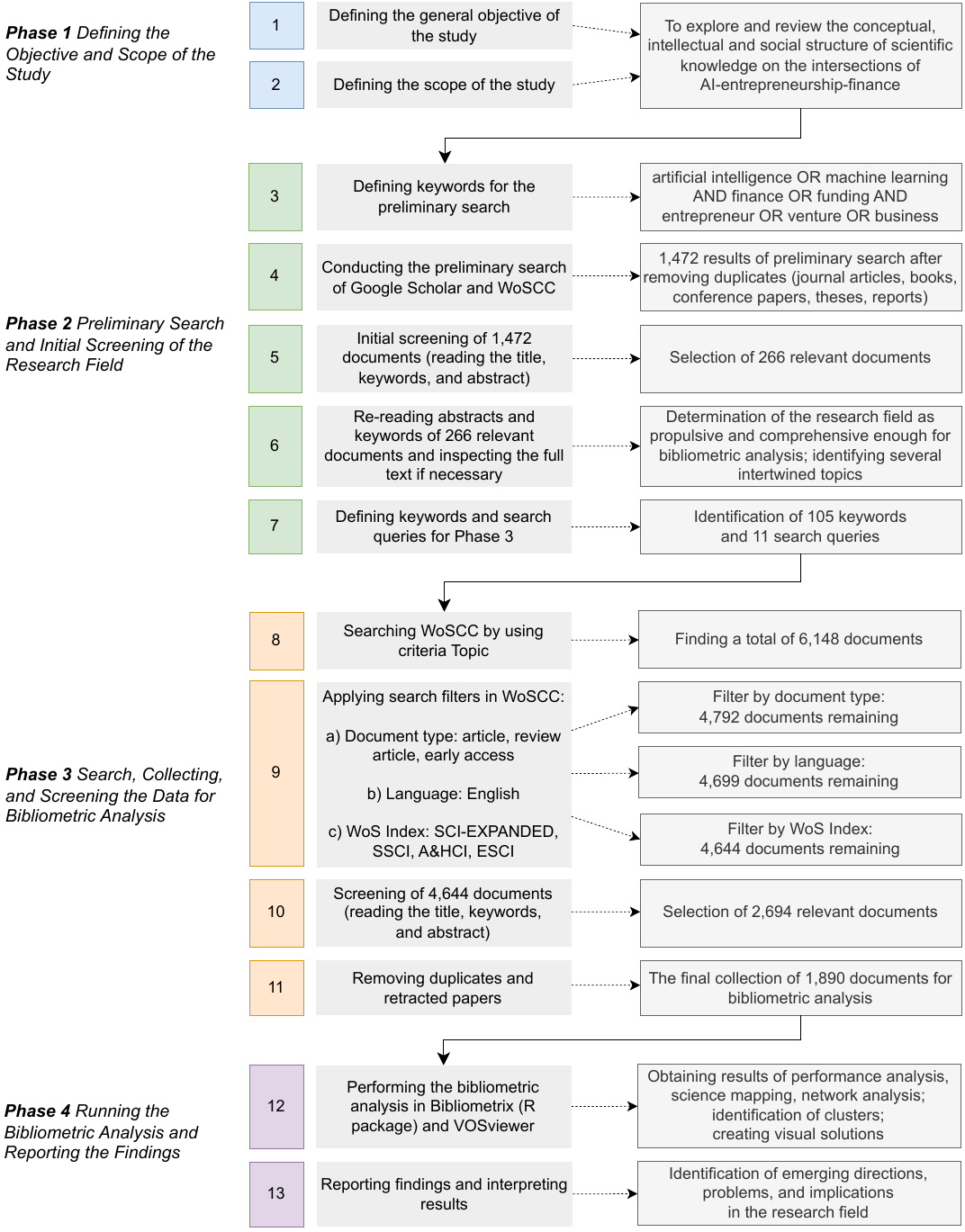}
	\caption{Flowchart detailing the bibliometric methodology steps and procedures in the present study (WoSCC denotes Web of Science Core Collection; SCI-EXPANDED denotes Science Citation Index Expanded; SSCI denotes Social Sciences Citation Index; A\&HCI denotes Arts and Humanities Citation Index; ESCI denotes Emerging Sources Citation Index)}\label{fig:protokol}
\end{figure}

Donthu et al. (2021) suggest four steps in the implementation of bibliometrics: (1) "define the aims and scope of the bibliometric study, (2) choose the techniques for bibliometric analysis, (3) collect the data for bibliometric analysis, and (4) run the bibliometric analysis and report the findings" (pp. 291-293). In order to carry out the latter, researchers have at their disposal two main categories of bibliometric tools: performance analysis and scientific mapping. Main tools can be enriched with network analyses which include a number of network metrics, as well as clustering and visualization \cite{Donthu2021}. The bibliometric analysis in the present study is grounded on general recommendations for the bibliometric methodology procedure \cite{Donthu2021,Aria2017} and is based on the application of well-known and established analytical tools, as described in Subsection~\ref{subsec8}. Before that, a detailed elaboration of the procedures of the data search, collection, and screening is given. The bibliometric methodology of the study is shown in the flowchart \ref{fig:protokol}.

\subsection{Preliminary Search and Initial Screening of the Research Field}
\label{subsec6}

The bibliometric analysis began by defining the general objective and scope of the study (Subsection~\ref{subsec4}), which was followed by a preliminary search and initial screening of the research field. The purpose of the preliminary search phase was to define an appropriate set of keywords and search queries as key inputs for the final data collection phase (Subsection~\ref{subsec7}). Proper selection of search keywords is crucial since even a small variation in terms and queries changes the data set, potentially generating different bibliometric results \cite{Hire2021}.

The preliminary search phase was conducted by searching the Google Scholar database using broadly defined keywords generated by the researchers. Google Scholar was selected as suitable for the preliminary search since it is the largest scientific database that offers the widest insight into a specific scientific field. In addition, the Web of Science Core Collection (WoSCC) was searched to ensure insight into the relevant literature of the area. From March 16, 2023, to March 29, 2023, a total of 1427 documents from Google Scholar (1053 after removing duplicates) and 419 documents from WoSCC were searched and screened. The search was conducted using different combinations of keywords shown in Tables \ref{tab:PSIS1} and \ref{tab:PSIS2}. Since it was a preliminary phase to get a general overview of the area, the initial screening included different types of documents (journal articles, books, conference papers, theses, and reports). By reading the titles, keywords and abstracts, 266 relevant documents were selected. The selected documents were further examined in detail by re-reading abstracts and keywords and, in some cases, by inspecting the full text of the document in order to ensure the validity of the research.

\begin{table}[h] 
	\caption{Keywords and Search Results in the Phase of Preliminary Search and Initial Screening of the Research Field -- Part I\textsuperscript{1}}\label{tab:PSIS1}%
	\begin{tabularx}{\textwidth}{@{}>{\RaggedRight}p{.5cm}>{\RaggedRight}p{2cm}>{\RaggedRight}X>{\RaggedRight}p{2cm}>{\RaggedRight}p{1cm}}
		\toprule
		\multicolumn{5}{@{}c@{}}{Google Scholar} \\
		\cmidrule{1-5}
		No. & Search Date & Search Keywords & Documents\textsuperscript{2} & Selected\textsuperscript{3} \\		
		\midrule
		1	& 3/16/2023 & artificial intelligence; finance; entrepreneur & $ 127 $ & $ 35 $ \\
		2	& 3/17/2023 & artificial intelligence; finance; venture & $ 91 $ & $ 19 $ \\
		3	& 3/22/2023 & artificial intelligence; finance; business & $ 113 $ & $ 36 $ \\
		4	& 3/22/2023 & artificial intelligence; funding; entrepreneur & $ 77 $ & $ 12 $ \\
		5	& 3/23/2023 & artificial intelligence; funding; venture & $ 67 $ & $ 5 $ \\
		6	& 3/25/2023 & artificial intelligence; funding; business & $ 74 $ & $ 3 $ \\
		7	& 3/25/2023 & machine learning; finance; entrepreneur & $ 96 $ & $ 19 $ \\
		8	& 3/27/2023 & machine learning; finance; venture & $ 87 $ & $ 15 $ \\
		9	& 3/27/2023 & machine learning; finance; business & $ 101 $ & $ 41 $ \\
		10	& 3/28/2023 & machine learning; funding; entrepreneur & $ 72 $ & $ 23 $ \\
		11	& 3/28/2023 & machine learning; funding; venture & $ 61 $ & $ 9 $ \\
		12	& 3/29/2023 & machine learning; funding; business & $ 87 $ & $ 6 $ \\
		\multicolumn{3}{@{}l@{}}{$\sum$} & $ 1053 $ & $ 223 $ \\
		\botrule
	\end{tabularx}
	\footnotetext[]{Sorted according to search date, in ascending order}
	\footnotetext[1]{Continuation of the analysis presented in the table can be seen in Table \ref{tab:PSIS2}}
	\footnotetext[2]{No. of documents in the initial screening after removing duplicates}
	\footnotetext[3]{No. of selected relevant documents}
\end{table}

\begin{table}[h] 
	\caption{Keywords and Search Results in the Phase of Preliminary Search and Initial Screening of the Research Field -- Part II\textsuperscript{1}}\label{tab:PSIS2}%
	\begin{tabularx}{\textwidth}{@{}>{\RaggedRight}p{.5cm}>{\RaggedRight}p{2cm}>{\RaggedRight}X>{\RaggedRight}p{2cm}>{\RaggedRight}p{1cm}}
		\toprule
		\multicolumn{5}{@{}c@{}}{Web of Science Core Collection\textsuperscript{2}} \\
		\cmidrule{1-5}
		No. & Search Date & Search Keywords & Documents\textsuperscript{3} & Selected\textsuperscript{4} \\		
		\midrule
		1	& 3/29/2023 & artificial intelligence AND finance AND entrepreneur & $ 19  $ & $ 2 $ \\
		2	& 3/29/2023 & artificial intelligence AND finance AND business & $ 200 $ & $ 20 $ \\
		3	& 3/29/2023 & machine learning AND finance AND business & $ 200 $ & $ 21 $ \\
		\multicolumn{3}{@{}l@{}}{$\sum$} & $ 419 $ & $ 43 $ \\
		\midrule
		\multicolumn{3}{@{}l@{}}{$\sum G + \sum W$} & $ 1472 $ & $ 266 $ \\
		\botrule
	\end{tabularx}
	\footnotetext[]{Sorted according to search date, in ascending order. $ G $ and $ W $ represent total sums for Google (Table \ref{tab:PSIS1}) and Web of Science (Table \ref{tab:PSIS2}), respectively}
	\footnotetext[1]{First part of the analysis presented in the table can be seen in Table \ref{tab:PSIS1}}
	\footnotetext[2]{The WoSCC search was conducted using three combinations of keywords that gave the largest number of relevant results in the Google Scholar search}
	\footnotetext[3]{No. of documents in the initial screening after removing duplicates}
	\footnotetext[4]{No. of selected relevant documents}
\end{table}

A preliminary search and screening of 266 relevant documents indicated that the intersection of AI-entrepreneurship-finance is a propulsive area with a large number of recent publications (from the last 3 years) mainly within two research areas: Computer Science and Business. The scope of the research field was assessed as sufficient for conducting a bibliometric analysis, and several intertwined topic niches (or branches) were found:

\renewcommand{\labelenumii}{\arabic{enumi}.\alph{enumii}}
\begin{enumerate}
	\item AI as support for entrepreneurial financing decisions
	\begin{enumerate}
		\item Investment success/business performance and entrepreneur's behavior and presentation
		\item Sources of entrepreneurial finance
		\item Valuation of an entrepreneurial venture/Prediction of performance and/or bankruptcy
	\end{enumerate}
	\item FinTech in the context of entrepreneurship
	\item Management of entrepreneurial finance
	\begin{enumerate}
		\item AI and accounting, auditing and detecting financial frauds
		\item Financial planning and other aspects of financial management
	\end{enumerate}
\end{enumerate}

Within each topic niche, different combinations of keywords and search queries were defined as inputs for the next stage of the literature search. The list of keywords and queries is shown in Tables \ref{tab:QSCS1}, \ref{tab:QSCS2} and \ref{tab:QSCS3} with the ordinal numbers of the corresponding topic niches, according to the enumeration above.

\subsection{Searching, Collecting, and Screening the Data for Bibliometric Analyses}\label{subsec7}

The data for bibliometric analysis was gathered from the Web of Science Core Collection database (WoSCC). Clarivate PLC's database was selected for the study as one of the most relevant and comprehensive collection of peer-reviewed scientific material. The bibliographic data and metadata it provides are suitable and sufficiently comprehensive for bibliometric analysis and can be exported in the appropriate format. The export of the data was carried out on May 5, 2023. The search was performed using the criteria Topic (searches title, abstract, author keywords, and Keywords Plus), and the following search filters were applied:

\begin{enumerate}
	\item Document Type: Article, Review Article, Early Access
	\item Language: English
	\item Web of Science Index: Science Citation Index Expanded (SCI-EXPANDED), Social Sciences Citation Index (SSCI), Arts \& Humanities Citation Index (A\&HCI), and Emerging Sources Citation Index (ESCI)
\end{enumerate}

By limiting the search to the WoSCC and the listed indexes and document types, data was retrieved only from "journals that demonstrate high levels of editorial rigor and best practice" \cite{2023}. Papers published in conference proceedings and other forms of scientific material were excluded from the data set. The goal was to form a corpus with only peer-reviewed and highest-quality scientific work \cite{Kumari2023}. Furthermore, no filter on the time range was placed (the data includes papers in the database from the year of publication of the first paper to the date of the search). Also, no filter on research areas was used. Such an approach ensured a complete collection of data across different periods and areas, providing a temporal and disciplinary comprehensive view of the research field \cite{Shi2019}.

The search was conducted using a large number of keywords and 11 different search queries (Tables \ref{tab:QSCS1}, \ref{tab:QSCS2} and \ref{tab:QSCS3}). The query syntax was adapted to the query formatting rules of the Web of Science. Accordingly, the search was based on operators: OR (to find records containing any of the search terms), AND (to find records containing all of the search terms), and NOT (to exclude records containing certain words). Wherever it was meaningful, the wildcard character "*" was applied, in order to control the retrieval of plurals, variant spellings etc. Quotation marks were used to search for exact phrases such as "artificial intelligence" \cite{2021}. The selection of keywords related to AI was aimed at covering as many different AI methods as possible. At the same time, keywords that could result in papers based solely on statistical methods, without an AI component, were avoided (examples are keywords such as "text mining" or "data mining") -- therefore if such a paper was mentioned AI, it was included, no inclusion otherwise.

\begin{table}[h] 
	\caption{Search Queries and Number of Records in Procedures of Searching, Collecting, and Screening the Data for Bibliometric Analysis -- Part I\textsuperscript{1}}\label{tab:QSCS1}%
	\begin{tabularx}{\textwidth}{@{}>{\RaggedRight}p{1cm}>{\RaggedRight}X>{\RaggedRight}p{1.5cm}}
		\toprule
		Topic\textsuperscript{2} & Search Query\textsuperscript{3} & Records\textsuperscript{4} \\		
		\midrule
		1 (a) 	& {\footnotesize ("artificial intelligence" OR "machine learning" OR "deep learning" OR "soft computing" OR "neural network*" OR "natural language processing") AND (text OR emotion OR nonverbal OR facial OR speech OR signal* OR pitch OR persuasion OR present* OR video OR trait* OR narrat* OR motivat* OR impress*) AND ("SME" OR "SMEs*" OR "small business*" OR entrepreneur* OR crowdfund*)} & $ 555 $ \\
		1 (b) 	& {\footnotesize ("artificial intelligence" OR "machine learning" OR "deep learning" OR "soft computing" OR "neural network*" OR "natural language processing") AND ("debt*" OR "loan*" OR "venture capital*" OR "venture fund*" OR "angel*" OR "equit*" OR "bootstrap financ*" OR "bootstrapping") AND ("SME" OR "SMEs*" OR enterprise* OR business* OR compan* OR firm* OR entrepreneur*)} & $ 736 $ \\
		1 (b) 	& {\footnotesize ("artificial intelligence" OR "machine learning" OR "deep learning" OR "soft computing" OR "neural network*" OR "natural language processing")  AND ("security token offer*" OR "initial coin offer*" OR crowdfund* OR kickstart* OR "peer-to-peer lending" OR "peer-to-peer loan")} & $ 196 $ \\
		1 (c) 	& {\footnotesize ("artificial intelligence" OR "machine learning" OR "deep learning" OR "soft computing" OR "neural network*" OR "natural language processing") AND ("SME failure" OR "SMEs* failure" OR "enterprise* failure" OR "compan* failure" OR "business* failure" OR "firm* failure" OR "entrepreneur* failure" OR bankruptcy)} & $ 1184 $ \\
		1 (c) 	& {\footnotesize ("artificial intelligence" OR "machine learning" OR "deep learning" OR "soft computing" OR "neural network*" OR "natural language processing") AND ("SME valuat*" OR "SMEs* valuat*" OR "enterprise* valuat*" OR "business* valuat*" OR "compan* valuat*" OR "firm* valuat*" OR "entrepreneur* valuat*" OR "SME success*" OR "SMEs* success*" OR "enterprise* success*" OR "business* success*" OR "compan* success*" OR "firm* success*" OR "entrepreneur* success*" OR "SME performance*" OR "SMEs* performance*" OR "enterprise* performance*" OR "business* performance*" OR "compan* performance*" OR "firm* performance*" OR "entrepreneur* performance*")} & $ 547 $ \\
		\botrule
	\end{tabularx}
	\footnotetext[]{Sorted according to search query, in ascending order}
	\footnotetext[1]{Continuation of the analysis presented in the table can be seen in Table \ref{tab:QSCS2}}
	\footnotetext[2]{Topic niches identified in the phase of preliminary search and initial screening of the research field presented in detail in Subsection~\ref{subsec6} of the paper}
	\footnotetext[3]{The search was updated and finalized, with the data being exported, on May 5, 2023}
	\footnotetext[4]{Total number of records}
\end{table}

\begin{table}[h] 
	\caption{Search Queries and Number of Records in Procedures of Searching, Collecting, and Screening the Data for Bibliometric Analysis -- Part II\textsuperscript{1}}\label{tab:QSCS2}%
	\begin{tabularx}{\textwidth}{@{}>{\RaggedRight}p{1cm}>{\RaggedRight}X>{\RaggedRight}p{1.5cm}}
		\toprule
		Topic\textsuperscript{2} & Search Query\textsuperscript{3} & Records\textsuperscript{4} \\		
		\midrule
		2 		& {\footnotesize ("artificial intelligence" OR "machine learning" OR "deep learning" OR "soft computing" OR "neural network*" OR "natural language processing")  AND (fintech OR "financ* technology") AND ("SME" OR "SMEs*" OR enterprise* OR business* OR compan* OR firm* OR entrepreneur*)} & $ 166 $ \\
		3 (a) 	& {\footnotesize ("artificial intelligence" OR "machine learning" OR "deep learning" OR "soft computing" OR "neural network*" OR "natural language processing") AND audit* AND ("SME" OR "SMEs*" OR enterprise* OR business* OR compan* OR firm*)} & $ 375 $ \\
		3 (a) 	& {\footnotesize ("artificial intelligence" OR "machine learning" OR "deep learning" OR "soft computing" OR "neural network*" OR "natural language processing") AND ("accounting" OR "accountant" OR "financ* statement*") AND ("SME" OR "SMEs*" OR enterprise* OR business* OR compan* OR firm* OR entrepreneur*)} & $ 696 $ \\
		3 (a) 	& {\footnotesize ("artificial intelligence" OR "machine learning" OR "deep learning" OR "soft computing" OR "neural network*" OR "natural language processing") AND ("fraud detect*" OR "financ* fraud" OR "accounting fraud") AND ("SME" OR "SMEs*" OR enterprise* OR business* OR compan* OR firm* OR entrepreneur*) NOT "credit card*"} & $ 258 $ \\
		3 (b) 	& {\footnotesize ("artificial intelligence" OR "machine learning" OR "deep learning" OR "soft computing" OR "neural network*" OR "natural language processing")  AND ("financ* management" OR "financ* planning" OR "financ* decision*" OR "financ* analys*" OR "financ* sustainability" OR "financ* distress" OR "financ* risk" OR "financ* constraints") AND ("SME" OR "SMEs*" OR enterprise* OR business* OR compan* OR firm* OR entrepreneur*)} & $ 745 $ \\
		3 (b) 	& {\footnotesize ("artificial intelligence" OR "machine learning" OR "deep learning" OR "soft computing" OR "neural network*" OR "natural language processing")  AND ("demand predict*" OR "demand forecast*" OR "predict* demand" OR "forecast* demand" OR "price* predict*" OR "price* forecast*" OR "predict* price*" OR "forecast* price*" OR "salar* predict*" OR "salar* forecast*" OR "wage* predict*" OR "wage* forecast*" OR "predict* salar*" OR "forecast* salar*" OR "predict* wage*" OR "forecast* wage*") AND ("SME" OR "SMEs*" OR enterprise* OR business* OR compan* OR firm* OR entrepreneur*)} & $ 690 $ \\
		\multicolumn{2}{@{}l@{}}{$\sum$} & $ 6148 $ \\
		\botrule
	\end{tabularx}
	\footnotetext[]{Sorted according to search query, in ascending order}
	\footnotetext[1]{First part of the analysis can be seen in Table \ref{tab:QSCS1}, with the continuation of the analysis presented in Table \ref{tab:QSCS3}}
	\footnotetext[2]{Topic niches identified in the phase of preliminary search and initial screening of the research field presented in detail in Subsection~\ref{subsec6} of the paper}
	\footnotetext[3]{The search was updated and finalized, with the data being exported, on May 5, 2023}
	\footnotetext[4]{Total number of records}
\end{table}

\begin{table}[h] 
	\caption{Search Queries and Number of Records in Procedures of Searching, Collecting, and Screening the Data for Bibliometric Analysis -- Part III\textsuperscript{1}}\label{tab:QSCS3}%
	\begin{tabular}{@{}lllll}
		\toprule
		& \multicolumn{4}{@{}c@{}}{Filtering Procedure} \\
		\cmidrule{2-5}
		Topic\textsuperscript{2} & Document Type & Language English & WoS Index\textsuperscript{3} & Screening\textsuperscript{4} \\		
		\midrule
		1 (a) 	& $ 385 $ & $ 377 $ & $ 369 $ & $ 121 $ \\
		1 (b) 	& $ 563 $ & $ 556 $ & $ 552 $ & $ 277 $ \\
		1 (b) 	& $ 146 $ & $ 145 $ & $ 143 $ & $ 120 $ \\
		1 (c) 	& $ 1021 $ & $ 997 $ & $ 986 $ & $ 740 $ \\
		1 (c) 	& $ 455 $ & $ 450 $ & $ 445 $ & $ 224 $ \\
		2 		& $ 129 $ & $ 124 $ & $ 121 $ & $ 39 $ \\
		3 (a) 	& $ 301 $ & $ 295 $ & $ 288 $ & $ 162 $ \\
		3 (a)	& $ 540 $ & $ 524 $ & $ 519 $ & $ 370 $ \\
		3 (a)	& $ 160 $ & $ 159 $ & $ 158 $ & $ 67 $ \\
		3 (b)	& $ 617 $ & $ 606 $ & $ 602 $ & $ 442 $ \\
		3 (b)	& $ 475 $ & $ 466 $ & $ 461 $ & $ 132 $ \\
		$\sum$	& $ 4792 $ & $ 4699 $ & $ 4644 $ & $ 2694 $ \\
		\multicolumn{4}{@{}l@{}}{$\sum D$} & $ 804 $ \\
		\multicolumn{4}{@{}l@{}}{$\sum B$} & $ 1890 $ \\
		\botrule
	\end{tabular}
	\footnotetext[]{Sorted according to search query, in ascending order.  $ D $ and $ B $ are representing duplicates and retracted papers, and number of papers for bibliometric data analysis ($ 2694 - 804 $), respectively.}
	\footnotetext[1]{Previous part of the analysis can be seen in Table \ref{tab:QSCS2}}
	\footnotetext[2]{Topic niches identified in the phase of preliminary search and initial screening of the research field presented in detail in Subsection~\ref{subsec6}. The search was updated and finalized, with the data being exported, on May 5, 2023.}	
	\footnotetext[3]{Selected Web of Science indexes, presented in Figure \ref{fig:protokol}}
	\footnotetext[4]{Performing reading of the title, keywords and abstract}
\end{table}

The data search yielded a total of $ 4644 $ results, which were subjected to a screening procedure. The screening was carried out by looking at the title, keywords, and abstract of the paper. In contrast to a systematic literature review, in the bibliometric methodology, screening of the abstract and the full text is carried out only if necessary \cite{Guo2020}. However, as a precaution, reading of abstracts in this study was performed for almost all of the $ 4644 $ papers -- excluding those documents for which the title was overwhelmingly clear. The screening strategy was based on a broad understanding of Entrepreneurial Finance, taking into account its overlap with related disciplines such as Corporate Finance, Management, Business Planning, Accounting, and Auditing. Therefore, in addition to entrepreneurial finance literature, the corpus of data for analysis included the body of literature related to corporations and the financial sector, which contains implications for the financing of entrepreneurship or the financial management of new entrepreneurs and small and medium-sized enterprises. Scientific material with implications only for financial institutions and financial markets was excluded from consideration (examples are studies dealing with banks, financial markets, or insurance companies in topics such as bank failure prediction, stock price movements forecasting, or credit card fraud detection). 

The screening procedure of $ 4644 $ records resulted in a set of $ 2694 $ relevant documents. The data set was then reduced by removing duplicates and retracted papers. These operations created a final corpus of 1890 documents which were subjected to analysis.

\subsection{Bibliometric Data Analyses}\label{subsec8}

Our bibliometric analysis was meant to be comprehensive, from the very beginning, as can be seen from research methodology and the aforementioned. In view of that fact, there are three tools selected for the analysis: RStudio (2023.03.0 Build 386), Bibliometrix (4.1.2) and VOSviewer (1.6.19). RStudio was a supporting tool as Bibliometrix is a package within the program. Bibliometrix was used as it is the only tool that supports the entire bibliometric process, and the VOSviewer was selected as an extension to the research and support to Bibliometrix, as in network analyses it was more reliable and had options and features not found in Bibliometrix, and therefore was a useful complementary tool. In this kind of research it is customary to list all techniques, analyses and metrics used and produced by the research, however as the research is vast this will be skipped, as the list would be quite big -- therefore we are directing the reader to consult Section~\ref{sec3} of the article where he can find section by section, methodologically conducted all analyses relevant for our research.

\begin{table}[h]
	\caption{Data Coverage (quality of population per variable)}\label{tab:dc}%
	\begin{tabular}{@{}lllll@{}}
		\toprule
		Key\textsuperscript{1} & Description & Missing {\footnotesize (no.)} & Missing {\footnotesize (\%)} & Sample Quality\textsuperscript{2}\\
		\midrule
		AU    & Author & $ 0 $ & $ 0.00 $ & Excellent\\
		CR    & Cited References & $ 0 $ & $ 0.00 $ & Excellent\\
		DT    & Document Type & $ 0 $ & $ 0.00 $ & Excellent\\
		SO	  &	Journal & $ 0 $ & $ 0.00 $ & Excellent\\
		LA	  &	Language & $ 0 $ & $ 0.00 $ & Excellent\\
		NR	  &	No. of Cited References & $ 0 $	& $ 0.00 $ & Excellent\\
		TI	  &	Title & $ 0 $ & $ 0.00 $ & Excellent\\
		TC	  &	Total Citation & $ 0 $ & $ 0.00 $ & Excellent\\
		AB	  &	Abstract & $ 1 $ & $ 0.05 $	& Good\\
		C1	  &	Affiliation & $ 3 $ & $ 0.16 $ & Good\\
		RP	  &	Corresponding Author & $ 3 $ & $ 0.16 $ & Good\\
		DI	  &	Digital Object Identifier & $ 101 $ & $ 5.34 $ & Good\\
		PY	  &	Publication Year & $ 103 $ & $ 5.45 $ & Good\\
		DE	  &	Keywords & $ 250 $ & $ 13.23 $ & Acceptable\\
		ID	  &	Keywords Plus\textsuperscript{3} & $ 294 $ & $ 15.56 $ & Acceptable\\
		WC	  &	Science Categories & $ 0 $ & $ 0.00 $ &	Excellent\\
		\botrule
	\end{tabular}
	\footnotetext[1]{Bibliometrix metadata identification keys}
	\footnotetext[2]{Scale by bibliometrix, in regard to missing percentage. $ 0.00 $: Excellent, $ [0.01, 10.00] $: Good, $ [10.01, 20.00] $: Acceptable, $ [20.01, 50.00] $: Poor, $ [50.01, 99.99] $: Critical, $ 100 $: Completely Missing \cite{Aria2023}}
	\footnotetext[3]{Frequent expressions that are algorithmically extracted from the article reference list. \cite{Garfield1993} Shown to be as effective as the author's keywords in investigating knowledge structure of scientific fields, but is less comprehensive in terms of representing the content of the article. \cite{Zhang2015}}
\end{table}

In Table \ref{tab:dc} the reader can observe the quality of data coverage. This data was produced by Bibliometrix as a part of the procedure of importing references into the program. Out of 16 items two are in an acceptable category, five are ranked as good, and 9 are not missing any information at all. This is important as it gave us the opportunity to make in-depth bibliometrics by not skipping any analysis for the reason of insufficient data quality. As the data is of quality, in the next section we are presenting bibliometrics and begin by presenting the results of preliminary data analysis.

\section{Research Results}\label{sec3}

This section presents the entire bibliometric analysis results, from preliminaries all the way to conceptual, intellectual and social analyses. Each analysis is clearly delineated by a subsection, within which one will find the interpretation of the data together with items of the analysis. In such a way the reader can be easily oriented and manage the content. Immediately after the section on research results, a section on discussion, implications and research constraints follows. The reader can therefore jump into a discussion, or follow the research and delve deep into analyses conducted. The first analysis deals with a standard descriptive statistic overview of the data, found in Subsection~\ref{sub:DDA}, and is a starting point for bigger things to come.

\subsection{Preliminary Data Analyses}
\label{sub:DDA}

The first year in terms of the time span of documents in the analysis is 1991. This is interesting, as this time frame coincides with the dawn of ever-increasing popularity and thrust of computers into masses, into the lives of a larger amount of people. And it seems as if this was happening, so was it more important to find out, and apply, what computers can do?

In Table \ref{tab:dds} descriptive statistic on main information and keywords is found. The number of sources is substantial, therefore authors of research papers have quite a number of avenues to choose from. This is also indicative of relevance to the scientific community if there are so many sources publishing these kinds of papers, and as was seen by performing a literature search, this relevance goes beyond economics, as a large number of research is published in computer science journals.

After final filtering was performed, the number of documents for bibliometrics was 1890 -- a large number and a number where bibliometrics is needed. Such a number of documents shows that we are far from the fledgling days, and gives a reason and a motivation to ascertain what is the current state of the field. It is also indicative of the amount of research results, authors etc., and information supportive of the reasoning that the number of questions in need of resolving is not small, and with the annual growth rate at $ 12.9\% $, with no slowing down in sight as will be seen from further data analysis, this is a place where both academia and industry can find their place in the light. The question always however is, what will be the next big thing? Will the link between quantum computation and algorithmics in the near future be the successor or a parallel art? Time will tell, as the question hangs on the balance of viability of quantum computation, which is not yet certain, but there are indications that quantum computation is here to stay.

Document average age of $ 5.6 $ years shows that the field is propulsive and produces a substantial amount of knowledge that is young, new. Such a situation confronts those wanting to enter the field with a steep learning curve and requires constant learning and self-improvement, but to be working in such an evolving field is also fascinating and rewarding. From all data in this table, but from later analyses as well, this trend will continue, and potentially even increase in speed and shorten the value for the average age of a document.

Considering that the number of documents is 1890, the average citation per document of $ 23.21 $ is high. This has been achieved either by a smaller number of highly cited articles (as later analysis shows a factor), or by other field characteristics such as number of authors, interdisciplinary nature of citing, number of research/industry projects in computing, and more specifically AI, etc. It is however possible, and it seems probable, that here it was a combination of factors that contributed to the situation as is now.

The number of references is also high, as according to these numbers it would mean that the expected value of references per document is 33. Since for a review paper one expects at least 30 references, preferably 50 references or more, the number of 33 is not so far from the review expected number of 50 references. The reasoning for such a situation can probably be projected from the reasoning for citation count, however, what will the future bring is more difficult to tell, as we are dealing here with a multidisciplinary environment -- for the foreseeable future this trend will most likely stay the same. On a different note, documents with this amount of references should be well grounded, and with this amount of citations, relevant for the field.

With this amount of keywords, the expected number for a document is 2, not quite the usually recommended number of 5. It is difficult to tell from the analysis why is this so, it however does not necessarily mean that the papers are not well supported by keywords, as later inspection shows, there might be nevertheless some extremes. Useful for future reference is a data point about the comparison of keywords and keywords plus (generated from references \cite{Garfield1993}), which is roughly $ 1 : 2 $ -- as it shows that it is possible to describe the structure of knowledge with half of the original keywords, but not so well the knowledge itself. \cite{Zhang2015}

\begin{table}[h]
	\caption{Descriptive Statistic on Main Information and Keywords}\label{tab:dds}
	\begin{tabularx}{\textwidth}{@{\extracolsep\fill}llll}
		\toprule%
		\multicolumn{2}{@{}c@{}}{Main information} & \multicolumn{2}{@{}c@{}}{Keywords} \\\cmidrule{1-2}\cmidrule{3-4}%
		Observed Variable & Variable Value & Keywords Plus\textsuperscript{1} & Keywords\textsuperscript{2} \\
		\midrule
		Time span	& 1991 -- 2023 & $ 2015 $ & $ 4342 $ \\
		Sources\textsuperscript{3}	& $ 637 $  &  & \\
		Documents 	& $ 1890 $ & & \\
		Annual Growth Rate {\footnotesize (\%)}	& $ 12.9 $ & & \\
		Document Average Age {\footnotesize (year)} & $ 5.6 $ & & \\
		Average citation {\footnotesize (per doc.)}	& $ 23.21 $ & & \\
		References	& $ 62567 $ & & \\
		\botrule
	\end{tabularx}
	\footnotetext[1]{Algorithmically generated by Web of Science.}
	\footnotetext[2]{Given by an author in the paper.}
	\footnotetext[3]{This research has narrowed down on the most complete and rigorous sources, therefore the category contains journals, and a large amount of those at that. Since there are more than enough documents when only journals are taken into consideration, there was no need to perhaps distort the results of the research with generally less relevant sources such as e.g. conferences.}
\end{table}

Further on, in Table \ref{tab:ac}, one can observe data on authors and collaborations. If we take into account the number of papers, then per paper there comes cca. 2 authors, so the papers are not saturated with authors, with only 225 authors producing a single document, that is $ 5.8\% $ -- indicating that those that are working in the field are typically here for a longer period, with intent to leave a more lasting contribution.

As for collaborations, it can be seen that there are 266 documents that have only one author, a number close to the number of authors that have only one document published, it is possible that a substantial part here comes from those single document authors, as these are less likely to have developed collaboration group. This number also indicates that we are dealing with a field where collaboration is high, as saturation with authors is not high, and there is cca. $ 14\% $ of documents with one author only.

A confirmation of this is seen in the number of authors per document, as the average is $ 2.91 $, a highly collaborative for computer science and economics it would seem, expected and a positive sign for a field that is at the crossroads. International collaboration is here as well, with $ 25.1\% $ of these co-authorships being international, which is not a necessity, but present nevertheless -- this means that one in four collaborations is international, which is on a global scale, with vast differences in conducting research, we would argue a substantial percentage, that likely hides other elements as well, like networking, international collaborative groups, international research projects etc. If further analyses are any indication, the number of authors and international collaboration will probably continue to grow at a significant pace.

\begin{table}[h]
	\caption{Descriptive Statistics on Authors and Collaborations}\label{tab:ac}
	\begin{tabularx}{\textwidth}{@{\extracolsep\fill}lllll}
		\toprule%
		\multicolumn{2}{@{}c@{}}{Authors} & \multicolumn{3}{@{}c@{}}{Collaborations} \\\cmidrule{1-2}\cmidrule{3-5}%
		No. & Single doc. & One author {\footnotesize (per doc.)} & Coauthors {\footnotesize (per doc.)} & International {\footnotesize (\%)}\\
		\midrule
		$ 3879  $ & $ 225 $ & $ 266 $ & $ 2.91 $ & $ 25.08 $ \\
		\botrule
	\end{tabularx}
\end{table}

\begin{figure}[h]%
	\centering
	\begin{tikzpicture}[]
		\begin{axis}[
			ylabel near ticks,
			xlabel near ticks,
			minor xtick={1990, 1992, ..., 2024},
			xtick={1991, 1993, ..., 2023},
			xmin=1990, xmax=2025,
			ymin=-20, ymax=450,
			xlabel={Year of Publication},
			ylabel={Number of Articles},
			grid=major,
			x tick label style={/pgf/number format/1000 sep=},
			width=.9\textwidth,
			xticklabel style={rotate=90},
			axis lines=left,
			]
			\addplot[thick, color=violet]
			table[x=Year, y=Articles, col sep=comma]{podaci/Annual_Production.csv};\label{asp}
			\node[font=\scriptsize, color=brown] at (6, 45){$ 9 $};
			\node[font=\scriptsize, color=brown] at (16, 60){$ 23 $};
			\node[font=\scriptsize, color=brown] at (19, 80){$ 41 $};
			\node[font=\scriptsize, color=brown] at (22, 105){$ 64 $};
			\node[font=\scriptsize, color=brown] at (26, 105){$ 66 $};
			\node[font=\scriptsize, color=brown, fill=white, minimum size=.35cm, label={[color=brown,font=\scriptsize]center:$ 136 $}] at (29, 160){};
			\node[font=\scriptsize, color=brown, fill=white, minimum size=.35cm, label={[color=brown,font=\scriptsize]center:$ 185 $}] at (30, 205){};
			\node[font=\scriptsize, color=brown, fill=white, minimum size=.35cm, label={[color=brown,font=\scriptsize]center:$ 279 $}] at (31, 300){};
			\node[font=\scriptsize, color=brown] at (32, 460){$ 416 $};
		\end{axis}
	\end{tikzpicture}
	\caption{Annual Scientific Production. The year 2023 is not relevant for discussion and interpretation as it is not over and full results are not in yet, however unlike for conferences where it can take quite some time for them to be indexed in Web of Science, journals are indexed quickly and the year 2022 is relevant, which the data itself confirms, as the peak is exactly in 2022 and follows the curve trend, only in 2023 there is an unusual drop in a number of published articles. This drop is very likely only due to incomplete data for a year in progress, and the increasing trend will probably continue, pegged on an ever-increasing explosion of application of artificial intelligence methods.}\label{fig:asp}
\end{figure}
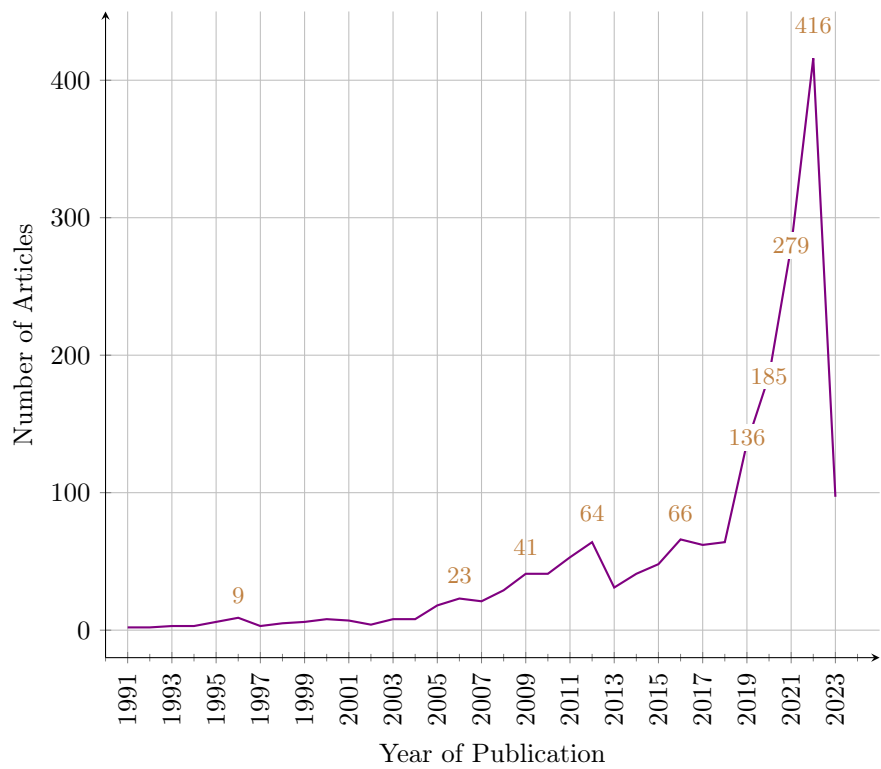

Next, we have an analysis of annual scientific production, in Figure \ref{fig:asp}. Things started slowly, with two articles published in the year of 1991, and this trend lasted until about 2002, only then did AI start getting speed. It took a long time for AI technology to increase enough in maturity, with the additional difficulty of interdisciplinarity, such a development is expected -- so the research was scarce, and contributions were not abundant.

Then in 2003 a difference, the start of a new era, the constant influx of papers, year by year, with an approximately linear trend. As it sometimes is in science, after a technology is discovered it is not immediately recognized, and even less applied, but in the 2000s things are changing and researchers are starting to realize the potential of AI methods, techniques and algorithms. This had a secondary effect as well, the influx of scientists and practitioners to the field, making the field more robust and faster advancing of knowledge.

It was however something totally unheard of when the number of papers started to grow exponentially around 2018, the same year when OpenAI published its first Generative Pretrained Transformer (GPT, often called ChatGPT) \cite{OpenAI2018} -- with 2017 being the year when deep learning transformer architecture was published by a team predominantly from Google Research and Google Brain \cite{Vaswani2017}. This has placed AI on the map, not only in economics, but in many other areas, and some would argue in almost every research area there is -- perhaps all? Such an extremely steep trend in the number of publications has increased the application of AI tremendously, and innovation follows, whether through technology or through methods optimization. From all the information this graph offers, as well as from further analyses below, the trend will continue for the foreseeable future and we will for some time not be seeing a concept drift.

\begin{figure}[h]%
	\centering
	\begin{tikzpicture}[]
		\begin{axis}[
			ylabel near ticks,
			xlabel near ticks,
			minor xtick={1990, 1992, ..., 2024},
			xtick={1991, 1993, ..., 2023},
			xmin=1990, xmax=2025,
			ymin=-.5, ymax=10.5,
			xlabel={Citable Years},
			ylabel={Citation Per Elapsed Years},
			grid=major,
			x tick label style={/pgf/number format/1000 sep=},
			width=.9\textwidth,
			xticklabel style={rotate=90},
			axis lines=left,
			legend style={at={(.62,1.12)},
				anchor=north,legend columns=1,legend cell align={left},draw=none},
			]
			\addlegendentry{\scriptsize Average Citation Per Elapsed Years}
			\addplot[thick, color=violet]
			table[x=Year, y=MeanTCperYear, col sep=comma]{podaci/Annual_Total_Citation_per_Year.csv};\label{tcpy}
			\addlegendentry{\scriptsize Influence of Average Citation Per Elapsed Years}
			\addplot[thick, color=purple]
			table[x=Year, y=Ponderirano, col sep=comma]{podaci/Annual_Total_Citation_per_Year_ponderirano.csv};\label{tcpyp}
			\node[font=\tiny, color=violet] at (2, 4){$ 0.3 $};
			\node[font=\tiny, color=violet] at (4, 103){$ 9.3 $};
			\node[font=\tiny, color=violet] at (7, 24){$ 2.5 $};
			\node[font=\tiny, color=violet] at (10, 87){$ 7.7 $};
			\node[font=\tiny, color=violet] at (13.2, 17){$ 1.4 $};
			\node[font=\tiny, color=violet] at (15, 62){$ 5.1 $};
			\node[font=\tiny, color=violet] at (18, 34){$ 3.4 $};
			\node[font=\tiny, color=violet] at (22, 54){$ 4.4 $};
			\node[font=\tiny, color=violet] at (25, 58){$ 4.8 $};
			\node[font=\tiny, color=violet] at (30, 62){$ 5.1 $};
			\node[font=\tiny, color=violet, fill=white, minimum size=.35cm, label={[color=violet,font=\tiny]center:$ 2.8 $}] at (31, 34){};
			\node[font=\tiny, color=violet] at (33, 10){$ 1.0 $};
			\node[font=\tiny, color=purple] at (4, 30){$ 2.17 $};			
			\node[font=\tiny, color=purple] at (7, 7){$ 0.64 $};			
			\node[font=\tiny, color=purple] at (10, 32){$ 2.24 $};			
			\node[font=\tiny, color=purple] at (13, 7){$ 0.46 $};			
			\node[font=\tiny, color=purple] at (15, 29){$ 1.87 $};			
			\node[font=\tiny, color=purple] at (18, 17){$ 1.47 $};			
			\node[font=\tiny, color=purple] at (22.2, 34){$ 2.54 $};			
			\node[font=\tiny, color=purple] at (25.2, 34){$ 3.76 $};			
			\node[font=\tiny, color=purple] at (30, 101){$ 8.97 $};			
			\node[font=\tiny, color=purple, fill=white, minimum size=.35cm, label={[color=purple,font=\tiny]center:$ 6.60 $}] at (31, 71){};			
			\node[font=\tiny, color=purple] at (33, 81){$ 7.21 $};
		\end{axis}
	\end{tikzpicture}
	\caption{Average Citation Per Elapsed Years (mean citation per article for a particular year divided by the number of citable years, e.g. for 2021 average citation per article is $ 8.5 $ and there are $ 3 $ citable years, therefore one has $ \frac{8.5}{3} = 2.83 $). Average Citations started very slowly, then there was an explosion that was an indicator of future events. The last three years are irrelevant for interpretation, as citations need cca. at least two years in order to accumulate to substantial amount, which is general knowledge in the scientific community, and especially known among journal editors, with some fields needing more time than others. Until 2007 the situation was somewhat erratic, afterward, it seemed as if a calm came, and the field had matured. However, this is deceptive, as the influence of average citation clearly reveals (as citations close to publication date are exceedingly more difficult to accumulate than those that will arrive later on -- calculation was performed as per amortization described in Appendix \ref{secA2}, with the result scaled by a factor of 7 so as to improve readability and try to reveal points of interest).}\label{fig:tcpy}
\end{figure}
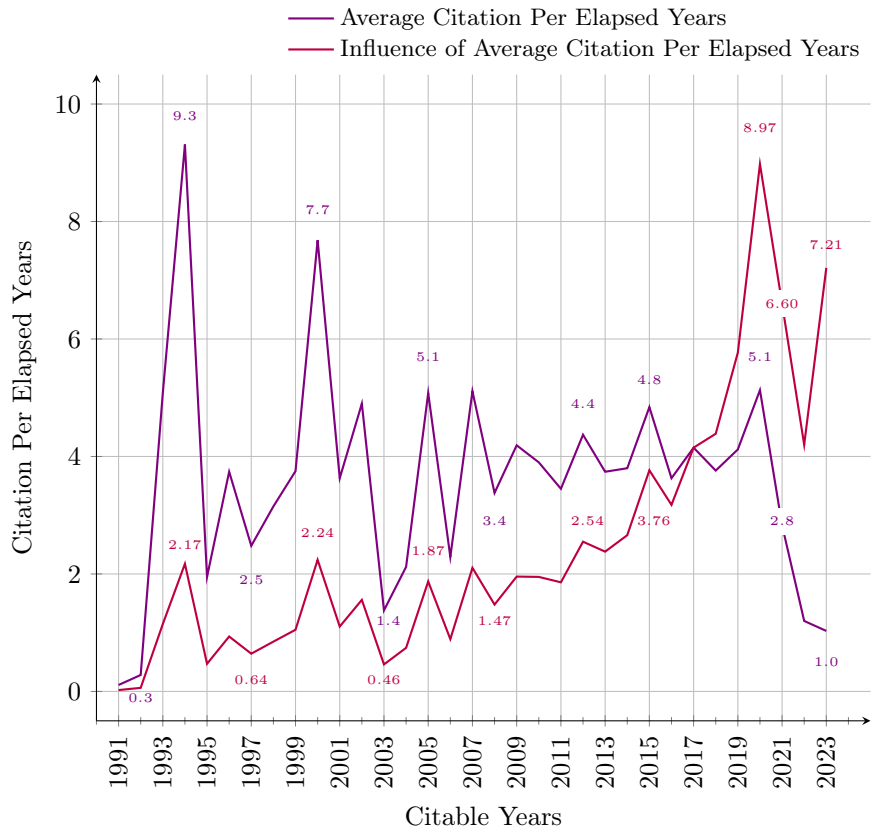

By looking at the influence of average citation per elapsed years, in Figure \ref{fig:tcpy}, after the original explosion, citations grew ever more rapidly, until they ended in the extreme peak of 2020 -- corresponding to AI's rapid development and growth during the cca. last decade. More concretely, we have points of interest, the first is in 1994, however since the number of documents is scarce during preceding years, it is more favorable to select the period until 1999 for inspection. The next period of interest would then be from 2000 until 2004, with the need to specifically determine the situation in the year 2000. Afterward, we have two prominent peaks, in 2005 and then in 2007, therefore the period from 2005 until 2008 needs to be looked at. Then we come to a period of exponential explosion of both published documents and citations. During this time there are three relevant periods that need to be looked at more closely, 2015 -- 2018, 2019 -- 2020, and the last speaking of things to come, from 2021 -- 2023. 

All of these have been analyzed with lesser or greater depth at different stages of the research, but for the most definite conclusion, one should go to Figures \ref{fig:TM2006}, and \ref{fig:TM2019}, with the accompanying interpretation text. Alongside these points, 2017 is the first year where for the first time AI citations grew to such an extent that the influence of those citations was more worthwhile than average citations (influence of average citation was scaled to the average citation range having in mind minimum and maximum value, in a conservative manner), and as per the data, the last leg of quite rapid AI development began around 2015, with the first ending in 2003, 2004.

Furthermore, in 2019 the world was confronted with COVID-19 epidemic conditions, and then pandemic conditions in 2020, which seems had quite an influence on research in a world that become predominantly online and digitally oriented, both in a business environment as well as in private life. The last three years of influence data indicate that the field will continue in, as it seems, exponential growth in the future as well, at least in terms of the influence of average citation -- and if that is any indication of innovation, there are great developments in the works for AI and entrepreneurship-finance, with other areas likely following the same trend.

Average citation per elapsed years shows how at the beginning a small number of papers has through time accumulated a large number of citations and gained substantial relevance. The further down through time one goes average citation stabilizes into a "line", giving an impression that these documents are less relevant in terms of citations -- this however is deceptive, as the average calculation does not take into account that recent years are more significant, in terms of cutting edge of the field, and that it is far more difficult to accumulate citations in a short time. 

To solve this issue we are suggesting the influence of average citation per elapsed year -- a detailed description of the calculation can be found in Appendix \ref{secA2}. The basic idea behind the influence of average citation is that this link between average document citation and the number of elapsed years is exponentially inversely proportional, giving more importance to recent years -- a measure going beyond only taking into account how many years have passed, but at the same time asking what was the strength of those years in terms of citation accumulation and cutting edge of the field.

If we compare the influence of average citation with the Figure \ref{fig:asp} and accompanying events it is clear that such a curve more closely describes the general situation, and therefore it is advisable to take into account both measures, average citation and influence of average citation per elapsed years. 

By observing the influence of the average citation we can discern three periods, and one emergence point. The first period ends around 2003, the second ends around 2015 (while in 2014 the soft-search neural machine mechanism for translation was published \cite{Bahdanau2014}, an important piece in later GPT models), with the last proceeding from around 2015 -- the last two periods are delineated by an emergence point of 2016-2017 transition. Here average citation influence has overtaken average citation, with documents becoming influential in terms of the mentioned fixed point, and highly influential from that point onward (influence of average citation was scaled to average citation range with having in mind minimum and maximum value, in a conservative manner).

Lastly for this analysis, before the emergence point, there were two periods that were influential during their time, the first during 1994-1995, and the second during 2002-2007 -- even though by looking at individual points they are far apart, generally speaking, extending to neighboring points, they are influential, which corresponds to the increased interest in AI, coming of age, and then applicatory and scientific breakthrough during the last few years. As for the foreseeable future, documents will most likely stay in the range of influential to highly influential since AI developments are far from becoming stagnant.

\begin{figure}[h] 
	\centering
	\includegraphics[width=1\textwidth]{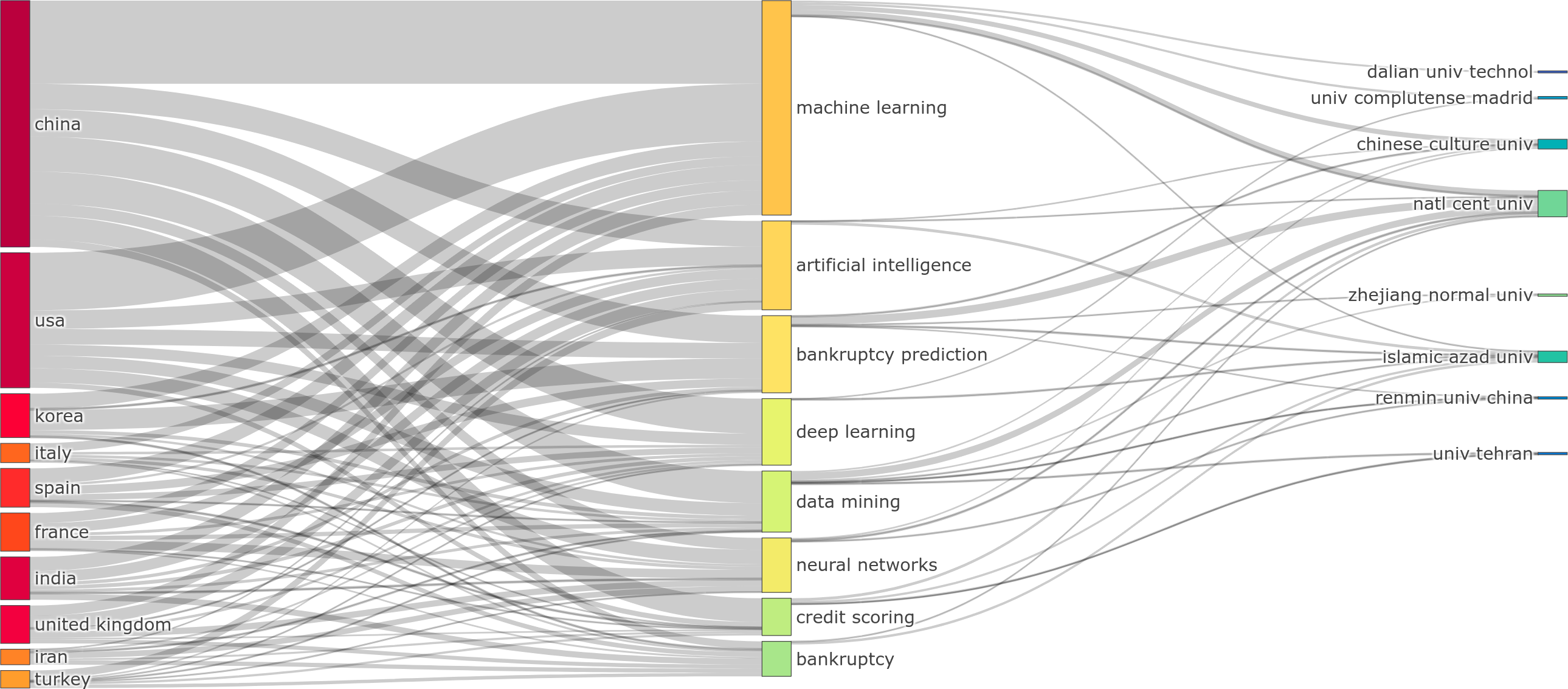}
	\caption{Sankey Diagram for Countries, Keywords and Affiliations.}\label{fig:ZKI}
\end{figure}

As per research objectives, the focus is topics, methods, techniques and algorithms in the context of AI, Finance and entrepreneurship, therefore a number of Sankey diagrams were created so as to place these in relation to other items of interest -- all diagrams were created with the same settings, defining the maximum number of items so as to leverage depth and breadth. In Figure \ref{fig:ZKI} we see, as per central pillar, prevalent topics, in general Artificial Intelligence paired with performance evaluation and credit assessment. These themes are overly general in order to dive deep, but they do give a broad overview of what is being researched. If bankruptcy and bankruptcy prediction are merged, then bankruptcy is a bigger element than artificial intelligence -- essentially making the field predominantly about machine learning and performance evaluation.

On the left, countries producing those themes are placed, with China and the US in the lead, followed by India, the United Kingdom, Korea and Spain. China and the USA have the biggest footprint in machine learning, approximately equal, and the same goes for AI as well, while Korea has a substantial footprint in bankruptcy prediction. The computer science side of things is dominant, with performance evaluation having a significant but substantially smaller piece of relevance. These three countries are in the lead -- other countries are significant but more dispersed across topics.

On the right one can observe affiliations, and as the data shows, National Central University, Islamic Azad University, and Chinese Culture University are the biggest. National Central University's largest contribution is in machine learning, bankruptcy prediction, and data mining, without having an impact in deep learning. Islamic Azad University's impact is dispersed, approximately evenly, without having an impact in neural networks. Chinese Culture University's greatest impact is in machine learning and bankruptcy prediction, without having an impact in deep learning and credit scoring. Out of 8 universities, five are Chinese, according to the data on the left. No university covers all topics, and perhaps an interesting point, both National Central University and Chinese Culture University are having a substantial impact in machine learning, and bankruptcy prediction, without having an impact in deep learning, perhaps an indication of aligned focus. Out of the universities making the list, Dalian University of Technology has strict specialization in machine learning, the only such case here, it however does not have the greatest impact, regardless of the aforementioned, as well perhaps an indication of a focus, reason of which is unclear.

\begin{figure}[h] 
	\centering
	\includegraphics[width=1\textwidth]{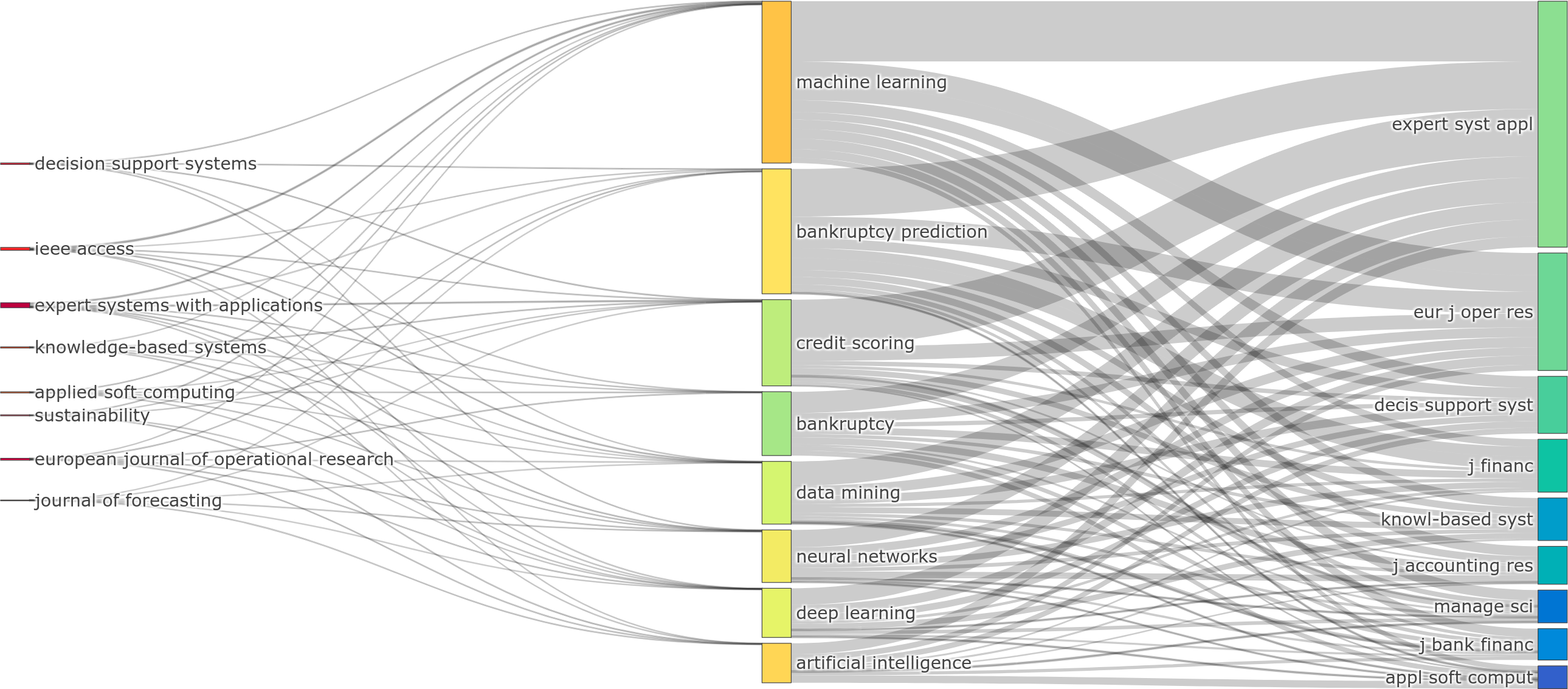}
	\caption{Sankey Diagram for Sources, Keywords and Source Citations.}\label{fig:IKC}
\end{figure}

As we are primarily interested in concepts, that is what will the content reveal, we will again keep as a central pillar keywords, and place them into juxtaposition with sources and source citations, Figure \ref{fig:IKC}. Topics are the same as in Figure \ref{fig:ZKI}, with the size of the footprint being the difference, and aside from the fact of machine learning domination, computer science is in the back of the queue -- bankruptcy and machine learning are again at the forefront, just in reversed order this time, being an indication of a larger picture. Thus it seems that top countries are more geared towards the technological aspect, as observed in Figure \ref{fig:ZKI}, while top sources are geared more towards the practical domain, as seen in Figure \ref{fig:IKC}.

On the left, we see sources, Expert Systems with Applications, IEEE Access, and the European Journal of Operational Research (EJOR) as being the most prominent ones. All the sources are dispersed in the themes they cover, and it is difficult to single anyone out, with not all the sources covering every topic -- so there is a certain amount of specialization there. If we look on the right side, we observe sources again, but citations in this case.

One would expect, when we compare sources and their citations, for these lists to have substantial overlap, and they have, but there are surprises as well. Out of the 8 journals on the left, three are missing on the right: IEEE Access, Sustainability, and Journal of Forecasting. Their output was not enough to earn them citation relevance on the right side, not to put excessive emphasis on this and discredit these journals, but could this kind of analysis be relevant, as one of the factors, in trying to detect predatory journals? The absence of IEEE Access is of special question, as its imprint on the left is significant, of course a broader analysis would probably include that journal as well, but it is strange enough to ask the question, what has happened so that these three journals are missing on the other side.

As for the situation with source citations, Expert Systems with Applications reigns supreme, a first one on the left side as well, with European Journal of Operational Research, and Decision Support Systems (a third place here, but not as relevant on the left, showing that output is not a foolproof way for high citation count, but not irrelevant either). Newcomers here, not found on the left, are the Journal of Finance, Journal of Accounting Research, Management Science, and Journal of Banking \& Finance, in spite of not being so highly relevant in terms of output and citation they are very impactful.

When we look at the journals themselves, we see that they correspond well with the topics in the middle, with computer science journals having the lead, indicating that the central point of the papers is the method itself, with the application domain having a supportive role. These kinds of analyses can be used by authors to decide in which journal they want to publish, e.g. Expert Systems with Applications publishes in all topics and very highly contributes in terms of citation count to machine learning, bankruptcy and credit scoring, with 2nd and 3rd journal of the right following a similar pattern.

\begin{figure}[h] 
	\centering
	\includegraphics[width=1\textwidth]{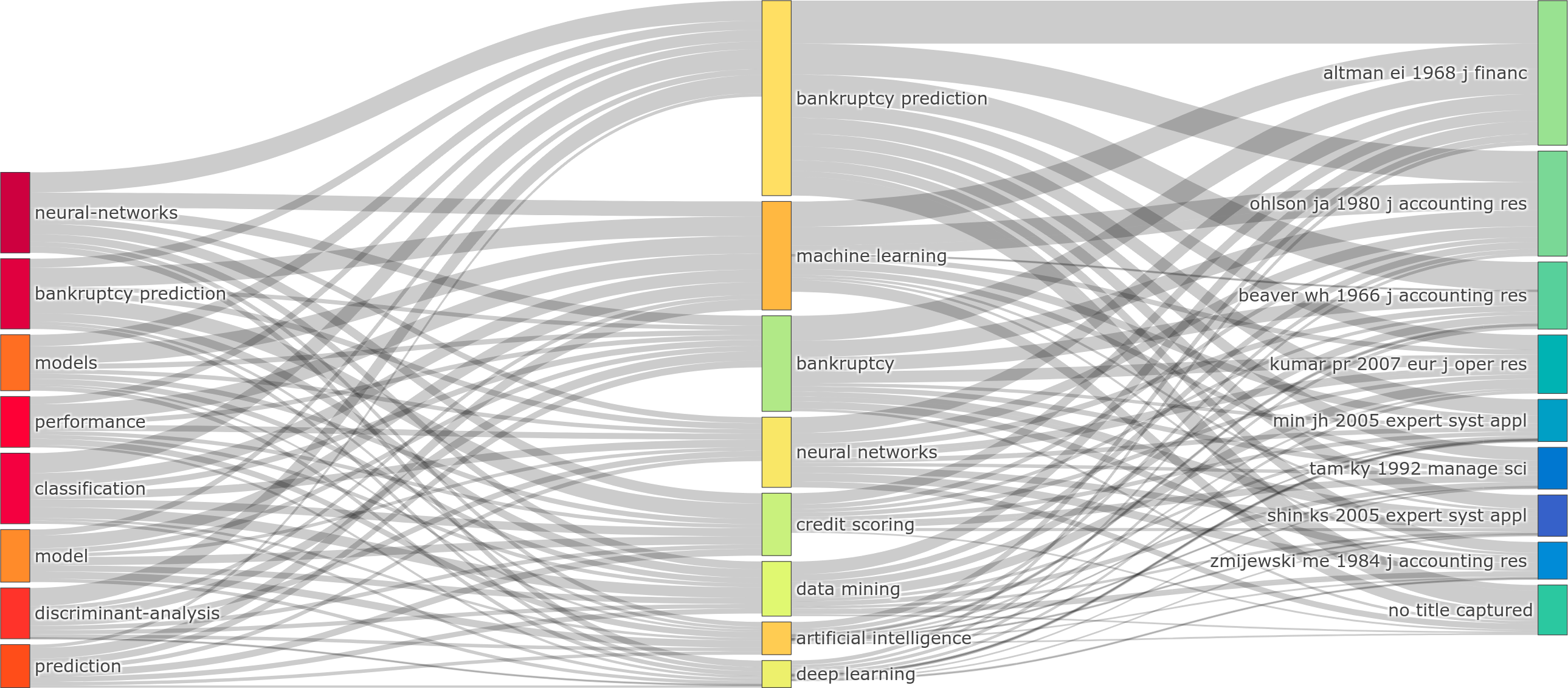}
	\caption{Sankey Diagram for Keywords Plus, Keywords and References.}\label{fig:KKR}
\end{figure}

The last analysis in this subsection is in Figure \ref{fig:KKR}, a Sankey diagram and a continuation of the previous ones. Keywords are in the middle, we are on purpose having the same fixed point for every Sankey, as this allows for a comparison between diagrams and pillars. Topics are the same, performance themes are again dominant, and bankruptcy prediction is vastly superior, a consequence of selected variables to create the diagram from, with machine learning and neural networks following. There is a pattern here, it seems that in a general case, performance evaluation is in the lead, followed by a substantial presence of artificial intelligence, machine learning in particular. A logical consequence of analyzing a field that is an intersection of AI, finance, and entrepreneurship -- the field is geared towards economic themes and is strongly drawing on AI, essentially the application of AI in finance and entrepreneurship, with the most prominent journal being Expert Systems with Applications, a fitting name for the purpose, aligning with field research activities.

All topics are quite strongly connected to all the elements on the left, keywords plus. Considering that keywords plus are algorithmically generated from document references, they represent foundational knowledge on which contributions rest, and as it can be seen, that foundational knowledge predominantly is bankruptcy prediction and neural networks, together with a number of themes as, far as it can be gathered, from AI, economics and statistics. In order to produce contributions in the middle, researchers have stood on the existing models, knowledge on prediction, classification, performance evaluation and statistics, coupled with the influence of neural nets -- this is what the analysis of top items reveals.

On the right, references can be seen, these represent the intellectual roots of the field. The top three references contributing the most to topics of interest in the middle are Altman, Ohlson, and Beaver -- however "none of their work is based on an artificial intelligent-based approach, due to the fact that all aforementioned work are pioneer studies in the bankruptcy prediction field, the posterior authors tend to cite them in their papers with high frequency." \cite{Shi2019} Out of all other references those that are AI-based are: Kumar, Min, Tam, Shin; while Zmijewski is dealing with methodological issues and financial distress prediction. There are an equal number of economics-based (also combined with other aspects) vs economics AI-based papers, with the constraint of one paper not having their title captured, indicating that both fields have approximately the same relevance.

When one compares intellectual roots on the right and foundational knowledge on the left there is correspondence, but things are changing. On the left it seems that prevalent themes are largely in economics, reasonable results as economics are the insider here, while on the right we see the same number of references for those more leaning toward economics and those that are AI-based, indicating a potential change in direction as the recent references are all AI-based. 

Computer science is increasingly entering economics thought, and it seems that references, through which foundational knowledge also, are becoming AI-based, and intellectual roots of the field are being modified. This will most likely continue and will produce even more citations and scientific contributions where AI and economics go hand in hand. The question is, will there come a time when AI will be so intertwined with economics and society that AI-based papers will be all to see here? As is on the left, so is similarly on the right, references are linked to all or almost all topics in the middle, indicating a broad relevance.

\subsection{Sources Data Analyses}
\label{sub:SDA}

In the sources data analyses the first insight to consider is in Figure \ref{fig:MRSleg}, here we see sources and their respective output in terms of number of documents. By far the most published source is Expert Systems with Applications, after which with a huge falling behind there comes Computational Intelligence and Neuroscience, others are decreasingly following with every previous one being not that far behind. ESA clearly shows a great deal of specialization for the field at hand and is a great venue for authors to publish in, with other sources also being a potential place for research publication. These are also publications that one can read and be well-informed about the subject.

Out of all the sources, the Journal of Forecasting is the only one that has an economic, social and behavioral focus, with not that strong computer science leaning. All others range from small to very large computer science footprints, indicating how computer science sources present a fertile ground for application-based and interdisciplinary research. Economics journals, and perhaps most others as well, are in a difficult situation in ascertaining the merit of research that has strong computer science, and especially artificial intelligence, elements. This situation is also potentially an indication that computer experts are dominant authors, with them leading and carrying out the research.

\definecolor{ceil}{rgb}{0.57, 0.63, 0.81}
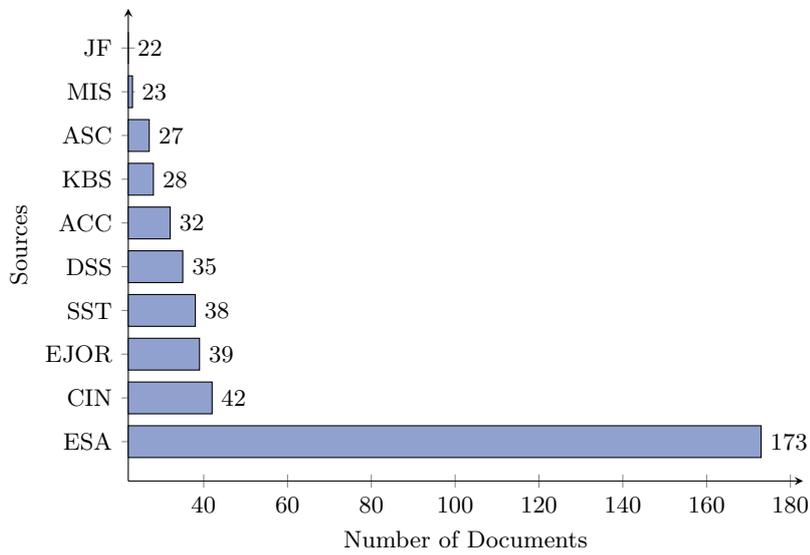
\begin{figure}[h]
	\centering
	\begin{tikzpicture}[font=\small]
		\begin{axis}[
			tickwidth = 5pt,
			ylabel near ticks,
			xlabel near ticks,
			xbar,
			bar width=12pt,
			xmax=183,
			xlabel={Number of Documents},
			ylabel={Sources},
			axis x line=bottom,
			axis y line=left,
			nodes near coords,
			enlarge y limits=0.1,
			width=.8\textwidth,
			height=.6\textwidth,
			symbolic y coords={
				ESA,
				CIN,
				EJOR,
				SST,
				DSS,
				ACC,
				KBS,
				ASC,
				MIS,
				JF
			},
			ytick = data,
			]
			\addplot[fill=ceil] coordinates {			
				(173,ESA)
				(42,CIN)
				(39,EJOR)
				(38,SST)
				(35,DSS)
				(32,ACC)
				(28,KBS)
				(27,ASC)
				(23,MIS)
				(22,JF)		
			};
		\end{axis}
	\end{tikzpicture}
	\caption{Most Relevant (in terms of number of documents published) Sources, ending in May 2023. As source titles are excessively long, they are abbreviated, with a legend being given below the figure.}\label{fig:MRS}
	\footnotesize
	\begin{tabular}{rl}
		& \\
		ESA		& Expert Systems with Applications \\
		CIN 	& Computational Intelligence and Neuroscience \\
		EJOR	& European Journal of Operational Research \\
		SST 	& Sustainability \\
		DSS 	& Decision Support Systems \\
		ACC 	& IEEE Access \\
		KBS 	& Knowledge-Based Systems \\
		ASC 	& Applied Soft Computing \\
		MIS 	& Mobile Information Systems \\
		JF		& Journal of Forecasting
	\end{tabular}\label{fig:MRSleg}
\end{figure}

In \ref{fig:MLCSleg} we are continuing with most local cited sources. ESA is in the strong lead again, considering the journal's output in the number of documents and it seems high specialization this is far from surprising. However there are surprises, as a number of sources from Figure \ref{fig:MRSleg} are here missing, and new sources have entered the door. This time economics journals have a more commanding presence in terms of relevance, and there could be multiple causes. It might be that some sources are not outputting relevant content, it might also be that citation practice is influencing these results, it could also happen that lack of expertise in computer science is producing lower citation count for certain journals, or it might be that in a predominantly economic field economics takes the center stage.

These sources can also be inspected in terms of intellectual roots, as they are the ones that are being cited in so many documents, and represent the publication branch of the field. By having economics sources more relevant, there is a coupling of themes, a consequence of which is so strong a link between AI and performance evaluation, as seen in previous analyses. Data in this figure, as well as many others, follows the Pareto distribution, and it is interesting how many phenomena follow such or similar distribution, and how many potential errors can be detected by inspecting whether or not something is following such a curve.

\definecolor{antiquebrass}{rgb}{0.8, 0.58, 0.46}
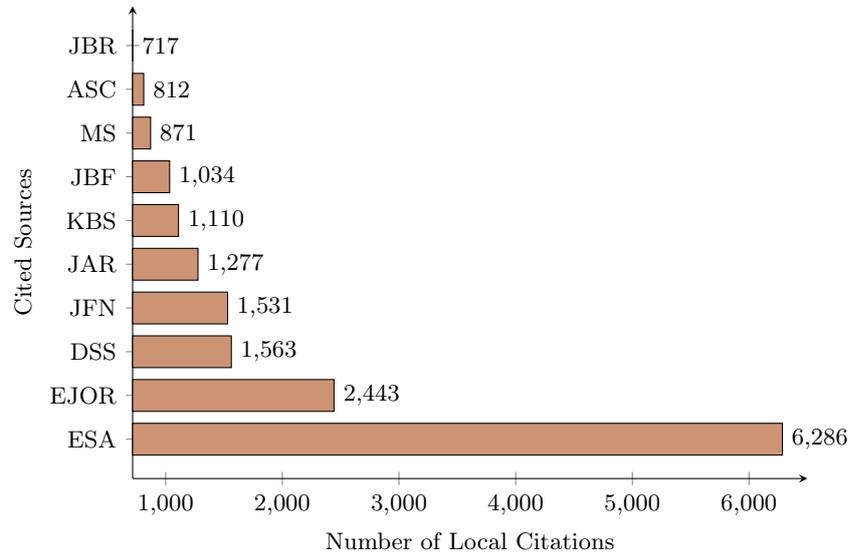
\begin{figure}[h]
	\centering
	\begin{tikzpicture}[font=\small]
		\begin{axis}[
			tickwidth = 5pt,
			ylabel near ticks,
			xlabel near ticks,
			xbar,
			bar width=12pt,
			xmax=6500,
			xlabel={Number of Local Citations},
			ylabel={Cited Sources},
			axis x line=bottom,
			axis y line=left,
			nodes near coords,
			enlarge y limits=0.1,
			width=.8\textwidth,
			height=.6\textwidth,
			symbolic y coords={
				ESA,
				EJOR,
				DSS,
				JFN,
				JAR,
				KBS,
				JBF,
				MS,
				ASC,
				JBR,
			},
			ytick = data,
			]
			\addplot[fill=antiquebrass] coordinates {			
				(6286,ESA)
				(2443,EJOR)
				(1563,DSS)
				(1531,JFN)
				(1277,JAR)
				(1110,KBS)
				(1034,JBF)
				(871,MS)
				(812,ASC)
				(717,JBR)
			};
		\end{axis}
	\end{tikzpicture}
	\caption{Most Local (within our own data, not including every single citation in existence) Cited Sources, ending in May 2023. As cited source titles are excessively long, they are abbreviated, with a legend being given below the figure.}\label{fig:MLCS}
	\footnotesize
	\begin{tabular}{rl}
		& \\
		ESA		& Expert Systems with Applications \\		
		EJOR	& European Journal of Operational Research \\
		DSS 	& Decision Support Systems \\
		JFN		& Journal of Finance \\
		JAR		& Journal of Accounting Research \\
		KBS 	& Knowledge-Based Systems \\
		JBF		& Journal of Banking \& Finance \\
		MS		& Management Science \\
		ASC 	& Applied Soft Computing \\
		JBR		& Journal of Business Research
	\end{tabular}\label{fig:MLCSleg}
\end{figure}

With that being said we can turn our attention to Table \ref{tab:CSBL}, determining core sources as per Bradford's Law. A more detailed look than before, 20 sources, with ESA being in great lead and being the number one source. 

ESA is followed by other sources -- here as well computer science journals are dominating, or at least the ones having a computer science component. Per the number of articles, after ESA, sources are close to one another, most likely a consequence of source scope, journal relevance, publisher, prominence, etc. 

If one takes a closer look at the total sum, it can be seen that the sum equals 629, a substantial amount of papers, since the number of papers analyzed is 1890, this makes $ 33.28 \% $ -- in accord with Bradford's Law, a high amount of documents, published in a small number of sources, if we recall that total number of sources was 637, this makes only $ 3.13\% $. A classical example of a center of power so to speak, often seen in nature, and the world we live in, that naturally leads authors to gravitate to these sources.

\begin{table}[h]
	\caption{Core Sources by Bradford's Law -- a well-known law in science that approximates exponentially diminishing returns in a subsequent action for substantially larger corpus (frequently stated as $ 1 : n : n^{2} $) \cite{Bradford1934,Brookes1985}, sometimes called Pareto Distribution \cite{Arnold2008}. Below we are giving the first zone for a Bradford's Law calculation in regard to the number of articles for sources of those articles.}\label{tab:CSBL}
	\begin{tabular}{@{}lll}
		\toprule
		No. of Articles (NA) & Source & $\sum$ NA	\\
		\midrule
		173	& Expert Systems with Applications	& 173	\\
		42	& Computational Intelligence and Neuroscience & 215	\\
		39	& European Journal of Operational Research	& 254	\\
		38	& Sustainability	& 292	\\
		35	& Decision Support Systems	& 327	\\
		32	& IEEE Access	& 359	\\
		28	& Knowledge-Based Systems	& 387	\\
		27	& Applied Soft Computing	& 414	\\
		23	& Mobile Information Systems	& 437	\\
		22	& Journal of Forecasting	& 459	\\
		20	& Annals of Operations Research & 479	\\
		20	& Intelligent Systems in Accounting Finance \& Management & 499	\\
		20	& Journal of Risk and Financial Management	& 519	\\
		17	& Scientific Programming	& 536	\\
		16	& Computational Economics	& 552	\\
		16	& Mathematical Problems in Engineering	& 568	\\
		16	& Neural Computing \& Applications	& 584	\\
		16	& Neurocomputing	& 600	\\
		15	& Applied Sciences-Basel	& 615	\\
		14	& Information Sciences	& 629	\\
		\botrule
	\end{tabular}
\end{table}

Following a strain of thought, we are arriving at $ h $-index, in Figure \ref{fig:SLI}, a measure combining both the number of documents and citation count. If we compare this analysis with the one in Figure \ref{fig:MLCSleg}, there is a clear difference, some publications are perhaps in the extreme with a small number of highly cited documents, which index $ h $ will demote -- 5 out of 10 are missing, a substantial amount, a warning for ascertaining an object with one measure only.

ESA is once more on top, with others following, but far behind -- with one measure after another, ESA confirms its relevance, making a strong case for its top position. From EJOR onward, other publications are gradually following, most likely a situation similar to that before. As a whole, by observing all sources and their corresponding indexes, a Pareto distribution as well. When looking at not only citations but the spread of those citations as well, these are the sources that are relevant.

\definecolor{yellow-green}{rgb}{0.6, 0.8, 0.2}
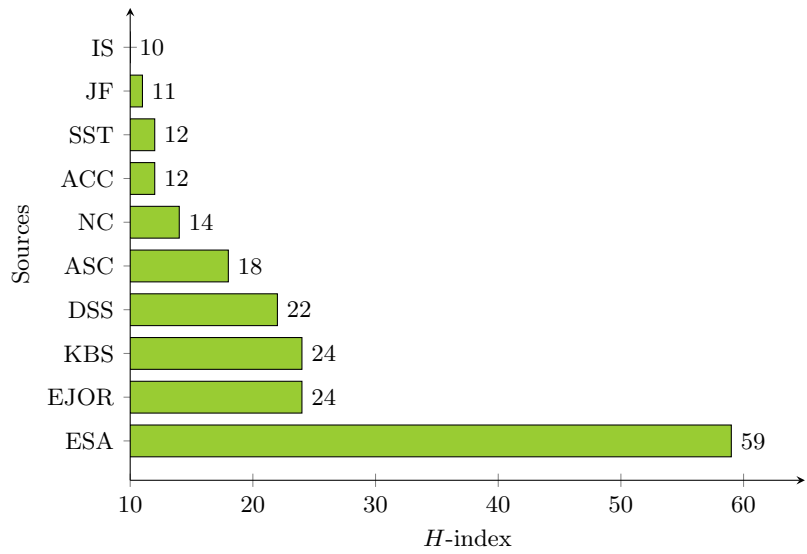
\begin{figure}[h]
	\centering
	\begin{tikzpicture}[font=\small]
		\begin{axis}[
			tickwidth = 5pt,
			ylabel near ticks,
			xlabel near ticks,
			xbar,
			bar width=12pt,
			xmax=65,
			xlabel={$ H $-index},
			ylabel={Sources},
			axis x line=bottom,
			axis y line=left,
			nodes near coords,
			enlarge y limits = 0.1,
			width=.8\textwidth,
			height=.6\textwidth,
			symbolic y coords={
				ESA,
				EJOR,
				KBS,
				DSS,
				ASC,
				NC,
				ACC,
				SST,
				JF,
				IS,
			},
			ytick = data,
			]
			\addplot[fill=yellow-green] coordinates {			
				(59,ESA)
				(24,EJOR)
				(24,KBS)
				(22,DSS)
				(18,ASC)
				(14,NC)
				(12,ACC)
				(12,SST)
				(11,JF)
				(10,IS)
			};
		\end{axis}
	\end{tikzpicture}
	\caption{Sources Local Impact (index $ h $, proposed by Hirsch in 2005 \cite{Hirsch2005} for measuring not only citations but output also, the definition of which is, "the number of papers with citation number $ \geq h $"), ending in May 2023. As cited source titles are excessively long, they are abbreviated, with a legend being given below the figure.}\label{fig:SLI}
	\footnotesize
	\begin{tabular}{rl}
		& \\
		ESA		& Expert Systems with Applications \\			
		EJOR	& European Journal of Operational Research \\
		KBS 	& Knowledge-Based Systems \\
		DSS 	& Decision Support Systems \\
		ASC 	& Applied Soft Computing \\
		NC		& Neurocomputing \\
		ACC 	& IEEE Access \\
		SST 	& Sustainability \\
		JF		& Journal of Forecasting \\
		IS		& Information Sciences
	\end{tabular}\label{fig:SLIleg}
\end{figure}

\begin{figure}[h]
	\centering
	\begin{tikzpicture}[]
		\begin{axis}[
			ylabel near ticks,
			xlabel near ticks,
			minor xtick={1990, 1992, ..., 2024},
			xtick={1991, 1993, ..., 2023},
			xmin=1990, xmax=2025,
			ymin=-.5, ymax=190,
			xlabel={Publication Years},
			ylabel={Number of Documents},
			grid=major,
			x tick label style={/pgf/number format/1000 sep=},
			width=1\textwidth,
			xticklabel style={rotate=90},
			axis lines=left,
			legend style={at={(0.5,-0.17)},
				anchor=north,legend columns=2,legend cell align={left},draw=none},
			]
			\addlegendentry{\scriptsize Expert Systems with Applications}
			\addplot[thick, color=red] table[x=Year, y={EXPERT SYSTEMS WITH APPLICATIONS}, col sep=comma]{podaci/Source_Dynamics.csv};\label{eswa}
			\addlegendentry{\scriptsize IEEE Access}
			\addplot[thick, color=gray] table[x=Year, y={IEEE ACCESS}, col sep=comma]{podaci/Source_Dynamics.csv};\label{acc}
			\addlegendentry{\scriptsize European Journal of Operational Research}
			\addplot[thick, color=blue] table[x=Year, y={EUROPEAN JOURNAL OF OPERATIONAL RESEARCH}, col sep=comma]{podaci/Source_Dynamics.csv};\label{ejor}
			\addlegendentry{\scriptsize Sustainability}
			\addplot[thick, color=yellow-green] table[x=Year, y={SUSTAINABILITY}, col sep=comma]{podaci/Source_Dynamics.csv};\label{sst}
			\addlegendentry{\scriptsize Decision Support Systems}
			\addplot[thick, color=magenta] table[x=Year, y={DECISION SUPPORT SYSTEMS}, col sep=comma]{podaci/Source_Dynamics.csv};\label{dss}
			\addlegendentry{\scriptsize Knowledge-Based Systems}
			\addplot[thick, color=black] table[x=Year, y={KNOWLEDGE-BASED SYSTEMS}, col sep=comma]{podaci/Source_Dynamics.csv};\label{kbs}
			\addlegendentry{\scriptsize Computational Intelligence and Neuroscience}
			\addplot[thick, color=green] table[x=Year, y={COMPUTATIONAL INTELLIGENCE AND NEUROSCIENCE}, col sep=comma]{podaci/Source_Dynamics.csv};\label{cins}
			\addlegendentry{\scriptsize Applied Soft Computing}
			\addplot[thick, color=orange] table[x=Year, y={APPLIED SOFT COMPUTING}, col sep=comma]{podaci/Source_Dynamics.csv};\label{asc}
			\addlegendentry{\scriptsize Mobile Information Systems}
			\addplot[thick, color=brown] table[x=Year, y={MOBILE INFORMATION SYSTEMS}, col sep=comma]{podaci/Source_Dynamics.csv};\label{mis}
			\addlegendentry{\scriptsize Journal of Forecasting}
			\addplot[thick, color=teal] table[x=Year, y={JOURNAL OF FORECASTING}, col sep=comma]{podaci/Source_Dynamics.csv};\label{jf}
		\end{axis}
	\end{tikzpicture}
	\caption{Sources Cumulative Document Production Over Time.}\label{fig:SCDMOT}
\end{figure}
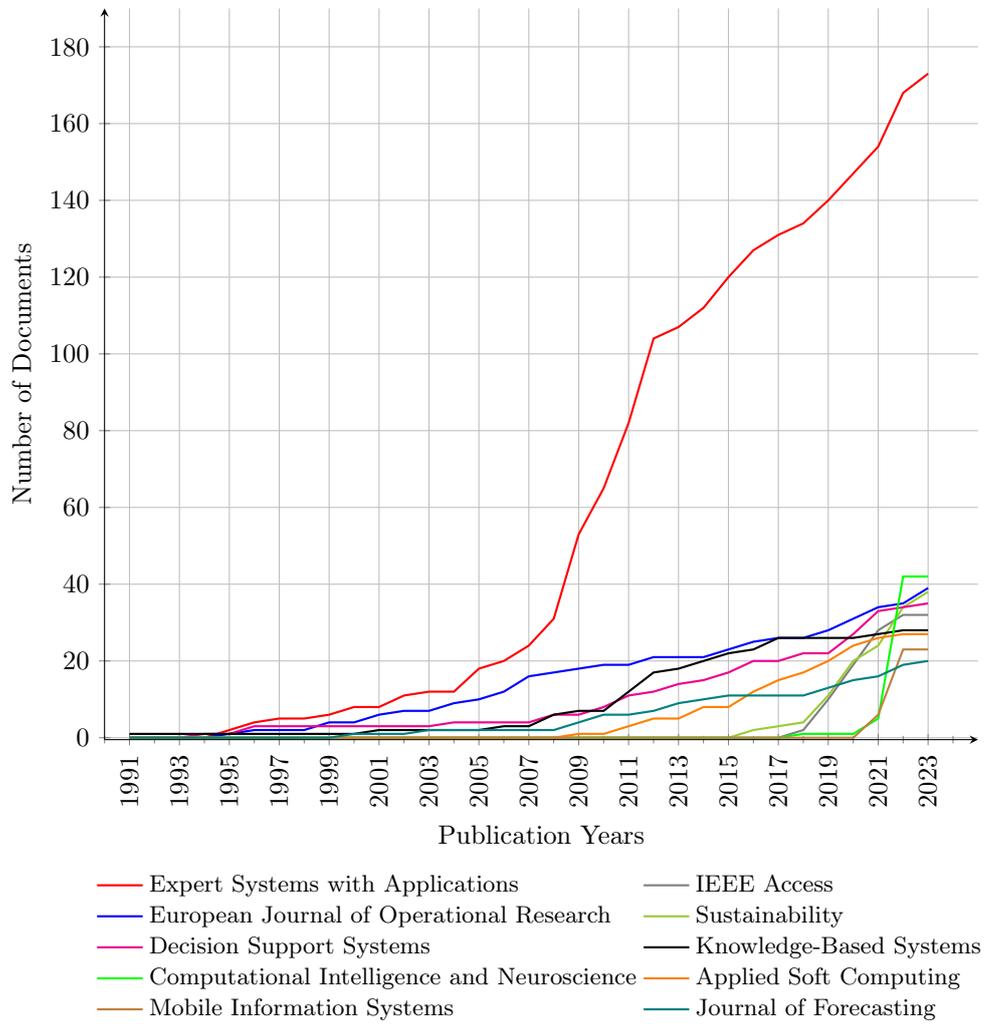

In Figure \ref{fig:SCDMOT} we can observe sources of cumulative production of documents. All sources started slowly in the early 1990s, after which as per field development and an increasing AI presence document production has for the most part steadily increased. There are however a number of points of interest.

ESA approximately follows the same trend as other sources, but then in cca. 2008 there is a takeoff that has continued to this day, indicating that this journal has very early on recognized AI relevance in finance and entrepreneurship, and it seems established a strong venue for these kinds of papers. During this time, other journals have also increased their output, there is an obvious change here, but far less than ESA.

There are other moments in time, with Sustainability, Computational Intelligence and Neuroscience, Mobile Information Systems, and IEEE Access. A long time has passed since these have started to publish a substantial amount of documents in this crossroad field, with one of the factors being the lifetime of the journal. With that in mind, both authors and the journals needed to recognize the importance of the area scope, and as well so to speak find each other.

During the first three years of the entire period, there was only one journal that was published, and that was Knowledge-Based Systems. This journal has published one document every year of those first three, after which this trend has varied from low to moderate, and is at the moment 7th journal in terms of the number of documents published. It is perhaps a strange fact that this journal that was a pioneer in publishing such papers is not publishing more, on the other hand, neither document output, nor citations, etc. is an indication of quality, which can be ascertained only by reading the paper, and it is possible that the journal has rigorous criteria for accepting a document for publication -- of course with other possibilities for this situation also.

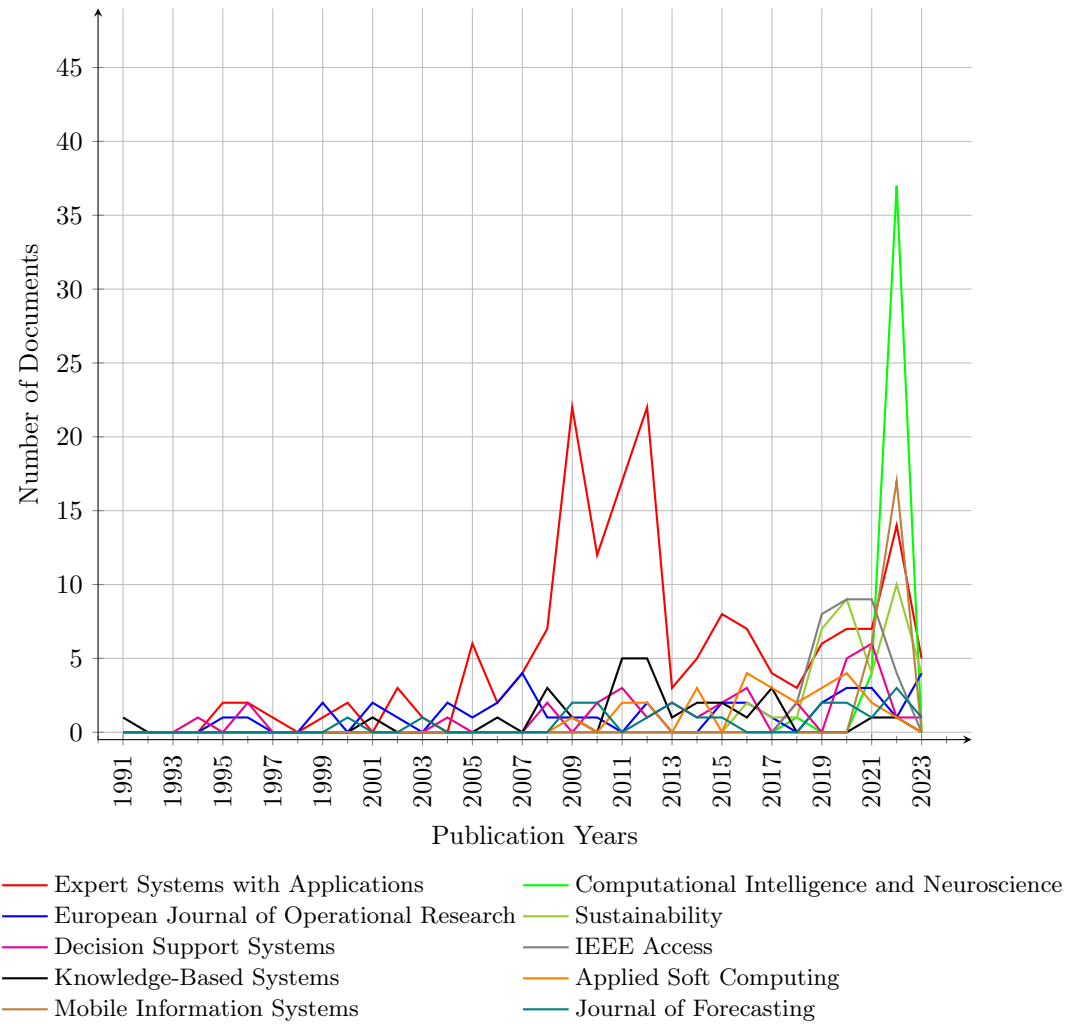
\begin{figure}[h]
	\centering
	\begin{tikzpicture}[]
		\begin{axis}[
			ylabel near ticks,
			xlabel near ticks,
			minor xtick={1990, 1992, ..., 2024},
			xtick={1991, 1993, ..., 2023},
			xmin=1990, xmax=2025,
			ymin=-.5, ymax=49,
			xlabel={Publication Years},
			ylabel={Number of Documents},
			grid=major,
			x tick label style={/pgf/number format/1000 sep=},
			width=1\textwidth,
			xticklabel style={rotate=90},
			axis lines=left,
			legend style={at={(0.5,-0.17)},
				anchor=north,legend columns=2,legend cell align={left},draw=none},
			]
			\addlegendentry{\scriptsize Expert Systems with Applications}
			\addplot[thick, color=red] table[x=Year, y={EXPERT SYSTEMS WITH APPLICATIONS}, col sep=comma]{podaci/Source_Dynamics_Year.csv};\label{esway}
			\addlegendentry{\scriptsize Computational Intelligence and Neuroscience}
			\addplot[thick, color=green] table[x=Year, y={COMPUTATIONAL INTELLIGENCE AND NEUROSCIENCE}, col sep=comma]{podaci/Source_Dynamics_Year.csv};\label{cinsy}
			\addlegendentry{\scriptsize European Journal of Operational Research}
			\addplot[thick, color=blue] table[x=Year, y={EUROPEAN JOURNAL OF OPERATIONAL RESEARCH}, col sep=comma]{podaci/Source_Dynamics_Year.csv};\label{ejory}
			\addlegendentry{\scriptsize Sustainability}
			\addplot[thick, color=yellow-green] table[x=Year, y={SUSTAINABILITY}, col sep=comma]{podaci/Source_Dynamics_Year.csv};\label{ssty}
			\addlegendentry{\scriptsize Decision Support Systems}
			\addplot[thick, color=magenta] table[x=Year, y={DECISION SUPPORT SYSTEMS}, col sep=comma]{podaci/Source_Dynamics_Year.csv};\label{dssy}
			\addlegendentry{\scriptsize IEEE Access}
			\addplot[thick, color=gray] table[x=Year, y={IEEE ACCESS}, col sep=comma]{podaci/Source_Dynamics_Year.csv};\label{accy}
			\addlegendentry{\scriptsize Knowledge-Based Systems}
			\addplot[thick, color=black] table[x=Year, y={KNOWLEDGE-BASED SYSTEMS}, col sep=comma]{podaci/Source_Dynamics_Year.csv};\label{kbsy}
			\addlegendentry{\scriptsize Applied Soft Computing}
			\addplot[thick, color=orange] table[x=Year, y={APPLIED SOFT COMPUTING}, col sep=comma]{podaci/Source_Dynamics_Year.csv};\label{ascy}
			\addlegendentry{\scriptsize Mobile Information Systems}
			\addplot[thick, color=brown] table[x=Year, y={MOBILE INFORMATION SYSTEMS}, col sep=comma]{podaci/Source_Dynamics_Year.csv};\label{misy}
			\addlegendentry{\scriptsize Journal of Forecasting}
			\addplot[thick, color=teal] table[x=Year, y={JOURNAL OF FORECASTING}, col sep=comma]{podaci/Source_Dynamics_Year.csv};\label{jfy}
		\end{axis}
	\end{tikzpicture}
	\caption{Sources Yearly Document Production Over Time.}\label{fig:SCDMOTy}
\end{figure}

By looking at sources' yearly production of documents, in Figure \ref{fig:SCDMOTy}, the image can be additionally cleared up. When ESA had a document renaissance, both the European Journal of Operational Research, and Knowledge-Based Systems started to join the trend, but only Expert Systems with Applications has continued it.

By the end of the period, around 2016 and onward, ESA is not as drastically ahead as it seemed in cumulative analysis, but it is in the top three nonetheless, with the Decision Support Systems' peak much more elucidated here. If we observe 2022, ESA is strong, however, Mobile Information Systems is a small amount ahead, with Computational Intelligence and Neuroscience jumping to an enormous lead, indicating that perhaps a change is ahead, with other journals taking center stage, and ESA behind, but following -- such a situation could be a consequence of ESA saturation, and other journals taking more of a prominent role in AI, entrepreneurship and finance.

Journals are typically trying to be of quality and on the cutting edge, this can potentially lead to an overproduction, has this happened here we can't say for certain, but considering the data, for a number of journals it is a possibility.

\subsection{Authors Data Analyses}
\label{sub:ADA}

When conducting bibliometric analysis of scientific data it is natural to perform an analysis of the author's data and interpret the author's relevant metrics (number of published documents, citations, $ h $-index, etc.), so as to see who's who in the field. 

It was however not possible to perform such an analysis, as during research it was detected that Bibliometrix \cite{Aria2017,Aria2023} has some serious issues in that respect, as the unique identification of authors is not working as expected -- in one situation it was so extreme where for a particular author's complete name abbreviation the program used about $ 10 $ different authors in order to calculate a metric\footnote{We have not noticed that this was mentioned in the literature, it might be possible that this is not an isolated incident, as we would also not detect it if we weren't so meticulous in conducting the research and checked if everything was in order, with the unfortunate fact that it was not. As we have conducted our research with BibTeX file, we have tried with the textual file as well, supported by Bibliometrix, but the results were again not correct, and not only that, they were to a degree different. It seems that the program has some combination of preprocessing, parsing, detection, identification, and calculation bugs.}.

In such a situation confidence in the results is seriously shaken and the analyses would be useless, and as every analysis depends on the unique identification of an author, we had to move over these -- an unfortunate fact as it is, but necessary nevertheless, at least when using Bibliometrix \cite{Aria2017,Aria2023}.

Aside from Bibliometrix \cite{Aria2017,Aria2023} there are other tools \cite{Donthu2021} with which one could conduct at least some part of the analysis if not everything, and as per our inspection VOSviewer \cite{Eck2006,PerianesRodriguez2016} is one of the more prominent ones and it seems trustworthy \cite{VianaLora2022}. Thus in order to analyze authors and their prospective metrics we have used this tool, as the tool calculates the data we need, even though it does not present it as Bibliometrix \cite{Aria2017,Aria2023}, however, the data is all we need, as analysis and visualization can be done by the researcher.

The first analysis was performed on the most relevant author as per the number of documents\footnote{The data was checked, just as was for the Bibliometrix, and this tool was working well, there was one minor mistake and a few potential ones (e.g. author changed or working on multiple institutions etc.), it seems that unique identification of a subject is a pressing issue, if however there would be a change in the results, it seems that change would be a minor one.} and is seen in Figure \ref{fig:MRA}. Here we see three groups, the first is made of two authors with 37 and 35 documents, then there is a substantial fall to 21, but still high, after which there is a group of 7 authors ranging in terms of the number of documents from 8 to 12, a group with authors close to one another that could be additionally divided into four groups, as per the number of documents. The data, and corresponding curve that could be drawn by entire dataset data points, approximately delineates Lotka's law, stating that as the number of documents increases, the number of authors that have published that amount decreases by a power inversely proportional (for $ n $ documents one has $ \frac{1}{n^{2}} $ authors, out of the authors that have one document published, e.g. if 10 authors have written 1 document then 1 author has written 3 documents: $ 10\times \frac{1}{3^2} \approx 1 $) \cite{Lotka1926}.

\definecolor{eggplant}{rgb}{0.38, 0.25, 0.32}
\begin{figure}[h]
	\centering
	\begin{tikzpicture}[font=\small]
		\begin{axis}[
			tickwidth = 5pt,
			ylabel near ticks,
			xlabel near ticks,
			xbar,
			bar width=12pt,
			xmax=42,
			xlabel={Number of Documents},
			ylabel={Authors},
			axis x line=bottom,
			axis y line=left,
			nodes near coords,
			enlarge y limits=0.1,
			width=.8\textwidth,
			height=.6\textwidth,
			symbolic y coords={
				{Sun, Jie},
				{Li, Hui},
				{Tsai, Chih-Fong},
				{Xu, Wei},
				{Hsu, Ming-Fu},
				{Ribeiro, Bernardete},
				{Wu, Chong},
				{Du Jardin, Philippe},
				{Chen, Ning},
				{Jones, Stewart}
			},
			ytick = data,
			]
			\addplot[fill=eggplant] coordinates {			
				(37,{Sun, Jie})
				(35,{Li, Hui})
				(21,{Tsai, Chih-Fong})
				(12,{Xu, Wei})
				(10,{Hsu, Ming-Fu})
				(9,{Ribeiro, Bernardete})
				(9,{Wu, Chong})
				(9,{Du Jardin, Philippe})
				(8,{Chen, Ning})
				(8,{Jones, Stewart})		
			};
		\end{axis}
	\end{tikzpicture}
	\caption{Most Relevant (in terms of number of documents published) Authors, exported in July 2023 -- as because of the problems with Bibliometrix \cite{Aria2017,Aria2023} we had to again export the same data from Web of Science into textual file (for the VOSviewer \cite{Eck2006,PerianesRodriguez2016}, since analysis was conducted in BibTeX, yet the VOSviewer does not support that file format). New export of the data should have no influence on the results, these and others, as the same corpus of documents is exported, therefore all the data values, aside from global indicators, should be the same -- instances of data update or completion are possible, yet one does not expect substantial volume of these.}\label{fig:MRA}
\end{figure}
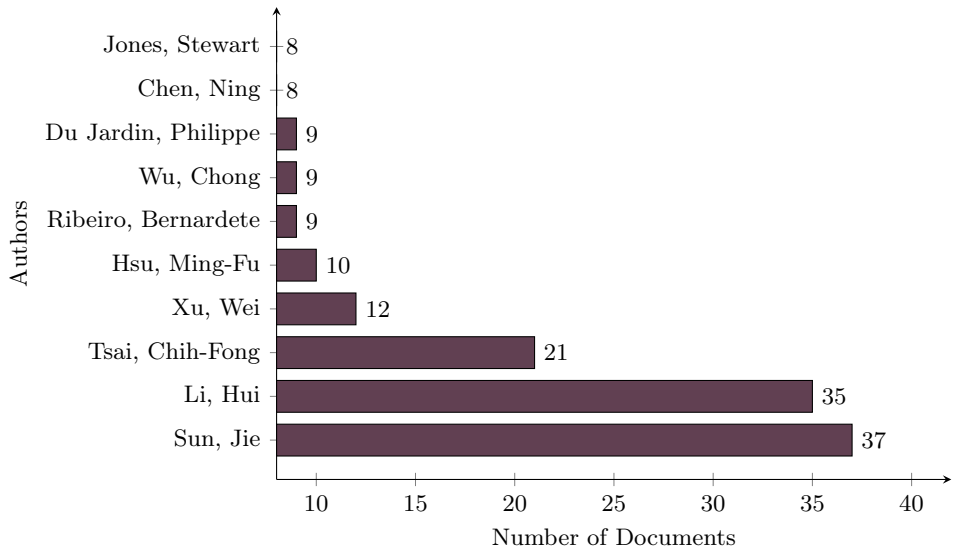

After a number of documents were published we are interested in citations, presented in Figure \ref{fig:MLCA}. Top authors have having substantial local citation count, indicating high relevance for the field, brought about by researching prominent themes, such as prediction, AI algorithms, etc., as bibliographic data reveals. By looking at the data for the top 10 authors, three groups are clearly observable -- with the first two having three authors, and the last with four authors. Such a situation is potentially an indication of collaboration, or research that is closely related. As for distribution, it is somewhat more linear than the one before, however, when we consider the entire dataset, a Pareto-like distribution is also in place here. When we compare with Figure \ref{fig:MRA} only three authors here are the same. A number of authors in Figure \ref{fig:MRA} does not have as long a publication history, however, a number of authors in Figure \ref{fig:MLCA} has with a few papers, and some are of an older nature, achieved high relevance -- a clear indication that a number of papers is not everything, but is a factor.

\definecolor{etonblue}{rgb}{0.59, 0.78, 0.64}
\begin{figure}[h]
	\centering
	\begin{tikzpicture}[font=\small]
		\begin{axis}[
			tickwidth = 5pt,
			ylabel near ticks,
			xlabel near ticks,
			xbar,
			bar width=12pt,
			xmax=1585,
			xlabel={Local Citations},
			ylabel={Authors},
			axis x line=bottom,
			axis y line=left,
			nodes near coords,
			x tick label style={/pgf/number format/1000 sep=},
			enlarge y limits=0.1,
			width=.8\textwidth,
			height=.6\textwidth,
			symbolic y coords={
				{Sun, Jie},
				{Li, Hui},
				{Tsai, Chih-Fong},
				{Pan, Wen-Tsao},
				{Zhang, GQP},
				{Ravi, V.},
				{Shin, KS},
				{Kumar, P. Ravi},
				{Varetto, F},
				{Hu, Yong}
			},
			ytick = data,
			]
			\addplot[fill=etonblue, style={/pgf/number format/1000 sep=}] coordinates {			
				(1579,{Sun, Jie})
				(1508,{Li, Hui})
				(1469,{Tsai, Chih-Fong})
				(1022,{Pan, Wen-Tsao})
				(1011,{Zhang, GQP})				
				(930,{Ravi, V.})				
				(683,{Shin, KS})				
				(619,{Kumar, P. Ravi})				
				(583,{Varetto, F})				
				(522,{Hu, Yong})	
			};
		\end{axis}
	\end{tikzpicture}
	\caption{Most Local (meaning within our own data) Cited Authors, ending in July 2023 -- this analysis has different ending date, as because of the problems with Bibliometrix \cite{Aria2017,Aria2023} we had to again export the same data from Web of Science into textual file (for the VOSviewer \cite{Eck2006,PerianesRodriguez2016}, since the analysis was conducted in BibTeX, but the VOSviewer does not support that file format). Compared to citations output by Bibliometrix \cite{Aria2017,Aria2023} there is a substantial difference, cca. from $ 3 $ to $ 5 $ times larger difference in citation count. Complete authors' names are not available for some of the authors, they are therefore given as they appear in the data itself. It should be noted that when compared to the number of published documents in Figure \ref{fig:MRA} only $ 3 $ authors are the same, indicating that there are other factors as well influencing citations.}\label{fig:MLCA}
\end{figure}
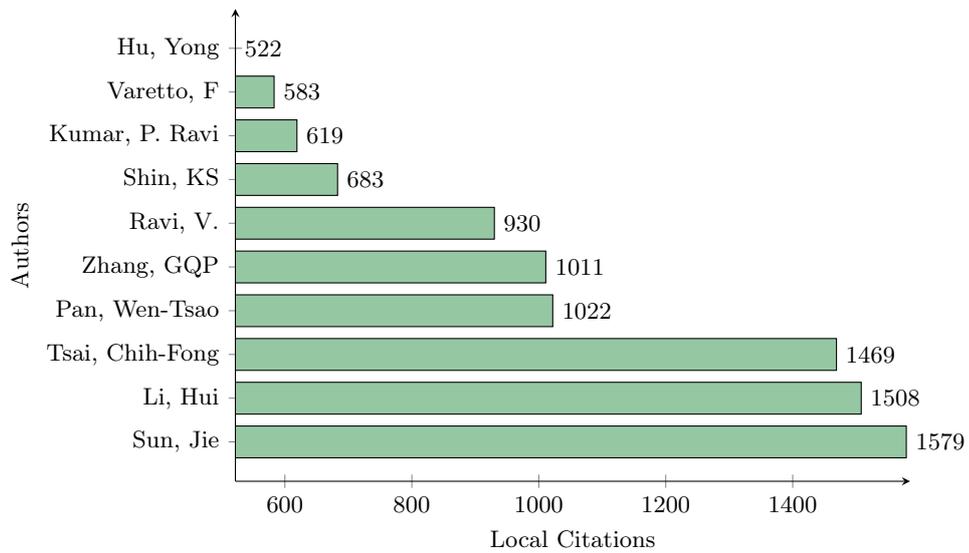

The last thing we wanted to achieve is to calculate standard and amortized $ h $-index for authors, so as to evaluate impact. Considering problems with the tools it was decided that we would try to uniquely identify authors via Google Scholar, and the author corpus for this task was the authors analyzed for the most number of documents published in Figure \ref{fig:MRA} and most locally cited authors in Figure \ref{fig:MLCA} -- as those authors are logical candidates for the entire dataset, and while analysis of the entire dataset is not possible, these logical picks are most likely being correct or close to correct. 

Nevertheless, it was not possible to make a unique Google Scholar identification as all the authors either do not have an account or perhaps they have changed affiliation and it was difficult to be sure if that is the right author. An effort was also made with a well-known tool Publish or Perish \cite{Harzing2023}, but to no avail. As a last resort to make the research as complete and rigorous as possible, we have searched the authors directly in Web of Science and obtained index $ h $ in such a way -- one should take note that such $ h $-index was calculated from Web of Science classification\footnote{Some records were generated algorithmically, while others were author verified records}, and is a few months newer than our own data.

This was however the best possible solution and was worth taking since it does represent the correct state in science, it is just not directly comparable with our own analyses, it does give authors relation to each other, just in a general manner on a larger corpus of documents. Index $ h $ calculation for the aforementioned corpus of authors, from which the top $ 10 $ were taken, following set methodology, is presented in Figure \ref{fig:ALIm}.

\definecolor{jazzberryjam}{rgb}{0.65, 0.04, 0.37}
\definecolor{jonquil}{rgb}{0.98, 0.85, 0.37}
\begin{figure}[h]
	\centering
	\begin{tikzpicture}[font=\small]
		\begin{axis}[
			tickwidth = 5pt,
			ylabel near ticks,
			xlabel near ticks,
			xbar,
			bar width=6pt,
			xmax=49.9,
			xlabel={$ H $-index},
			ylabel={Authors},
			axis x line=bottom,
			axis y line=left,
			nodes near coords,
			enlarge y limits = 0.1,
			height=.7\linewidth,
			legend style={legend columns=1,legend cell align={left},draw=none},
			legend image code/.code={
				\draw [#1] (0cm,-0.1cm) rectangle (0.2cm,0.25cm);},
			symbolic y coords={
				{Zhang, GQP},
				{Tsai, Chih-Fong},
				{Ravi, V.},
				{Li, Hui},
				{Sun, Jie},
				{Wu, Chong},
				{Ribeiro, Bernardete},
				{Xu, Wei},
				{Jones, Stewart},
				{Shin, KS},	
			},
			ytick = data,
			]
			\addlegendentry{\scriptsize Standard}
			\addplot[fill=jazzberryjam] coordinates {
				(47,{Zhang, GQP})
				(31,{Ravi, V.})
				(13,{Shin, KS})
				(26,{Ribeiro, Bernardete})				
				(17,{Jones, Stewart})
				(27,{Wu, Chong})
				(30,{Li, Hui})				
				(28,{Sun, Jie})
				(34,{Tsai, Chih-Fong})
				(21,{Xu, Wei})
			};
			\addlegendentry{\scriptsize Amortized}
			\addplot[fill=jonquil, bar width=6pt] coordinates {			
				(16.02,{Zhang, GQP})
				(17.22,{Ravi, V.})
				(7.80,{Shin, KS})
				(18.57,{Ribeiro, Bernardete})				
				(12.75,{Jones, Stewart})
				(20.25,{Wu, Chong})
				(23.68,{Li, Hui})				
				(23.33,{Sun, Jie})
				(28.33,{Tsai, Chih-Fong})
				(21.00,{Xu, Wei})
			};
		\end{axis}
	\end{tikzpicture}
	\caption{Authors Impact (index $ h $, proposed by Hirsch in 2005 \cite{Hirsch2005} for measuring not only citations but output also, the definition of which is, "the number of papers with citation number $ \geq h $"), ending in July 2023. Amortized $ h $-index is presented so as to impede the problem of authors having different track records starting year -- high $ h $-index for a shorter publication time period is exceedingly more difficult to accumulate than for a longer period of time. Complete author names are not available for some of the authors, they are therefore given as they appear in the data itself. For a description of how to calculate amortized $ h $-index, one can consult Appendix \ref{secA2}.}\label{fig:ALIm}
\end{figure}
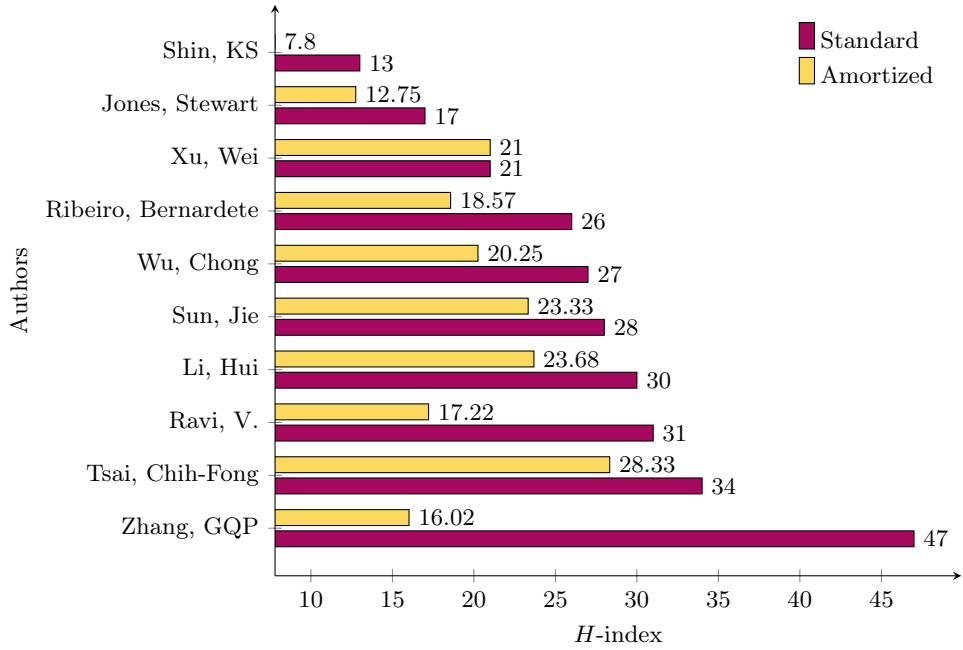

With the mentioned constraints, out of the authors in Figure \ref{fig:ALIm} 6 appear in the most cited analysis, while four are not, with those four still appearing in a number of documents analysis -- the results are close, it could however indicate that if one has large citation count, a chance to fare well in index $ h $ is greater than if only document count is high. As for the authors in Figure \ref{fig:ALIm} that appear in both citation and document analysis, there are three authors falling into that category (with the overlap of citation and document analysis being those same three authors), indicating the importance of balance between document count and citations, as extremes will induce low/lower $ h $-index.

It should be noted that these extremes can be unfairly penalized by $ h $-index, e.g. an author can have 10 papers, with 5 papers having 100 citations each. Such a situation would produce index $ h $ of 5, presenting that an author has 5 documents with 5 or more citations, which is true, but a gross understatement of an author's performance. Such an author could potentially be a candidate for some of the most prestigious scientific awards, with $ h $-index not giving such an indication.

When the top 10 authors as per $ h $-index are considered, distribution is for the most part linear, a logical consequence of index $ h $ cutting the extremes, therefore these authors are most likely well-balanced in terms of the number of documents and relevance. However, with a substantial number of low indexes appearing at the bottom of the entire dataset, distribution as a whole would be similar to Pareto. The future will most likely see these authors at the same place, for some time at least, as the field will most likely stably develop and grow in the present direction for a foreseeable period of time.

With amortized $ h $-index here so as to impede the problem of sources having different starting years the following is observed. Zhang's relevance has substantially diminished, as well as Ravi's who is known in front of Zhang -- with Tsai, Chih-Fong; Li, Hui; and Sun, Jie taking top three places, respectively. So when we take into consideration the idea behind amortized $ h $-index and make scaling of standard $ h $-index, there are changes that reveal a somewhat different ranking, namely in Zhang, GQP'; Ravi, V.'; and Xu, Wei' repositioning. Relevant information, as some authors have been overly emphasized by standard index $ h $, with the other authors being of higher importance than it originally seemed.

\subsection{Affiliation Data Analyses}
\label{sub:AfDA}

Then we moved on to affiliation analysis, however it again turned out that Bibliometrix \cite{Aria2017,Aria2023} has problems uniquely identifying institutions, a problem that seems the same as the one with the authors, as per our manual analysis the number of affiliations was inflated, therefore for this analysis we have used VOSviewer \cite{Eck2006,PerianesRodriguez2016}, that is the calculated data from the tool, the tool more capable in that regards.

The first analysis was on the most relevant affiliations by the total number of published articles and can be seen in Figure \ref{fig:MRAf}. Distribution here as well is one like Pareto, with two institutions being very dominant, while others approximately linearly decreasing in terms of number of publications. Zhejiang Normal University is as it seems far superior than any other institution, and has high specialization in the observed field, while National Central University follows. Institutions not making the ranking are involved in the research about AI, finance, and entrepreneurship, but to a substantially lesser degree and presumably with other fields of research taking more prominent/equal roles.

\definecolor{darkterracotta}{rgb}{0.8, 0.31, 0.36}
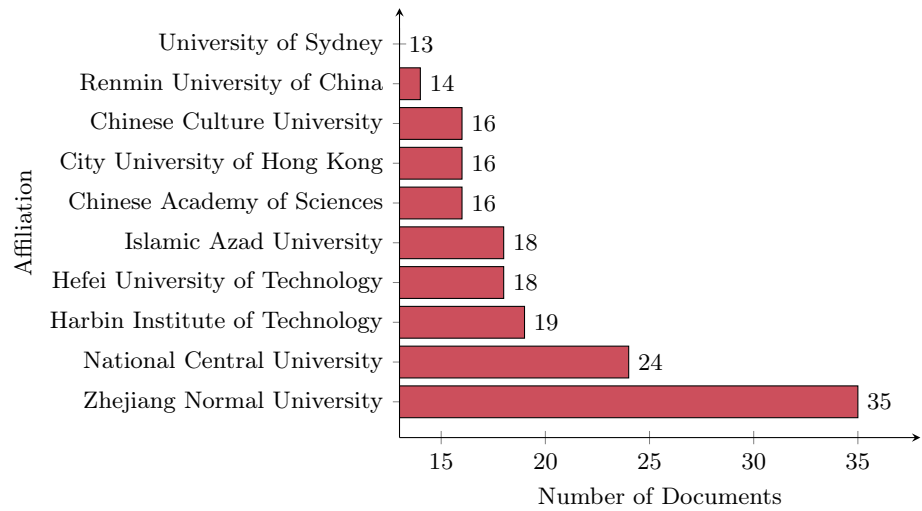
\begin{figure}[h]
	\centering
	\begin{tikzpicture}[font=\small]
		\begin{axis}[
			tickwidth = 5pt,
			ylabel near ticks,
			xlabel near ticks,
			xbar,
			bar width=12pt,
			xmax=38,
			xlabel={Number of Documents},
			ylabel={Affiliation},
			axis x line=bottom,
			axis y line=left,
			nodes near coords,
			enlarge y limits=0.1,
			symbolic y coords={
				{Zhejiang Normal University},
				{National Central University},
				{Harbin Institute of Technology},
				{Hefei University of Technology},
				{Islamic Azad University},
				{Chinese Academy of Sciences},
				{City University of Hong Kong},
				{Chinese Culture University},
				{Renmin University of China},
				{University of Sydney},
			},
			ytick = data,
			]
			\addplot[fill=darkterracotta] coordinates {		
				(35,{Zhejiang Normal University})
				(24,{National Central University})
				(19,{Harbin Institute of Technology})
				(18,{Hefei University of Technology})
				(18,{Islamic Azad University})
				(16,{Chinese Academy of Sciences})
				(16,{City University of Hong Kong})
				(16,{Chinese Culture University})
				(14,{Renmin University of China})
				(13,{University of Sydney})
			};
		\end{axis}
	\end{tikzpicture}
	\caption{Most Relevant (in terms of number of documents published) Affiliations, ending in July 2023 -- this analysis has a different ending date, as because of the problems with Bibliometrix \cite{Aria2017,Aria2023} we had to again export the same data from Web of Science into a textual file (for the VOSviewer \cite{Eck2006,PerianesRodriguez2016}, since the analysis was conducted in BibTeX, but the VOSviewer does not support that file format).}\label{fig:MRAf}
\end{figure}

When one knows document output, the next useful information is citation count and this is presented in Figure \ref{fig:MLCAFleg}. Institutions are more closer to one another, in terms of the number of citations than in the previous analysis, indicating that the number of documents is not the only criterion for achieving a high citation count. 

If we look at all institutions, aside from these 10, the distribution would also conform to a Pareto-like curve, just with a more competitive edge in this instance -- as if there is a battle for influence. It is also possible that such grouping is a consequence of collaboration or some other linking factor. 

When compared to document output in Figure \ref{fig:MRAf} there are only two overlapping affiliations, Zhejiang Normal University and National Central University, which is both telling and surprising. It would not be strange for this to happen for institutions at the back of the line, however when affiliation has high document output, not appearing in citation count raises questions. The top two affiliations in terms of document output are also highly relevant in citation count, while other affiliations are missing altogether as if there is a gap, either in quality or perhaps temporal delay of relevance caused by other factors, some of which could be affiliation renown, track record, current employees, alumni, prominence, country of origin, the main discipline of research, policy alignment, state affiliation, grants, industry collaboration, size, etc.

\definecolor{deeppeach}{rgb}{1.0, 0.8, 0.64}
\begin{figure}[h]
	\centering
	\begin{tikzpicture}[font=\small]
		\begin{axis}[
			tickwidth = 5pt,
			ylabel near ticks,
			xlabel near ticks,
			xbar,
			bar width=12pt,
			xmax=1580,
			xlabel={Local Citations},
			ylabel={Affiliation},
			axis x line=bottom,
			axis y line=left,
			nodes near coords,
			enlarge y limits=0.1,
			width=.8\textwidth,
			height=.6\textwidth,
			x tick label style={/pgf/number format/1000 sep=},
			symbolic y coords={
				{ZNU},
				{GSU},
				{NCU},
				{KAIST},
				{OIT},
				{IDRBT},
				{SU},
				{NYU},
				{DU},
				{EWU},
			},
			ytick = data,
			]
			\addplot[fill=deeppeach, style={/pgf/number format/1000 sep=}] coordinates {
				(1504,{ZNU})
				(1420,{GSU})
				(1248,{NCU})
				(1043,{KAIST})
				(1001,{OIT})
				(974,{IDRBT})
				(833,{SU})
				(810,{NYU})
				(765,{DU})
				(723,{EWU})
			};
		\end{axis}
	\end{tikzpicture}
	\caption{Most Local (meaning within our own data) Cited Affiliations, ending in July 2023 -- this analysis has a different ending date, as because of the problems with Bibliometrix \cite{Aria2017,Aria2023} we had to again export the same data from Web of Science into textual file (for the VOSviewer \cite{Eck2006,PerianesRodriguez2016}, since analysis was conducted in BibTeX, but the VOSviewer does not support that file format).}\label{fig:MLCAF}
	\footnotesize
	\begin{tabular}{rl}
		& \\
		ZNU		& Zhejiang Normal University \\
		GSU		& Georgia State University \\
		NCU		& National Central University \\
		KAIST	& Korea Advanced Institute of Science \& Technology \\
		OIT 	& Oriental Institute of Technology \\
		IDRBT	& Institute for Development and Research in Banking Technology \\
		SU	 	& Sogang University \\
		NYU 	& New York University \\
		DU		& Dongguk University \\
		EWU		& Ewha Womans University
	\end{tabular}\label{fig:MLCAFleg}
\end{figure}
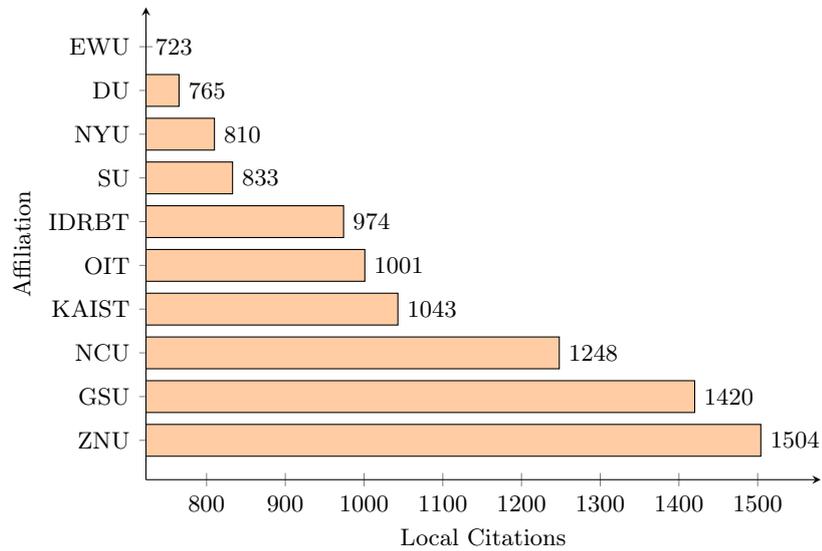

There is one more analysis that would be of interest to present the results from, and that is $ h $-index for affiliations. Neither Bibliometrix \cite{Aria2017,Aria2023} nor VOSviewer \cite{Eck2006,PerianesRodriguez2016} present this measure in the aforementioned context, and in order to achieve this we have used the tool Publish or Perish \cite{Harzing2023} and extracted the data via Crossref \cite{Crossref1999}, a well known and respectable knowledge metadata database that is heavily used by publishers and institutions -- in this way analysis will not be dependent on any particular scientific/professional database, and free from potential skewing.

Such $ h $-index will not be directly comparable with analyses conducted from our own data, yet it will give an insight into how local data institutions compare with global reach in mind, and that is useful and interesting. As a basis for the analysis, affiliations as the most relevant in terms of the number of documents published in Figure \ref{fig:MRAf} and citation count in Figure \ref{fig:MLCAF} were used. These institutions are in focus of the research and therefore a focus of index $ h $ calculation as well. Index $ h $ calculation for the aforementioned corpus of affiliations, from which the top $ 10 $ were taken, following set methodology, is presented in Figure \ref{fig:ALImp}.

\definecolor{denim}{rgb}{0.08, 0.38, 0.74}
\definecolor{deepskyblue}{rgb}{0.0, 0.75, 1.0}
\begin{figure}[h]
	\centering
	\begin{tikzpicture}[font=\small]
		\begin{axis}[
			tickwidth = 5pt,
			ylabel near ticks,
			xlabel near ticks,
			xbar,
			bar width=12pt,
			xmax=84,
			xlabel={$ H $-index},
			ylabel={Affiliations},
			axis x line=bottom,
			axis y line=left,
			nodes near coords,
			enlarge y limits = 0.1,
			width=.8\textwidth,
			height=.7\textwidth,
			symbolic y coords={
				NYU,
				GSU,
				CUHK,
				SU,
				RUC,
				US,
				ZNU,
				CCU,
				KAIST,
				OIT
			},
			ytick = data,
			]
			\addplot[fill=denim] coordinates {
				(80,NYU)
				(59,GSU)
				(51,CUHK)
				(51,SU)
				(50,RUC)
				(48,US)
				(45,ZNU)
				(41,CCU)
				(41,KAIST)
				(41,OIT)
			};
		\end{axis}
	\end{tikzpicture}
	\caption{Affiliations Impact (index $ h $, proposed by Hirsch in 2005 \cite{Hirsch2005} for measuring not only citations but output also, the definition of which is, "the number of papers with citation number $ \geq h $"), ending in July 2023 with data obtained from Crossref \cite{Crossref1999} and standard index $ h $ calculated by Publish or Perish \cite{Harzing2023}. As affiliation names are excessively long, they are abbreviated, with a legend being given below the figure. This analysis was performed for the entire publication period -- analysis for the last $ 10 $ years, from $ 2013 $ -- $ 2023$, can be seen in Figure \ref{fig:ALI2013mp}.}\label{fig:ALImp}
	\footnotesize
	\begin{tabular}{rl}
		& \\
		NYU 	& New York University \\
		GSU		& Georgia State University \\
		CUHK	& City University of Hong Kong \\
		SU	 	& Sogang University \\
		RUC		& Renmin University of China \\
		US		& University of Sydney \\
		ZNU		& Zhejiang Normal University \\	
		CCU		& Chinese Culture University \\
		KAIST	& Korea Advanced Institute of Science \& Technology \\	
		OIT 	& Oriental Institute of Technology \\
	\end{tabular}\label{fig:ALIleg}
\end{figure}
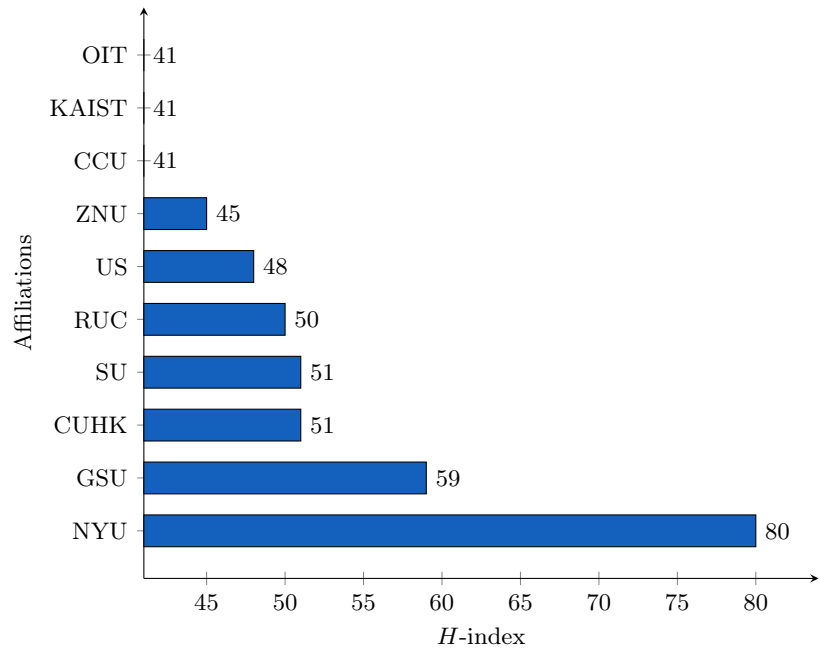

As it is difficult to compare this analysis with those before, we will forego such discussion, but what we can do is comment on the analysis itself. Thus we can articulate that NYU is in the lead, far and beyond the rest of affiliations. After NYU, GSU and CUHK are following, respectively. The rest of the affiliations flow in a close descending order.

When we observe the general picture we see a Pareto-like distribution, with NYU reigning supreme, followed by far behind GSU, and then the rest of affiliations following in a close manner to one another.

It is however not only about the entire period of time, but about the latest developments as well, and this can be seen in Figure \ref{fig:ALI2013mp}. NYU is still highly relevant and far in the lead, though this time ZNU has jumped from 7th position to the second, in line with previous analyses, indicating recent developments and ZNUs latest prominence. Afterward, we have five affiliations sequencing a group, perhaps an indication of interrelationship, but without deeper analysis, it is difficult to tell.

\definecolor{richmaroon}{rgb}{0.69, 0.19, 0.38}
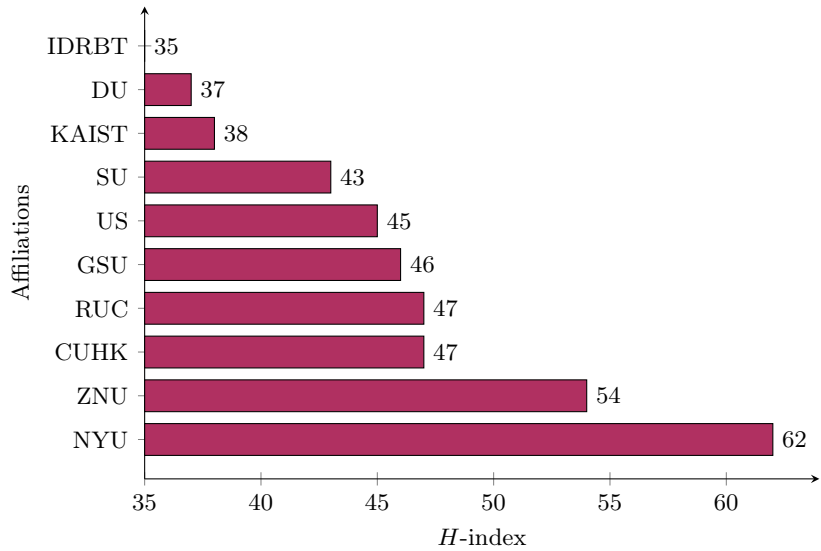
\begin{figure}[h]
	\centering
	\begin{tikzpicture}[font=\small]
		\begin{axis}[
			tickwidth = 5pt,
			ylabel near ticks,
			xlabel near ticks,
			xbar,
			bar width=12pt,
			xmax=64,
			xlabel={$ H $-index},
			ylabel={Affiliations},
			axis x line=bottom,
			axis y line=left,
			nodes near coords,
			enlarge y limits = 0.1,
			width=.8\textwidth,
			height=.6\textwidth,
			symbolic y coords={
				NYU,
				ZNU,
				CUHK,
				RUC,
				GSU,
				US,
				SU,
				KAIST,
				DU,
				IDRBT,
			},
			ytick = data,
			]
			\addplot[fill=richmaroon] coordinates {
				(62,NYU)
				(54,ZNU)
				(47,CUHK)
				(47,RUC)
				(46,GSU)
				(45,US)
				(43,SU)
				(38,KAIST)
				(37,DU)
				(35,IDRBT)
			};
		\end{axis}
	\end{tikzpicture}
	\caption{Affiliations Impact, $ 2013 $ -- $ 2023$, ending in July with data obtained from Crossref \cite{Crossref1999} and standard index $ h $ calculated by Publish or Perish \cite{Harzing2023}. As affiliation names are excessively long, they are abbreviated, with a legend being given below the figure.}\label{fig:ALI2013mp}
	\footnotesize
	\begin{tabular}{rl}
		& \\
		NYU 	& New York University \\
		ZNU		& Zhejiang Normal University \\
		CUHK	& City University of Hong Kong \\
		RUC		& Renmin University of China \\
		GSU		& Georgia State University \\
		US		& University of Sydney \\
		SU	 	& Sogang University \\	
		KAIST	& Korea Advanced Institute of Science \& Technology \\
		DU		& Dongguk University \\
		IDRBT	& Institute for Development and Research in Banking Technology \\
	\end{tabular}\label{fig:ALI2013leg}
\end{figure}

A last point of interest here, there are two new affiliations in this analysis, as compared to Figure \ref{fig:ALImp}, with Chinese Culture University and Oriental Institute of Technology missing, while Dongguk University and Institute for Development and Research in Banking Technology are making an appearance. An indication of current and potentially future developments as well, in terms of not only citations but the spread and broader relevance of publications. This situation could be a consequence of the number of document counts and strong technological presence in today's world, notwithstanding potential other factors as well, yet without the ability to directly compare to local research dataset and with Bibliometrics constraints it would not be prudent to overly project.

\subsection{Countries Data Analyses}
\label{sub:DoDA}

Sources, authors and affiliations are linked, and if we generalize the picture, it is about countries that we speak about -- as revealed by authors, and affiliations also. Such analysis is first presented in Table \ref{tab:CSP}, with countries scientific production. 

The most prominent countries are China, the United States of America, and India, followed closely by the United Kingdom. China is by far the most productive (not surprising considering its population size and number of scientists \cite{Ikarashi2023}), followed by the United States of America which is also well very high producing country. Afterward, the relative measure as per baseline is slowly going down, without such a high difference. These results indicate the high focus and interest of China and the United States of America in AI, entrepreneurship and finance, and if further innovation comes, it will most likely predominately be from these countries. Most of the countries in the list in Table \ref{tab:CSP} represents those wealthy of the world with a strong economy and high focus on innovation, and thus such results are not surprising, with the future most likely bringing similar trend. \cite{Ikarashi2023}

\begin{table}[h]	
	\caption{Countries Scientific Production}\label{tab:CSP}%
	\begin{tabular}{@{}lll}
		\toprule
		No. of Documents & Country & Relative\textsuperscript{1} to Baseline (---)  \\
		\midrule
		2063	& China	& --- \\
		717		& United States of America	& 35\\
		279		& India	& 14\\
		266		& United Kingdom	& 13\\
		239		& South Korea	& 12\\
		234		& Spain	& 11\\
		163		& France	& 8\\
		159		& Italy	& 8\\
		138		& Iran	& 7\\
		111		& Turkey	& 5\\
		\botrule
	\end{tabular}
	\footnotetext[]{Sorted according to number of documents}
	\footnotetext[1]{\%, rounded}	
\end{table}

In Figure \ref{fig:CAC} we can see corresponding authors and collaborations. Before interpretation, one would need to know what it means to be a corresponding author. Unfortunately, this definition is not so clear cut, and it can have various meanings, from simply being the one in charge of communication, or having a substantial research impact, to having the greatest research impact and being a problem bearer. Depending on the definition, so does the interpretation change. In our instance, we will adopt the meaning of being a problem bearer, the one who has scientifically contributed the most, as out of all options this meaning is most heavily leaned towards -- it is a general statement, supposition most likely of being correct, yet is a constraint and should be viewed as one as deviation from such definition could be extremely pronounced in various fields and countries also.

And so the most contributing nations it would seem are China, the USA, and Korea, with China having a strong lead and the USA also being substantially more prominent than other nations. Other nations are more closely related in terms of a number of documents, and respective contributions to the scientific field of focus. It is possible that the state of China and the USA, but Korea as well in a sense, is at least partially produced by cooperation between these countries. Here as well high percentage of the Western world and wealthy nations can be observed, as everything has a price, and those that have the right conditions are those that are producing, although there are other countries, like India for example.

Subsequently, the collaboration aspect tells a bit different story. Out of the three most published, that is China, the USA and Korea, the USA is in the lead in terms of multiple country publications with $ 24.3\% $, indicating the environment most geared toward collaboration with the outside world. This however is not the top, as the UK with $ 59.2\% $ outmatches everyone by far, possibly a consequence of the commonwealth and highly multicultural society -- while France and Italy are following strongly, and even beating the USA. There are other countries as well with rather high collaboration numbers so to speak, nevertheless with substantially fewer documents, and therefore perhaps not that significant.

\definecolor{antiquebrass}{rgb}{0.8, 0.58, 0.46}
\definecolor{antiquefuchsia}{rgb}{0.57, 0.36, 0.51}
\definecolor{blue(pigment)}{rgb}{0.2, 0.2, 0.6}
\begin{figure}[h]
	\centering
	\begin{tikzpicture}[font=\small]
		\begin{axis}[
			tickwidth = 5pt,
			ylabel near ticks,
			xlabel near ticks,
			xbar,
			bar width=6pt,
			xmax=550,
			xlabel={Number of Documents},
			ylabel={Countries},
			axis x line=bottom,
			axis y line=left,
			nodes near coords,
			enlarge y limits = 0.04,
			height=1.2\linewidth,
			width=.8\linewidth,
			legend style={legend columns=1,legend cell align={left},draw=none},
			legend image code/.code={
				\draw [#1] (0cm,-0.1cm) rectangle (0.2cm,0.25cm);},
			symbolic y coords={
				{China},
				{USA},
				{Korea},
				{India},
				{Spain},
				{UK},
				{Italy},
				{France},
				{Iran},
				{Australia},
				{Turkey},
				{Germany},
				{Greece},
				{Poland},
				{CZ},
				{Romania},
				{Canada},
				{Finland},
				{Brazil},
				{Portugal},
			},
			ytick = data,
			]
			\addlegendentry{\scriptsize Single Country Publications}
			\addplot[fill=antiquebrass] coordinates {
				(538,China)
				(162,USA)
				(74,Korea)
				(58,India)
				(56,Spain)
				(29,UK)
				(34,Italy)
				(28,France)
				(38,Iran)
				(27,Australia)
				(28,Turkey)
				(23,Germany)
				(21,Greece)
				(22,Poland)
				(20,CZ)
				(24,Romania)
				(15,Canada)
				(14,Finland)
				(10,Brazil)
				(9,Portugal)
			};
			\addlegendentry{\scriptsize Multiple Country Publications}
			\addplot[fill=antiquefuchsia, bar width=6pt] coordinates {
				(123,China)
				(52,USA)
				(11,Korea)
				(17,India)
				(16,Spain)
				(42,UK)
				(15,Italy)
				(14,France)
				(3,Iran)
				(6,Australia)
				(5,Turkey)
				(6,Germany)
				(7,Greece)
				(6,Poland)
				(5,CZ)
				(1,Romania)
				(7,Canada)
				(8,Finland)
				(9,Brazil)
				(8,Portugal)
			};
			\node[font=\footnotesize, color=blue(pigment)] at (210, 2.3){$ 18.6\% $};
			\node[font=\footnotesize, color=blue(pigment)] at (129, 12.3){$ 24.3\% $};
			\node[font=\footnotesize, color=blue(pigment)] at (129, 22.3){$ 12.9\% $};
			\node[font=\footnotesize, color=blue(pigment)] at (129, 32.3){$ 22.7\% $};
			\node[font=\footnotesize, color=blue(pigment)] at (129, 42.3){$ 22.2\% $};
			\node[font=\footnotesize, color=blue(pigment)] at (129, 52.3){$ 59.2\% $};
			\node[font=\footnotesize, color=blue(pigment)] at (129, 62.3){$ 30.6\% $};
			\node[font=\footnotesize, color=blue(pigment)] at (129, 72.3){$ 33.3\% $};
			\node[font=\footnotesize, color=blue(pigment)] at (70, 82.3){$ 7.3\% $};
			\node[font=\footnotesize, color=blue(pigment)] at (70, 92.3){$ 18.2\% $};
			\node[font=\footnotesize, color=blue(pigment)] at (70, 102.3){$ 15.2\% $};
			\node[font=\footnotesize, color=blue(pigment)] at (70, 112.3){$ 20.7\% $};
			\node[font=\footnotesize, color=blue(pigment)] at (70, 122.3){$ 25.0\% $};
			\node[font=\footnotesize, color=blue(pigment)] at (70, 132.3){$ 21.4\% $};
			\node[font=\footnotesize, color=blue(pigment)] at (70, 142.3){$ 20.0\% $};
			\node[font=\footnotesize, color=blue(pigment)] at (70, 152.3){$ 4.0\% $};
			\node[font=\footnotesize, color=blue(pigment)] at (70, 162.3){$ 31.8\% $};
			\node[font=\footnotesize, color=blue(pigment)] at (70, 172.3){$ 36.4\% $};
			\node[font=\footnotesize, color=blue(pigment)] at (70, 182.3){$ 47.4\% $};
			\node[font=\footnotesize, color=blue(pigment)] at (70, 192.3){$ 47.1\% $};
		\end{axis}
	\end{tikzpicture}
	\caption{Corresponding Authors Countries per Number of Documents Published. Considering we are dealing here with collaboration, it does make methodological sense to expand the number of countries we are focusing on, therefore here the number observed for interpretation will be doubled, concretely $ 20 $ most published countries per corresponding author criteria. For a more compact presentation United States of America was abbreviated USA, the United Kingdom was abbreviated UK and the Czech Republic was abbreviated CZ. Figure percentage represents multiple country publications in the whole, e.g. for China calculation would be $ \frac{123}{538+123}*100=18.6 $. The list of countries is sorted according to the total number of documents for each country, that is the sum of single-country publications and multiple-country publications.}\label{fig:CAC}
\end{figure}
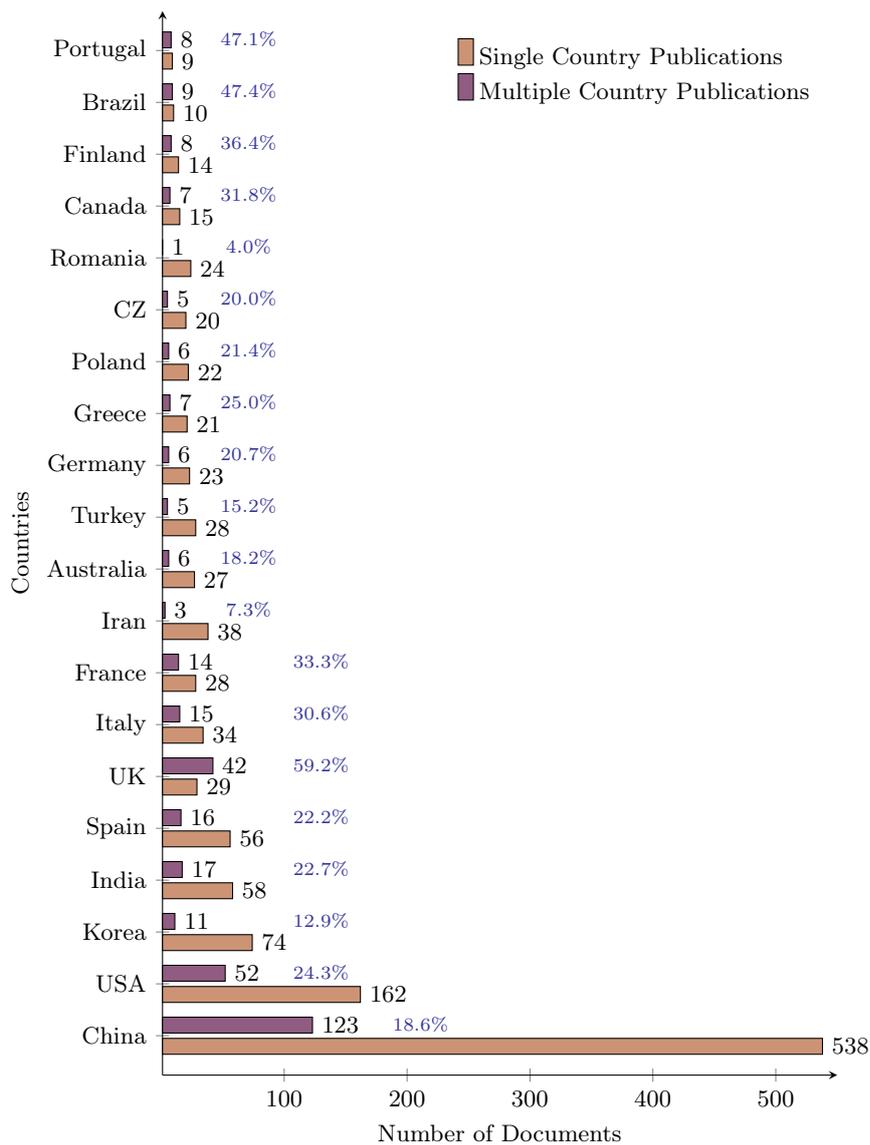

\begin{figure}[h]
	\centering
	\begin{tikzpicture}[]
		\begin{axis}[
			ylabel near ticks,
			xlabel near ticks,
			minor xtick={1990, 1992, ..., 2024},
			xtick={1991, 1993, ..., 2023},
			xmin=1990, xmax=2025,
			ymin=-.5, ymax=2100,
			xlabel={Publication Years},
			ylabel={Number of Documents},
			grid=major,
			x tick label style={/pgf/number format/1000 sep=},
			width=1\textwidth,
			xticklabel style={rotate=90},
			axis lines=left,
			legend style={at={(0.5,-0.17)},
				anchor=north,legend columns=2,legend cell align={left},draw=none},
			]
			\addlegendentry{\scriptsize United States of America}
			\addplot[thick, color=red] table[x=Year, y=Articles, col sep=comma]{podaci/Countries_Production_Over_Time/USA.csv};\label{cpot:USA}
			\addlegendentry{\scriptsize United Kingdom}
			\addplot[thick, color=green] table[x=Year, y=Articles, col sep=comma]{podaci/Countries_Production_Over_Time/UK.csv};\label{cpot:UK}
			\addlegendentry{\scriptsize Turkey}
			\addplot[thick, color=blue] table[x=Year, y=Articles, col sep=comma]{podaci/Countries_Production_Over_Time/TUR.csv};\label{cpot:TUR}
			\addlegendentry{\scriptsize Spain}
			\addplot[thick, color=cyan] table[x=Year, y=Articles, col sep=comma]{podaci/Countries_Production_Over_Time/SPA.csv};\label{cpot:SPA}
			\addlegendentry{\scriptsize Korea}
			\addplot[thick, color=magenta] table[x=Year, y=Articles, col sep=comma]{podaci/Countries_Production_Over_Time/KOR.csv};\label{cpot:KOR}
			\addlegendentry{\scriptsize Italy}
			\addplot[thick, color=yellow-green] table[x=Year, y=Articles, col sep=comma]{podaci/Countries_Production_Over_Time/IT.csv};\label{cpot:IT}
			\addlegendentry{\scriptsize Iran}
			\addplot[thick, color=gray] table[x=Year, y=Articles, col sep=comma]{podaci/Countries_Production_Over_Time/IRA.csv};\label{cpot:IRA}
			\addlegendentry{\scriptsize India}
			\addplot[thick, color=orange] table[x=Year, y=Articles, col sep=comma]{podaci/Countries_Production_Over_Time/IND.csv};\label{cpot:IND}
			\addlegendentry{\scriptsize France}
			\addplot[thick, color=purple] table[x=Year, y=Articles, col sep=comma]{podaci/Countries_Production_Over_Time/FR.csv};\label{cpot:FR}
			\addlegendentry{\scriptsize China}
			\addplot[thick, color=teal] table[x=Year, y=Articles, col sep=comma]{podaci/Countries_Production_Over_Time/CHI.csv};\label{cpot:CHI}
		\end{axis}
	\end{tikzpicture}
	\caption{Countries Production over Time. It seems that there is a discrepancy in Bibliometrix on the number of documents, as compared to analysis in Table \ref{tab:CSP} -- upon inspection, there are documents that do not have publication date specified, a result of Web of Science data export and incomplete data, these are naturally not in the analysis.}\label{fig:CPOT}
\end{figure}
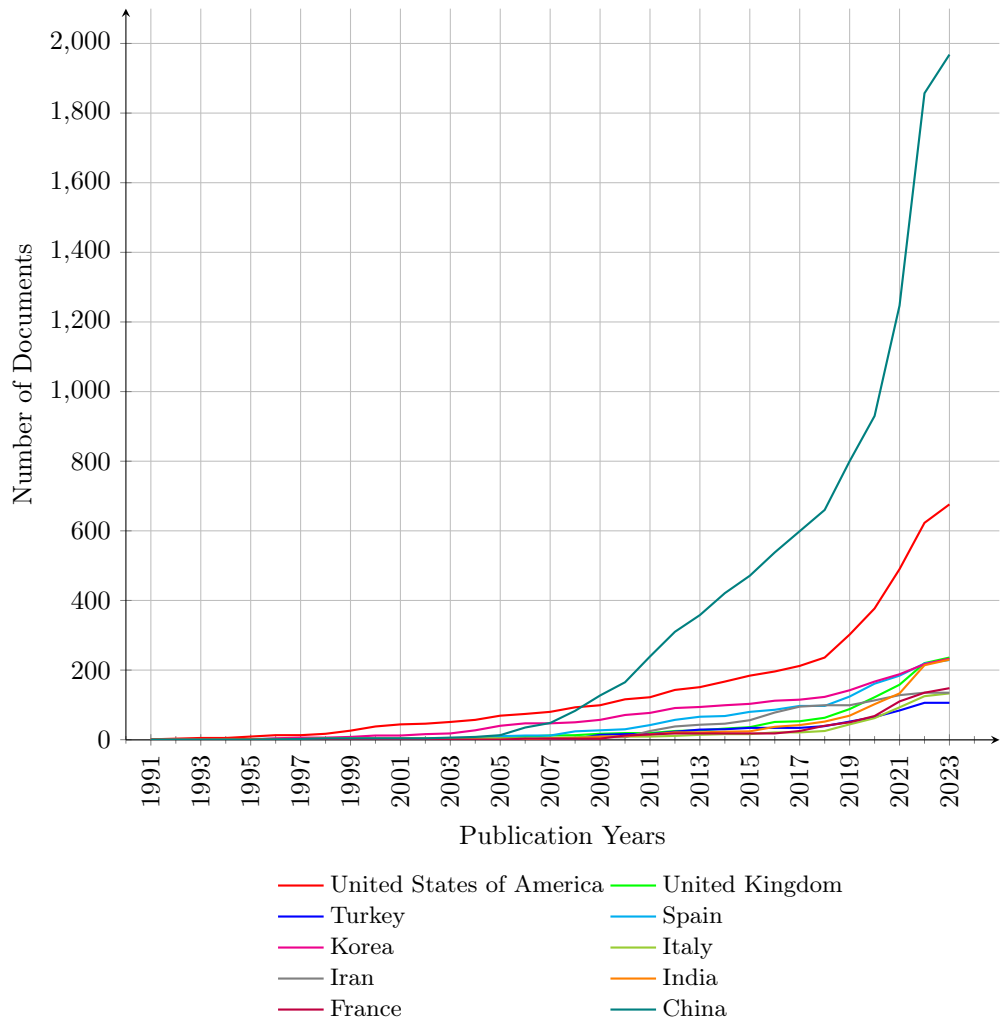

Countries' production over time is presented in Figure \ref{fig:CPOT}, an important analysis so as to ascertain the trend, and potential future impact. There are two curves that obviously stand out, China and the United States of America, all the others are substantially below and could in a way be grouped together, in spite of time and count differences.

The USA had an important role for quite a number of years before China started its ascent around the year 2005. By 2008 these countries were equal, and in 2009 China made an overtaking move. Since then, in absolute terms, China has been publishing substantially more in the number of documents. It is however fascinating to compare these results in terms of other factors, e.g. population size, number of scientists \cite{Ikarashi2023}, time available for research \cite{Ikarashi2023}, etc. China has a population of cca. $ 1,413,659,000 $ \cite{Lewis2023}, while the USA has a population of cca. $ 339,277,000 $ \cite{Gopnik2023}, which makes China cca. $ 4.17 $ times larger than the USA. Therefore as China's document count is $ 2063 $ (as stated in Table \ref{tab:CSP}), one would expect from the USA cca. $ 495 $ documents, as opposed to $ 717 $. Which makes the USA extremely productive, cca. $ 45\% $ above expectations and China -- the same calculation can be performed for other countries as well. Thus it seems that there is more here than meets the eye and that special care needs to be taken into account when interpreting bibliometrics results.

All the countries have had a positive trend, during the last few years at least, with the USA and China being on the extreme side of things (in line with the enormous funding of science \cite{Ikarashi2023}), and as it seems highly relevant. As confirmed by other analyses, an explosion driven by the huge leaps in AI, and one that will most likely continue for the foreseeable future, with other countries being in a supportive role.

It is of course a part of the picture of how cited all of these documents are, and this we answer in Figure \ref{fig:MCC}. Not surprisingly, China, the USA and Korea are leading the group of most cited countries (in line with the enormous funding of science \cite{Ikarashi2023}), with a Pareto-like distribution being observed. Here well can be ascertained that the USA is substantially more relevant than China, not in terms of absolute value, but when e.g. population size would be a factor, with Korea beating both of them, as Korea (South Korea) has a population size of cca. $ 51,268,000 $ \cite{Yu2023} -- which makes Korea extremely successful and a powerhouse in terms of relevance given by citation sum (in South Korea "research spending in the university sector has quadrupled" over the past several decades \cite{Ikarashi2023}).

This rule of being bigger when smaller might be deceptive, one might say that population might be an aggravating factor, nevertheless in the case of India for example it seems that population size is not a factor, as that country has a population size of cca. $ 1,370,695,000 $ \cite{Wolpert2023}, a number quite close to China's, yet without results of China -- indicating that population size is not the constraint, but that there are other factors at play \cite{Ikarashi2023}. The same can be seen with Spain, population size cca. $ 47,900,000 $ \cite{Viguera2023}, but far less relevant than Korea, with approximately the same population size as Korea. Although it is not excluded that e.g. population size can be a factor, and it probably is one of the factors, it seems that this factor can be accommodated \cite{Ikarashi2023}.

On the other hand, the absolute number shows the overall strength, regardless of different issues, and is something that must be taken into account and can't be disregarded. It not only establishes trajectory, but makes strong fixed points or points, for citation, relevance, collaboration, etc., and this strength is difficult to ignore, both globally and locally.

\definecolor{camouflagegreen}{rgb}{0.47, 0.53, 0.42}
\begin{figure}[h]
	\centering
	\begin{tikzpicture}[font=\small]
		\begin{axis}[
			ylabel near ticks,
			xlabel near ticks,
			xbar,
			bar width=12pt,
			xmax=12950,
			xlabel={Number of Citations},
			ylabel={Countries},
			axis x line=bottom,
			axis y line=left,
			nodes near coords,
			enlarge y limits = 0.1,
			width=.8\textwidth,
			height=.6\textwidth,
			symbolic y coords={
				China,
				USA,
				Korea,
				India,
				Spain,
				UK,
				France,
				Italy,
				Germany,
				Turkey
			},
			ytick = data,
			]
			\addplot[fill=camouflagegreen, 
			style={/pgf/number format/1000 sep=}
			] coordinates {
				(12584,China)
				(6791,USA)
				(4262,Korea)
				(2270,India)
				(2205,Spain)
				(1654,UK)
				(1340,France)
				(1252,Italy)
				(1132,Germany)
				(1059,Turkey)
			};
		\end{axis}
	\end{tikzpicture}
	\caption{Most Cited Countries, total citation sum. So as to make the figure compact, the United Kingdom was abbreviated as the UK and the United States of America was abbreviated as USA.}\label{fig:MCC}
\end{figure}
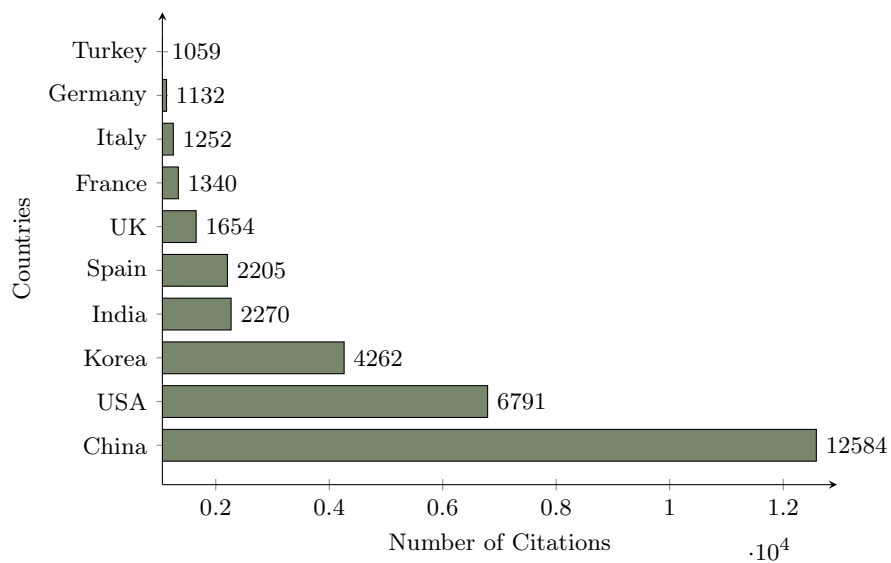

To make a full circle and try to ascertain the issue as rigorously as large research allows, we have contextualized citations and presented an expected value, as seen in Figure \ref{fig:MCCA}, with average article citation following Pareto-like distribution. The results are both interesting and in a way surprising. Out of the countries in Figure \ref{fig:MCC}, representing total citation sum, China, India, UK, and Italy are not represented in the average citation per article analysis -- while Malaysia, Norway, Qatar, and Cyprus have made an entry.

This indicates that the missing countries have produced relevance by a large number of articles that are not as relevant, which is then revealed when the average is taken into consideration, with the reverse being indicated for countries that are new. So the top three countries then are, Korea, Malaysia, and Germany, with both Korea and Malaysia being contenders for which one would not expect such high relevance, yet here they are. The same could be said for some other countries as well, with the 9th place of the USA representing quite a surprise, while the absence of China in the top 10 is an even bigger astonishment.

Pareto distribution, or Weibull perhaps, is found to be describing the phenomena, and so many others also. This is significant in at least two points. The first is that it can be a detection tool for an analysis, a way to right at the beginning try to determine whether what one has produced is a correct/complete picture of a situation. While the second point is of a more informative nature, telling us that it is a typical case having a few best ones, and then come all the others that are perhaps also solid, but are nowhere near the top.

\definecolor{cambridgeblue}{rgb}{0.64, 0.76, 0.68}
\begin{figure}[h]
	\centering
	\begin{tikzpicture}[font=\small]
		\begin{axis}[
			tickwidth = 5pt,
			ylabel near ticks,
			xlabel near ticks,
			xbar,
			bar width=12pt,
			xmax=54,
			xlabel={Average Article Citation},
			ylabel={Countries},
			axis x line=bottom,
			axis y line=left,
			nodes near coords,
			width=.8\textwidth,
			height=.6\textwidth,
			enlarge y limits = 0.1,
			symbolic y coords={
				Korea,
				Malaysia,
				Germany,
				Norway,
				Qatar,
				Cyprus,
				Turkey,
				France,
				USA,
				Spain
			},
			ytick = data,
			]
			\addplot[fill=cambridgeblue] coordinates {
				(50.10,Korea)
				(42.70,Malaysia)
				(39.00,Germany)
				(35.40,Norway)
				(33.70,Qatar)
				(33.40,Cyprus)
				(32.10,Turkey)
				(31.90,France)
				(31.70,USA)
				(30.60,Spain)
			};
		\end{axis}
	\end{tikzpicture}
	\caption{Most Cited Countries by Average Article Citation. So as to make the figure compact, the United States of America was abbreviated as USA.}\label{fig:MCCA}
\end{figure}
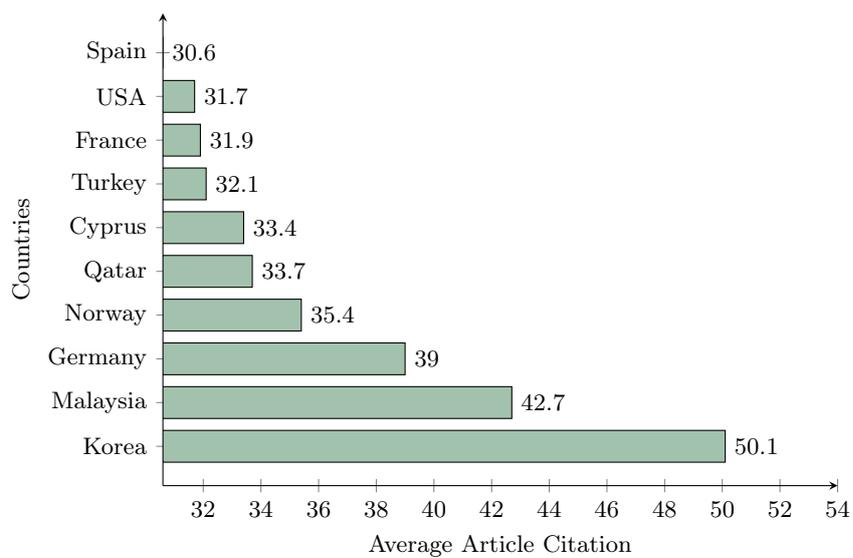

\subsection{Documents Data Analysis}
\label{sec:DTA}

As one ascertains the documents it can be observed that many publications are in operations research, a discipline well known and highly researched in the past. As we are nearing the present day, the research is shifting to computing, information and algorithms. Time will eventually tell, but it seems that computer science will deal with the issue more favorably and bring the solutions into the lives of a large number of people in an accessible manner.

Continuing from analyses for countries, we are coming to an analysis of the most locally cited documents, as seen in Figure \ref{fig:MLCD}. The highest citation count is 187, the three most cited documents are not those that are quite old, therefore indicating that relevance was not a result of old age so to speak. Notwithstanding that fact, aside from one document published in 2017, the youngest document is 15 years old, published in 2008, with the oldest documents going all the way to 1994. Indicating that generally speaking one needs to wait for a number of years, far more than the usual citation accumulation period of two to five years, before he can expect to shine in the top 10 of the subject area published in.

There is however that paper of 2017, an exception, yes, but worthy of notice nonetheless, as the exceptions are being looked at here, are they not. A much younger paper than the rest, but it entered into the top 10 most cited documents nevertheless. Published in Expert Systems with Applications, as already established, quit the influential journal in the observed field, and thus perhaps an expected outcome. The paper itself deals with machine learning and bankruptcy prediction, as later analyses will reveal themes of high interest, indicating relevance in the hereunder, as well as prominence of the topics in not the distant past and most likely in the near future at least. 

The citation distribution, at least for the presented data, is approximately linear, and documents are not that far apart in regard to citation count. Indicating competitiveness and that a group of close relevance documents is to be expected for the top cited documents of the field at hand -- a logical result, as a discipline, or a field, is comprised of a corpus of knowledge, not a few all-containing documents, and it seems that document citations are following that rule. 

As an added information to the analyses, there is also an expected value, so as to complete the picture of the situation. For the most part, the average reveals the same order of relevance, with two standouts. The first is that same document of 2017, with a value substantially higher than that of the first document -- additionally confirming the relevance of the paper and the themes, while at the same time indicating future developments. The second is a paper from 2008, and by closer inspection, one can observe neural networks, bankruptcy prediction and credit scoring, subjects quite similar to the first document in regard to the average value. Considering the aforementioned it is perhaps possible to predict future developments by finding such patterns and projecting them to later years, a separate research topic worthy of pursuing.

\definecolor{wildblueyonder}{rgb}{0.64, 0.68, 0.82}
\definecolor{yaleblue}{rgb}{0.06, 0.3, 0.57}
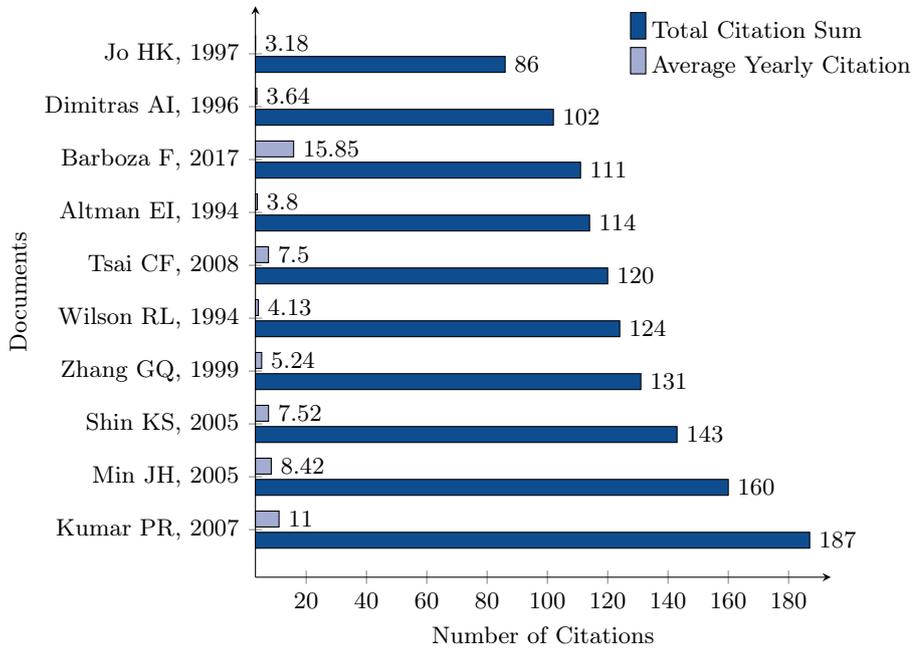
\begin{figure}[h] 
	\centering
	\begin{tikzpicture}[font=\small]
		\begin{axis}[
			tickwidth = 5pt,
			ylabel near ticks,
			xlabel near ticks,
			xbar,
			bar width=6pt,
			xmax=194,
			xlabel={Number of Citations},
			ylabel={Documents},
			axis x line=bottom,
			axis y line=left,
			nodes near coords,
			enlarge y limits = 0.1,
			width=.7\textwidth,
			height=.7\textwidth,
			legend style={at={(.9,1)},
				anchor=north, legend columns=1,legend cell align={left},draw=none},
			legend image code/.code={
				\draw [#1] (0cm,-0.1cm) rectangle (0.2cm,0.25cm);},
			symbolic y coords={
				{Kumar PR, 2007},
				{Min JH, 2005},
				{Shin KS, 2005},
				{Zhang GQ, 1999},
				{Wilson RL, 1994},
				{Tsai CF, 2008},
				{Altman EI, 1994},
				{Barboza F, 2017},
				{Dimitras AI, 1996},
				{Jo HK, 1997},
			},
			ytick = data,
			]
			\addlegendentry{\scriptsize Total Citation Sum}
			\addplot[fill=yaleblue] coordinates {
				(187,Kumar PR, 2007)
				(160,Min JH, 2005)
				(143,Shin KS, 2005)
				(131,Zhang GQ, 1999)
				(124,Wilson RL, 1994)
				(120,Tsai CF, 2008)
				(114,Altman EI, 1994)
				(111,Barboza F, 2017)
				(102,Dimitras AI, 1996)
				(86,Jo HK, 1997)
			};
			\addlegendentry{\scriptsize Average Yearly Citation}
			\addplot[fill=wildblueyonder, bar width=6pt] coordinates {
				(11.0,Kumar PR, 2007)
				(8.42,Min JH, 2005)
				(7.52,Shin KS, 2005)
				(5.24,Zhang GQ, 1999)
				(4.13,Wilson RL, 1994)
				(7.50,Tsai CF, 2008)
				(3.80,Altman EI, 1994)
				(15.85,Barboza F, 2017)
				(3.64,Dimitras AI, 1996)
				(3.18,Jo HK, 1997)
			};
		\end{axis}
	\end{tikzpicture}
	\caption{Most Local (within our own data) Cited Documents. In order to make the figure more compact, the publication source was transferred to the figure legend below -- with the link presented as well, so as to make the transition from the analysis to the publication as easy as possible. Entries are sorted according to the total citation sum. The average yearly citation was calculated in a conservative manner by adding one additional year to the time span of the calculation, e.g. for Kumar PR, 2007 equation is $ \frac{187}{2023 - 2007 + 1} = 11 $.}\label{fig:MLCD}
	\footnotesize
	\begin{tabular}{rl}
		& \\
		P. Ravi Kumar, 2007 	& \href{https://doi.org/10.1016/j.ejor.2006.08.043}{EUR J OPER RES} \\
		
		Jae H. Min, 2005	& \href{https://doi.org/10.1016/j.eswa.2004.12.008}{EXPERT SYST APPL} \\
		
		Kyung-Shik Shin, 2005	& \href{https://doi.org/10.1016/j.eswa.2004.08.009}{EXPERT SYST APPL} \\
		
		Guoqiang Zhang, 1999	& \href{https://doi.org/10.1016/S0377-2217(98)00051-4}{EUR J OPER RES} \\
		
		Rick L. Wilson, 1994	& \href{https://doi.org/10.1016/0167-9236(94)90024-8}{DECIS SUPPORT SYST} \\
		
		Chih-Fong Tsai, 2008	& \href{https://doi.org/10.1016/j.eswa.2007.05.019}{EXPERT SYST APPL} \\
		
		Edward I. Altman, 1994	& \href{https://doi.org/10.1016/0378-4266(94)90007-8}{J BANK FINANC} \\
		
		Flavio Barboza, 2017	& \href{https://doi.org/10.1016/j.eswa.2017.04.006}{EXPERT SYST APPL} \\
		
		A. I. Dimitras, 1996	& \href{https://doi.org/10.1016/0377-2217(95)00070-4}{EUR J OPER RES} \\
		
		Hongkyu Jo, 1997		& \href{https://doi.org/10.1016/S0957-4174(97)00011-0}{EXPERT SYST APPL} \\
	\end{tabular}\label{fig:MLCDleg}
\end{figure}

Alongside document relevance, it is also important to determine the same for the references, the information presented in Figure \ref{fig:MLCR}. The number of references is large, 62517 in total, indicating extensive foundational knowledge, which is not surprising considering we are dealing here with the intersection of three separate disciplines, AI, finance, and entrepreneurship. Coming from references to documents, in regard to sources, we can also observe a transition from let's say economics sources, to sources that are more of a computing nature, a transition most likely made possible via AI entering into finance and entrepreneurship.

The total citation sum is substantially higher here than in document analysis, as is the age of references, an expected outcome, and older and foundational knowledge if more relevant. The distribution is Pareto-like, an obvious transition from the linear distribution present in document analysis, indicating crystallization of the best through time. The expected value reveals that three references are more relevant than it seems, and these are also the youngest references of the entire group, which are two from 2005 and one from 2007 (a review document). A closer inspection reveals bankruptcy prediction in the context of computing and Artificial Intelligence -- aligning with the findings in document analysis in Figure \ref{fig:MLCD}, the future here is AI, and it could have been predicted for some time now.

\definecolor{wenge}{rgb}{0.39, 0.33, 0.32}
\definecolor{verdigris}{rgb}{0.26, 0.7, 0.68}
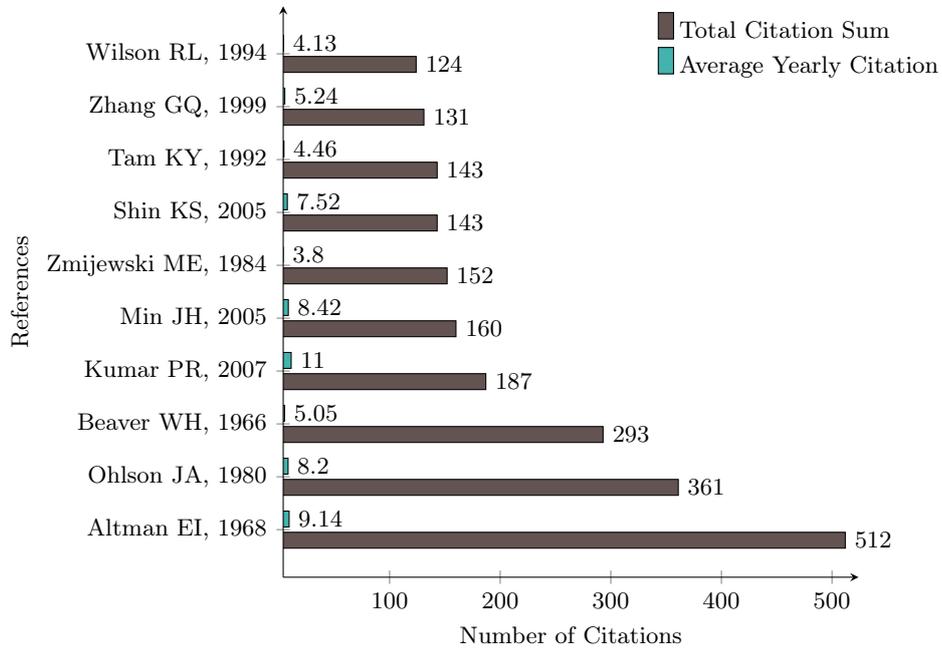
\begin{figure}[h] 
	\centering
	\begin{tikzpicture}[font=\small]
		\begin{axis}[
			tickwidth = 5pt,
			ylabel near ticks,
			xlabel near ticks,
			xbar,
			bar width=6pt,
			xmax=524,
			xlabel={Number of Citations},
			ylabel={References},
			axis x line=bottom,
			axis y line=left,
			nodes near coords,
			enlarge y limits = 0.1,
			width=.7\textwidth,
			height=.7\textwidth,
			legend style={at={(.9,1)},
				anchor=north, legend columns=1,legend cell align={left},draw=none},
			legend image code/.code={
				\draw [#1] (0cm,-0.1cm) rectangle (0.2cm,0.25cm);},
			symbolic y coords={
				{Altman EI, 1968},
				{Ohlson JA, 1980},
				{Beaver WH, 1966},
				{Kumar PR, 2007},
				{Min JH, 2005},
				{Zmijewski ME, 1984},
				{Shin KS, 2005},
				{Tam KY, 1992},
				{Zhang GQ, 1999},
				{Wilson RL, 1994},
			},
			ytick = data,
			]
			\addlegendentry{\scriptsize Total Citation Sum}
			\addplot[fill=wenge] coordinates {
				(512,Altman EI, 1968)
				(361,Ohlson JA, 1980)
				(293,Beaver WH, 1966)
				(187,Kumar PR, 2007)
				(160,Min JH, 2005)
				(152,Zmijewski ME, 1984)
				(143,Shin KS, 2005)
				(143,Tam KY, 1992)
				(131,Zhang GQ, 1999)
				(124,Wilson RL, 1994)
			};
			\addlegendentry{\scriptsize Average Yearly Citation}
			\addplot[fill=verdigris, bar width=6pt] coordinates {
				(9.14,Altman EI, 1968)
				(8.20,Ohlson JA, 1980)
				(5.05,Beaver WH, 1966)
				(11.00,Kumar PR, 2007)
				(8.42,Min JH, 2005)
				(3.80,Zmijewski ME, 1984)
				(7.52,Shin KS, 2005)
				(4.46,Tam KY, 1992)
				(5.24,Zhang GQ, 1999)
				(4.13,Wilson RL, 1994)
			};
		\end{axis}
	\end{tikzpicture}
	\caption{Most Local (within our own data) Cited References. In order to make the figure more compact, the reference source was transferred to the figure legend below -- with the link presented as well, so as to make the transition from the analysis to the reference as easy as possible. Entries are sorted according to total citation sum, with the analysis being performed on $ 62517 $ reference corpus. The average yearly citation was calculated in a conservative manner by adding one additional year to the time span of the calculation, e.g. for Altman EI, 1968 equation is $ \frac{512}{2023 - 1968 + 1} = 9.14 $.}\label{fig:MLCR}
	\footnotesize
	\begin{tabular}{rl}
		& \\
		Edward I. Altman, 1968	& \href{https://doi.org/10.1111/j.1540-6261.1968.tb00843.x}{J FINANC, V23, P589} \\
		
		James A. Ohlson, 1980	& \href{https://doi.org/10.2307/2490395}{J ACCOUNTING RES, V18, P109} \\
		
		William H. Beaver, 1966	& \href{https://doi.org/10.2307/2490171}{J ACCOUNTING RES, V4, P71} \\
		
		P. Ravi Kumar, 2007	& \href{https://doi.org/10.1016/j.ejor.2006.08.043}{EUR J OPER RES, V180, P1} \\
		
		Jae H. Min, 2005	& \href{https://doi.org/10.1016/j.eswa.2004.12.008}{EXPERT SYST APPL, V28, P603} \\
		
		Mark E. Zmijewski, 1984	& \href{https://doi.org/10.2307/2490859}{J ACCOUNTING RES, V22, P59} \\
		
		Kyung-Shik Shin, 2005	& \href{https://doi.org/10.1016/j.eswa.2004.08.009}{EXPERT SYST APPL, V28, P127} \\
		
		Kar Yan Tam, 1992	& \href{https://doi.org/10.1287/mnsc.38.7.926}{MANAGE SCI, V38, P926} \\
		
		Guoqiang Zhang, 1999	& \href{https://doi.org/10.1016/S0377-2217(98)00051-4}{EUR J OPER RES, V116, P16} \\
		
		Rick L. Wilson, 1994	& \href{https://doi.org/10.1016/0167-9236(94)90024-8}{DECIS SUPPORT SYST, V11, P545} \\
	\end{tabular}\label{fig:MLCRleg}
\end{figure}

\begin{figure}[h]
	\centering
	\begin{tikzpicture}[]
		\begin{axis}[
			ylabel near ticks,
			xlabel near ticks,
			minor xtick={1765, 1766, ..., 2023},
			xtick={1764, 1774, ..., 2024},
			xmin=1759, xmax=2029,
			ymin=-.5, ymax=6500,
			xlabel={Year of Publication},
			ylabel={Citations/No. of References},
			grid=major,
			x tick label style={/pgf/number format/1000 sep=},
			width=1\textwidth,
			xticklabel style={rotate=90},
			axis lines=left,
			legend style={at={(0.5,1.15)},
				anchor=north,legend columns=1,legend cell align={left},draw=none},
			]
			\addlegendentry{\scriptsize Citations for References in that Year}
			\addplot[thick, color=magenta] table[x=Year, y=Citations, col sep=comma]{podaci/RPYS.csv};\label{cpot:CfD}
			\addlegendentry{\scriptsize Number of References Published in that Year}
			\addplot[thick, color=cyan] table[x=Godine, y={Br. referenci}, col sep=comma]{podaci/RFYS_doc_year.csv};\label{cpot:CfDdok}
		\end{axis}
	\end{tikzpicture}
	\caption{Reference Spectroscopy -- time slicing references by grouping them per year of publication, and measuring the number of references per group-year and citation count per year of the group. The first reference is in $ 1764 $, and the last is in $ 2023 $. With this analysis, one can determine reference distribution and relevance there, together with the information about a particular time slice.}\label{fig:RSpec}
\end{figure}
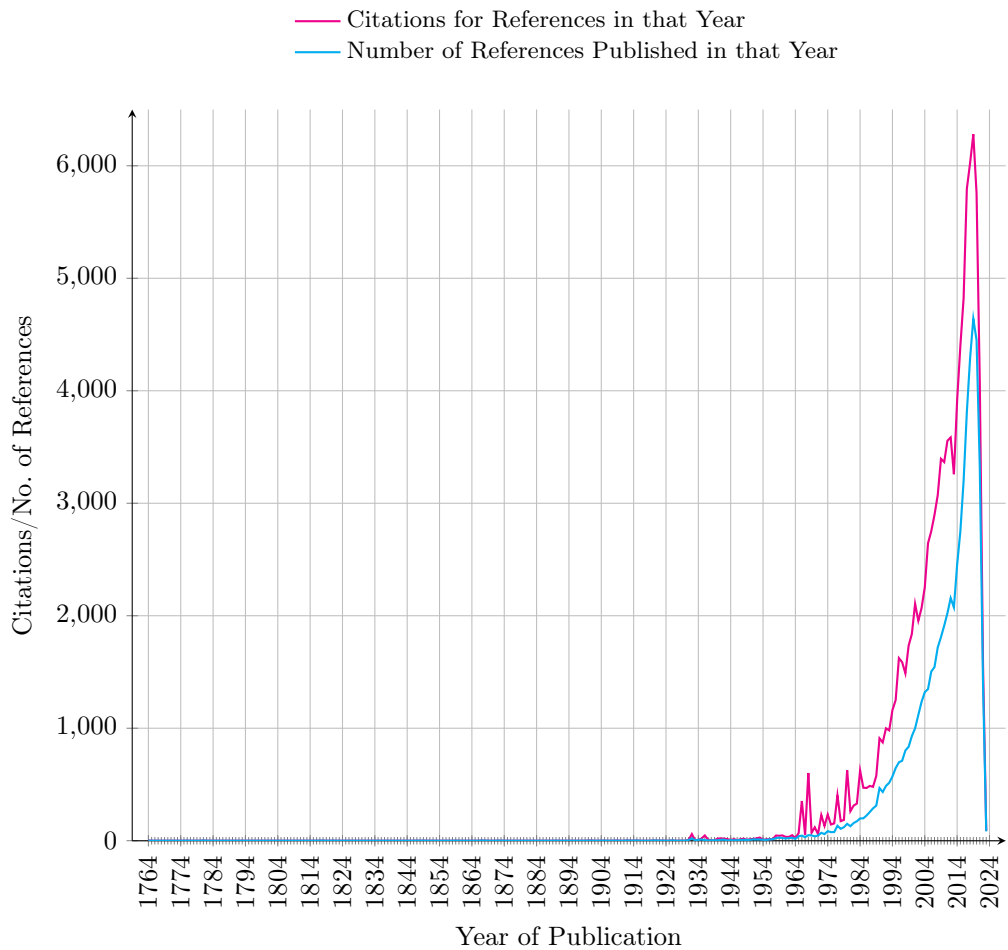

The last analysis in this context will be reference Spectroscopy, found in Figure \ref{fig:RSpec}. In 1764 the first reference is found, most likely referring to a document of Richard Price \cite{Price1764} communicating Thomas Bayes's doctrine of chances to John Canton, as Bayes never published this great work before he died, while Price to whom the manuscripts of Bayes were passed found it of high merit and well worthy of preservation with hopes of communicating the find to the Royal Society of which Bayes was a member, and thus this great work was published. \cite{Price1764,McGrayne2011,Bellhouse2004,Bayes1763} Considering that we are dealing here with economics, finance, machine learning and generally speaking with everything AI, it is of no surprise, especially with the importance of probability in AI and computing in general, that foundational knowledge would begin with Bayes and 1764.

After that first paper, there is a reference here and there, yet considering future developments, nothing of greater note happened until around the 1930s when both numbers of references and citations started a greater growth -- and it took another cca. 30 years, until around 1964, which was the beginning of the growth never seen before, firstly in an approximately linear fashion, after which there came an exponential explosion that continued to the present day. 

It took significantly more than half the period, from 1764 until 2023, for references to start higher and higher growth, indicating that aside from some older foundational knowledge, this discipline is vastly based on recent knowledge, which is most likely, at least in part, a consequence of timely advances in economics and computer science, with then consequently new discoveries transferring to the shoulders of giants (supported by comparison of Figures \ref{fig:asp} and \ref{fig:RSpec}). The citation curve follows the reference curve and presents a case for relevance, references are moderately to highly cited, and as time passes increasingly highly cited, indicating relevance and most likely practice of the authors working in the discipline.

\subsection{Words Data Analysis}
\label{sec:WDA}

One of the objectives of the research is to find prominent topics at the intersection of AI, finance, and entrepreneurship, and this analysis, analysis of words, will allow us to accomplish that goal, together with other related analyses of the research.

As seen in Table \ref{tab:MFW}, and out of those that are of a more concrete nature, an indication is that the most relevant topics in terms of computing are deep learning, neural network, support vector machine, blockchain, and decision tree. In regard to topics that would be classified into an economic group, we have bankruptcy prediction, credit scoring, firm performance, business failure, and fraud detection. 

The aforementioned topics indicate the focus of the published documents and authors, which are most likely linked to research projects with corresponding institutions and as such speak about a much wider field of thought, interest, and influence. Such a time slice is also useful as an indication of potential future developments as well as the basis for information for authors to ascertain hot topics, areas of potential major developments, and other enabling factors.

\begin{table}[h]	
	\caption{Most Frequent Terms}\label{tab:MFW}
	\begin{tabular}{@{}llll}
		\toprule
		Term\textsuperscript{1} & Frequency\textsuperscript{2} & Term & Frequency \\
		\midrule
		\textbf{machine learning} & 306	& \textbf{artificial intelligence} &	146\\
		bankruptcy prediction & 144	& \textbf{neural networks} &	93\\
		\textbf{deep learning} & 91 & \textbf{data mining} &	90\\
		credit scoring & 80	& bankruptcy &	77\\
		\textbf{neural network} & 66	& learning & 65\\
		prediction & 59	& financial distress &	58\\
		classification & 54	& credit risk &	52\\
		forecasting & 52 & financial distress prediction & 49\\
		\textbf{support vector machine} & 47	& crowdfunding & 45\\
		\textbf{artificial neural networks} & 44	& \textbf{logistic regression} &	43\\
		machine & 43 & demand forecasting &	41\\
		financial ratios & 41 &\textbf{big data} & 38\\
		\textbf{feature selection} & 36 & \textbf{genetic algorithm} & 35\\
		finance & 33 & networks	&	31\\
		analysis & 29 &	\textbf{artificial neural network} &	29\\
		\textbf{blockchain} & 29	& accounting &	27\\
		firm performance & 27 &	auditing &	26\\
		\textbf{natural language processing} & 25 & artificial &	24\\
		fintech & 24 & \textbf{support vector machines} & 24\\
		\textbf{decision trees} & 22	& financial	& 22\\
		model & 22 & peer-to-peer lending &	22\\
		business failure & 21 & default prediction & 21\\
		fraud detection & 21 & \textbf{random forest} &	21\\
		\textbf{regression} & 21	& business & 20\\
		\textbf{decision tree} & 20 & smes &	20\\
		\botrule
	\end{tabular}
	\footnotetext[]{Sorted according to frequency -- from left to right. Frequency is presented for $ 50 $ most occurring authors' own keywords. Terms that are emphasized with bold letters represent data points of interest as per the research objective, to extract algorithms, methods and techniques in AI found beneficial and applicable to entrepreneurial finance at the intersection of AI-entrepreneurship-finance, with here broadening the focus a bit for a more general overview.}
	\footnotetext[1]{Expression}
\end{table}

\begin{figure}[h]
	\centering
	\begin{tikzpicture}[]
		\begin{axis}[
			ylabel near ticks,
			xlabel near ticks,
			minor xtick={1990, 1992, ..., 2024},
			xtick={1991, 1993, ..., 2023},
			xmin=1990, xmax=2024,
			ymin=-.5, ymax=299,
			xlabel={Occurance Year},
			ylabel={Cumulative Term Occurrence},
			grid=major,
			x tick label style={/pgf/number format/1000 sep=},
			width=1\textwidth,
			xticklabel style={rotate=90},
			axis lines=left,
			legend style={at={(0.5,-0.17)},
				anchor=north,legend columns=2,legend cell align={left},draw=none},
			]
			\addlegendentry{\scriptsize Machine Learning}
			\addplot[thick, color=red] table[x=Year, y={MACHINE LEARNING}, col sep=comma]{podaci/Word_Dynamics.csv};\label{cpot:ML}
			\addlegendentry{\scriptsize Bankruptcy Prediction}
			\addplot[thick, color=green] table[x=Year, y={BANKRUPTCY PREDICTION}, col sep=comma]{podaci/Word_Dynamics.csv};\label{cpot:BP}
			\addlegendentry{\scriptsize Artificial Intelligence}
			\addplot[thick, color=blue] table[x=Year, y={ARTIFICIAL INTELLIGENCE}, col sep=comma]{podaci/Word_Dynamics.csv};\label{cpot:AI}
			\addlegendentry{\scriptsize Neural Networks}
			\addplot[thick, color=cyan] table[x=Year, y={NEURAL NETWORKS}, col sep=comma]{podaci/Word_Dynamics.csv};\label{cpot:NN}
			\addlegendentry{\scriptsize Data Mining}
			\addplot[thick, color=magenta] table[x=Year, y={DATA MINING}, col sep=comma]{podaci/Word_Dynamics.csv};\label{cpot:DM}
			\addlegendentry{\scriptsize Deep Learning}
			\addplot[thick, color=yellow-green] table[x=Year, y={DEEP LEARNING}, col sep=comma]{podaci/Word_Dynamics.csv};\label{cpot:DL}
			\addlegendentry{\scriptsize Credit Scoring}
			\addplot[thick, color=gray] table[x=Year, y={CREDIT SCORING}, col sep=comma]{podaci/Word_Dynamics.csv};\label{cpot:CS}
			\addlegendentry{\scriptsize Bankruptcy}
			\addplot[thick, color=orange] table[x=Year, y={BANKRUPTCY}, col sep=comma]{podaci/Word_Dynamics.csv};\label{cpot:BR}
			\addlegendentry{\scriptsize Neural Network}
			\addplot[thick, color=purple] table[x=Year, y={NEURAL NETWORK}, col sep=comma]{podaci/Word_Dynamics.csv};\label{cpot:NNT}
			\addlegendentry{\scriptsize Learning}
			\addplot[thick, color=teal] table[x=Year, y={LEARNING}, col sep=comma]{podaci/Word_Dynamics.csv};\label{cpot:LN}
		\end{axis}
	\end{tikzpicture}
	\caption{Terms Cumulative Frequency Over Time -- calculated over the authors' own keywords.}\label{fig:TFOT}
\end{figure}
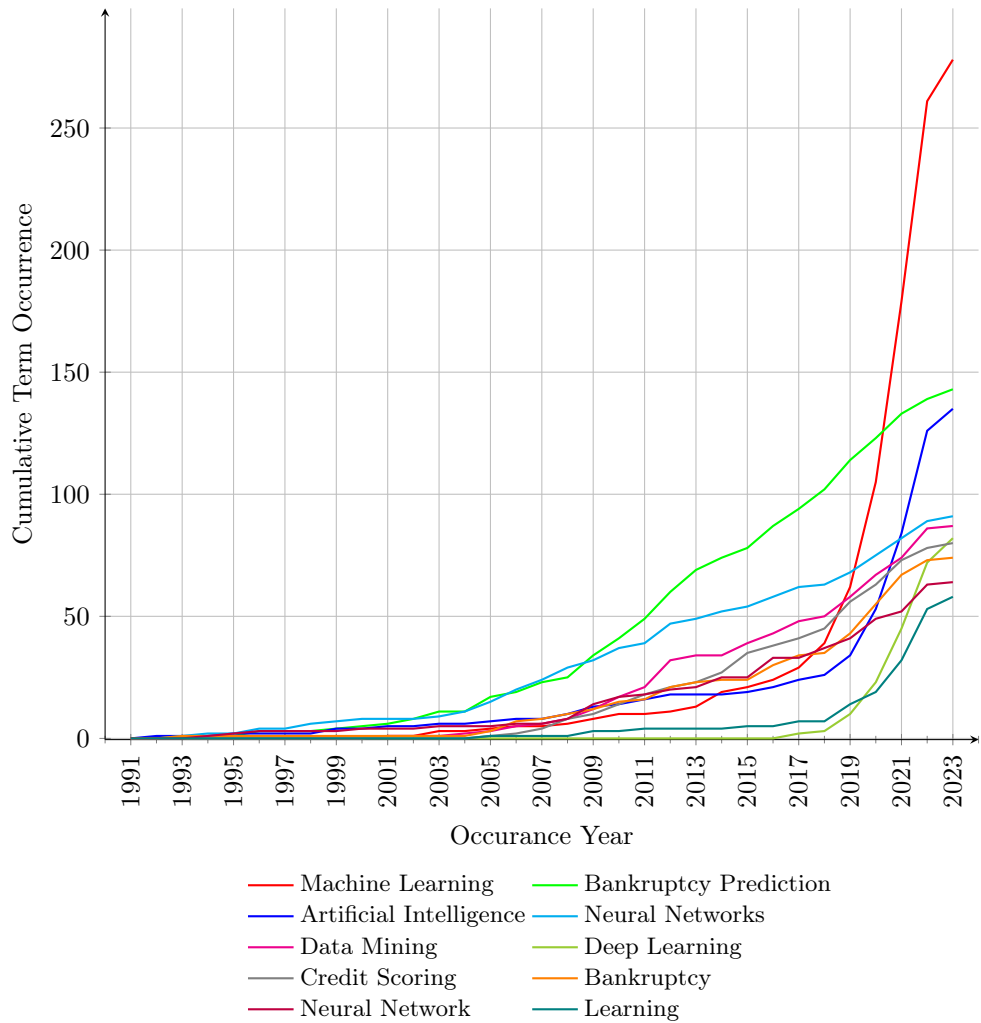

As useful as a point in time is, one would also like to ascertain the entire period of research with having more robust information and a foundation for reasoning about prediction as well, for top 10 terms and in a cumulative manner this can be seen in Figure \ref{fig:TFOT}. Computer science terms are in a substantial lead, indicating that even though the application part is in economics, it seems that documents have a heavy focus on computing and most likely improvement thereof.

Up until about 2003, listed terms are seeing slow but steady growth, indicating development and interest. From then onward there are two main periods, to around 2018, and from then until the present day. All terms are seeing high growth of occurrence, just at different stages in the timeline, most likely fueled by research contribution and the world we live in influence.

There are some obvious extremes, two of which coincide with bankruptcy prediction (appearing two times, these need to be taken together) and neural network (appearing two times, these need to be taken together). Both of these have at around the start of the century seen substantial interest, with the occurrence skyrocketing at around the year 2008 -- most likely governed by the well-known financial crisis of 2007-2009 together with the increase in real estate prices and accumulating debt of the decade \cite{Acharya2009}. Such research trajectory could have been an indication of things to come.

Another such event is enormous interest in AI and Machine Learning around the year 2018, as articulated before, a cause of which most likely is huge technological advances in that particular area of computing coupled with increased interest and availability of such research results in various forms. The third extreme point, linked to the second, happening at about the same time is the rise of deep learning (which needs to be merged with term learning) -- this rise is extremely high, especially considering that it began from a very low point, a logical consequence considering that deep learning is a building block of neural network and as such linked to its relevance.

Even though deep learning is a paradigm that is far from new and its beginning stretches all the way to 1962 \cite{Hopfield1982,Tappert2019}, it took a long time for the approach to mature, at least in this discipline, to find its right application and a way into society, referring of course to the recent advances of deep learning in natural language, communication, image generation, programming, etc.

As a consequence of the events spoken of, other areas are being researched and influenced by, such as data mining, and credit scoring, which themselves command substantial occurrence, just not as high as other terms, indicating that not everything is in aforementioned and likely supportive/collaborative role of these and as we saw from Table \ref{tab:MFW} other tools and areas of research as well, considering a need for data combing and scoring this was expected.

\begin{figure}[t!] 
	\centering
	\includegraphics[width=1.5\textwidth, angle=90]{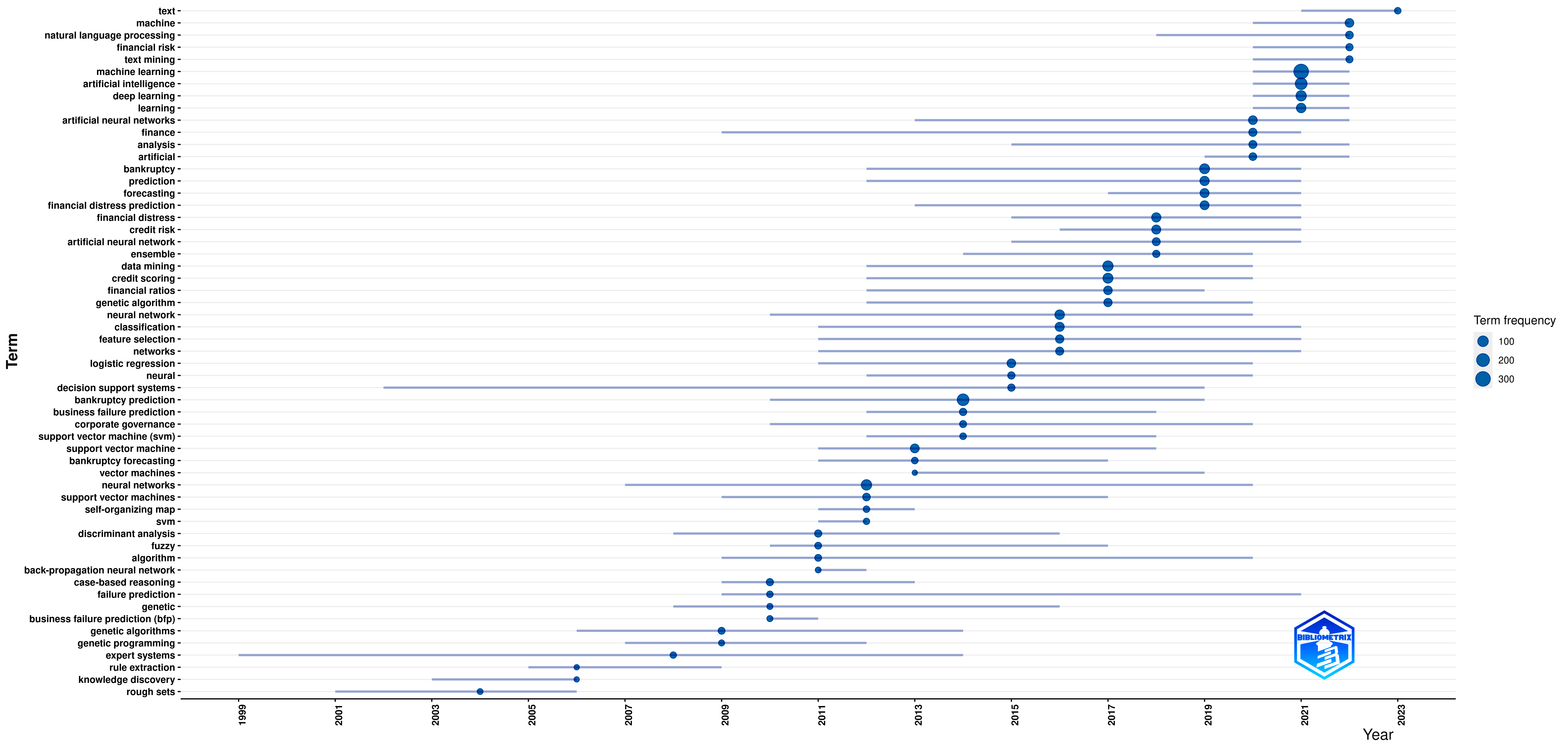}
	\caption{Trending Terms over Time. The list was generated using authors' own keywords for the entire data document lifespan, with the resulting data starting in $ 1999 $ through the term, expert systems. In order for a word to appear in the figure, occurrence needs to equal at least $ 5 $, so as to collect terms most relevant, with the maximum number of words per year being $ 4 $, as a high number of words results in nonrelevant terms or terms already appearing in the analysis -- furthermore, a high number of terms makes the analysis overly crowded with terms, and difficult to analyze and perceive what are the results indicating. The blue blob represents the frequency of occurrence for a particular term, while the vertical light blue line represents the period of occurrence (start, focus point, end; Q1, median, Q3, respectively).}\label{fig:TTOT}
\end{figure}

A continuation of the previous analysis, giving insight for a particular year is presented in Figure \ref{fig:TTOT}. On the x-axis one can observe a specific year and on the y-axis particular term can be observed, and so as not to repeat what has already been said we are directing the reader to the Figure \ref{fig:TTOT} and previous analyses.

Aside from already stated, perhaps there is a need for mentioning case-based reasoning, that deals with solving new problems while employing knowledge from problems previously solved, an indication of a field coming of age; ensemble that is most likely referring to ensemble learning, that uses multiple models/algorithms to solve a problem, an indication of a field substantially developed; self-organizing map, an artificial neural network used for reducing dimensionality of data, invented in 1982 \cite{Kohonen1981,Kohonen1982,Kohonen1990} and as such an indication of practical relevance in the observed discipline; discriminant analysis, a method used to find features that separate classes of objects, an indication of, together with a number of other terms, a statistical relevance in the discipline; and corporate governance, management term describing governance concerned with delivering long term success to a company, an indication of technology being recognized as one of the factors of importance to the subject.

Analysis in Figure \ref{fig:TTOT} also allows for longitudinal inspection, and as seen, considering that the starting year of the analyzed corpus is 1991, it took almost a decade for terms to achieve the desired frequency of relevance, a logical consequence of a beginning and a developing field trying to crystallize and establish its foundations.

In regard to median frequency, the first stretch of the timeline was marked by expert systems, a simpler form of artificial intelligence based on if-then rules, small beginnings considering the state of affairs in the year 2023, but a start, with the median high point in 2008. Then we come to a period where various disparate approaches have found their way into the fray: genetic algorithms, case-based reasoning, discriminant analysis, neural networks, support vector machine, etc., with the median high point being 2012 with neural network term -- this period is also seeing entering of economic aspect with a business failure prediction, a start of a deeper interconnection, found together with case-based reasoning, most likely linked with various cases and problems both in economics and computer science.

Further on in the timeline, we see a proliferation and expansion of the terms of the past, with decision support systems appearing here as well in the year 2015, and also corporate governance in 2014, expansion of the knowledge and rising to a governance level we see, important milestones scientifically and practically -- the high point here is in the year 2014 with bankruptcy prediction, a strong indication of interdisciplinarity, most likely collaboration and perhaps other elements as well.

From 2016 onward we have a strong presence of computing, economics and statistics mingling together, with ensemble methods appearing and elucidating now substantial maturation of the discipline and research contributions. This period ends in 2019 with bankruptcy and financial distress prediction, while it has a median high of 90 in 2017 with data mining, a value not that far behind of 2019 with bankruptcy and a value of 77. This period started with computing and statistical terms but ended with economics and financial distress. If previous analyses taken together with this one are of any indication, it seems that the financial waters of the future are wavy and muddy.

The last period of time it seems starts around 2020, and as here we most likely have a substantial data gap we can project it until 2023 and be uncertain about the interpretation of it. What can be said with some certainty is that the discipline is continuing in the same/similar direction, with deep learning becoming a strong factor and natural language processing having a role as well -- these will most likely have a significant impact on the technological side, but as the history and necessity tells us their inclusion into finance and entrepreneurship will not be abstained from.

\newgeometry{vmargin=2.5cm} 
\restoregeometry

\begin{figure}[h] 
	\centering
	\includegraphics[width=1\textwidth]{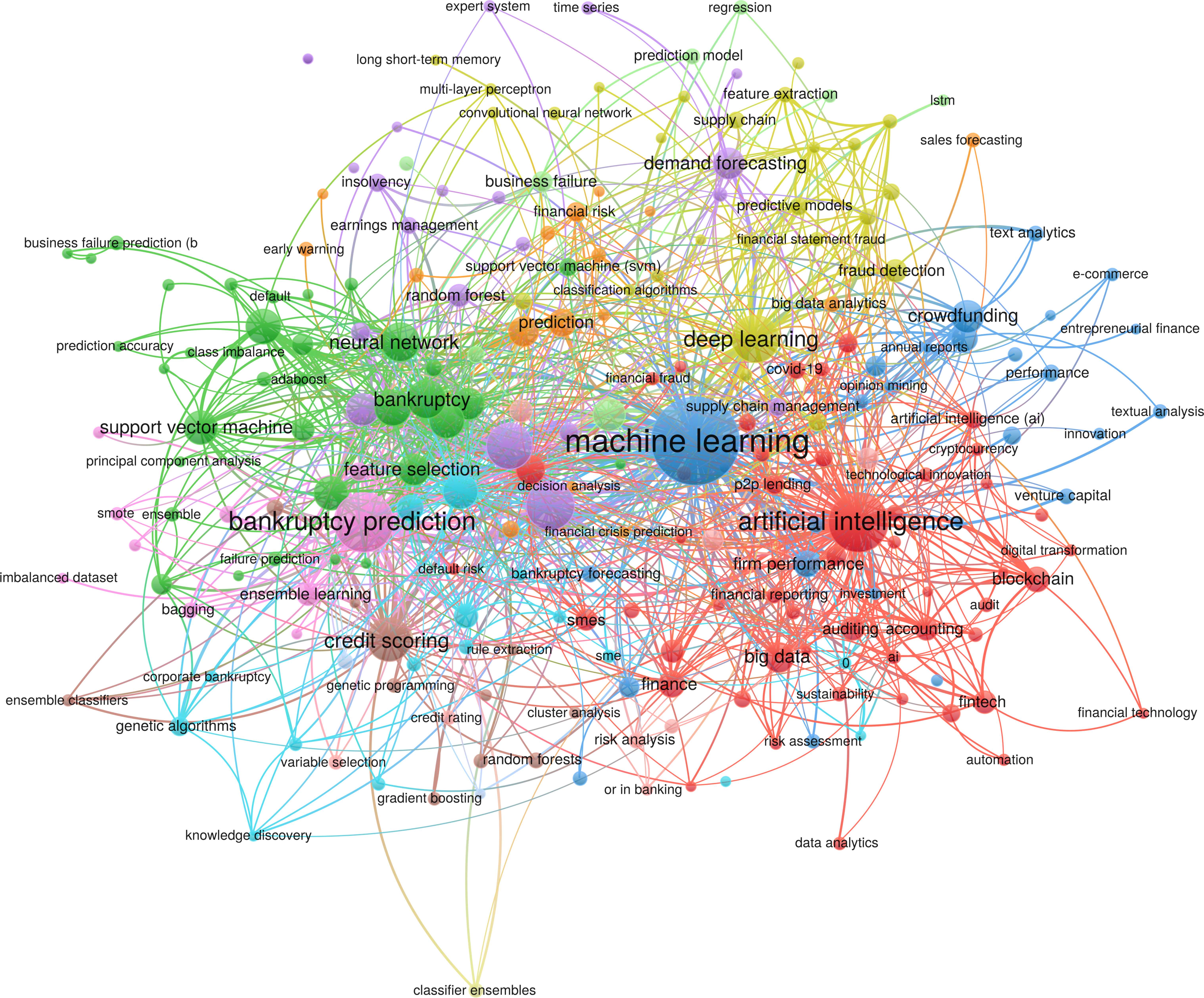}
	\caption{Authors Keywords Co-occurrence (determined according to documents in which items co-occur). The list of relevant terms was generated using authors' own keywords, by VOSviewer \cite{Eck2006,PerianesRodriguez2016} (logo was removed so as to make the figure content larger and easier to see), for the entire data document lifespan. In order for a word to appear in the figure, occurrence needs to equal to at least $ 5 $ -- links between keywords were fully counted (every co-occurrence is equal in weight). The analysis extracted $ 231 $ relevant terms, in $ 13 $ clusters, with $ 2490 $ links totaling $ 4258 $ in strength. Every colored blob represents a term and its size depends on the occurrence count, while graph edges represent co-occurrence with thickens representing how strong the link between terms is. Colors represent clusters, a tightly connected subgraph of nodes and edges, that is terms and links between them.}\label{fig:AKCO}
\end{figure}

\begin{figure}[h] 
	\centering
	\includegraphics[width=1\textwidth]{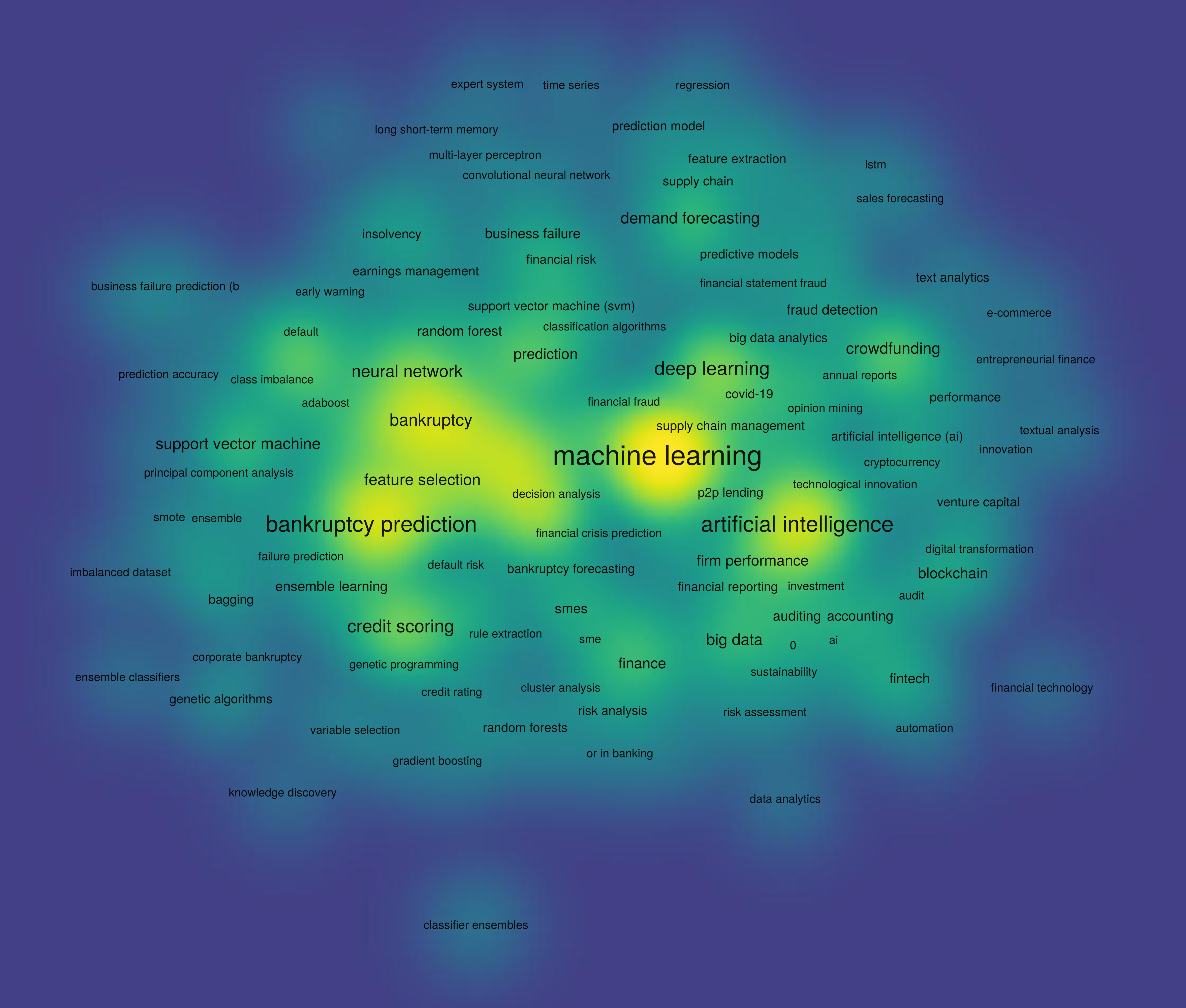}
	\caption{Authors Keywords Co-occurrence Density (determined according to documents in which items co-occur). The list of relevant terms was generated using authors' own keywords, by VOSviewer\protect\footnotemark \cite{Eck2006,PerianesRodriguez2016} (the logo was removed so as to make the figure content larger and easier to see), for the entire data documents lifespan. In order for a word to appear in the figure, occurrence needs to equal to at least $ 5 $ -- links between keywords were fully counted (every co-occurrence is equal in weight). The analysis extracted $ 231 $ relevant terms, in $ 13 $ clusters, with $ 2490 $ links totaling $ 4258 $ in strength. A density map represents a heat map where areas shifted to the blue are cold and weak in occurrence, while areas shifted to the red are hot and strong in occurrence.}\label{fig:AKCOD}
\end{figure}

In order to ascertain main themes and achieve a more fine-grained result an analysis of co-occurrence is also presented in Figure \ref{fig:AKCO}, together with density analysis in Figure \ref{fig:AKCOD}. Main themes, in line with analyses performed thus far, are machine learning, artificial intelligence, and bankruptcy prediction (which needed to be merged with bankruptcy). In itself knowing only these themes is not that useful, fortunately, via co-occurrence one can determine clusters and links as well.

Within the machine learning cluster prominent themes are firm performance, crowdfunding, natural language processing, sentiment analysis, and decision support systems -- indicating it seems the need on the one side, and context on the other. Within artificial intelligence prominent themes are big data, artificial neural network, blockchain, peer-to-peer lending, fin-tech, accounting, auditing, finance, and SMEs (small and medium-sized enterprises) -- combining new technologies and new ways of conducting business, with entrepreneurship and old questions of interest, indicating a changing economic environment, an adaptation to the new. While within bankruptcy prediction prominent themes are ensemble learning, boosting, and imbalanced data -- cluster concerned with using state of the art for business purposes, while battling with data issues.

There are other large themes of note, like credit scoring (part of a cluster are genetic programming, random forests, xgboost (most likely extreme gradient boosting), etc.), neural networks (should be merged with neural network cluster, part of these are data mining, demand forecasting, classification, random forest, financial distress prediction, support vector machine, feature selection, logistic regression, etc.), artificial neural networks (mergeable with other neural networks, part of a cluster are support vector machines, decision trees, genetic algorithms, rough sets, corporate failure, etc.), deep learning (part of a cluster are optimization, fraud detection, long short-term memory, predictive models, feature extraction, etc.), and genetic algorithm (part of a cluster are prediction, financial risk, supply chain finance, big data analytics etc.).

A diverse set of clusters, having a strong neural net presence (out of 13 clusters, in 7 of them one can find a neural network in one form or the other). There are also clusters more geared toward computing, but also those geared perhaps more toward economics, most likely a result of scientific contribution and author background. It is also a fact of note that clusters are generally quite intertwined, both through co-occurrence and theme appearance in different clusters, thus making a theme a part of one cluster, or multiple clusters, but then perhaps having a cluster where that theme is the dominant one -- indicating relevance to cluster and the field (e.g. neural network, genetic algorithm, support vector machine, bankruptcy prediction, business failure, fraud detection, credit risk, decision support).

Out of 1890 documents, 231 terms were extracted, with the effort taken to strike a balance between quantity and quality. Considering $ 2490 $ links, totaling $ 4258 $ in strength, and with term occurrence going as high as over 300, the resulting network makes a compact occurrence network that describes this discipline at the intersection of AI, finance, and entrepreneurship in a relevant manner, generally but in-depth as well, enabled by the extensive document corpus.

\begin{figure}[h] 
	\centering
	\includegraphics[width=1\textwidth]{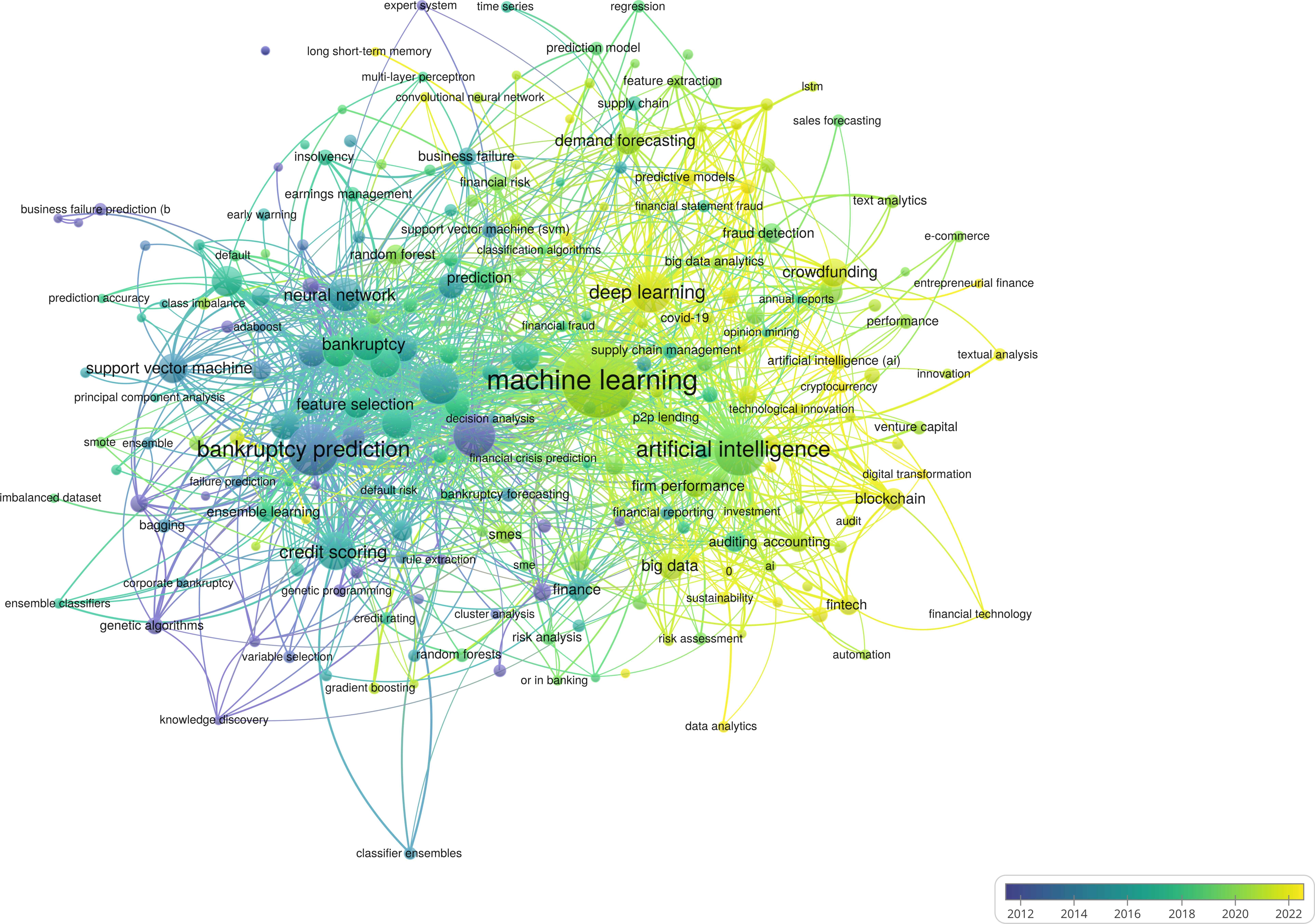}
	\caption{Authors Keywords Co-occurrence (determined according to number of documents in which items co-occur) Overlay. The list of relevant terms was generated using authors own keywords, by VOSviewer\protect\footnotemark \cite{Eck2006,PerianesRodriguez2016} (the logo was removed so as to make the figure content larger and easier to see), for the entire data document lifespan, with the overlay focusing on the last decade. In order for a word to appear in the figure, occurrence needs to equal to at least $ 5 $ -- links between keywords were fully counted (every co-occurrence is equal in weight). The analysis extracted $ 231 $ relevant terms, in $ 13 $ clusters, with $ 2490 $ links totaling $ 4258 $ in strength. The overlay map represents a graph timeline with nodes and edges shifted to the purple being old and part of the history, while areas shifted to the red are new/novel and part of the present.}\label{fig:AKCOT}
\end{figure}

Before this section is concluded we still want to determine the age overlay, presented in Figure \ref{fig:AKCOT}. Focusing on the last decade, as per set parameters, themes have seemed to overcome a change, from more contextualized at the beginning of the decade to more technological at the end of the decade, most likely a consequence of the popularity of artificial intelligence and related themes, and so a focus went from application in economics to emphasis on computing. General terms like machine learning, and artificial intelligence have a dominant presence, perhaps indicating substantial hype, with the emphasis on methods being less of a factor than one would hope.

The network can also be deceptive if not meticulously reasoned about, as for example, it seems that the presence of neural network is diminished, but the deep learning theme is very strong, leaning strongly towards the end of the decade. Bankruptcy has lost it luster, but prediction and themes related to performance, and analytics as well, are still relevant today.

Aside from deep learning, highly occurring recent themes are crowdfunding, blockchain, fin-tech, big data, and text mining. Still a mix of computing, economics, and statistics, an indication of discipline direction but development and changing environment as well. When looked at as a whole, it seems as of late that there is overly strong of an emphasis on computer science, and statistics, with too little time spent on a problem domain, yet AI is not its own end goal, at least it should not be. There is also a possibility that certain themes need maturation, or that collaboration between theory and practice, or computing and economics experts is not on a level. If recent themes are an indication, then there is a new kind of economics in development, both financial and social.

\subsection{Conceptual Data Analysis}
\label{sec:CDA}

\begin{figure}[h] 
	\centering
	\includegraphics[width=1\textwidth]{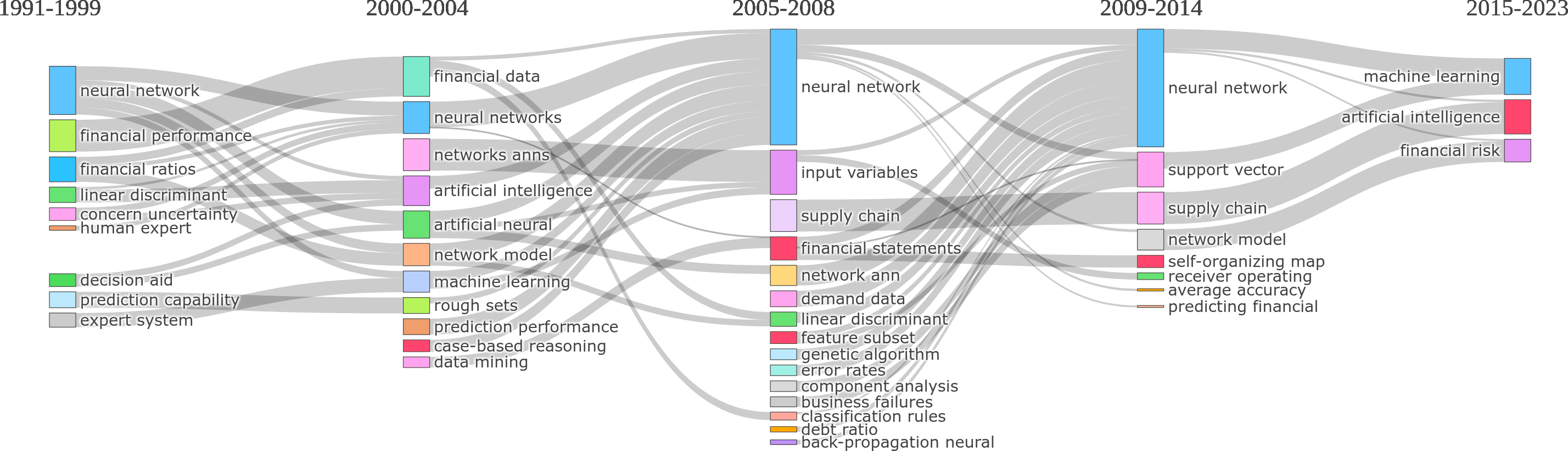}
	\caption{Thematic Evolution of Concepts -- from $ 1991 $ until $ 2014 $, with the last time slice, $ 2015 $ until $ 2023 $, representing the future to come. Time periods were selected for the reason of being one of the periods of interest from analysis in Figure \ref{fig:tcpy}. As we are interested in concepts, authors' abstracts were used for the analysis, since by choosing bi-grams we were able to achieve improved descriptiveness of the clusters, compared to keywords, with the number of words being $ 1000 $ and minimum cluster frequency set at $ 20 $, so as to have high confidence in the results and pick those terms that are prevalent. The minimum weight index was set to the lowest value of $ 0.02 $, wanting to explore in-depth, while the clustering algorithm used was Walktrap, a random walk-based algorithm known for its ability to capture with quality community structure of a network (based on the idea that short random walks belong to the same community). \cite{Pons2005,Yang2016} Colored rectangles represent clusters that are observed during specified periods, period is noted above the clusters, with terms representing an overarching theme of the cluster -- size of the cluster represents significance during the period, the bigger the cluster, the more importance for the period the cluster has. Clusters are typically linked to other clusters, either in the vicinity, that is right next to, or to the ones farther apart, jumping the periods, depending on which clusters the influence was exhorted. The thickness of the edge represents how strong the influence is, with thicker the edge, the stronger the influence. For most highly cited documents per period with data on the specific peaks one should consult Appendix \ref{secA3}. A word of caution, analysis in for example Figure \ref{fig:AKCO} has been conducted with authors' own keywords, while thematic evolution of concepts was performed with abstracts, therefore one is not to compare terms in such a situation, but themes and ideas.}\label{fig:TM2006}
\end{figure}

\begin{figure}[h] 
	\centering
	\includegraphics[width=1\textwidth]{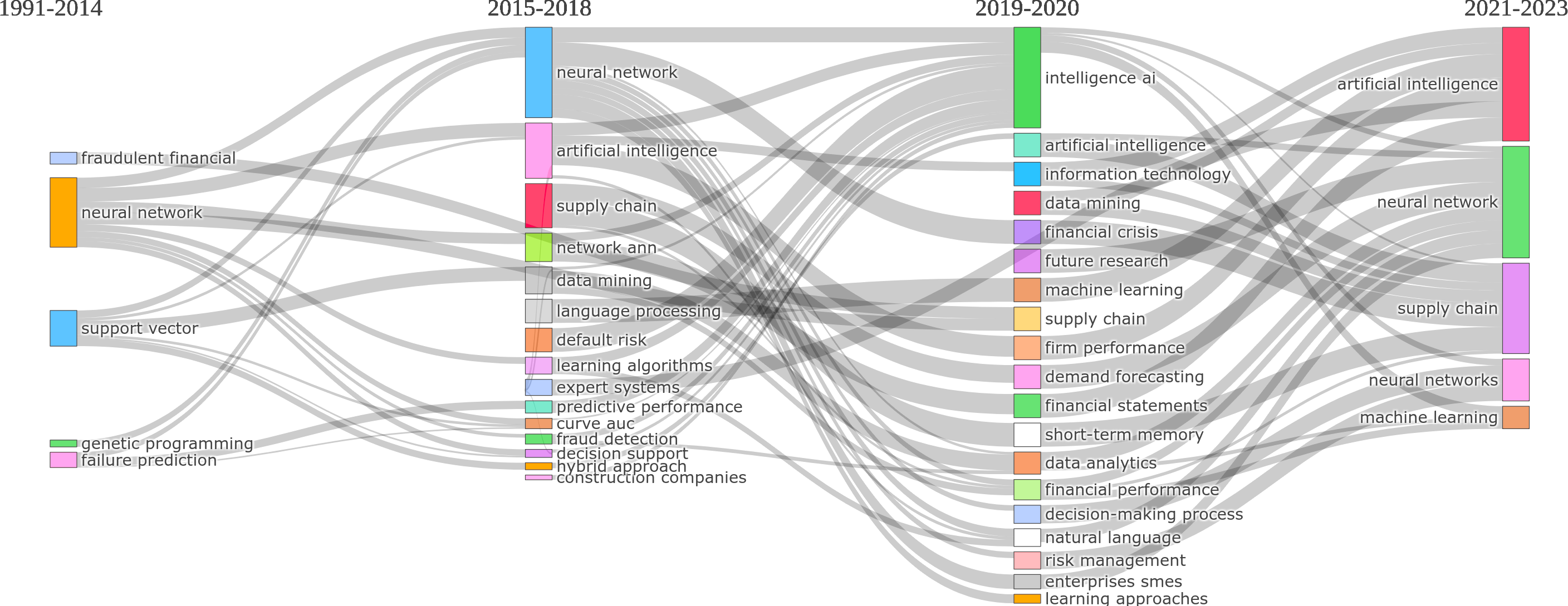}
	\caption{Thematic Evolution of Concepts -- from $ 2015 $ until $ 2023 $, with the last time slice, $ 2021 $ until $ 2023 $, representing the future to come for which the data is not yet complete. Time periods were selected for the reason of being one of the periods of interest from analysis in Figure \ref{fig:tcpy}. As we are interested in concepts, authors' abstracts were used for the analysis, since by choosing bi-grams we were able to achieve improved descriptiveness of the clusters, compared to keywords, with the number of words being $ 1000 $ and minimum cluster frequency set at $ 20 $, so as to have high confidence in the results and pick those terms that are prevalent. The minimum weight index was set to the lowest value of $ 0.02 $, wanting to explore in-depth, while the clustering algorithm used was Walktrap, a random walk-based algorithm known for its ability to capture with quality community structure of a network (based on the idea that short random walks belong to the same community). \cite{Pons2005,Yang2016} Colored rectangles represent clusters that are observed during specified periods, period is noted above the clusters, with terms representing an overarching theme of the cluster -- size of the cluster represents significance during the period, the bigger the cluster, the more importance for the period the cluster has. Clusters are typically linked to other clusters, either in the vicinity, that is right next to, or to the ones farther apart, jumping the periods, depending on which clusters the influence was exhorted. The thickness of the edge represents how strong the influence is, with thicker the edge, the stronger the influence. For most highly cited documents per period with data on the specific peaks one should consult Appendix \ref{secA3}. A word of caution, analysis in for example Figure \ref{fig:AKCO} has been conducted with authors' own keywords, while thematic evolution of concepts was performed with abstracts, therefore one is not to compare terms in such a situation, but themes and ideas.}\label{fig:TM2019}
\end{figure}

Analysis of authors' own keywords has been performed, it is now time to do the same for abstracts, and this can be seen in Figures \ref{fig:TM2006} and \ref{fig:TM2019}, representing longitudinal theme analysis. As we are dealing here with over 30 years of history, the analysis is divided into two parts, with the first part spanning from 1991 until 2014, and the second part continuing until the end, that is until 2023.

The first period is from 1991-1999, with themes being concerned with neural networks, performance, prediction, and expert systems, in line with previous analyses, and with a substantial prevalence of expert systems during the beginnings of the corpus data, while neural networks and financial performance make the most dominance, likely a consequence of later research outcomes.

At the beginning of the 21st century, we can observe a very strong presence of dealing with financial data in tandem with artificial intelligence, primarily machine learning and neural networks -- with prediction staying its course, and after years of research case-based reasoning becoming a factor.

Then from 2005-2008, neural networks became quite the powerhouse, with nothing else coming even close, and this state will become a trend from then onward. It also seems that in this period research was also concerned with broadness, as the number of themes is higher and more disparate than in previous periods, and we see input variables as substantial factors, supply chain, continuation of some statistical themes, appearance of genetic algorithm, etc.

The last period analyzed in detail from Figure \ref{fig:TM2006} is 2009-2014. Here we observe a consolidation of themes with the mentioned neural network trend. Here for the first time support vector machine became one of the dominant themes, with a self-organizing map, a type of neural network, being noticeable also. The period from 2015-2023, analyzed in detail in Figure \ref{fig:TM2019}, is what comes next, and from the looks of it three postulates are awaiting, machine learning, artificial intelligence, and financial risk, thus building on the aforementioned.

In Figure \ref{fig:TM2019} we begin by looking back to a period marked by neural networks and financial predictions and it seems to support vector machine, coming out of hiding, an indication of how important it is of importance to analyze data at different levels of abstraction.

The first period analyzed in detail, meaning of shorter time span, is from 2015-2018, and this period is seeing past themes receiving life again, the research is continuing, and improvements are being made. There are however two additions of note, language processing, and hybrid approach. With all the new developments, hybridization is naturally something of interest, and the research outcomes taken together with the need have most likely made language processing an invention sought after.

Before a conclusion can be made, there is yet the period of 2019-2020, with quite a large number of themes, a logical consequence of a period for which data is perhaps not yet complete and where an explosion of AI developments has taken place. As such, the enormous prevalence of a general theme of AI is present, not that useful in terms of finding specifics, however an indication of a situation. The research is continuing, with information technology, SMEs (small and medium-sized enterprises), and short-term memory (most likely long short-term memory) indicating a role of information, innovation, technology, and entrepreneurship in the period, and probably in the years to come.

The last period of the entire analysis is from 2021-2023, a period of significant data gaps, it is however not expected that the picture painted will be substantially different. Artificial intelligence is the norm, and specifically neural networks it would seem, a trend that started during the last part of the last decade of the 20th century, and still going strong, maturation, innovation and the environment have made AI a force to be reckoned with, a state most likely entrenched for a time. The supply chain is unusually strong, likely a result of the COVID-19 pandemic and the starting point in Wuhan, China \cite{Wu2020}. Machine learning is still a factor, but it seems that the statistical approach is waning.

Before we present intellectual structure data analysis there is one more thing to present here, namely artificial intelligence method occurrence by topic niches, in Table \ref{tab:MPG}. With this analysis, we have some sort of a lower bound of methods used, as some uses might be hidden in the text itself. Artificial intelligence has been mentioned an enormous number of times, with methods being mentioned far less, perhaps an indication of authors projecting AI onto an area but not dealing concretely with any method and thus producing research and application that is substantially below AI term occurrence.

If one reads bibliometric literature with documents trying to ascertain what methods are being published, it is found that such analyses are only partially useful, as they include terms such as artificial intelligence, machine learning, etc., broad terms describing either entire AI or large subdivisions, thus inhibiting specific information that is sought after. In this paper, we have taken a different approach by avoiding altogether such terms that are overly broad and in the end not that useful, keeping in mind that sometimes it is a struggle to differentiate between method, technique, algorithm, and even discipline subdivision, as Bibliometrics deals with general data analysis, and so as to resolve the issue and not miss certain uses and present an analysis that is not unpopulated, if something is close to a process/activities, then it is mentioned in the analysis in Table \ref{tab:MPG}.

The focus was methods, techniques, and algorithms that are not from statistics, that is the focus was on those items that are from computing, or computer science, e.g. randomized algorithms, neural nets, etc., as such items seem of most relevance today, as seen from Figure \ref{fig:TM2019}, and will most likely be dominant in the near future as well -- with having in mind that these methods also rely to a certain degree on statistics. If an item occurrence is below a minimum number of keyword occurrences set to $ 3 $, it is not included in the analysis, as such an item is on a level of rumor.

As confirmed by the analyses performed thus far, the most prominent is branch 1(c), which deals with valuation, and prediction of performance, with the largest corpus of documents being here as well. If a number of documents is taken into consideration then branch 3(b), financial planning, is also prominent in AI methods use, a branch linked to 1(c) -- performance, finance, and planning, items of most importance to every private company, and is therefore expected that companies would be most interested to leverage new technologies in these areas first, with the branch 3(a) following and 1(b) also being in the mix.

As far as methods are concerned, neural networks, and items linked to them, together with support vector machine, are extremely dominant, with neural networks being far ahead of everything else -- these two are most likely driven by their success in achieving results, and likely the hype also. Nevertheless, the number of items in the analysis is substantial, with 20 approaches in total, indicating an effort to tackle issues from different angles. These are the methods, using the term methods to encompass items sought after, predominantly neural networks, with statistical approaches, but with heuristics and meta-heuristics as well, all the way to programming paradigms. A number of approaches has a small footprint, especially if one looks at individual branches, yet these might be the first sparks, or perhaps an indication of the need to try the method in a different area. 

The lack of application e.g. in 1(a) and 2 (topic niches that have started to develop only recently), might also be an indication that collaboration of experts from computer science and economics is needed, as without that cooperation modernization of economics, and specifically entrepreneurship, will be difficult, while some methods are naturally more in line with certain problems. There are potentially many reasons why an item is present and present to a certain degree, and why it is not, with this analysis presenting state of the art, while at the same time a potential enabler for future developments. The severe lack of application, aside from a few deep learning attempts, in FinTech (in the context of entrepreneurship) most definitely presents a point of interest and future research.

Both neural networks, and support vector machine have the most prominence in branches 1(c) and 3(b), making them a natural combination of branches as per financial interest and methods that are propulsive and provide it seems the best results. Alongside these there is also the matter of ensemble learning and genetic algorithm, on the one side combining multiple approaches in order to solve a problem, and on the other using a metaheuristic to optimize and to search, borrowing from nature so as to solve a problem brought about by nature.

\newgeometry{vmargin=6cm}
\pdfpagewidth 14in
\begin{table}[t!] 
	\caption{Artificial Intelligence Method Occurrence by Topic Niches}\label{tab:MPG}
	\begin{tabularx}{2\textwidth}{@{\extracolsep\fill}>{\RaggedRight}p{.4cm}>{\RaggedRight}p{.4cm}>{\RaggedRight}p{5cm}lllllllllllllllllllll}
		\toprule%
		ON\textsuperscript{1} & ND\textsuperscript{2} & Topic Niches\textsuperscript{3} & ANN & DNN & BPNN & DT & EL & GA & RNN & RF & GB & TM & SVM & CNN & SOM & CBR & PSO & GP & LSTM & FNN & FZNN & GHSOM & $\sum$ \\
		\midrule
		1(a) & $ 121 $ & \textsc{Investment success/business performance and entrepreneur's behavior and presentation} & $ 11 $ & $ 7 $ & $  $ & $  $ & $  $ & $  $ & $  $ & $  $ & $  $ & $  $ & $  $ & $  $ & $  $  & $  $  & $  $ & $  $ & $  $ & $  $ & $  $ & $  $ & $ 18 $ \\
		1(b) & $ 384 $	& \textsc{Sources of entrepreneurial finance} & $ 48 $ & $ 22 $ & $ 3 $ & $ 11 $ & $ 8 $ & $ 3 $ & $  $ & $ 16 $ & $ 3 $ & $ 3 $ & $ 13 $ & $  $ & $  $ & $  $ & $  $ & $  $ & $  $ & $  $ & $  $ & $  $ & $ 130 $ \\
		1(c) & $ 950 $	& \textsc{Valuation of an entrepreneurial venture/Prediction of performance and/or bankruptcy} & $ 177 $ & $ 37 $ & $ 4 $ & $ 33 $ & $ 48 $ & $ 39 $ & $  $ & $ 18 $ & $ 16 $ & $  $ & $ 76 $ & $ 3 $ & $ 11 $ & $ 16 $ & $ 7 $ & $ 7 $ & $ 4 $ & $ 3 $ & $ 3 $ & $  $ & $ 502 $ \\
		2 & $ 39 $	& \textsc{FinTech in the context of entrepreneurship} & $  $ & $ 3 $ & $  $ & $  $ & $  $ & $  $ & $  $ & $  $ & $  $ & $  $ & $  $ & $  $ & $  $ & $  $ & $  $ & $  $ & $  $ & $  $ & $  $ & $  $ & $ 3 $ \\
		3(a) & $ 476 $	& \textsc{AI and accounting, auditing and detecting financial frauds} & $ 56 $ & $ 21 $ & $  $ & $ 11 $ & $  $ & $ 7 $ & $ 4 $ & $ 5 $ & $ 6 $ & $  $ & $ 17 $ & $ 4 $ & $ 6 $ & $  $ & $  $ & $  $ & $ 5 $ & $  $ & $  $ & $ 3 $ & $ 145 $ \\
		3(b) & $ 573 $	& \textsc{Financial planning and other aspects of financial management} & $ 77 $ & $ 36 $ & $ 12 $ & $ 10 $ & $ 12 $ & $ 15 $ & $ 6 $ & $ 9 $ & $ 10 $ & $  $ & $ 38 $ & $ 5 $ & $  $ & $ 7 $ & $ 6 $ & $  $ & $ 7 $ & $ 3 $ & $  $ & $  $ & $ 253 $ \\
		$\sum$ & $  $ & $  $ & $ 369 $ & $ 126 $ & $ 19 $ & $ 65 $ & $ 68 $ & $ 64 $ & $ 10 $ & $ 48 $ & $ 35 $ & $ 3 $ & $ 144 $ & $ 12 $ & $ 17 $ & $ 23 $ & $ 13 $ & $ 7 $ & $ 16 $ & $ 6 $ & $ 3 $ & $ 3 $ & $  $ \\
		\botrule
	\end{tabularx}
	\footnotetext{AI methods, techniques and algorithms were extracted from documents, categorized according to aforementioned branches, by VOSviewer \cite{Eck2006,PerianesRodriguez2016} via co-occurrence (determined according to the number of documents in which items co-occur) analysis for author keywords (so as to descriptively capture the content of the documents, as keywords plus are less comprehensive in terms of the actual content \cite{Zhang2015}) with the minimum number of keyword occurrence set to $ 3 $, so as to leverage precision and obscurity. Such analysis has produced clusters of term co-occurrences from which specific AI methods were then manually extracted and occurrence obtained.}
	\footnotetext{ANN -- Artificial Neural Network; DNN -- Deep Neural Network; BPNN -- Back Propagation Neural Network; DT -- Decision Tree; EL -- Ensemble Learning; GA -- Genetic Algorithm; RNN -- Recurrent Neural Network; RF -- Random Forest; GB -- Gradient Boost; TM -- Topic Modeling; SVM -- Support Vector Machine; CNN -- Convolutional Neural Network; SOM -- Self-organizing Map; CBR -- Case-based Reasoning; PSO -- Particle Swarm Optimization; GP -- Genetic Programming; LSTM -- Long Short-term Memory; FNN -- Feed-forward Neural Network; FZNN -- Fuzzy Neural Network; GHSOM -- Growing Hierarchical Self-organizing Map}
	\footnotetext[1]{Ordinal number identifying the particular branch.}
	\footnotetext[2]{Number of documents per specific branch, total sum higher than $ 1890 $ (number of documents in bibliometric analysis) as papers can belong to different categories at the same time, detected retracted papers are excluded -- documents analyzed span the entire period of the research.}
	\footnotetext[3]{Topic Niches, or branches, as identified during preliminary literature search. Branches of type 1 are a part of the subject, AI as support for entrepreneurial financing decisions; branches of type 3 are a part of the subject, Management of entrepreneurial finance.}
	\footnotetext{For a heat map of the data in the table one can consult Figure \ref{fig:HM1} in appendix. If one is interested in the classification of AI methods, techniques and algorithms, then e.g. \cite{Dhall2019,Das2020,Zhang2021,Tapeh2022} can be consulted.}
\end{table}
\restoregeometry
\pdfpagewidth 8.25in

As a last few words, explainable AI (XAI) can be brought into the foreground, as it is one thing to obtain a quality result, while it is another to know how the algorithm got there, and sometimes that is important, reasoning from cause to effect. This is especially evident when one works in a situation with potentially devastating consequences, or in an area of high uncertainty, e.g. one might try to optimize business processes so as to achieve greater financial gain, yet without knowing why is the result the way it is, it is problematic to be sure in the result of a method and also more difficult to predict what will real consequences be.

On the algorithmic side, it also might be possible to design improved algorithms or get an idea of how to make something improved, or even design an algorithm from the beginning when one has more information about the problem being tackled, if one can understand how AI agent went about an issue and why it made such a move.

\subsection{Intellectual Structure Data Analysis}
\label{sec:ISDA}

\begin{figure}[h] 
	\centering
	\includegraphics[width=1\textwidth]{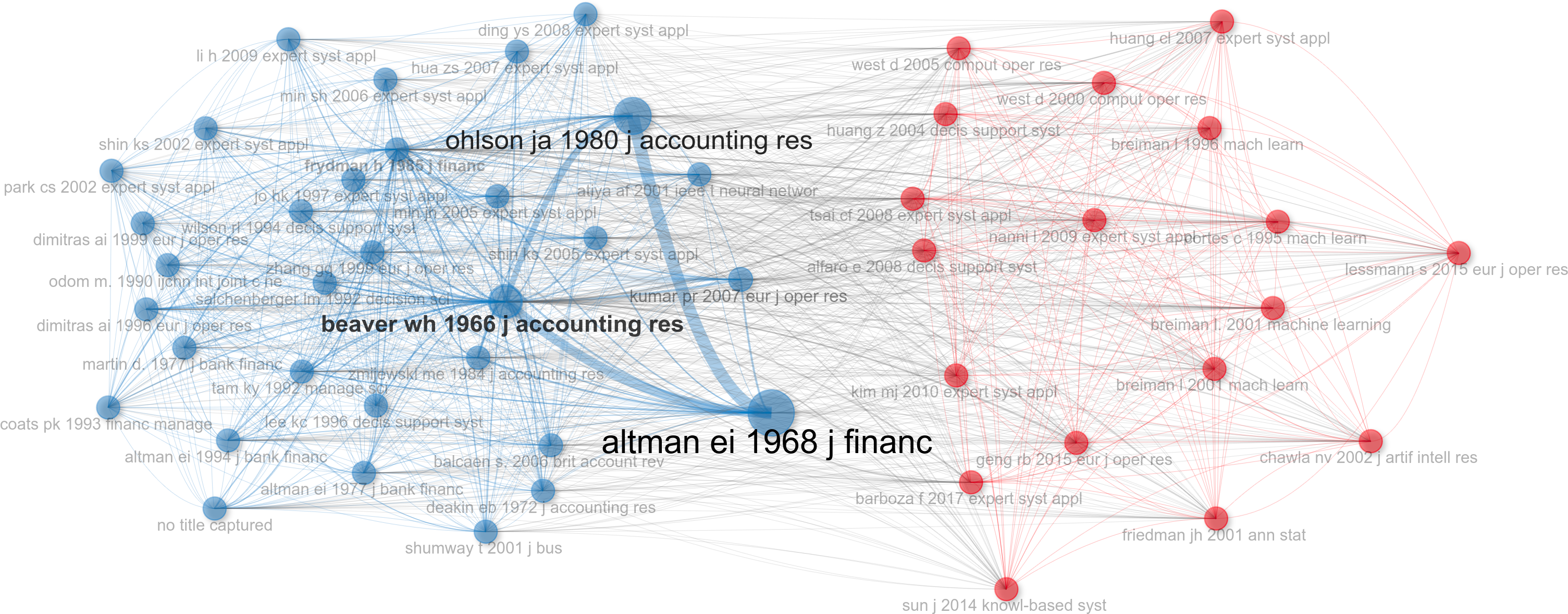}
	\caption{References Co-citation Network (determined according to the number of times references have been co-cited, i.e. cited together in a third item). As in Bibliometrix, it is difficult to ascertain relationships on large graphs, the network above was generated on $ 50 $ nodes, with isolated nodes removed and the minimum number of edges set to $ 1 $, in order to take into account the entire subgraph but without islands which are not being part of the greater field. The clustering algorithm used was Walktrap, a random walk-based algorithm known for its ability to capture with quality community structure of a network (based on the idea that short random walks belong to the same community). \cite{Pons2005,Yang2016} Nodes are colored according to the cluster they belong to, while labels denote a reference (author, year, source). Reference size is determined according to the number of citations, with a higher number of citations represented by a bigger node, coupled with an edge, the size of which is being determined as per co-citation, thicker edge means there is a greater degree of co-citation.}\label{fig:RCCN}
\end{figure}

In order to ascertain the authors' contribution to the field and determine which references (both references and documents of the research corpus) are of foundational knowledge, a references co-citation network was performed, and presented in Figure \ref{fig:RCCN}. At first glance, one can observe that Altman (1968), Ohlson (1980), and Beaver (1966) are of the highest relevance, there is however a caveat here.

Indeed, "through the analysis of reference co-citation, the most frequently cited" references "are Altman (1968), Ohlson (1980) and Beaver (1966)." \cite{Shi2019} Yet "none of their work is based on an artificial intelligence approach," but "due to the fact that all aforementioned works are pioneer studies in the bankruptcy prediction field, the posterior authors tend to cite them in their papers with high frequency." \cite{Shi2019}

\begin{figure}[h] 
	\centering
	\includegraphics[width=1\textwidth]{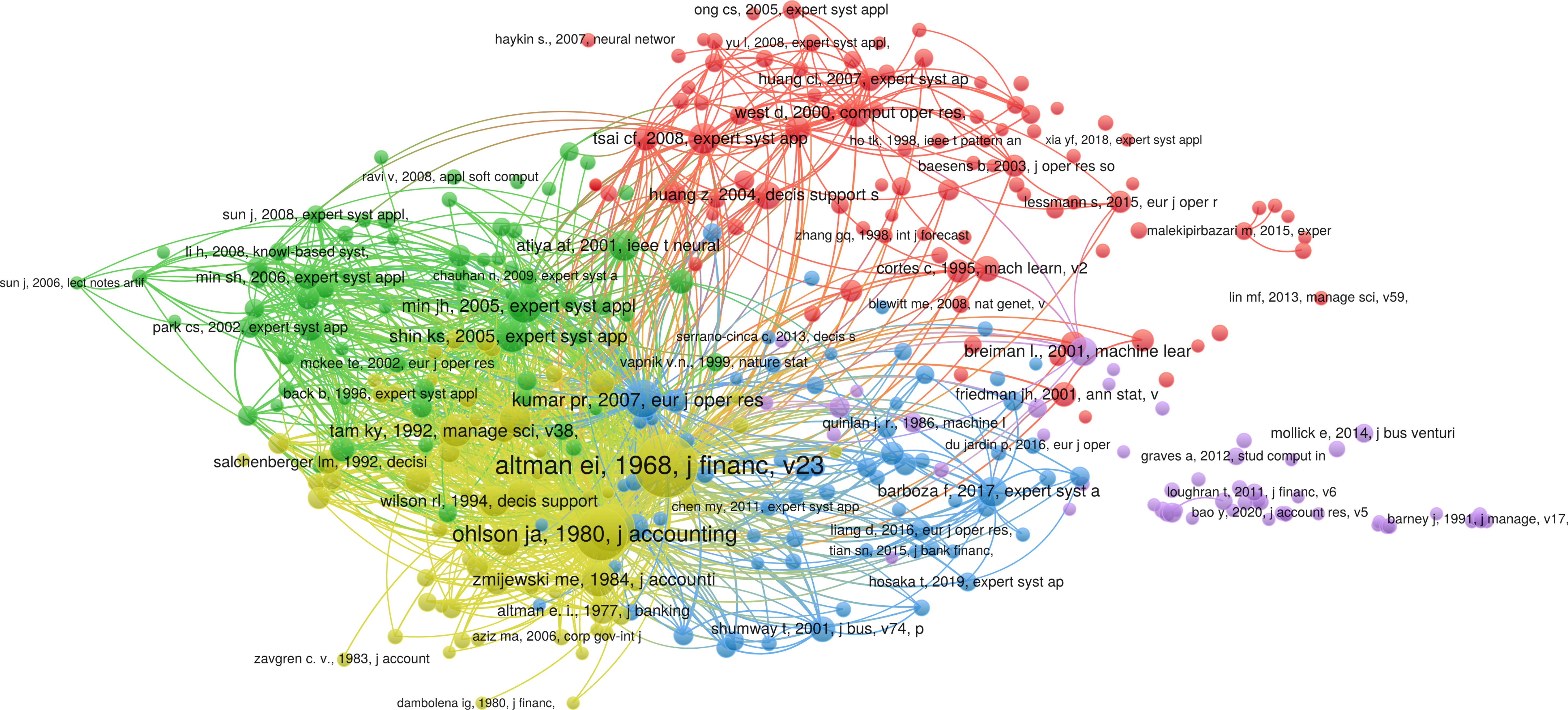}
	\caption{References Co-citation Network (determined according to the number of times references have been co-cited, i.e. cited together in a third item). The figure was generated by VOSviewer \cite{Eck2006,PerianesRodriguez2016} (logo was removed so as to make the figure content larger and easier to see) with the citation data ending in July 2023 -- this analysis has a different ending date, as because of the problems with Bibliometrix \cite{Aria2017,Aria2023} we had to again export the same data from Web of Science into textual file (for the VOSviewer \cite{Eck2006,PerianesRodriguez2016}, since analysis was conducted in BibTeX, but the VOSviewer does not support that file format). Compared to citations output by Bibliometrix \cite{Aria2017,Aria2023}, and within the top $ 10 $ references, there is almost no difference in reference citation count, only one citation difference here and there, with one position exchange for neighboring references which in Bibliometrix had the same citation count -- results from both tools are almost identical. Sub-network colors denote clusters, while node size depends on the number of citations, the bigger the citation count the bigger the node itself, with edges representing co-citations, with thicker links representing stronger co-citation. Node labels, naturally, denote a reference (author, year, source). Co-citations are fully counted (every co-occurrence is equal in weight), and as now plotting the full network, the minimum number of citations was set to $ 20 $, as outliers are of no interest here -- results of which was $ 353 $ focus references. The resulting network has $ 5 $ clusters, $ 37069 $ links, and a total link strength of $ 146241 $.}\label{fig:RCCNvos}
\end{figure}

It is therefore of interest to present a broader outlook, and with that in mind analyses in Tables \ref{tab:MRRCNbib} and \ref{tab:MRRCNbib1} are conducted also. As Bibliometrix is inferior to VOSviewer in terms of large network visualization, both tools were used, first we present analysis from Bibliometrix in Figure \ref{fig:RCCN}, and afterwards from VOSviewer in Figure \ref{fig:RCCNvos}.

Analysis in Figure \ref{fig:RCCN} presents two clusters, the first blue, with Altman (1968), Ohlson (1980), and Beaver (1966), and the second red -- indicating two subdivisions within the observed discipline at the intersection of AI, finance, and entrepreneurship. In the blue cluster, other references of note are presented in Table \ref{tab:MRRCNbib}, and as seen an overall analysis shows the dominance of the blue cluster. All the references are of an older type, and the journals published correspond to analyses performed thus far. 

Half of the references are not a part of the research corpus, but are present in a reference form only, indicating relevance of both the old and the new, but also the relevance of a broader knowledge. Of the references part of the research corpus, in every instance branch VP (1(c) in Table \ref{tab:MPG}) is present, thus indicating why the blue cluster has such dominance overall, as described when interpreting Table \ref{tab:MPG}. With FP (3(b) in Table \ref{tab:MPG}) and AIF (3(a) in Table \ref{tab:MPG}) also being a factor and contributing to the relevance of the blue cluster, which was also of note in the methods analysis performed in Table \ref{tab:MPG}.

As in the overall analysis, all the most relevant items are in the blue cluster, a special analysis of the red cluster is warranted and is thus presented in Table \ref{tab:MRRCNbib1}. The top three authors here are Tsai (2008), Nanni (2009), and West (2005). In the red cluster references are of a more recent nature, with one reference from 2017, indicating a discipline subdivision that is more of a continuation of the roots -- confirmation of which is a list of journals, where Expert Systems with Application is a dominant force, in alignment with previous analyses. The prevalent theme here is also VP (1(c) in Table \ref{tab:MPG}), with the absence of FP (3(b) in Table \ref{tab:MPG}) and AIF (3(a) in Table \ref{tab:MPG}), instead SEF (1(b) in Table \ref{tab:MPG}) is present, a strong branch, but not to par with other prominent ones, thus such a cluster is not as relevant as the blue one is. The red cluster is sitting more on the recent achievements, both via year of publication and via branch of economics, and is therefore a cluster less relevant than the blue cluster.

\begin{table}[h]	
	\caption{Most Relevant References Overall in Co-citation Network from Figure \ref{fig:RCCN}}\label{tab:MRRCNbib}%
	\begin{tabular}{@{}lll}
		\toprule
		Reference\textsuperscript{1} & Cluster\textsuperscript{2} & Branch\textsuperscript{3} \\
		\midrule
		Altman EI, 1968, J FINANC 				& 2 & ref \\
		Ohlson JA, 1980, J ACCOUNTING RES 		& 2 & ref \\
		Beaver WH, 1966, J ACCOUNTING RES 		& 2 & ref \\
		Kumar PR, 2007, EUR J OPER RES 			& 2 & VP, FP \\
		Min JH, 2005, EXPERT SYST APPL 			& 2 & VP \\
		Shin KS 2005, EXPERT SYST APPL 			& 2 & VP \\
		Zmijewski ME, 1984, J ACCOUNTING RES 	& 2 & ref \\
		Tam KY, 1992, MANAGE SCI 				& 2 & ref \\
		Wilson RL, 1994, DECIS SUPPORT SYST 	& 2 & VP, AIF \\
		Zhang GQ, 1999, EUR J OPER RES 			& 2 & VP \\
		\botrule
	\end{tabular}
	\footnotetext[]{Sorted according to reference relevance in terms of PageRank (approximating importance, assuming more important nodes have more links to them from other sources while taking into account that not every link has the same weight \cite{Bianchini2005}) calculated by Bibliometrix, in decreasing order}
	\footnotetext[1]{Reference label from Figure \ref{fig:RCCN}}
	\footnotetext[2]{Clusters from Figure \ref{fig:RCCN} -- 2 (blue), 1 (red)}
	\footnotetext[3]{Subject branches as defined in Subsection~\ref{subsec6} -- if stated as 'ref', the reference is not part of document corpus, but rather only a reference, and is thus not clustered into any branch}
	\footnotetext[VP]{Valuation of an entrepreneurial venture/Prediction of performance and/or bankruptcy}
	\footnotetext[AIF]{AI and accounting, auditing and detecting financial frauds}
	\footnotetext[FP]{Financial planning and other aspects of financial management}
\end{table}

\begin{table}[h]	
	\caption{Most Relevant References for Cluster Red (identification 1) in Co-citation Network from Figure \ref{fig:RCCN}}\label{tab:MRRCNbib1}%
	\begin{tabular}{@{}lll}
		\toprule
		Reference\textsuperscript{1} & Cluster\textsuperscript{2} & Branch\textsuperscript{3} \\
		\midrule
		Tsai CF, 2008, EXPERT SYST APPL 	& 1 & VP \\
		Nanni L, 2009, EXPERT SYST APPL 	& 1 & VP \\
		West D, 2005, COMPUT OPER RES 		& 1 & VP \\
		West D, 2000, COMPUT OPER RES 		& 1 & ref \\
		Alfaro E, 2008, DECIS SUPPORT SYST 	& 1 & VP \\
		Huang Z, 2004, DECIS SUPPORT SYST 	& 1 & ref \\
		Breiman L, 2001, MACHINE LEARNING 	& 1 & ref \\
		Huang CL, 2007, EXPERT SYST APPL 	& 1 & ref \\
		Barboza F, 2017, EXPERT SYST APPL 	& 1 & SEF, VP \\
		Kim MJ, 2010, EXPERT SYST APPL 		& 1 & VP \\
		\botrule
	\end{tabular}
	\footnotetext[]{Sorted according to reference relevance in terms of PageRank (approximating importance, assuming more important nodes have more links to them from other sources while taking into account that not every link has the same weight \cite{Bianchini2005}) calculated by Bibliometrix, in decreasing order}
	\footnotetext[1]{Reference label from Figure \ref{fig:RCCN}}
	\footnotetext[2]{Clusters from Figure \ref{fig:RCCN} -- 2 (blue), 1 (red)}
	\footnotetext[3]{Subject branches as defined in Subsection~\ref{subsec6} -- if stated as 'ref', the reference is not part of document corpus, but rather only a reference, and is thus not clustered into any branch}
	\footnotetext[VP]{Valuation of an entrepreneurial venture/Prediction of performance and/or bankruptcy}
	\footnotetext[SEF]{Sources of entrepreneurial finance}
\end{table}

\begin{figure}[h] 
	\centering
	\includegraphics[width=1\textwidth]{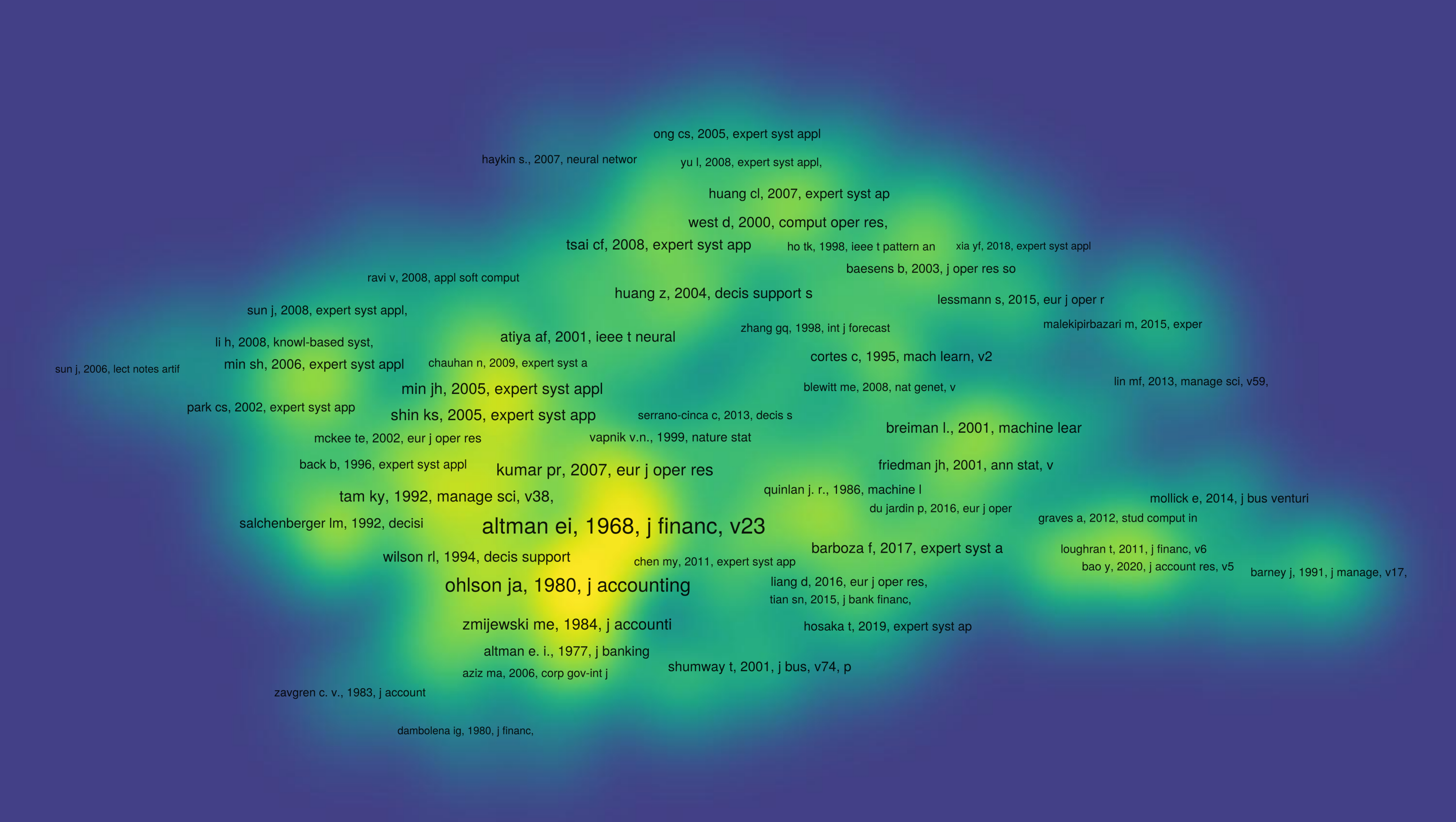}
	\caption{References Co-citation Density Map (determined according to the number of times references have been co-cited, i.e. cited together in a third item). The figure was generated by VOSviewer \cite{Eck2006,PerianesRodriguez2016} (logo was removed so as to make the figure content larger and easier to see) with the citation data ending in July 2023 -- this analysis has a different ending date, as because of the problems with Bibliometrix \cite{Aria2017,Aria2023} we had to again export the same data from Web of Science into textual file (for the VOSviewer \cite{Eck2006,PerianesRodriguez2016}, since analysis was conducted in BibTeX, but the VOSviewer does not support that file format). Compared to citations output by Bibliometrix \cite{Aria2017,Aria2023}, and within the top $ 10 $ references, there is almost no difference in reference citation count, only one citation difference here and there, with one position exchange for neighboring references which in Bibliometrix had the same citation count -- results from both tools are almost identical. Labels, naturally, denote a reference (author, year, source). Co-citations are fully counted (every co-occurrence is equal in weight), and as now plotting the full network, the minimum number of citations was set to $ 20 $, as outliers are of no interest here -- results of which was $ 353 $ focus references. The resulting network has $ 5 $ clusters, $ 37069 $ links, and a total link strength of $ 146241 $. A density map represents a heat map where areas shifted to the blue are cold and weak in citations, while areas shifted to the red are hot and strong in citations.}\label{fig:RCCNvosD}
\end{figure}

To ascertain a broader image, an analysis of the entire dataset was performed, together with the density analysis, and is presented in Figures \ref{fig:RCCNvos} and \ref{fig:RCCNvosD}. Considering a different tool was used here, some specific results are different, however, a general direction and conclusion is the same -- confirmed as well by Table \ref{tab:MRRCNvos}.

The top three references overall are still Altman (1968), Ohlson (1980), and Beaver (1966), as per the mentioned circumstances, no surprise here, an expected result. This time though, we are analyzing the entire dataset, which has resulted in five clusters.

Considering that the network has 353 nodes (that is references) number of links (that is edges) is substantial, with the link strength also indicating a compact network. The reason for this lies in the fact that four out of five clusters predominantly cover one thematic niche - VP (1 (c) in Table 14). In the yellow cluster are references related to authors who generally made a significant contribution to the development of bankruptcy prediction models, while the green and blue clusters cover newer references with the application of AI in the VP domain. The red cluster includes references with credit scoring topics, also dominantly leaning on the VP niche.

Finally, the purple cluster (cluster with identification 5) is weakly linked to the rest of the network and in fact within itself as well (some links are not seen as per the defined variable of the VOSviewer to draw 1000 lines). It is a cluster with references covering newer topics in the intersection of AI-entrepreneurship-finance, such as for example crowdfunding, and the application of text analysis methods in finance.

The previous conclusions are confirmed by the References Co-citation Density Map in Figure \ref{fig:RCCNvosD}. The lower left quadrant stands out the most and coincides with the yellow cluster in Figure \ref{fig:RCCNvos}. As said, these are influential references related to the development of bankruptcy prediction models throughout history.

Out of all references in Table \ref{tab:MRRCNvos} only one is not from the blue cluster in Figure \ref{fig:RCCN}, Tsai CF (2008), a reference that is most relevant in the red cluster, as seen from Table \ref{tab:MRRCNbib1}. Confirming the high relevance of the blue cluster in the entire network, foundational for the entire network, with the additional knowledge so to speak slowly making its place -- also indicating how difficult it is to compete with the classics. In terms of year of publication and reference source, the situation in Table \ref{tab:MRRCNvos} closely resembles information found in Table \ref{tab:MRRCNbib} with conclusions being matched, aside from one mentioned exception.

When results from Tables \ref{tab:MRRCNbib}, \ref{tab:MRRCNbib1}, and \ref{tab:MRRCNvos} are combined, the trend from economics, prediction, etc. to the same in the strong (depending on the branch) context of computing is clearly seen, there is a transformation ongoing, the result of which is still in expectancy it would seem. Although not a focus of this research, every transformation brings its challenges, especially if digital technology is involved, and with such an emphasis in mind, an obvious lack of security and privacy content (as per authors own keywords) is present, thus being a warranted research area in the context of individual, societal, social, and entrepreneurial aspect.

References are classified in different clusters, as per co-citation criteria and tool settings of course, with the obvious absence of the purple cluster 5. The dominance of cluster 4, a cluster we could say corresponds to the blue cluster in Figure \ref{fig:RCCN}, is obvious, with $ 6/10 $ of a presence. Outside of references in this cluster, a number of references were in this larger network dispersed among clusters 2, 3, and 1 -- identifying a subdivision and dispersion as per co-citation. However, as per branches of interest in Subsection~\ref{subsec6}, and as defined for the research, the prevalence of branches 1(c) and 3(b), with the highest document count as well, is obvious, especially so for 1(c) with its almost unanimous presence -- confirming conclusions from the analysis in Figure~\ref{fig:RCCN}. Half of the references are not a part of the research corpus, but are present in a reference form only, indicating relevance of both the old and the new, but also the relevance of a broader knowledge.

\begin{table}[h]	
	\caption{Most Relevant References Overall in Co-citation Network from Figure \ref{fig:RCCNvos}}\label{tab:MRRCNvos}%
	\begin{tabular}{@{}lll}
		\toprule
		Reference\textsuperscript{1} & Cluster\textsuperscript{2} & Branch\textsuperscript{3} \\
		\midrule
		Altman EI, 1968, J FINANC, V23 				& 4 & ref \\
		Ohlson JA, 1980, J ACCOUNTING RES, V18 		& 4 & ref \\
		Beaver WH, 1966, J ACCOUNTING RES, V4 		& 4 & ref \\
		Kumar PR, 2007, EUR J OPER RES, V180 		& 3 & VP, FP \\
		Min JH, 2005, EXPERT SYST APPL, V28 		& 2 & VP \\
		Shin KS, 2005, EXPERT SYST APPL, V28 		& 2 & VP \\
		Zmijewski ME, 1984, J ACCOUNTING RES, V22 	& 4 & ref \\
		Zhang GQ, 1999, EUR J OPER RES, V116 		& 4 & VP \\
		Tsai CF, 2008, EXPERT SYST APPL, V34 		& 1 & VP \\
		Tam KY, 1992, MANAGE SCI, V38 				& 4 & ref \\
		\botrule
	\end{tabular}
	\footnotetext[]{Sorted according to reference relevance in terms of Total Link Strength (full counting method used for links) calculated by VOSviewer, in decreasing order. As VOSviewer depicts more information about the references these are included, as it complements previous data and it makes the reference more easily identifiable. Data from the table are retrieved from analysis in Figure \ref{fig:RCCNvos} with $ 353 $ references and $ 5 $ clusters.}
	\footnotetext[1]{Reference label from Figure \ref{fig:RCCNvos}}
	\footnotetext[2]{Clusters from Figure \ref{fig:RCCNvos} -- 5 (purple), 4 (yellow), 3 (blue), 2 (green), 1 (red)}
	\footnotetext[3]{Subject branches as defined in Subsection~\ref{subsec6} -- if stated as 'ref', the reference is not part of document corpus, but rather only a reference, and is thus not clustered into any branch}
	\footnotetext[VP]{Valuation of an entrepreneurial venture/Prediction of performance and/or bankruptcy}
	\footnotetext[FP]{Financial planning and other aspects of financial management}
	\footnotetext[Comment]{As compared to Table \ref{tab:MRRCNbib}, Wilson RL, 1994, DECIS SUPPORT SYST reference is missing, as in Total Link Strength calculation performed by VOSviewer this reference ranked 11th, with having 9th place in Table \ref{tab:MRRCNbib}}
\end{table}

\begin{figure}[h] 
	\centering
	\includegraphics[width=1\textwidth]{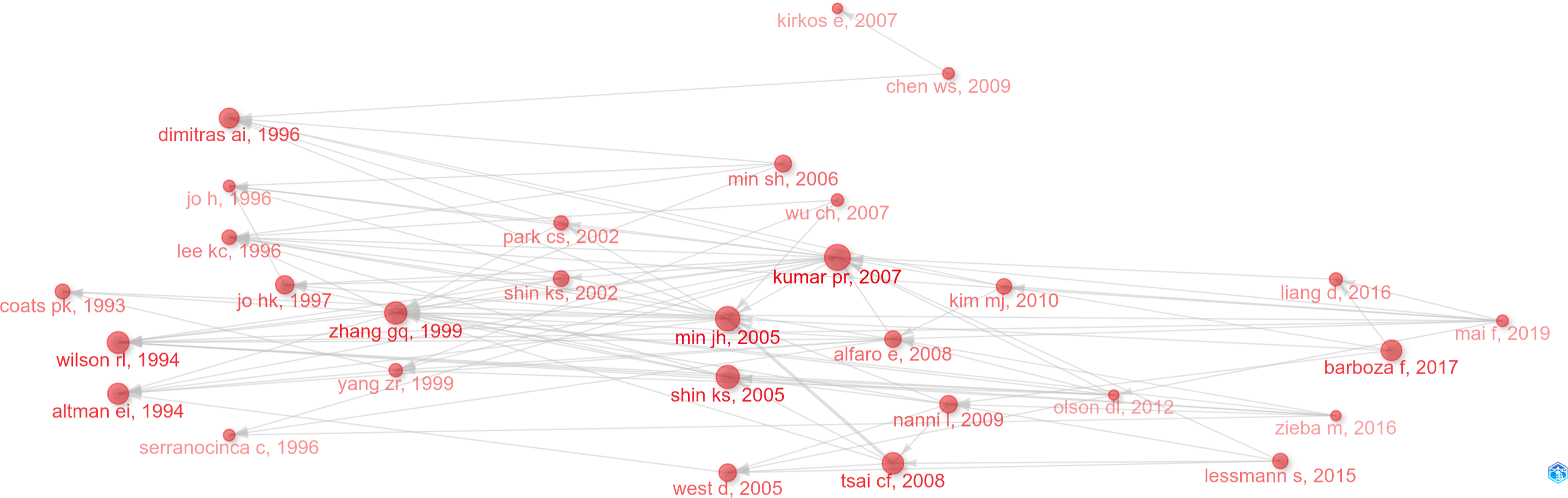}
	\caption{Historiograph. This kind of analysis, although not often seen in the literature, is a useful tool in assessing the number of different topics and their core authors and documents, within a bibliographic corpus and in a chronological sequence. In this instance, there is one topic pervading the corpus. Labels denote the author and the year of the topic document, with documents being sorted in ascending order, from left to right. In this chronological sequence core documents are related via direct citations, this relationship is presented by edges connecting nodes. These edges are directed, and therefore more appropriately called arcs, since they need to depict chronological relationship, and as such directed from right to left. The chosen number of nodes was $ 30 $, so as to have an appropriate sample for the core knowledge of the topic, with labels being in a short identification (first author, year of publication) format.}\label{fig:Historiograph}
\end{figure}

In order to present the most relevant direct citations and permeating topics, a Historiograph analysis was performed, as seen in Figure \ref{fig:Historiograph}. In spite of various different subdivisions of research, there is only one topic as per Historiograph analysis, in the research document corpus. 

It crystallized sometime around 1993, with Coats (1993), and two significant documents in 1994, Wilson, and Altman. From then onward it jumped through a number of more prominent periods, with the most relevant documents being, Dimitras (1996), Zhang (1999), Min (2005), Shin (2005), Kumar (2007), Tsai (2008), and Barboza (2017).

In order to obtain central keywords, themes for the definition of the topic, it is useful to sort historiograph documents by date, in both orders, to ascertain the beginning and what is happening at the cutting edge -- alongside which citation count in descending order is also useful, so as not to miss anything in the middle. Thus by analyzing metadata of the top 10 in every group of the three, it is possible to get the gist of the matter.

By doing the aforementioned one will observe the following aspects of note: financial distress, neural networks, bankruptcy prediction, financial diagnosis, business failure, industrial application, probabilistic approach, deep learning, machine learning, corporate governance, bench-marking, credit scoring, data mining, experimental approach, ensemble approach, statistical and intelligent techniques, support vector machine, etc.

Therefore a sentence like definition of the topic might be: Artificial intelligence in the service of entrepreneurial finance with application; almost a fitting title for this research article, and quite the resemblance with the journal Expert Systems with Applications published so many of the documents in the corpus and in the Historiograph analysis ($ 16/30 $).

\subsection{Social Structure Data Analysis}
\label{sec:SSDA}

\begin{figure}[h] 
	\centering
	\includegraphics[width=1\textwidth]{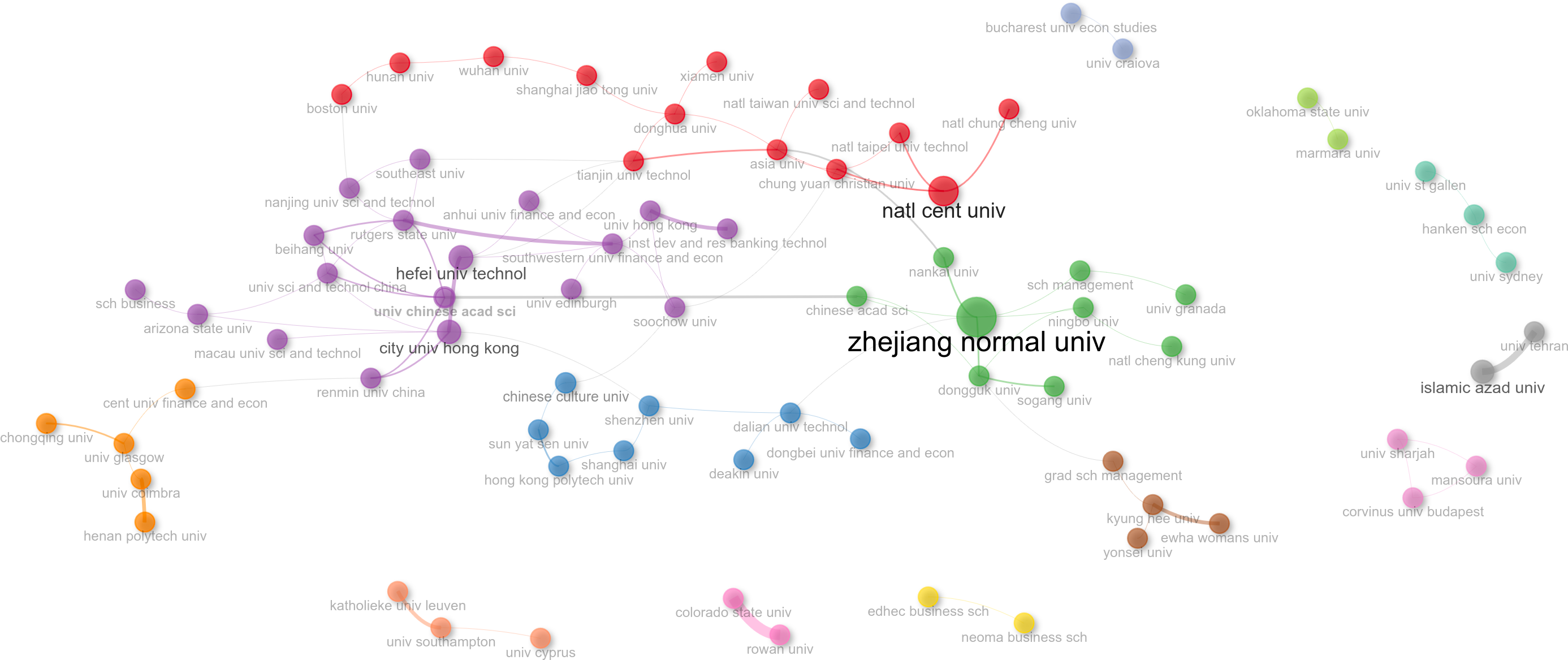}
	\caption{Collaboration Network by Institution. With this analysis, one determines clusters and strengths of collaboration's -- relatedness of items is calculated by the number of co-authored documents. The network was created for $ 100 $ nodes, so as to delve beyond just core clusters, and still make the figure useful, while removing isolated nodes and with the minimum number of edges being $ 1 $ -- this will ensure not to focus on those that are of no interest on the one side, and to consider all those that are collaborating. The clustering algorithm used was Walktrap, a random walk based algorithm known for its ability to capture with quality community structure of a network (based on the idea that short random walks belong to the same community). \cite{Pons2005,Yang2016} As for the visual matter, the colored blob represents the institution with a label denoting the institution's shortened name. The color of the blobs defines cluster, while the size of the blob informs about the institution's document output, the bigger the blob, the more documents an institution has produced. Institutions are collaborating, and this has been represented by graph edges, the thicker the edge is, the more co-authorships there are, and the stronger the link is.}\label{fig:CNI}
\end{figure}

Before we close, and take into account the whole research there is one more thing that needs to be done, and that is social structure analysis, so as to shine a light upon those that have produced the research, and this is found in Figure \ref{fig:CNI}. Institutions outputting the most documents are Zhejiang Normal University, National Central University, Hefei University of Technology, City University of Hong Kong, and Islamic Azad University -- with Harbin Institute of Technology, and Chinese Academy of Sciences not being as relevant as in previous analysis on number of documents, a discrepancy potentially caused by analyzing with different tools, which makes a case for analyzing with different tools in order to ascertain the issue more completely.

Around Zhejiang Normal University, National Central University, and the Hefei University of Technology established the biggest collaboration clusters: purple, red, and green. Which is a logical consequence of high document output. Out of all the others, the blue cluster, with Chinese Culture University, is closely following the top three. Link collaboration strength is not of the highest value in the biggest clusters, as seen from pink, and gray clusters -- indicating other factors to a collaboration link strength, aside from document output, potentially factors like national, religious, possibly social, or perhaps state or project factors, etc. 

There is also a question of the bridges between clusters, widening a collaboration further. The bridges themselves can represent a strong collaboration, e.g. Chinese Academy of Sciences and University of the Chinese Academy of Sciences (university under control of the Chinese Academy of Sciences), or Nankai University and Asia University -- indicating that for bridges of such strength perhaps a factor beyond a scientific one is present, which could possibly have a negative influence on science and the advancement thereof. It is also evident that a large number of institutions are Chinese, or with a Chinese aspect, with the bridges also being dominated by such universities, indicating Chinese relevance to collaboration networks.

If we rank institutions by PageRank, taking into account number but also prestige, approximating importance, the situation is somewhat different and is presented in Table \ref{tab:MRICN}. With such important calculation, City University of Hong Kong is at the top, with Zhejiang Normal University and Rutgers State University following, indicating the same issue as with the analysis of document output, multiple factors need to be taken into consideration if one is to ascertain the situation, and here it seems that some institutions are in a more prestigious, so to say, collaboration than others. When clusters are observed 4 (purple in Figure \ref{fig:CNI}) and 1 (red in Figure \ref{fig:CNI}) are both of frequency 3, together being $ 6/10 $ of a presence, making those clusters in demand in the PageRank eyes -- while cluster 3 (green in Figure \ref{fig:CNI}) is following with frequency of 2.

By looking more closely into bridges and shortest paths, that is betweenness of which more is presented in Table \ref{tab:MRICNB}, the top three institutions are Zhejiang Normal University, Asia University, and City University of Hong Kong, with $ 7/10 $ universities being Chinese one, confirming previous analysis of high relevance of Chinese universities, it seems that these universities are on collaboration points of influence. While Asia University, Dongguk University, and Rutgers State University are making an international presence so to speak. On the cluster side, clusters 4 (purple in Figure \ref{fig:CNI}) and 3 (green in Figure \ref{fig:CNI}) are almost completely dominant, with 4 having a frequency of 6, and 3 having a frequency of 3, indicating that these clusters are holding collaboration network afloat, and if one looks at Figure \ref{fig:CNI} this is correct.

It seems that Zhejiang Normal University, Asia University, City University of Hong Kong, Dongguk University, and Rutgers State University are those that are of most relevance generally, both in terms of collaboration influence and in terms of being at the crossroads of collaboration. In terms of the intersection of clusters, it is the purple (4 in the tables) cluster that is a constant, and it seems the most prominent of the three forces, as this is the cluster without which the collaboration network would be most disturbed -- it is a cluster comprised of a mix of international universities with strong Chinese presence, while the green cluster (3 in the tables) is tightly following. 

The aforementioned is significant because if other institutions that are on the margins of collaboration in Figure \ref{fig:CNI} want to improve on that particular point, they can either expand their own cluster on an individual basis, or link with institution/cluster of dominance and more quickly expand its reach, thus making those clusters/institutions that are of highest relevance desirable and influential.

\begin{table}[h]	
	\caption{Most Relevant Institutions by Collaboration Network PageRank (collaboration network in Figure \ref{fig:CNI})}\label{tab:MRICN}%
	\begin{tabular}{@{}lll}
		\toprule
		Rank\textsuperscript{1} & Institution & Cluster\textsuperscript{2} \\
		\midrule
		1	& \textsc{city univ hong kong} 	& 4 \\
		2	& \textsc{zhejiang normal univ}	& 3 \\
		3	& \textsc{rutgers state univ} 	& 4 \\
		4	& \textsc{asia univ} 			& 1 \\
		5	& \textsc{dongguk univ} 		& 3 \\
		6	& \textsc{southwestern univ finance and econ}	& 4 \\
		7	& \textsc{natl cent univ} 		& 1 \\
		8	& \textsc{kyung hee univ} 		& 6 \\
		9	& \textsc{tianjin univ technol}	& 1 \\
		10	& \textsc{univ southampton} 	& 10 \\
		\botrule
	\end{tabular}
	\footnotetext[]{Sorted according to PageRank (approximating importance, assuming more important nodes have more links to them from other sources while taking into account that not every link has the same weight \cite{Bianchini2005}), in decreasing order}
	\footnotetext[1]{Paired inversely (decreasing $\rightarrow$ increasing) with PageRank calculated by Bibliometrix, so as to enable ease of use and reduce complexity}
	\footnotetext[2]{Clusters from Figure \ref{fig:CNI} -- the network consists of a total of $ 14 $ clusters, paired through the institution's name}
\end{table}

\begin{table}[h]	
	\caption{Most Relevant Institutions by Collaboration Network Betweenness (collaboration network in Figure \ref{fig:CNI})}\label{tab:MRICNB}%
	\begin{tabular}{@{}lll}
		\toprule
		Rank\textsuperscript{1} & Institution & Cluster\textsuperscript{2} \\
		\midrule
		1	& \textsc{zhejiang normal univ} & 3 \\
		2	& \textsc{asia univ}			& 1 \\
		3	& \textsc{city univ hong kong} 	& 4 \\
		4	& \textsc{univ chinese acad sci}	& 4 \\
		5	& \textsc{dongguk univ} 		& 3 \\
		6	& \textsc{rutgers state univ} 	& 4 \\
		7	& \textsc{renmin univ china} 	& 4 \\
		8	& \textsc{chinese acad sci} 	& 3 \\
		9	& \textsc{southwestern univ finance and econ}	& 4 \\
		10	& \textsc{hefei univ technol} 	& 4 \\
		\botrule
	\end{tabular}
	\footnotetext[]{Sorted according to betweenness (measuring the number of times an institution is on the shortest path in between other institutions, showing network bridges \cite{White1994}), in decreasing order}
	\footnotetext[1]{Paired inversely (decreasing $\rightarrow$ increasing) with betweenness calculated by Bibliometrix, so as to enable ease of use and reduce complexity}
	\footnotetext[2]{Clusters from Figure \ref{fig:CNI} -- the network consists of a total of $ 14 $ clusters, paired through the institution's name}
\end{table}

\begin{figure}[h] 
	\centering
	\includegraphics[width=1\textwidth]{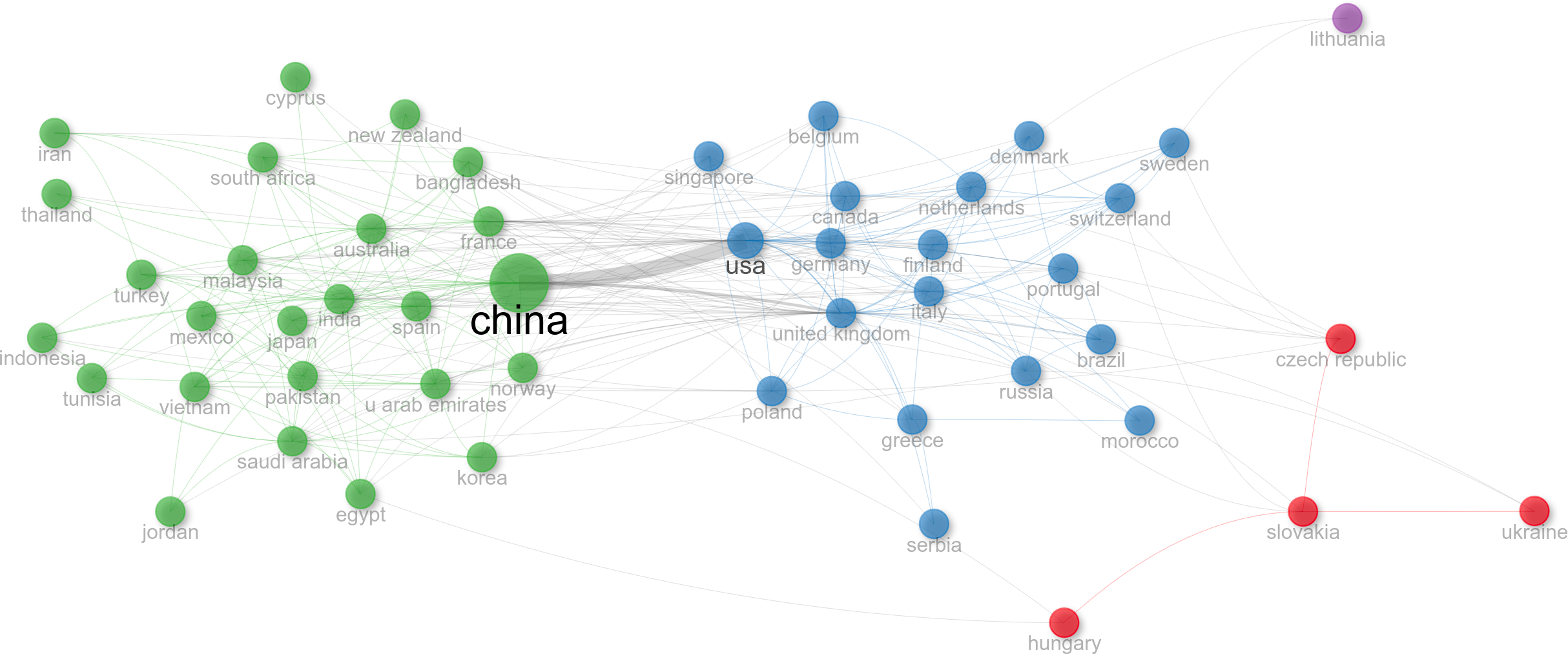}
	\caption{Collaboration Network by Country -- a subset of countries. With this analysis, one determines clusters and strength of collaboration -- relatedness of items is calculated by the number of co-authored documents. The network was created for $ 50 $ nodes, so as to firstly determine the network in a situation where the choice of belonging to a cluster is more binary,  at the same time removing isolated nodes and with the minimum number of edges being $ 1 $ -- this will ensure not to focus on those that are of no interest on the one side, and to consider all those that are collaborating. The clustering algorithm used was Walktrap, a random walk based algorithm known for its ability to capture with quality community structure of a network (based on the idea that short random walks belong to the same community). \cite{Pons2005,Yang2016} Similarly to a collaboration network for institutions, a colored blob represents a country with a label denoting the name. The color of the blob defines cluster, while the size of the blob informs about country document output, the bigger the blob, the more documents a country has produced. Countries are collaborating, and this has been represented by graph edges, the thicker the edge is, the more co-authorships there are, and the stronger the link is.}\label{fig:CNC}
\end{figure}

The next analysis is by country, presented in Figure \ref{fig:CNC}. This analysis was performed on a smaller number of countries, 50 to be exact, as smaller analyses can give insight that a larger picture cannot. By looking only at those most prominent, one can ascertain the relationships between those elements only, without other items interfering. Therefore by comparing smaller and larger networks, we can determine how strongly a country is in a particular cluster, and what are countries leanings, while also determining the most influential players with a small network, and general collaboration with a large network.

By observing Figure \ref{fig:CNC} it is obvious that there are two big collaboration clusters, green and blue, and that there are countries around which it would seem much of the collaboration revolves, mainly China and the United States of America. Therefore on this level of detail, in a situation where these countries are involved, collaborations will most likely behave in accord with the presented manner.

It is also clear that collaboration between the green and blue clusters is substantial, with many paths for jumping from one cluster to the other, there is no country without which collaboration would be severely diminished, as one might go the other route and establish collaboration that way. Indeed there is an elephant in the room that can't be ignored, namely collaboration between China and the USA, which is so strong that it dwarfs every other collaboration in comparison -- indicating an enormous amount of cooperation between these two countries, regardless of the involvement of other countries.

If one looks at the clusters and countries' geographical location, and perhaps even geopolitical aspects, it seems that the network resembles the divide between East and the West (there is a significant presence of the West in the green cluster also), with Russia surprisingly being in the blue cluster -- which might change over time? Is it possible that collaboration network by country has geopolitical significance? Could such an analysis perhaps be a tool in a prediction of future events, or is a scientific collaboration only a result of the events of the past, and what influence does such an element have on science?

In addition to China's and the USA's clusters, we have the issue of Lithuania, whose collaboration was not of a nature where the country would belong to a larger cluster, and therefore sits alone with collaboration to Denmark and Sweden, indicating a geographical link, and collaboration of a lower intensity. The last cluster of the network is the red one, with four countries quite close to one another in terms of geography, with Slovakia interestingly enough being the center of the cluster. This cluster is linked mostly to the blue cluster, with a number of connections to the green one also -- an extension of sorts it seems to the blue cluster trying to make collaborations with others as well.

By turning to PageRank, taking into account not only the number of links but the relevance of the sources as well, presented in Table \ref{tab:MRCCNP}, aligning situation follows, with China and the USA, followed by the United Kingdom, being in the top three, respectively -- with the United Kingdom having the high number of collaborations within and without its own blue cluster, an indication perhaps of the history and the Commonwealth. These are the countries that are as per PageRank involved in most so to say prestigious collaborations, and are thus a desirable collaboration unit of interest with potentially other benefits.

The number one cluster is 3 (green in Figure \ref{fig:CNC}), with the next two positions belonging to cluster 2 (blue in Figure \ref{fig:CNC}), however afterward there is a series of cluster 3 appearances starting with India following immediately after United Kingdom, with cluster 3 making $ 6/10 $ of a presence. It is therefore cluster 3 that is generally it would seem more influential, yet cluster 2 has $ 2/3 $ of a presence in the top three, an indication of the strength of both clusters, the influence of which is perhaps derived through disparate methods.

Before we present the global network of countries, a look at betweenness is warranted, presented in Table \ref{tab:MRCCNB}, and especially so as here it is more difficult to ascertain link-bridges via network analysis. Unsurprisingly United Kingdom is the top country here, positioning itself between the greats, likely a consequence of a Commonwealth and present environment. With China and the USA following, Italy and Spain not making the list, while the Netherlands and United Arab Emirates are significant it would seem in terms of presenting themselves as a bridge for collaboration -- some are therefore in a good company but not as central, while others are not as influential but are central for collaboration fostering.

The overview now more clearly shows the dominant cluster, and it is the green one, number 3, with $ 6/10 $ of a presence, yet in PageRank analysis of Table \ref{tab:MRCCNP} cluster 2 is strong in rank, with placements 1 (United Kingdom) and 3 (USA) being of its set of countries, while cluster 3 is by betweenness more at the back than before. It seems that cluster 3 has strength in numbers so to speak, while cluster 2 has strength in top positions, and so they are collaborating together, which seems logical for the reason of science, necessity, competition, etc. 

Western countries are dominant, as in Table \ref{tab:MRCCNP} there are 7 of them and in Table \ref{tab:MRCCNB} there are 6 of them. This presence can also be seen in Figure \ref{fig:CNC} by observing the whole picture, an indication of past success and present circumstances, and no doubt a result of collaboration positioning. The question however is, what the future holds, is leading of the West still forthcoming, or is the trend changing, partially or more fully? What we can gather from this research is that future collaboration will be of a more multilateral nature.

\begin{table}[h]	
	\caption{Most Relevant Countries by Collaboration Network PageRank (collaboration network in Figure \ref{fig:CNC})}\label{tab:MRCCNP}%
	\begin{tabular}{@{}p{2cm}p{3cm}p{2cm}}
		\toprule
		Rank\textsuperscript{1} & Country & Cluster\textsuperscript{2} \\
		\midrule
		1	& \textsc{china} 	& 3 \\
		2	& \textsc{usa}		& 2 \\
		3	& \textsc{united kingdom} 	& 2 \\
		4	& \textsc{india} 	& 3 \\
		5	& \textsc{france} 	& 3 \\
		6	& \textsc{australia}		& 3 \\
		7	& \textsc{spain} 	& 3 \\
		8	& \textsc{saudi arabia} 	& 3 \\
		9	& \textsc{italy}	& 2 \\
		10	& \textsc{canada} 	& 2 \\
		\botrule
	\end{tabular}
	\footnotetext[]{Sorted according to PageRank (approximating importance, assuming more important nodes have more links to them from other sources while taking into account that not every link has the same weight \cite{Bianchini2005}), in decreasing order}
	\footnotetext[1]{Paired inversely (decreasing $\rightarrow$ increasing) with PageRank calculated by Bibliometrix, so as to enable ease of use and reduce complexity}
	\footnotetext[2]{Clusters from Figure \ref{fig:CNC} -- the network consists in total of $ 4 $ clusters, paired through country name (USA denoting the United States of America)}
\end{table}

\begin{table}[h]	
	\caption{Most Relevant Countries by Collaboration Network Betweenness (collaboration network in Figure \ref{fig:CNC})}\label{tab:MRCCNB}%
	\begin{tabular}{@{}p{2cm}p{3cm}p{2cm}}
		\toprule
		Rank\textsuperscript{1} & Country & Cluster\textsuperscript{2} \\
		\midrule
		1	& \textsc{united kingdom} 	& 2 \\
		2	& \textsc{china}			& 3 \\
		3	& \textsc{usa} 				& 2 \\
		4	& \textsc{france}			& 3 \\
		5	& \textsc{india} 			& 3 \\
		6	& \textsc{australia} 		& 3 \\
		7	& \textsc{netherlands} 		& 2 \\
		8	& \textsc{u arab emirates} 	& 3 \\
		9	& \textsc{canada}			& 2 \\
		10	& \textsc{saudi arabia} 	& 3 \\
		\botrule
	\end{tabular}
	\footnotetext[]{Sorted according to betweenness (measuring the number of times an institution is on the shortest path in between other institutions, showing network bridges \cite{White1994}), in decreasing order}
	\footnotetext[1]{Paired inversely (decreasing $\rightarrow$ increasing) with betweenness calculated by Bibliometrix, so as to enable ease of use and reduce complexity}
	\footnotetext[2]{Clusters from Figure \ref{fig:CNC} -- the network consists in total of $ 4 $ clusters, paired through country name (USA denoting United States of America, and U ARAB EMIRATES denoting United Arab Emirates)}
\end{table}

\begin{figure}[h] 
	\centering
	\includegraphics[width=1\textwidth]{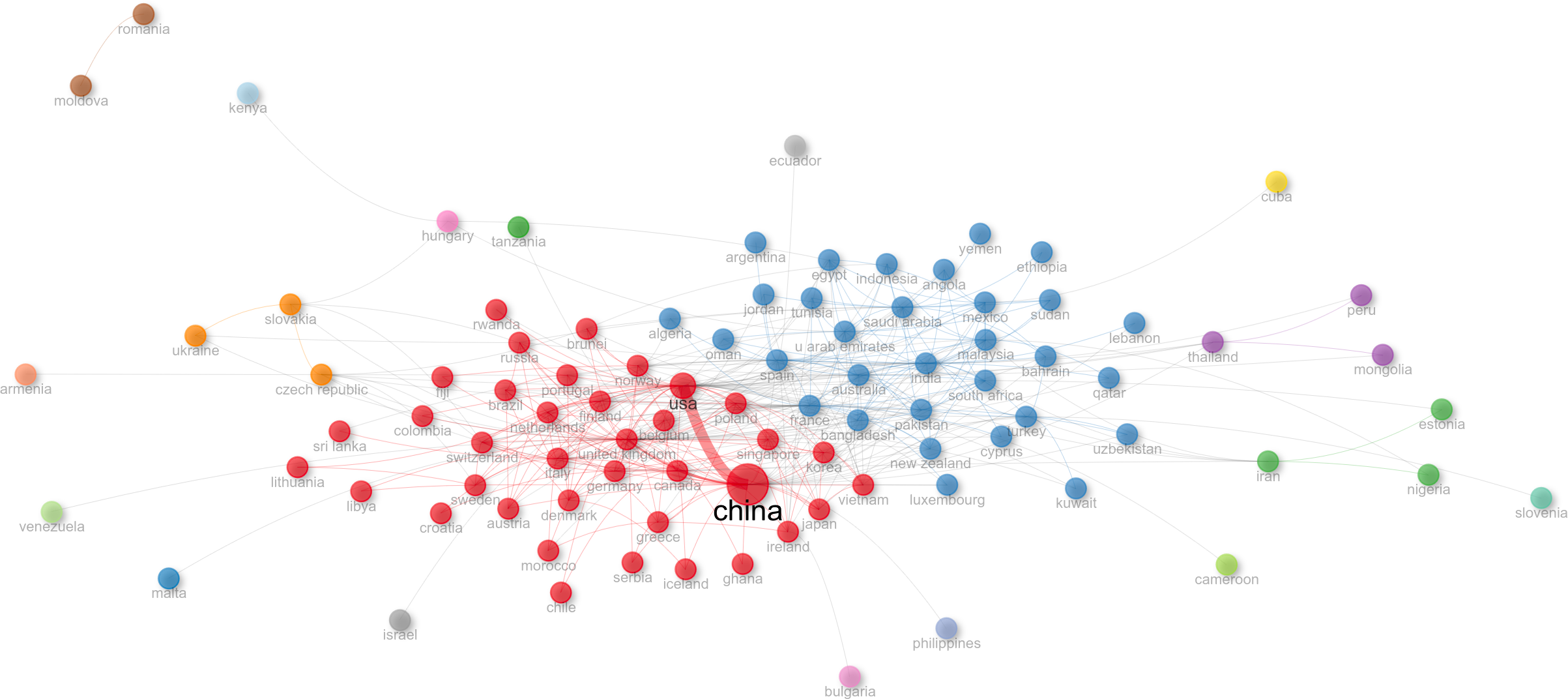}
	\caption{Collaboration Network by Country -- all countries. With this analysis, one determines clusters and strength of collaboration -- relatedness of items is calculated by the number of co-authored documents. A network was created for all the nodes with the following setup, so as to generate a general picture, removing isolated nodes and with the minimum number of edges being $ 1 $, this will ensure no focus on those that are of no interest on the one side, and to consider all those that are collaborating ($ 92 $ countries in total as per Bibliometrix calculation). The clustering algorithm used was Walktrap, a random walk based algorithm known for its ability to capture with quality community structure of a network (based on the idea that short random walks belong to the same community). \cite{Pons2005,Yang2016} Colored blob represents a country with a label denoting the name. The color of the blob defines cluster, while the size of the blob informs about country document output, the bigger the blob, the more documents a country has produced. Countries are collaborating, and this has been represented by graph edges, the thicker the edge is, the more co-authorships there are, and the stronger the link is.}\label{fig:CNCall}
\end{figure}

The following analysis is the final analysis of the paper, presented in Figure \ref{fig:CNCall}, Table \ref{tab:MRCCNPsve} and Table \ref{tab:MRCCNBsve}, and builds upon the previous analysis with a broader context in mind -- here we analyze all the countries, 92 countries in total, and the entire collaboration network, with the results being especially interesting when compared to the analysis conducted on 50 nodes.

There are still two overly dominant clusters of collaboration, blue and red, with no other cluster being larger than three, indicating a generally large amount of collaboration divided primarily into two camps -- and if there are other factors to this division it is difficult to state, yet geographical, and most like geopolitical, factors are obvious.

Comparing those two most prominent clusters, the red cluster with USA and China is larger, with 37 countries in total, while the blue cluster with India has 31 countries in total, at least in part a consequence of countries repositioning, from the previous analysis those would primarily be China, USA, and the United Kingdom -- with substantial collaboration between these three clearly seen.

The global collaboration differs as compared to local, that is as compared only to those countries of the highest weight, presented in Figure \ref{fig:CNC}. By trying to map clusters, the blue cluster from the previous analysis would correspond to the red, while the green cluster would correspond to the blue one here, and now by analyzing the changes we are coming to some striking realizations, with some being more easily explainable than others.

For example, China has made a move, from the blue (green in the previous analysis) cluster to the red (blue in the previous analysis) cluster, and is now in the same cluster as the USA, a move to the cluster of the USA and the United Kingdom, which would indicate a difference in local policy as compared to global policy. The same change was made by Korea and Japan as well, where the change could be perhaps explained by geopolitical reasons. There are however changes where additional information would be needed, as in the example of Norway, although it could be argued that Norway, together with some other countries, was at the edge, and so when additional nodes entered calculation, the country was more closely aligned in terms of collaboration with the other cluster.

Cooperation between these two largest clusters is strong, especially with China, the USA, and the United Kingdom on the one side, and with France, India, and the United Arab Emirates on the other, etc. Aside from the cluster of Moldova and Romania, the collaboration network is it seems strongly connected, with collaboration being easily accessible. 

Both clusters have a number of co-clusters on their side of the network, with the leaning being towards the red cluster, while Hungary is for example even more dislocated than in the previous analysis, standing alone. There is also an obvious extreme of Israel, standing behind the red cluster and linked to it by the collaboration with the USA, the only link this country has in this discipline and corpus of knowledge -- with the overtone of using such analysis for ascertaining current state of affairs and possibly future events in the air, as it seems that science which should be independent in the search for the truth is showing signs of other influences, which need not be necessarily of a negative nature, yet there could be a negative influence.

The importance of collaborating countries ascertained by PageRank, as presented in Table \ref{tab:MRCCNPsve}, reveals almost identical results as in the previous analysis of Table \ref{tab:MRCCNP}, with Australia's somewhat diminished importance as per widening the scope, while Spain and Saudi Arabia have climbed the ladder. On the cluster side of things, both clusters, speaking of those largest ones naturally, are equal in presence, as China has changed the collaboration cluster, with cluster 1 (red in Figure \ref{fig:CNCall}) holding the top three positions -- an equilibrium of sorts, with the red cluster holding the high ground, while the blue cluster (with identification 2) following in a series.

By bringing link-bridges into the foreground with betweenness calculation of Table \ref{tab:MRCCNBsve}, the situation is similar, but with the Netherlands and Canada not making the list this time, at least in the top 10, while Spain and Italy have shown themselves as highly relevant in terms of centrality -- as compared to Table \ref{tab:MRCCNB}. The top three countries are the same, with this time all belonging to the cluster of the United Kingdom, red cluster in Figure \ref{fig:CNCall} with identification 1 in the Table \ref{tab:MRCCNBsve}, thus further improving the collaboration strength of the blue cluster from the previous analysis. 

The frequency of clusters is the same for both previous and current analyses, with a change happening in the positioning of clusters, and this time United Kingdom's cluster holds the front and back of the queue, while India's cluster has taken command of the middle. As difficult as it is to make a conclusion about the more relevant countries and clusters, the data shows that the United Kingdom's cluster always commands top positions in terms of both PageRank relevance and centrality, while the other hand, while cluster second to that one is strong in terms of collaboration, yet it is always around the middle of the list of top 10 -- consequently indicating that whether close or far, the blue cluster in Figure \ref{fig:CNC} which approximately maps to the red cluster in Figure \ref{fig:CNCall} represents a cluster that is more relevant as per document output, importance, and centrality to the collaboration network -- in essence, there is a reason why this is so, collaboration is a result of activities preceding, with which we have dealt with in other sections of the paper.

\begin{table}[h]	
	\caption{Most Relevant Countries by Collaboration Network PageRank (collaboration network in Figure \ref{fig:CNCall}, all countries included)}\label{tab:MRCCNPsve}%
	\begin{tabular}{@{}p{2cm}p{3cm}p{2cm}}
		\toprule
		Rank\textsuperscript{1} & Country & Cluster\textsuperscript{2} \\
		\midrule
		1	& \textsc{china} 	& 1 \\
		2	& \textsc{usa}		& 1 \\
		3	& \textsc{united kingdom} 	& 1 \\
		4	& \textsc{india} 	& 2 \\
		5	& \textsc{france} 	& 2 \\
		6	& \textsc{spain}	& 2 \\
		7	& \textsc{saudi arabia} 	& 2 \\
		8	& \textsc{australia}		& 2 \\
		9	& \textsc{italy}	& 1 \\
		10	& \textsc{canada} 	& 1 \\
		\botrule
	\end{tabular}
	\footnotetext[]{Sorted according to PageRank (approximating importance, assuming more important nodes have more links to them from other sources while taking into account that not every link has the same weight \cite{Bianchini2005}), in decreasing order}
	\footnotetext[1]{Paired inversely (decreasing $\rightarrow$ increasing) with PageRank calculated by Bibliometrix, so as to enable ease of use and reduce complexity}
	\footnotetext[2]{Clusters from Figure \ref{fig:CNCall} -- the network consists of a total of $ 19 $ clusters, paired through country name (USA denoting the United States of America)}
\end{table}

\begin{table}[h]	
	\caption{Most Relevant Countries by Collaboration Network Betweenness (collaboration network in Figure \ref{fig:CNCall}, all countries included)}\label{tab:MRCCNBsve}%
	\begin{tabular}{@{}p{2cm}p{3cm}p{2cm}}
		\toprule
		Rank\textsuperscript{1} & Country & Cluster\textsuperscript{2} \\
		\midrule
		1	& \textsc{united kingdom} 	& 1 \\
		2	& \textsc{china}			& 1 \\
		3	& \textsc{usa} 				& 1 \\
		4	& \textsc{india}			& 2 \\
		5	& \textsc{spain} 			& 2 \\
		6	& \textsc{france} 			& 2 \\
		7	& \textsc{saudi arabia} 	& 2 \\
		8	& \textsc{australia} 		& 2 \\
		9	& \textsc{u arab emirates}	& 2 \\
		10	& \textsc{italy} 			& 1 \\
		\botrule
	\end{tabular}
	\footnotetext[]{Sorted according to betweenness (measuring the number of times an institution is on the shortest path in between other institutions, showing network bridges \cite{White1994}), in decreasing order}
	\footnotetext[1]{Paired inversely (decreasing $\rightarrow$ increasing) with betweenness calculated by Bibliometrix, so as to enable ease of use and reduce complexity}
	\footnotetext[2]{Clusters from Figure \ref{fig:CNCall} -- the network consists of a total of $ 19 $ clusters, paired through country name (the USA denoting the United States of America, and U ARAB EMIRATES denoting the United Arab Emirates)}
\end{table}

\section{From Bibliometrics to Showing Intelligent Behavior}
\label{sec:mc}

Can machines present intelligent behavior? A thesis postulated by Turing in the 1950s, as opposed to whether there is an actual intelligence, real mind, a consciousness \cite{Turing1950} -- which are questions for philosophy and theology, are extremely difficult to determine, and impossible to prove. However if one is dealing only with inputs and outputs, and has a focus on the result, the matter at hand is much more lighter, although difficult nonetheless.

By looking into AI and economics, the situation found is somewhat dire, although not unexpected, as the interdisciplinary link between new and innovative technology, and a field not accustomed to such rapid change, is difficult to make. The literature is filled with AI terms, and depending on the sub-field, algorithms implementing various techniques and methods, there is nonetheless a lack of empirical research into these, so as to critically appraise how well these financial technologies compare in terms of expected output. \cite{Biju2023} The Most pressing issues are pitfalls of Machine Learning and AI in the process of prediction for the reason of biases -- happening "mostly in the areas of insurance, credit scoring and mortgages." \cite{Biju2023} There is therefore a need for a turnaround, or a rethink, of how to employ these new technologies that are reshaping finance in a dramatic way. \cite{Biju2023}

Technology has actually always rapidly advanced in innovation from the dawn of digital computing in the 1940s \cite{Dyson2012}, and it seems that the speed of that innovation/knowledge discovery is increasing -- as either new technology or new knowledge discovered, often by that technology, is almost ubiquitous. Computer experts are accustomed to that, but rarely does a field of research embroil in these conditions. It is sometimes said that computing has not yet matured, as it constantly evolves at such speed, however after over $ 80 $ years of progress, whether in software or hardware, that argument is more difficult to make, as it seems that it is a feature of a field being uplifted not only by the influence from inside the field itself, but often from the outside as well -- with the latest great leap being quantum computing, brought on the map by interference of quantum mechanics. \cite{Dyson2012,Brunette2009,Gyongyosi2019}

The foundational paradigm of AI is randomization, with its weights, probabilities, and outputs. On that foundation other ideas are grafted, thus making the advance in algorithms, techniques, and methods. Whether evolutionary computation, machine learning, or some other sub-field of AI, randomization is hard to avoid, as it is that factor that brings dynamic behavior and makes learning, pattern recognition, and greater adaptability possible. Beyond AI, randomization made possible two most widely known breakthroughs in quantum computing, Grover's search algorithm \cite{Brickman2005} and Shor's factorization algorithm \cite{Monz2016}, only bringing randomized algorithms to the center stage more than ever. If quantum computers are to be a reality, it seems that randomized algorithms will be a force to wield.

Therefore in order to tackle the aforementioned problems of AI and finance, with the benefit not only for the wider economic field, but also for all interested parties, whether a computer expert, or someone entirely from another field, to bridge that gap, and further application of artificial intelligence in entrepreneurship as well, here we will review and establish this foundational paradigm in a concrete way. The most appropriate algorithm then to focus on is a Monte Carlo randomized algorithm. This type of algorithm corresponds well to the uncertain nature of difficult problems, with the added benefit of securing confidence in the solution.

Monte Carlo algorithm represents a series of steps, and in a full sense an algorithm, that can in polynomial time, and with arbitrary probability, find an optimal solution for a given problem -- such probability never reaches $ 1 $, but as we get ever closer to one, the confidence that optimal solution has been found grows. \cite{Kudelic2023}

By generalizing on the algorithm, we are coming to a paradigm of thought, a method for the creation of the Monte Carlo algorithm, whose implementation will depend on the problem being solved, and the steps are as follows.

\begin{enumerate}[I]
	\item An algorithm needs to work on the data, and that data needs to be organized, therefore we need to choose an appropriate data structure so as to tackle the problem.
	
	\item Decide on the distribution through which you will obtain the desired solution and arbitrary confidence thereof. This distribution is used for generating random numbers, with which one makes decisions. The distribution typically used is uniform, as it often conforms to the problem well.
	
	\item As per the chosen distribution, calculate the probability that a non-optimal solution has been achieved if an algorithm has been run only once.
	
	\item As per probability calculated for one run of the algorithm, calculate confidence that the optimal solution was output by an algorithm for $ n $ runs.
\end{enumerate}

The question is, how would one go about implementing such a procedure, and the best way to demonstrate this is to show an example of the algorithm in existence. There is a "quintessential problem of algorithmics and, more generally, of computer science", namely Minimum Feedback Arc Set (MFAS). \cite{Kudelic2022} The problem has many applications, in hardware design, machine learning, deadlock prevention, and cell apoptosis, just to name a few, and in spite of the importance and many attempts to find efficient and always optimal algorithm, this was not achieved. \cite{Kudelic2022} The problem is NP-complete, NP-hard, and APX-hard, with the definition of the problem as follows: "for a given
directed graph $ G = (V, A) $, find the smallest subset $ A'\subset A $ such that $ G' = (V, A\diagdown A') $ is acyclic." \cite{Kudelic2022}

\begin{figure}[h] 
	\centering
	\includegraphics[width=.8\textwidth]{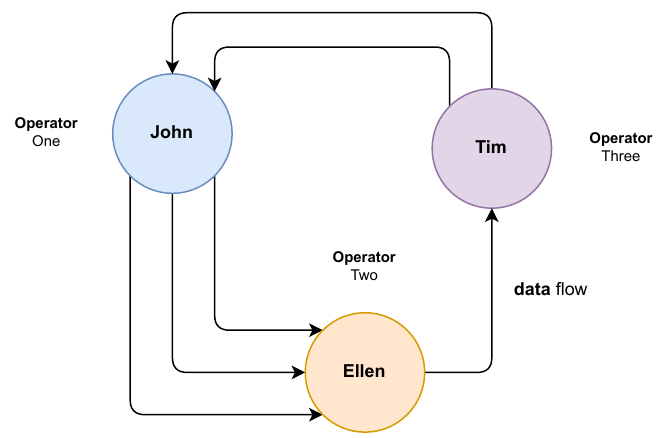}
	\caption{Illustrative Example of the Minimum Feedback Arc Set problem -- communication between three parties. Each person prepares his information based on the information received. The question however is, how to start communication by causing a minimal amount of upset in the data flow.}\label{fig:MFASprim}
\end{figure}

Let us observe Figure \ref{fig:MFASprim} and illustrate the issue of MFAS. In this figure, we have three people: John, Ellen and Tim. They are friends, and one day, they met, cordially greeted each other and struck up a conversation. After a while, when they exchanged enough information, it was decided that when they got home, they would use a landline to continue this communication, however, this communication would not proceed at will. John will communicate to Ellen three messages, but only when Tim sends him two. Ellen will communicate to Tim one message, but only when John sends her three. Tim will communicate to John two messages, but only when Ellen sends him one message. They agreed all was well, and then they said goodbye to each other.

When they came home they were ready to continue communication, but nothing was happening. Soon they all realized that a mistake had been made in designing the protocol for communication, if things stay as they are, there will be no communication at all, never. In order to transmit their messages, every one of them first needs to receive information, but that will never happen, as they are in a deadlock, waiting for each other. As communication would not be happening, they all went out of their homes, to the previous meeting place. When they met, they again greeted each other, with a bit of a laugh, and a question, what now? Tim said, nothing, this protocol is of no use, we need something that we will use. So they were thinking, discussing, and suggesting new ways of communicating, but no good idea struck. After a while, John said, wait, the old protocol is not that bad, and the problem of starting our communication can be resolved -- Ellen and Tim were all ears -- to which John continued, all we need to do is for Tim to start our communication, and everything else will fall into place. Ellen was curious and asked why was that John. Because in such a way we will retain our communication and make the least amount of upset to our communication network, John replied. Tim and Ellen were amazed at John's solution to the problem, they all agreed, yes, this is it, let's go home, and after goodbyes, they did.

If we again look at Figure \ref{fig:MFASprim} we can clearly observe that John's solution to the problem is the right way to solve the conundrum. As there is only one message that will be missed in the data flow at the beginning of their communication, any other solution would lose substantially more, and the communication channel would suffer to a greater extent. So this is the optimal solution to their problem, with the least amount of upset to the network. This same situation can be observed in many areas of science and the world we live in, which is strongly interconnected.

\begin{figure}[h] 
	\centering
	\includegraphics[width=.8\textwidth]{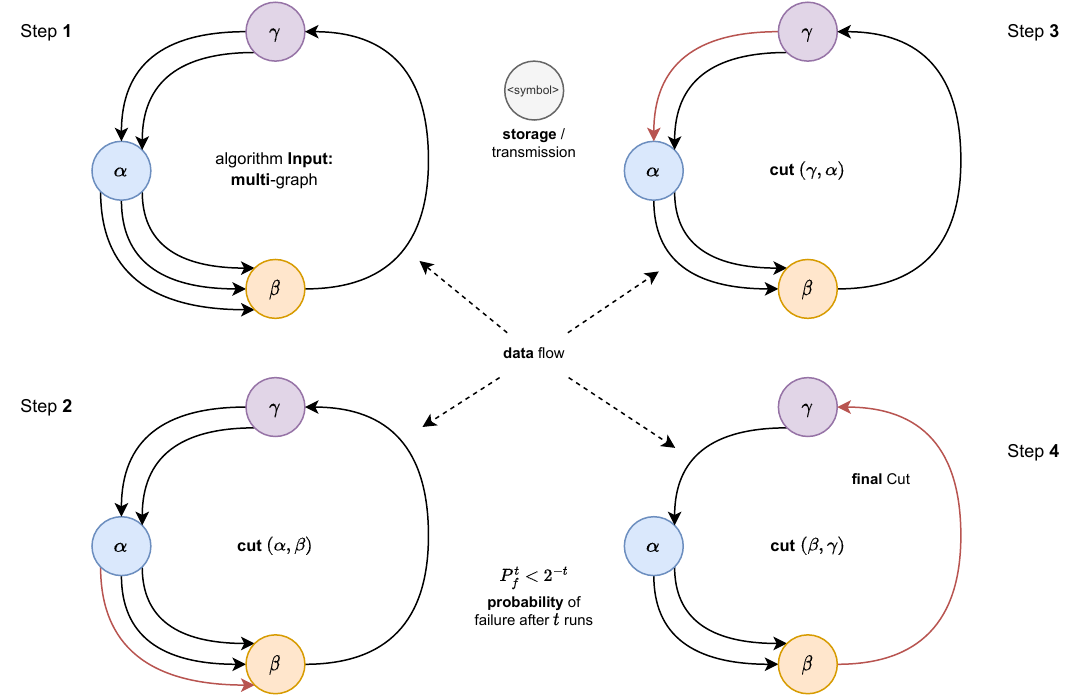}
	\caption{Monte Carlo for the Minimum Feedback Arc Set -- one algorithm run for illustrative example in Figure \ref{fig:MFASprim}. The input for the algorithm is the multi-graph. The algorithm chooses and breaks arcs in a uniform fashion until arcs have been broken and the graph is made acyclic.}\label{fig:MFASalgos}
\end{figure}

While dealing with complex systems it is clearly not an option to look at the network and determine manually what would be the best solution, nor would it be a good option to try to aimlessly guess what to do. Fortunately, this problem can be solved algorithmically \cite{Kudelic2023a} and then executed on a computing machine -- the process can be visually observed in Figure \ref{fig:MFASalgos}.

In order for the Monte Carlo algorithm for MFAS to work, input for the algorithm needs to be a multi-graph\footnote{A graph that "allows multiple arcs between every pair of nodes, has no loops, and has no arc weights" is a multi-graph \cite{Kudelic2022}}, and it will soon be clear why that is. On that multi-graph the algorithm is breaking arcs by choosing them as per uniform distribution, which means that the probability of breaking a set of arcs between any pair of nodes, $ A^{'}_{ij} \subseteq A' \subset A | i\neq j $, is inversely proportional to the probability that an arc from that set will be chosen. If one looks at step 2 of Figure \ref{fig:MFASalgos}, the highest probability of being chosen has the arc set $ \{\alpha, \beta\} $, however we are choosing arcs uniformly, therefore in the next iteration, step 3, an arc from the set $ \{\gamma, \alpha\} $, will be chosen, upon which we will again choose an arc, $ a_{ij} \subseteq A^{'}_{ij} | i\neq j $, and this time it will be from the set $ \{\beta, \gamma\} $. 

So what would happen if we were not only choosing these arcs but also breaking them, or reversing them? As it can be seen from the figure, in step 4, by making a cut $ (\beta, \gamma) $ we have broken the arc set, in this instance with only one arc, without which our graph has become acyclic, and this is exactly what we wanted. Even though this arc is in a minority when we consider other sets, and this is an idealized case, nevertheless because we are breaking arcs uniformly, every individual arc has the same probability of being broken, and therefore probability of breaking a cycle is the greatest at places where the number of arcs is minimal, just at the right spot.

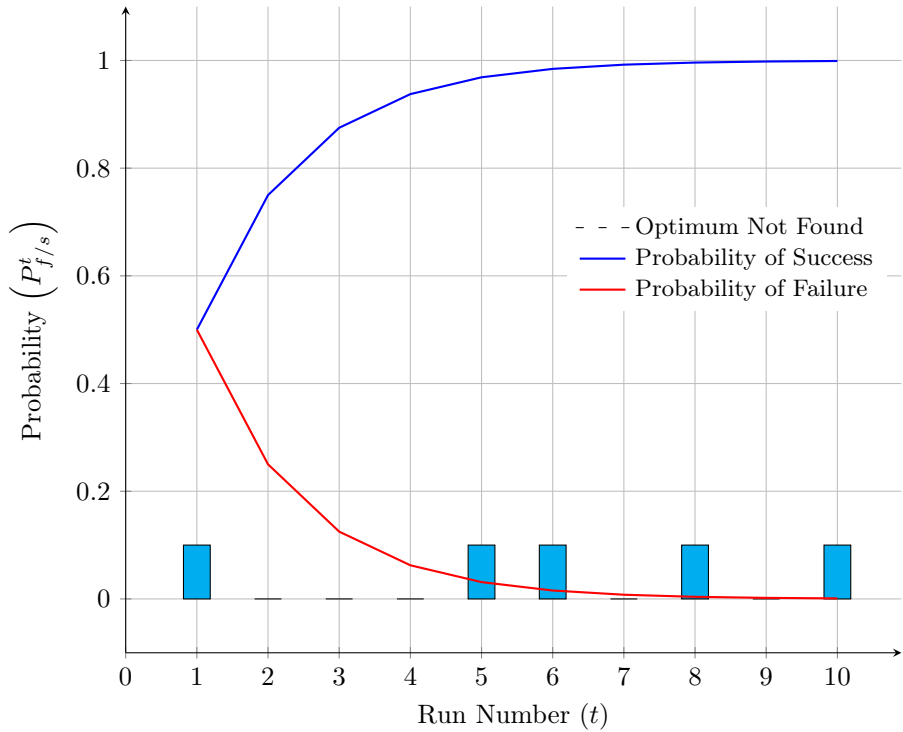
\begin{figure}[h]
	\centering
	\begin{tikzpicture}[]
		\begin{axis}[
			ylabel near ticks,
			xlabel near ticks,
			xmin=0, xmax=10.9,
			ymin=-.1, ymax=1.1,
			xlabel={Run Number ($ t $)},
			ylabel={Probability $\left(P^{t}_{f/s}\right)$},
			grid=major,
			width=.9\textwidth,
			axis lines=left,
			legend style={at={(.775,.691)},
				anchor=north,legend columns=1,legend cell align={left},draw=none},
			]
			\addlegendentry{\scriptsize Optimum Not Found}
			\addplot[ybar, color=black, fill=cyan] coordinates {(1, .1) (2, 0) (3, 0) (4, 0) (5, .1) (6, .1) (7, 0) (8, .1) (9, 0) (10, .1)};
			\addlegendentry{\scriptsize Probability of Success}
			\addplot[thick, color=blue] table[x=BrojIzvrsavanja, y=VjerojatnostOptimuma, col sep=comma]{podaci/MCalgos.csv};\label{mc:S}
			\addlegendentry{\scriptsize Probability of Failure}
			\addplot[thick, color=red] table[x=BrojIzvrsavanja, y=VjerojatnostPogreske, col sep=comma]{podaci/MCalgos.csv};\label{mc:F}
		\end{axis}
	\end{tikzpicture}
	\caption{Cumulative Probability of Failure/Success After $ t $ Runs. This figure succinctly depicts $P^{t}_{f/s}$ growth, and inverse proportionality thereof -- Monte Carlo algorithm probability parabola. Cyan colored bar graph shows when the algorithm produced the optimal solution, as the probability on individual run is $ \frac{1}{2} $ expected number of optimal solutions cumulatively needs to be around that value, as is the case here -- in this way it is possible to verify whether the algorithm functions as designed, for inputs with known solutions. If the bar is raised, optimum is found, no optimum otherwise.}\label{fig:PMC}
\end{figure}

When the algorithm arrives at that final cut and has broken all cycles, it would be of high use if one knew how good of a solution the Monte Carlo algorithm has output -- and fortunately, this is one of the strengths of Monte Carlo algorithms. This truly can be done, and it is done by calculating probabilities. In this particular instance, dealing with the Monte Carlo algorithm for minimum Feedback Arc Set, the probability of the algorithm producing a suboptimal solution is $ P^{t}_{f} < 2^{-t} $ \cite{Kudelic2019}, where $ t $ represents the number of times the algorithm has been run. Therefore after some $ t $ times probability that at least one of the solutions output by the algorithm is an optimal one is, classified as success, $ P^{t}_{s} \geq 1 - 2^{-t} $. For a visual representation of how the probability of success quickly grows as the algorithm is repeatedly run one can consult Figure \ref{fig:PMC}.

In a real-world situation, convergence towards an optimal solution might be a bit more difficult, if for no other reason, than because of the quality of the random number generator, and the dynamic nature of finding a solution, however optimal or close to the optimal solution will be found, it's a matter of probability. This is pure Monte Carlo\footnote{If one is interested in various types of Monte Carlo, complexity classes, open questions etc., then further reading would be \cite{Kudelic2023}, while for the problem of FAS one can consult monograph on the issue with three main chapters: 1) dealing with everything important FAS \cite{Kudelic2022a}, 2) dealing with FAS algorithms \cite{Kudelic2022b}, and 3) dealing with possible avenues for achieving result in different scenarios \cite{Kudelic2022c}}, and for pseudocode one can look into \cite{Kudelic2016}, with a more easier to grasp version being in \cite{Kudelic2023a}.

Yet the algorithm can be modified, and even improved, in convergence towards optimal solution. By borrowing the idea from Ant Colony Optimization \cite{Dorigo2006}, and implementing a learning mechanism, it is possible to further navigate in a more efficient manner towards a solution, and by applying probability calculation vertically (on the first run in every series of runs), instead of horizontally (throughout the series of runs) it is possible to retain confidence in a solution measure. \cite{Kudelic2019} What brings us to all the marvels of AI -- a different approach, but in a sense the same, an extension of the same idea, how to make algorithms navigate themselves. Artificial Intelligence brings this to a new level, with algorithm structure resembling nature, and with far more dynamic computation, various kernels, with ever-changing paths towards the desired goal, an approximation algorithm in its essence. \cite{Wang2003,Asteris2019,Zhang2019,Bubeck2023}

As conducted bibliometric analysis shows, AI is entering in, Finance and Entrepreneurship are being changed, and from the looks of it, this trend will continue to an even greater degree. The question of how to, and where to, implement these technologies is not an easy one, especially more so as society is not just numbers, it's a very complex environment, and if experts do not find a way of aligning with the essence of society and life, merging of AI and other areas will not work, as AI needs to be controlled, in the service of, and not being in control. On top of that, this technology is not simple, and it takes substantial effort to achieve a system that satisfies algorithmic, data, interface, user, societal, and legal requirements. In order to put a brick in that wall, and repair the breach exposed in \cite{Biju2023}, this short review is given, and with a valiant effort, via the collaboration of experts from disparate fields, the results should be achieved.

\section{Discussion and Research Implications}\label{sec4}

As a conclusion of the paper, here we will discuss the most important results, and link those results to articulated implications for parties of interest. Additionally to the aforementioned, research constraints will also be revealed and commented upon, while the section will be concluded with recommendations for improving Bibliometrics, and directions for future research.

With this bibliometric analysis we have given insight into the literature on entrepreneurship, finance and artificial intelligence, and this has been done through many facets. As to the question of what is of high quality and what is not, a question as difficult as that can not be answered by bibliometric analysis. Such a question can only be answered by reading through, and inferring how much significance the text is to the subject of life and to the subject of science. If it bears no significance on these, it is of no use. Yet, if insight into the field of AI-entrepreneurship-finance is a question, the job has been done, information is here, and the conclusions are as follows, itemized as per research objectives and cross-referenced with relevant parts of the paper which deal in part or entirely with the specific research objective, and also presented throughout this section of the paper.

\begin{enumerate}[1.]
	\item To determine the publication productivity and evolution of scientific knowledge on the intersection of AI-entrepreneurship-finance -- achieved in Subsection~\ref{sub:DDA}, Subsection~\ref{sub:SDA}, Subsection~\ref{sec:DTA}, Subsection~\ref{sec:WDA}, Subsection~\ref{sec:CDA}, and Subsection~\ref{sec:ISDA}.
	
	\item To identify the most influential articles, and the most prolific academic scholars, journals, institutions, and countries on the intersection of AI-entrepreneurship-finance, and to determine the degree of academic cooperation and multidisciplinarity in the knowledge field -- achieved in Subsection~\ref{sub:DDA}, Subsection~\ref{sub:SDA}, Subsection~\ref{sub:ADA}, Subsection~\ref{sub:AfDA}, Subsection~\ref{sub:DoDA}, Subsection~\ref{sec:DTA}, Subsection~\ref{sec:WDA}, Subsection~\ref{sec:CDA}, Subsection~\ref{sec:ISDA}, and Subsection~\ref{sec:SSDA}.
	
	\item To determine and interpret prominent topics on the intersection of AI-entrepreneurship-finance, and to identify the chronological development of prominent topics -- achieved in Subsection~\ref{sec:WDA}, and Subsection~\ref{sec:CDA}.
	
	\item To determine AI methods (method, algorithm, technique) used in the study of certain topics at the intersection of AI-entrepreneurship-finance, so as to determine the current state, and project future possibilities -- achieved in Subsection~\ref{sec:CDA}.
	
	\item To reach a more profound insight into the research field, and to reflect on the emerging research directions and the promising AI methods for future applications in entrepreneurial finance, with implications for the scientific community, computer experts, entrepreneurs, and investors in entrepreneurship -- achieved in Subsection~\ref{sec:DTA}, Subsection~\ref{sec:CDA}, Subsection~\ref{sec:ISDA}, Subsection~\ref{sec:SSDA}, and Section~\ref{sec:mc}.
	
	\item To give recommendations for future improvement of bibliometric methodology -- achieved in Appendix \ref{secA2} as well as in the application of suggested amortized $ h $-index in the research, as seen in presented analyses of the paper.
\end{enumerate}

Considering methods and branches in Table \ref{tab:MPG}, the data there, perhaps more than any other part of the paper, speaks about the current state of research, gaps, and potential future directions. One can clearly see the most prominent branches and the entirety of computing methods that were used throughout these sub-divisions of economics. It is thus more obvious where are current successes and where should or could future research endeavors head. Aside from the research perspective, there is also a question of application, entrepreneurship, and investing. The issue of application and entrepreneurship is clear, at least in regard to dominant branches and methods, with already possible avenues for those less prominent areas determined by other relevant methods and branches of economics.

Answers to the questions investors would ask are somewhat more elusive, primarily in regard to whether one is seeking a long-term or short-term investment, and then in terms of either investing in a branch where the computing landscape is more known or in a branch that leans more to an uncharted territory -- with the primary focus of the investment possibly being technological, or perhaps some sort of a diversification. When problems presented in \cite{Biju2023} and brought to the fore as well, there seem to be three medium to high gain paths. The first is focusing on a correct implementation in an economic context with a compelling product, while the second is heading beyond the current state to new technologies or new branches and the first bricks that will build the future. The third option leverages the old and gradual innovation, either by combining innovation with the most prominent branches, or less innovation with branches not so heavily saturated with computing. The first and the third paths are likely of the most appeal to investors, if for nothing else then for the time span which should here be of a shorter nature and such a project price coupled with that fact.

Following the research results, and in accordance with the objectives of the research, in the continuation of the article we provide the following implications for future field development:

\begin{itemize}
	\item  Stronger development of entrepreneurship research in the FinTech sector and further expansion of AI in the segment of alternative sources of financing for entrepreneurs (crowdfunding, peer-to-peer lending, robo-advisors). One example of interesting future research is examining how robo-advisors can help (1) business angels in making investment decisions regarding the financing of entrepreneurial ventures, and (2) portfolio entrepreneurs in making decisions about business expansion and new investment opportunities \cite{Goodell2021}. Our research shows that there is a smaller number of works in this topic niche and that the field started development only recently.
	
	\item Progress of research on AI as a support for preventing and detecting financial fraud. In the context of entrepreneurship, frauds are becoming more and more common when making financial payments in business-to-business operations (for example, false information about the bank account number of a business partner by attackers). Likewise, by integrating AI, blockchain and smart contracts, it is perhaps possible to overcome the shortcomings of auditing and financial reporting and to act positively in the direction of preventing financial frauds related to auditing, while keeping in mind privacy, security and societal concerns. \cite{Goodell2021,Kumar2022}
	
	\item The emergence of the field of blockchain in entrepreneurship. It is a new area of research since the application of AI methods in this area has yet to be explored (for example, the application of blockchain in business scaling and automation to improve the performance of an entrepreneurial venture). In the future, blockchain can become one of the key resources of entrepreneurs because it can contribute to the rationalization of business (through savings in production) but also in financial accounting, compliance requirements, and auditing \cite{Giuggioli2022}. On the other hand, this technology had its own share of frauds, privacy concerns, security issues, etc. that need to be dealt with -- the blockchain market is also quite volatile. \cite{Phan2019,Islam2021,Feng2019,Liu2019,Bouri2019,Katsiampa2019}
	
	\item The more common application of ensemble in finance and entrepreneurship. Despite the advantages of deep learning, classic machine learning approaches (such as decision trees, Random Forest, SVM, k-NN, and Bayesian models) are still widely used. The use of the ensemble approach in finance and entrepreneurship is still in its infancy, although according to the first findings, they show good performance \cite{Nazareth2023}.
	
	\item Stronger development of predictive analytics in business planning for entrepreneurs (for example, firm-level price forecasting). There are few works in this area since most of the existing papers are related to macroeconomic and microeconomic forecasting (for example, forecasting the price of oil or electricity, and forecasting stock prices on the capital market). Knowledge related to the performance of individual AI methods from the macroeconomic domain should be transferred to the area of business planning. According to Nazareth and Reddy (2023), LSTM models show outstanding performance in predicting financial time series on stock markets \cite{Nazareth2023} -- the possibilities of implementing these models in the field of business planning of entrepreneurs have yet to be explored.
	
	\item Expansion of the application of AI to improve communication strategies and impression management in obtaining financial resources for an entrepreneurial venture. Various NLP techniques play a key role in this area, and neuroscience in combination with AI also has potential. Research examples refer to the development of interpretive models that explain the reactions of the human brain during exposure to communication or presentation directed by an entrepreneur towards a potential investor and the examination of how these reactions affect the final financing decision. Investors can use this knowledge to rationalize funding decisions, and entrepreneurs to adjust behavior and ensure success in financing \cite{Giuggioli2022}. Our research shows fewer works in this topic niche and a growing interest in the area.
\end{itemize}

There are a number of issues that puzzle the mind, which are both potential future hurdles as well as topics for future research. When we were gathering documents for the research it was evident that there was a lack of cooperation between computer experts and economic experts, which then in turn resulted in research of a lesser quality, as economic authors lack the computing expertise that is not so easily acquired. \cite{Shi2019,Levesque2020} Such author multidisciplinary cooperation would use artificial intelligence state-of-the-art technologies and methods to test and build new entrepreneurial theories in a rigorous, relevant, and impactful way -- an effort must be therefore taken so as to achieve cooperation between computer and economic researchers, or the research will continue to suffer. \cite{Shi2019,Levesque2020}

An evident lack of a necessary policy framework for the application of AI is also an issue. The question of how to apply these technologies in a morally acceptable way, dealing with entrepreneurship and finance, is not completely solved nor is it entirely yet clear how that should look like a logical consequence of revolutionary technology, and a path for further discussion and research. \cite{Chen2023,Gupta2023,Levesque2020} Aside from necessary laws there is also a need for education, as people need to be acquainted with the new state of affairs, and after that adapt to it, a transition period is a must. \cite{Chen2023,Gupta2023,Levesque2020} One of the areas where this sensitive issue is particularly clear, both in research and practice, is using AI to identify facial expressions and emotion detection in entrepreneurial finance, as possible misconduct by such a technology and manipulation could be extraordinarily negative. \cite{Chen2023,Gupta2023,Levesque2020}

It is a necessity to take a constructive and critical stance and to take into consideration constraints, risks, and implications of AI usage in entrepreneurial finance, e.g. AI has biases, makes mistakes, etc., researchers and practitioners need to be aware of that, thus the issues can be tackled, with transparency -- otherwise we are doomed to blindly trust the algorithms to eventually lead us in a wrong way, algorithm's result needs to be checked and the final decision must be on a human being if we lose control, we will be controlled. \cite{Obschonka2019}

This research was conducted in an encompassing, methodological, rigorous, and in-depth way. There are however a number of constraints, some are a result of the reach Bibliometrics can take, while others are technological or generally scientific. These need to be taken into consideration when one thinks about the research results.

\begin{enumerate}[1.]
	\item Bibliometrix has it seems a number of bugs. When one uses BibTeX file some science categories are not being recognized, while author counting is not working properly/precisely.
	
	\item The database of the research is Web of Science. Other databases, like e.g. Scopus (we were planning initially on including Scopus also, there were however a number of issues that for the reason of time and complexity of the entire research we were not able to resolve -- Elsevier was very quick to respond and quite cooperative; if Scopus was on the other hand also included the research would be even larger, it is therefore for the time being including WoS only a proper step in an iterative nature of science) are not included in the analysis.
	
	\item Documents that are included in the Bibliometrics analysis are in the English language and are journal articles -- conferences and books are not included as the focus was only on the most encompassing, impactful and relevant documents, generally speaking, those are journal articles. Books are not the primary driving force of research and peer review and are not typical avenues for Bibliometrics research.
	
	\item Data for the last two to three years is not yet complete, with a number of documents also missing the final date of publication, this is therefore restrictive on the interpretation and projection of future events.
	
	\item As the database included in the research is WoS it is difficult to generalize, this constraint nevertheless results in gathering the most impactful documents, which in turn suggests that results from Scopus should generally align.
\end{enumerate}

Aside from various suggestions, interpretations, and methodological improvements made to Bibliometrics, there is also a specific improvement that should be specially mentioned, namely amortized $ h $-index. With this measure, we hope that the age shortcoming of $ h $-index can be inhibited, and thus obtain a more realistic result about the item being measured. With amortized $ h $-index we can therefore ascertain how an item fares with passed years, or decades, as compared to other items that are of a different time span, and with that get information on the successes/failures of the young, or successes/failures of the old -- on the details about amortized $ h $-index please consult Appendix \ref{secA2}.

During various Bibliometrics analyses, one often deals with counting documents, or with citations, and it would be useful if the tools would have an option for filtering surveys, reviews, Bibliometrics, and meta-analysis articles -- this option would allow for a fine-grained look into contributions, and contributions about contributions, with making interpretations easier and likely more precise. Such an option would be useful in other analyses, e.g. conceptual clustering, etc., and would improve the entire process in a substantial way.

The rise of AI, coupled with big data, has given birth to the Fin-tech sector, comprising of digital technology innovations with new business models for finance backed by that technology. \cite{Dixon2020} The following list therefore presents possible research projects to take and make a contribution to the field -- preferably by collaboration of computing and economics experts, and complementary knowledge thereof.

\begin{enumerate}[1.]
	\item Financial firms are in need of developing global authentication standards. \cite{Chaklader2023}
	
	\item Improvement of CIA (Confidentiality, Integrity, Availability) triad for Fin-tech is needed. \cite{Chaklader2023}
	
	\item Optimal introduction of security knowledge and training for financial sector professions is needed. \cite{Chaklader2023}
	
	\item Innovations in "blockchain and cryptocurrencies, digital advisory and trading systems, equity crowdfunding, peer-to-peer lending, and mobile payment services" are needed -- which are examples of areas central to Fin-tech sector. \cite{Goodell2021}
	
	\item Research is required in robo-advisors (providing e.g. "portfolio management services and financial advice without significant human intervention") so as to "understand the myriad aspects of Fin-tech and the impact of AI" on the area of interest. \cite{Goodell2021}
	
	\item Ascertaining the usefulness of machine learning techniques in cryptocurrency and blockchain is needed. \cite{Nazareth2023}
	
	\item Research in AI and financial crisis prediction, portfolio management, and detection of anomalies in financial statements by AI is needed. \cite{Nazareth2023}
	
	\item Research in explainable AI and its application in finance, and other areas, is needed -- indicated by this research.
	
	\item Research into constraints, risks, and implications of AI usage in entrepreneurial finance, and other areas, is needed -- e.g. AI has biases, makes mistakes, etc. and can lead not only to unnecessary actions but to harmful ones also. \cite{Obschonka2019}
	
	\item Research in branches 1(a) and 2 in Table \ref{tab:MPG} and adequate AI methods, techniques, and algorithms are needed -- indicated by this research.
\end{enumerate}

Entrepreneurs are tasked with multiple and various jobs that require hours upon hours to complete. Perhaps one of the greatest benefits of AI is its ability to allow small business owners to greatly reduce the time needed to complete tasks, especially those that often feel burdensome. Since the days of the industrial revolution, people have been concerned that machines would take their jobs and humans would become redundant. However, many jobs machines perform relieve humans of mundane tasks so they can focus their efforts on situations that require human intelligence and caring. For example, humans can operate help lines, but many workers would argue that such mind-numbing jobs do not take advantage of human skills and intelligence. AI systems can interact with customers at base levels so that humans can focus on those who truly need more personalized and interactive assistance.

Entrepreneurial finance and Fin-tech sector have seen, and are yet to see, substantial advances in terms of AI, especially by the application of various even disparate approaches to a field accustomed to doing things the old way. The change is it seems inevitable, it is up to us (society at large) to determine what will be the result of that change, with one part of that process being research, and discussion -- the part in which Academia is proficient -- the part with which this transformation should be backed.

\backmatter

\bmhead{Acknowledgments}

For the research the following was also used: Latex\footnote{https://www.latex-project.org/}, TexLive\footnote{https://tug.org/texlive/}, MikTeX\footnote{https://miktex.org/}, Draw.io\footnote{https://www.drawio.com/}, GIMP\footnote{https://www.gimp.org/}, Texstudio\footnote{https://www.texstudio.org/}, LibreOffice\footnote{https://www.libreoffice.org/}, JabRef\footnote{https://www.jabref.org/}, Rstudio\footnote{https://posit.co/}, Bibliometrix \cite{Aria2017,Aria2023}, VOSviewer \cite{Eck2006,PerianesRodriguez2016}, Publish or Perish \cite{Harzing2023}, Linux Mint\footnote{https://linuxmint.com/}.

\section*{Declarations}

The authors declare no conflict of interest.

\begin{appendices}

	\section{Amortized $ h $-index}\label{secA2}
	
	Here we will describe how one can calculate amortized $ h $-index, or in general terms a measure for calculating influence through time if the relationship between variables is exponentially inversely proportional (having in mind the neighboring sequence of years stretching to the current year, e.g. 2010, 2011, 2012,..., 2023) -- presented in Table \ref{tab:AHI}.

	\begin{table}[h] 
		\caption{Amortized $ h $-index Example Calculation}\label{tab:AHI}
		\begin{tabularx}{\textwidth}{@{}>{\RaggedRight}X>{\RaggedRight}X>{\RaggedRight}X>{\RaggedRight}X>{\RaggedRight}X>{\RaggedRight}p{1.4cm}}
			\toprule%
			Year\textsuperscript{1} & $ h $-index & Citable Years\textsuperscript{2} & Pondering Scalar\textsuperscript{3} (PS) & Normalized\textsuperscript{4} PS & Amortized $ h $-index\textsuperscript{5} \\
			\midrule
			$ 1991 $ & $ 51 $ & $ 10 $ & $ 1.0000 $  & $ 0.1000 $ & $ 5.1000 $ \\
			$ 1992 $ & $ 44 $ & $ 9 $  & $ 1.1111 $  & $ 0.1111 $ & $ 4.8888 $ \\
			$ 1993 $ & $ 42 $ & $ 8 $  & $ 1.2500 $  & $ 0.1250 $ & $ 5.2500 $ \\
			$ 1994 $ & $ 36 $ & $ 7 $  & $ 1.4285 $  & $ 0.1428 $ & $ 5.1428 $ \\
			$ 1995 $ & $ 34 $ & $ 6 $  & $ 1.6666 $  & $ 0.1666 $ & $ 5.6666 $ \\
			$ 1996 $ & $ 27 $ & $ 5 $  & $ 2.0000 $  & $ 0.2000 $ & $ 5.4000 $ \\
			$ 1997 $ & $ 21 $ & $ 4 $  & $ 2.5000 $  & $ 0.2500 $ & $ 5.2500 $ \\
			$ 1998 $ & $ 14 $ & $ 3 $  & $ 3.3333 $  & $ 0.3333 $ & $ 4.6666 $ \\
			$ 1999 $ & $ 11 $ & $ 2 $  & $ 5.0000 $  & $ 0.5000 $ & $ 5.5000 $ \\
			$ 2000 $ & $ 5 $  & $ 1 $  & $ 10.0000 $ & $ 1.0000 $ & $ 5.0000 $ \\
			\botrule
		\end{tabularx}
		\footnotetext{The data is sorted in ascending order according to year, which is not a prerequisite for amortized $ h $-index calculation.}
		\footnotetext[1]{Year of an item, e.g. journal, author, etc., years can have any time span between them and they can be the same as well, e.g. 2010, 2014, 2015, 2015,..., 2023. The current year does not need to be an ongoing real-world year, it is up to the expert to choose the appropriate year for the instance analyzed, although an ongoing real-world year would be the typical case.}
		\footnotetext[2]{Calculated as the difference of the last and current year, with the addition of one more year, so as to include the time passed in an incomplete year and make a conservative estimate.}
		\footnotetext[3]{Calculated as a quotient of the highest citable years and current citable years, e.g. for 1993 calculation would be $ \frac{10}{8} = 1.25 $.}
		\footnotetext[4]{As per maximum value of pondering scalar set -- thus making the last observed year a fixed point for comparative analysis.}
		\footnotetext[5]{Calculated as a product of normalized pondering scalar and $ h $-index, e.g. for the year 1997 calculation would be $ 0.25 \times 21 = 5.25 $.}
	\end{table}

	As the year 2000 is a fixed point (the current year for example calculation) the $ h $-index is unchanged, that is the value for the amortized $ h $-index is the same, while all the other values are amortized accordingly. The amortized $ h $-index for the year 1991 is $ 5.1 $, as opposed to 51 originally -- year by year, the original value is very close to the fixed point of 5, which is therefore taken into account with amortized value. The year 1999 is half a point higher from the fixed point, while the year by year also being very close -- however this instance is of a far younger nature than 1991, and is therefore more influential; with similar reasoning for the year 1998 being valid, where amortized value is somewhat lower, a consequence of lower original value and passing of more time, as per fixed point.
	
	There is a possibility that a number of or all of amortized $ h $-indexes turn equal, as a consequence of original values, time spans and pondering distribution presenting such a case. If there is a strong need for prioritization, such a turn of events can be dealt with. Every $ h $-index needs to be decreased by one point until all amortized $ h $-indexes are unique -- this kind of estimate then presents a potential recent state with which one can then rank items. This kind of move can also be performed just on equivalent items, with the ranking then valid only between those, other items do not change position. It is also possible to give preference to items that are more recent, as per the idea behind amortized $ h $-index, and rank items with that in mind, while other items do not change position. One can also additionally calculate an average between the amortized measure and the one not amortized, and try to resolve the issue that way.
	
	If an extreme year is a part of the data then it might be preferable to conduct analysis with and without an extreme data point (an instance where the current year would not be fixed, e.g. one might want to ascertain the historical state), so as to ascertain the impact of the extreme year, and produce an objective and relevant result in terms of items relation. There are a number of constraints that also need to be taken into consideration when one performs and interprets results with amortized $ h $-index, these are as follows.

	\begin{enumerate}[1.]
		\item History of science can be substantially different as compared to today, in terms of the number of authors, journals, citation practices, institutional practices, working environment, etc., all these and other factors can influence the metric observed.
		
		\item A high influence is given to recent items, for which time will show far less in terms of relevance.
		
		\item A low influence is given to old items, for which time is not a friend in an environment of substantial positive quantitative change.
	\end{enumerate}

	As far as pondering distribution is concerned, it is possible that there are instances where some other distribution would be more appropriate, and should therefore replace the distribution suggested in this paper. Thus with an amortized $ h $-index we can more precisely ascertain the relation between various items where the observed metric depends on time -- amortization can also be generalized with time replaced by another characteristic.

	\section{Thematic Evolution of Concepts Most Highly Cited Documents}\label{secA3}
	
	As an upgrade to the thematic evolution of concepts, we have additionally performed an analysis of the most highly cited documents per period of interest from Figures \ref{fig:tcpy}, \ref{fig:TM2006}, and \ref{fig:TM2019}, including analysis for peak years. 
	
	Thus the reader can get an insight into the most relevant documents together with meta-data, which is useful in itself. Performed analyses can be seen in Tables \ref{tab:MHCDPI99}, \ref{tab:MHCDPI04}, \ref{tab:MHCDPI08}, \ref{tab:MHCDPI14}, \ref{tab:MHCDPI18}, \ref{tab:MHCDPI20}, \ref{tab:MHCDPI23}. Aside from relevance, information about the influence on the specific cluster, that is the theme, is presented as well. In this way it is possible to, in a concrete way, link the evolution of themes and those who have built that significance via multiple perspectives enabled by document metadata.

	\begin{table}[h] 
		\caption{Most Highly Cited Documents Per Period of Interest from Figures \ref{fig:tcpy}, \ref{fig:TM2006} and \ref{fig:TM2019}}\label{tab:MHCDPI99}
		\begin{tabular*}{\textwidth}{@{\extracolsep\fill}lp{.7\textwidth}}
			\toprule%
			\multicolumn{2}{@{}c@{}}{1991 -- 1999 (Top Three)}
			\\\cmidrule{1-2}
			Total/Average\textsuperscript{1} Citation & Document Reference\textsuperscript{2}/DOI\textsuperscript{3} \\
			\midrule
			$ 440 $/$ 14.667 $ & \href{https://doi.org/10.1016/0378-4266(94)90007-8}{ALTMAN EI, 1994, J BANK FINANC} \\
			&	\footnotesize CORPORATE DISTRESS DIAGNOSIS - COMPARISONS USING LINEAR DISCRIMINANT-ANALYSIS AND NEURAL NETWORKS (THE ITALIAN EXPERIENCE) \\
			
			$ 356 $/$ 11.867 $ & \href{https://doi.org/10.1016/0167-9236(94)90024-8}{WILSON RL, 1994, DECIS SUPPORT SYST} \\
			& \footnotesize BANKRUPTCY PREDICTION USING NEURAL NETWORKS \\
			
			$ 348 $/$ 13.92 $ & \href{https://doi.org/10.1016/S0377-2217(98)00051-4}{ZHANG GQ, 1999, EUR J OPER RES} \\
			& \footnotesize ARTIFICIAL NEURAL NETWORKS IN BANKRUPTCY PREDICTION: GENERAL FRAMEWORK AND CROSS-VALIDATION ANALYSIS \\
			\cmidrule{1-2}
			\multicolumn{2}{@{}c@{}}{1994 -- peak year (Top Three)\textsuperscript{4}}\\
			\midrule
			$ 42 $/$ 1.4 $ & \href{https://doi.org/10.1016/0360-8352(94)90330-1}{TSUKUDA J, 1994, COMPUT IND ENG} \\
			& \footnotesize PREDICTING JAPANESE CORPORATE BANKRUPTCY IN TERMS OF FINANCIAL DATA USING NEURAL-NETWORK \\
			\botrule
		\end{tabular*}
		\footnotetext{Sorted according to total citation in descending order, rounded in two decimal places}
		\footnotetext{Of the clusters for the accompanying period, ALTMAN EI, 1994, J BANK FINANC very highly contributed to linear discriminant cluster; WILSON RL, 1994, DECIS SUPPORT SYST highest contribution was to neural network cluster; ZHANG GQ, 1999, EUR J OPER RES highest contribution was to neural network and human expert clusters; TSUKUDA J, 1994, COMPUT IND ENG highest contribution was to financial ratios and human expert clusters}
		\footnotetext[1]{Average citation per year}
		\footnotetext[2]{Abbreviated document metadata and title}
		\footnotetext[3]{Link of a Digital Object Identifier}
		\footnotetext[4]{ALTMAN EI, 1994, J BANK FINANC and WILSON RL, 1994, DECIS SUPPORT SYST are the first two, respectively}
	\end{table}

	\begin{table}[h] 
		\caption{Most Highly Cited Documents Per Period of Interest from Figures \ref{fig:tcpy}, \ref{fig:TM2006} and \ref{fig:TM2019}}\label{tab:MHCDPI04}
		\begin{tabular*}{\textwidth}{@{\extracolsep\fill}lp{.7\textwidth}}
			\toprule%
			\multicolumn{2}{@{}c@{}}{2000 -- 2004 (Top Three)}
			\\\cmidrule{1-2}
			Total/Average\textsuperscript{1} Citation & Document Reference\textsuperscript{2}/DOI\textsuperscript{3} \\
			\midrule
			$ 1002 $/$ 41.75 $ & \href{https://doi.org/10.1109/5326.897072}{ZHANG GQP, 2000, IEEE TRANS SYST MAN CYBERN PART C-APPL REV} \\
			&	\footnotesize NEURAL NETWORKS FOR CLASSIFICATION: A SURVEY \\
			
			$ 343 $/$ 14.91 $ & \href{https://doi.org/10.1109/72.935101}{ATIYA AF, 2001, IEEE TRANS NEURAL NETW} \\
			& \footnotesize BANKRUPTCY PREDICTION FOR CREDIT RISK USING NEURAL NETWORKS: A SURVEY AND NEW RESULTS \\
			
			$ 244 $/$ 10.17 $ & \href{https://doi.org/10.1016/S0957-4174(99)00053-6}{AHN BS, 2000, EXPERT SYST APPL} \\
			& \footnotesize THE INTEGRATED METHODOLOGY OF ROUGH SET THEORY AND ARTIFICIAL NEURAL NETWORK FOR BUSINESS FAILURE PREDICTION \\
			\cmidrule{1-2}
			\multicolumn{2}{@{}c@{}}{2000 -- peak year (Top Three)\textsuperscript{4}}\\
			\midrule
			$ 77 $/$ 3.21 $ & \href{https://doi.org/10.1023/A:1019292321322}{CHARALAMBOUS C, 2000, ANN OPER RES} \\
			& \footnotesize COMPARATIVE ANALYSIS OF ARTIFICIAL NEURAL NETWORK MODELS: APPLICATION IN BANKRUPTCY PREDICTION \\
			\botrule
		\end{tabular*}
		\footnotetext{Sorted according to total citation in descending order, rounded in two decimal places}
		\footnotetext{Of the clusters for the accompanying period, ZHANG GQP, 2000, IEEE TRANS SYST MAN CYBERN PART C-APPL REV has contributed to neural networks cluster exclusively; ATIYA AF, 2001, IEEE TRANS NEURAL NETW very highly contributed to neural networks cluster; AHN BS, 2000, EXPERT SYST APPL highest contribution was to neural networks and rough sets clusters; CHARALAMBOUS C, 2000, ANN OPER RES highest contribution was to financial data and neural networks clusters}
		\footnotetext[1]{Average citation per year}
		\footnotetext[2]{Abbreviated document metadata and title}
		\footnotetext[3]{Link of a Digital Object Identifier}
		\footnotetext[4]{ZHANG GQP, 2000, IEEE TRANS SYST MAN CYBERN PART C-APPL REV and AHN BS, 2000, EXPERT SYST APPL are the first two, respectively}
	\end{table}

	\begin{table}[h] 
		\caption{Most Highly Cited Documents Per Period of Interest from Figures \ref{fig:tcpy}, \ref{fig:TM2006} and \ref{fig:TM2019}}\label{tab:MHCDPI08}
		\begin{tabular*}{\textwidth}{@{\extracolsep\fill}lp{.7\textwidth}}
			\toprule%
			\multicolumn{2}{@{}c@{}}{2005 -- 2008 (Top Three)}
			\\\cmidrule{1-2}
			Total/Average\textsuperscript{1} Citation & Document Reference\textsuperscript{2}/DOI\textsuperscript{3} \\
			\midrule
			$ 609 $/$ 35.82 $ & \href{https://doi.org/10.1016/j.ejor.2006.08.043}{KUMAR PR, 2007, EUR J OPER RES} \\
			&	\footnotesize BANKRUPTCY PREDICTION IN BANKS AND FIRMS VIA STATISTICAL AND INTELLIGENT TECHNIQUES - A REVIEW \\
			
			$ 512 $/$ 26.95 $ & \href{https://doi.org/10.1016/j.eswa.2004.12.008}{MIN JH, 2005, EXPERT SYST APPL} \\
			&	\footnotesize BANKRUPTCY PREDICTION USING SUPPORT VECTOR MACHINE WITH OPTIMAL CHOICE OF KERNEL FUNCTION PARAMETERS \\
			
			$ 456 $/$ 24.00 $ & \href{https://doi.org/10.1016/j.eswa.2004.08.009}{SHIN KS, 2005, EXPERT SYST APPL} \\
			&	\footnotesize AN APPLICATION OF SUPPORT VECTOR MACHINES IN BANKRUPTCY PREDICTION MODEL \\
			\cmidrule{1-2}
			\multicolumn{2}{@{}c@{}}{2005, 2007 -- peak years (Top Three Each, respectively)\textsuperscript{4}}\\
			\midrule
			$ 221 $/$ 11.63 $ & \href{https://doi.org/10.1016/j.cor.2004.03.017}{WEST D, 2005, COMPUT OPER RES} \\
			&	\footnotesize NEURAL NETWORK ENSEMBLE STRATEGIES FOR FINANCIAL DECISION APPLICATIONS \\
			\cmidrule{1-2}
			$ 283 $/$ 16.65 $ & \href{https://doi.org/10.1016/j.eswa.2006.02.016}{KIRKOS E, 2007, EXPERT SYST APPL} \\
			&	\footnotesize DATA MINING TECHNIQUES FOR THE DETECTION OF FRAUDULENT FINANCIAL STATEMENTS \\
			
			$ 257 $/$ 15.12 $ & \href{https://doi.org/10.1016/j.eswa.2005.12.008}{WU CH, 2007, EXPERT SYST APPL} \\
			&	\footnotesize A REAL-VALUED GENETIC ALGORITHM TO OPTIMIZE THE PARAMETERS OF SUPPORT VECTOR MACHINE FOR PREDICTING BANKRUPTCY \\
			\botrule
		\end{tabular*}
		\footnotetext{Sorted according to total citation in descending order, rounded in two decimal places}
		\footnotetext{Of the clusters for the accompanying period, KUMAR PR, 2007, EUR J OPER RES very highly contributed to neural network cluster; MIN JH, 2005, EXPERT SYST APPL very highly contributed to neural network cluster; SHIN KS, 2005, EXPERT SYST APPL highest contribution was to neural network cluster; WEST D, 2005, COMPUT OPER RES exclusively contributed to neural network cluster; KIRKOS E, 2007, EXPERT SYST APPL very highly contributed to financial statements cluster; WU CH, 2007, EXPERT SYST APPL highest contribution was to neural network and genetic algorithm clusters}
		\footnotetext[1]{Average citation per year}
		\footnotetext[2]{Abbreviated document metadata and title}
		\footnotetext[3]{Link of a Digital Object Identifier}
		\footnotetext[4]{MIN JH, 2005, EXPERT SYST APPL and SHIN KS, 2005, EXPERT SYST APPL are the first two for 2005, respectively; KUMAR PR, 2007, EUR J OPER RES is the first for 2007}
	\end{table}

	\begin{table}[h] 
		\caption{Most Highly Cited Documents Per Period of Interest from Figures \ref{fig:tcpy}, \ref{fig:TM2006} and \ref{fig:TM2019}}\label{tab:MHCDPI14}
		\begin{tabular*}{\textwidth}{@{\extracolsep\fill}lp{.7\textwidth}}
			\toprule%
			\multicolumn{2}{@{}c@{}}{2009 -- 2014 (Top Three)}
			\\\cmidrule{1-2}
			Total/Average\textsuperscript{1} Citation & Document Reference\textsuperscript{2}/DOI\textsuperscript{3} \\
			\midrule
			$ 980 $/$ 81.67 $ & \href{https://doi.org/10.1016/j.knosys.2011.07.001}{PAN WT, 2012, KNOWLEDGE-BASED SYST} \\
			&	\footnotesize A NEW FRUIT FLY OPTIMIZATION ALGORITHM: TAKING THE FINANCIAL DISTRESS MODEL AS AN EXAMPLE \\
			
			$ 496 $/$ 33.07 $ & \href{https://doi.org/10.1016/j.eswa.2007.10.005}{PALIWAL M, 2009, EXPERT SYST APPL} \\
			&	\footnotesize NEURAL NETWORKS AND STATISTICAL TECHNIQUES: A REVIEW OF APPLICATIONS \\
			
			$ 489 $/$ 37.62 $ & \href{https://doi.org/10.1016/j.dss.2010.08.006}{NGAI EWT, 2011, DECIS SUPPORT SYST} \\
			&	\footnotesize THE APPLICATION OF DATA MINING TECHNIQUES IN FINANCIAL FRAUD DETECTION: A CLASSIFICATION FRAMEWORK AND AN ACADEMIC REVIEW OF LITERATURE \\
			\cmidrule{1-2}
			\multicolumn{2}{@{}c@{}}{No Extreme Peaks in this Period}\\
			\botrule
		\end{tabular*}
		\footnotetext{Sorted according to total citation in descending order, rounded in two decimal places}
		\footnotetext{Of the clusters for the accompanying period, PAN WT, 2012, KNOWLEDGE-BASED SYST highest contribution was to neural network and network model clusters; PALIWAL M, 2009, EXPERT SYST APPL highest contribution was to neural network and support vector clusters; NGAI EWT, 2011, DECIS SUPPORT SYST exclusively contributed to neural network cluster}
		\footnotetext[1]{Average citation per year}
		\footnotetext[2]{Abbreviated document metadata and title}
		\footnotetext[3]{Link of a Digital Object Identifier}
	\end{table}

	\begin{table}[h] 
		\caption{Most Highly Cited Documents Per Period of Interest from Figures \ref{fig:tcpy}, \ref{fig:TM2006} and \ref{fig:TM2019}}\label{tab:MHCDPI18}
		\begin{tabular*}{\textwidth}{@{\extracolsep\fill}lp{.7\textwidth}}
			\toprule%
			\multicolumn{2}{@{}c@{}}{2015 -- 2018 (Top Three)}
			\\\cmidrule{1-2}
			Total/Average\textsuperscript{1} Citation & Document Reference\textsuperscript{2}/DOI\textsuperscript{3} \\
			\midrule
			$ 468 $/$ 52.00 $ & \href{https://doi.org/10.1016/j.ejor.2015.05.030}{LESSMANN S, 2015, EUR J OPER RES} \\
			&	\footnotesize BENCHMARKING STATE-OF-THE-ART CLASSIFICATION ALGORITHMS FOR CREDIT SCORING: AN UPDATE OF RESEARCH \\
			
			$ 279 $/$ 39.86 $ & \href{https://doi.org/10.1016/j.eswa.2017.04.006}{BARBOZA F, 2017, EXPERT SYST APPL} \\
			&	\footnotesize MACHINE LEARNING MODELS AND BANKRUPTCY PREDICTION \\
			
			$ 209 $/$ 23.22 $ & \href{https://doi.org/10.1016/j.ejor.2014.08.016}{GENG R, 2015, EUR J OPER RES} \\
			&	\footnotesize PREDICTION OF FINANCIAL DISTRESS: AN EMPIRICAL STUDY OF LISTED CHINESE COMPANIES USING DATA MINING \\
			\cmidrule{1-2}
			\multicolumn{2}{@{}c@{}}{2015, 2017 -- peak years (Top Three Each, respectively)\textsuperscript{4}}\\
			\midrule
			$ 204 $/$ 22.67 $ & \href{https://doi.org/10.1016/j.eswa.2015.02.001}{MALEKIPIRBAZARI M, 2015, EXPERT SYST APPL} \\
			&	\footnotesize RISK ASSESSMENT IN SOCIAL LENDING VIA RANDOM FORESTS \\
			\cmidrule{1-2}
			$ 135 $/$ 19.29 $ & \href{https://doi.org/10.1016/j.engappai.2017.05.003}{WANG M, 2017, ENG APPL ARTIF INTELL} \\
			&	\footnotesize GREY WOLF OPTIMIZATION EVOLVING KERNEL EXTREME LEARNING MACHINE: APPLICATION TO BANKRUPTCY PREDICTION \\
			
			$ 117 $/$ 16.71 $ & \href{https://doi.org/10.1016/j.eswa.2016.12.020}{ABELIAN J, 2017, EXPERT SYST APPL} \\
			&	\footnotesize A COMPARATIVE STUDY ON BASE CLASSIFIERS IN ENSEMBLE METHODS FOR CREDIT SCORING \\
			\botrule
		\end{tabular*}
		\footnotetext{Sorted according to total citation in descending order, rounded in two decimal places}
		\footnotetext{Of the clusters for the accompanying period, LESSMANN S, 2015, EUR J OPER RES highest contribution was to artificial intelligence and neural network clusters; BARBOZA F, 2017, EXPERT SYST APPL highest contribution was to neural network and predictive performance clusters; GENG R, 2015, EUR J OPER RES highest contribution was to neural network and data mining clusters; MALEKIPIRBAZARI M, 2015, EXPERT SYST APPL exclusively contributed to neural network cluster; WANG M, 2017, ENG APPL ARTIF INTELL highest contribution was to curve auc and neural network clusters; ABELIAN J, 2017, EXPERT SYST APPL highest contribution was to neural network and artificial intelligence clusters}
		\footnotetext[1]{Average citation per year}
		\footnotetext[2]{Abbreviated document metadata and title}
		\footnotetext[3]{Link of a Digital Object Identifier}
		\footnotetext[4]{LESSMANN S, 2015, EUR J OPER RES and GENG R, 2015, EUR J OPER RES are the first two for 2015, respectively; BARBOZA F, 2017, EXPERT SYST APPL is the first for 2017}
	\end{table}

	\begin{table}[h] 
		\caption{Most Highly Cited Documents Per Period of Interest from Figures \ref{fig:tcpy}, \ref{fig:TM2006} and \ref{fig:TM2019}}\label{tab:MHCDPI20}
		\begin{tabular*}{\textwidth}{@{\extracolsep\fill}lp{.7\textwidth}}
			\toprule%
			\multicolumn{2}{@{}c@{}}{2019 -- 2020 (Top Three)}
			\\\cmidrule{1-2}
			Total/Average\textsuperscript{1} Citation & Document Reference\textsuperscript{2}/DOI\textsuperscript{3} \\
			\midrule
			$ 332 $/$ 83.00 $ & \href{https://doi.org/10.1016/j.ijforecast.2019.07.001}{SALINAS D, 2020, INT J FORECAST} \\
			&	\footnotesize DEEPAR: PROBABILISTIC FORECASTING WITH AUTOREGRESSIVE RECURRENT NETWORKS \\
			
			$ 220 $/$ 55.00 $ & \href{https://doi.org/10.1016/j.ijinfomgt.2019.08.005}{WONG LW, 2020, INT J INF MANAGE} \\
			&	\footnotesize TIME TO SEIZE THE DIGITAL EVOLUTION: ADOPTION OF BLOCKCHAIN IN OPERATIONS AND SUPPLY CHAIN MANAGEMENT AMONG MALAYSIAN SMES \\
			
			$ 160 $/$ 40.00 $ & \href{https://doi.org/10.1016/j.ijpe.2019.107599}{DUBEY R, 2020, INT J PROD ECON} \\
			&	\footnotesize BIG DATA ANALYTICS AND ARTIFICIAL INTELLIGENCE PATHWAY TO OPERATIONAL PERFORMANCE UNDER THE EFFECTS OF ENTREPRENEURIAL ORIENTATION AND ENVIRONMENTAL DYNAMISM: A STUDY OF MANUFACTURING ORGANISATIONS \\
			\cmidrule{1-2}
			\multicolumn{2}{@{}c@{}}{2019, 2020 -- peak years (Top Three Each, respectively)\textsuperscript{4,}\textsuperscript{5}}\\
			\midrule
			$ 148 $/$ 29.60 $ & \href{https://doi.org/10.1016/j.jclepro.2019.03.181}{RAUT RD, 2019, J CLEAN PROD} \\
			&	\footnotesize LINKING BIG DATA ANALYTICS AND OPERATIONAL SUSTAINABILITY PRACTICES FOR SUSTAINABLE BUSINESS MANAGEMENT \\
			
			$ 126 $/$ 25.20 $ & \href{https://doi.org/10.1016/j.ijpe.2019.01.032}{ZHU Y, 2019, INT J PROD ECON} \\
			&	\footnotesize FORECASTING SMES' CREDIT RISK IN SUPPLY CHAIN FINANCE WITH AN ENHANCED HYBRID ENSEMBLE MACHINE LEARNING APPROACH \\
			
			$ 107 $/$ 21.40 $ & \href{https://doi.org/10.1016/j.bar.2019.04.002}{MOLL J, 2019, BRIT ACCOUNT REV} \\
			&	\footnotesize THE ROLE OF INTERNET-RELATED TECHNOLOGIES IN SHAPING THE WORK OF ACCOUNTANTS: NEW DIRECTIONS FOR ACCOUNTING RESEARCH \\
			\botrule
		\end{tabular*}
		\footnotetext{Sorted according to total citation in descending order, rounded in two decimal places}
		\footnotetext{Of the clusters for the accompanying period, SALINAS D, 2020, INT J FORECAST highest contribution was to demand forecasting, short-term memory and data analytics clusters; WONG LW, 2020, INT J INF MANAGE highest contribution was to supply chain and enterprises smes; DUBEY R, 2020, INT J PROD ECON highest contribution was to artificial intelligence and data analytics clusters; RAUT RD, 2019, J CLEAN PROD very highly contributed to data analytics cluster; ZHU Y, 2019, INT J PROD ECON highest contribution was to machine learning cluster; MOLL J, 2019, BRIT ACCOUNT REV highest contribution was to future research and artificial intelligence clusters}
		\footnotetext[1]{Average citation per year}
		\footnotetext[2]{Abbreviated document metadata and title}
		\footnotetext[3]{Link of a Digital Object Identifier}
		\footnotetext[4]{Even tough the whole period consists of two consecutive years, both of them are high peak years, and are therefore of interest for separate analysis}
		\footnotetext[5]{SALINAS D, 2020, INT J FORECAST, WONG LW, 2020, INT J INF MANAGE and DUBEY R, 2020, INT J PROD ECON are the first three for 2020, respectively}
	\end{table}

	\begin{table}[h] 
		\caption{Most Highly Cited Documents Per Period of Interest from Figures \ref{fig:tcpy}, \ref{fig:TM2006} and \ref{fig:TM2019}}\label{tab:MHCDPI23}
		\begin{tabular*}{\textwidth}{@{\extracolsep\fill}lp{.7\textwidth}}
			\toprule%
			\multicolumn{2}{@{}c@{}}{2021 -- 2023 (Top Three)}
			\\\cmidrule{1-2}
			Total/Average\textsuperscript{1} Citation & Document Reference\textsuperscript{2}/DOI\textsuperscript{3} \\
			\midrule
			$ 113 $/$ 37.67 $ & \href{https://doi.org/10.1016/j.im.2021.103434}{MIKALEF P, 2021, INF MANAGE} \\
			&	\footnotesize ARTIFICIAL INTELLIGENCE CAPABILITY: CONCEPTUALIZATION, MEASUREMENT CALIBRATION, AND EMPIRICAL STUDY ON ITS IMPACT ON ORGANIZATIONAL CREATIVITY AND FIRM PERFORMANCE \\
			
			$ 91 $/$ 30.33 $ & \href{https://doi.org/10.1016/j.jbusres.2020.01.007}{KUMAR V, 2021, J BUS RES} \\
			&	\footnotesize INFLUENCE OF NEW-AGE TECHNOLOGIES ON MARKETING: A RESEARCH AGENDA \\
			
			$ 89 $/$ 44.50 $ & \href{https://doi.org/10.1080/09585192.2020.1871398}{VRONTIS D, 2022, INT J HUM RESOUR MANAG} \\
			&	\footnotesize ARTIFICIAL INTELLIGENCE, ROBOTICS, ADVANCED TECHNOLOGIES AND HUMAN RESOURCE MANAGEMENT: A SYSTEMATIC REVIEW \\
			\cmidrule{1-2}
			\multicolumn{2}{@{}c@{}}{Incomplete and Erratic  Data\textsuperscript{4}}\\
			\botrule
		\end{tabular*}
		\footnotetext{Sorted according to total citation in descending order, rounded in two decimal places}
		\footnotetext{Of the clusters for the accompanying period, MIKALEF P, 2021, INF MANAGE exclusively contributed to artificial intelligence cluster; KUMAR V, 2021, J BUS RES very highly contributed to artificial intelligence cluster; VRONTIS D, 2022, INT J HUM RESOUR MANAG exclusively contributed to artificial intelligence cluster}
		\footnotetext[1]{Average citation per year}
		\footnotetext[2]{Abbreviated document metadata and title}
		\footnotetext[3]{Link of a Digital Object Identifier}
		\footnotetext[4]{Las period of the research, relevant for analysis in regard to ascertaining potential future trend, however substantially lacking consistency and completeness of the data so as to be relevant for peak year analysis}
	\end{table}

	\section{Artificial Intelligence Method Occurrence Heat Map}\label{secA1}
	
\begin{figure}[h] 
		\centering
		\includegraphics[width=1\textwidth]{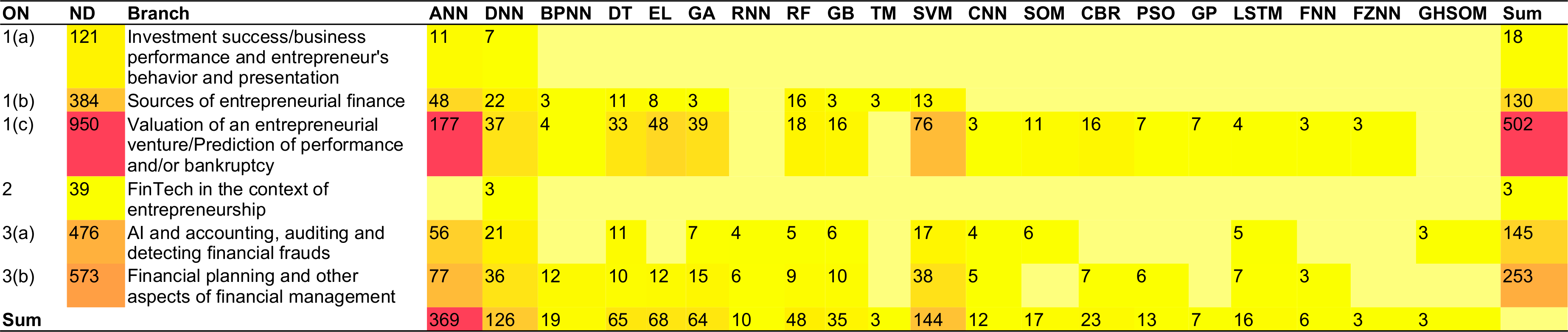}
		\caption{Heat map for Artificial Intelligence Method Occurrence by Topic Niches in Table \ref{tab:MPG}. The color of the map is tied to the numeric value and represents the intensity of occurrence. Bright yellow denotes no value, while if the color is shifted to yellow, occurrence is low, and if the color is shifted to red, occurrence is high.}\label{fig:HM1}
	\end{figure}

	As an addition to the analysis in Table \ref{tab:MPG} heat map of the data was made, presented in Figure \ref{fig:HM1}. With this helper figure one can more easily ascertain the relevance of every branch, as per document corpus, and also the relevance of every method, as per use in a specific branch -- with totals also being revealing.

	\section{Historiograph Topic Documents}\label{secA4}
	
	An addition to the analysis in Figure \ref{fig:Historiograph}, here we present historiograph documents as well as their metadata, in Tables \ref{tab:MHCHD15} and \ref{tab:MHCHD30}. With these, the reader can in more detail ascertain both the documents and the defined topic of the entire research corpus. As the historiograph includes 30 items the presented data is divided into two parts and sorted as per local citation count.

	\begin{table}[h] 
		\caption{Historiograph (Figure \ref{fig:Historiograph}) Documents -- Part I}\label{tab:MHCHD15}
		\begin{tabular*}{\textwidth}{@{\extracolsep\fill}lp{.7\textwidth}}
			\toprule%
			\multicolumn{2}{@{}c@{}}{First $ 15 $ Historiograph Documents Included}
			\\\cmidrule{1-2}
			Local/Global\textsuperscript{1} Citation & Document Reference\textsuperscript{2}/DOI\textsuperscript{3} \\
			\midrule
			$ 187 $/$ 609 $ & \href{https://doi.org/10.1016/j.ejor.2006.08.043}{Kumar PR, 2007, EUR J OPER RES} \\
			&	\footnotesize BANKRUPTCY PREDICTION IN BANKS AND FIRMS VIA STATISTICAL AND INTELLIGENT TECHNIQUES - A REVIEW \\
			
			$ 160 $/$ 512 $ & \href{https://doi.org/10.1016/j.eswa.2004.12.008}{Min JH, 2005, EXPERT SYST APPL} \\
			&	\footnotesize BANKRUPTCY PREDICTION USING SUPPORT VECTOR MACHINE WITH OPTIMAL CHOICE OF KERNEL FUNCTION PARAMETERS \\
			
			$ 143 $/$ 456 $ & \href{https://doi.org/10.1016/j.eswa.2004.08.009}{Shin KS, 2005, EXPERT SYST APPL} \\
			&	\footnotesize AN APPLICATION OF SUPPORT VECTOR MACHINES IN BANKRUPTCY PREDICTION MODEL \\
			
			$ 131 $/$ 348 $ & \href{https://doi.org/10.1016/S0377-2217(98)00051-4}{Zhang GQ, 1999, EUR J OPER RES} \\
			&	\footnotesize ARTIFICIAL NEURAL NETWORKS IN BANKRUPTCY PREDICTION: GENERAL FRAMEWORK AND CROSS-VALIDATION ANALYSIS \\
			
			$ 124 $/$ 356 $ & \href{https://doi.org/10.1016/0167-9236(94)90024-8}{Wilson RL, 1994, DECIS SUPPORT SYST} \\
			&	\footnotesize BANKRUPTCY PREDICTION USING NEURAL NETWORKS \\
			
			$ 120 $/$ 304 $ & \href{https://doi.org/10.1016/j.eswa.2007.05.019}{Tsai CF, 2008, EXPERT SYST APPL} \\
			&	\footnotesize USING NEURAL NETWORK ENSEMBLES FOR BANKRUPTCY PREDICTION AND CREDIT SCORING \\
			
			$ 114 $/$ 440 $ & \href{https://doi.org/10.1016/0378-4266(94)90007-8}{Altman EI, 1994, J BANK FINANC} \\
			&	\footnotesize CORPORATE DISTRESS DIAGNOSIS - COMPARISONS USING LINEAR DISCRIMINANT-ANALYSIS AND NEURAL NETWORKS (THE ITALIAN EXPERIENCE) \\
			
			$ 111 $/$ 279 $ & \href{https://doi.org/10.1016/j.eswa.2017.04.006}{Barboza F, 2017, EXPERT SYST APPL} \\
			&	\footnotesize MACHINE LEARNING MODELS AND BANKRUPTCY PREDICTION \\
			
			$ 102 $/$ 314 $ & \href{https://doi.org/10.1016/0377-2217(95)00070-4}{Dimitras AI, 1996, EUR J OPER RES} \\
			&	\footnotesize A SURVEY OF BUSINESS FAILURES WITH AN EMPHASIS ON PREDICTION METHODS AND INDUSTRIAL APPLICATIONS \\
			
			$ 86 $/$ 190 $ & \href{https://doi.org/10.1016/S0957-4174(97)00011-0}{Jo HK, 1997, EXPERT SYST APPL} \\
			&	\footnotesize BANKRUPTCY PREDICTION USING CASE-BASED REASONING, NEURAL NETWORKS, AND DISCRIMINANT ANALYSIS \\
			
			$ 79 $/$ 221 $ & \href{https://doi.org/10.1016/j.cor.2004.03.017}{West D, 2005, COMPUT OPER RES} \\
			&	\footnotesize NEURAL NETWORK ENSEMBLE STRATEGIES FOR FINANCIAL DECISION APPLICATIONS \\
			
			$ 79 $/$ 171 $ & \href{https://doi.org/10.1016/j.eswa.2008.01.018}{Nanni L, 2009, EXPERT SYST APPL} \\
			&	\footnotesize AN EXPERIMENTAL COMPARISON OF ENSEMBLE OF CLASSIFIERS FOR BANKRUPTCY PREDICTION AND CREDIT SCORING \\
			
			$ 74 $/$ 260 $ & \href{https://doi.org/10.1016/j.eswa.2005.09.070}{Min SH, 2006, EXPERT SYST APPL} \\
			&	\footnotesize HYBRID GENETIC ALGORITHMS AND SUPPORT VECTOR MACHINES FOR BANKRUPTCY PREDICTION \\
			
			$ 72 $/$ 179 $ & \href{https://doi.org/10.1016/j.dss.2007.12.002}{Alfaro E, 2008, DECIS SUPPORT SYST} \\
			&	\footnotesize BANKRUPTCY FORECASTING: AN EMPIRICAL COMPARISON OF ADABOOST AND NEURAL NETWORKS \\
			
			$ 67 $/$ 212 $ & \href{https://doi.org/10.1016/S0957-4174(02)00051-9}{Shin KS, 2002, EXPERT SYST APPL} \\
			&	\footnotesize A GENETIC ALGORITHM APPLICATION IN BANKRUPTCY PREDICTION MODELING \\
			\botrule
		\end{tabular*}
		\footnotetext{Sorted according to local citation in descending order. For the reason of the substantial amount of data, this analysis has two parts, with the continuation being in Table \ref{tab:MHCHD30}.}	
		\footnotetext[1]{Local citations are those from the research data, while global citations are those in the entire Web of Science database (included in the exported data from Web of Science)}
		\footnotetext[2]{Abbreviated document metadata and title}
		\footnotetext[3]{Link of a Digital Object Identifier}	
	\end{table}

	\begin{table}[h] 
		\caption{Historiograph (Figure \ref{fig:Historiograph}) Documents -- Part II}\label{tab:MHCHD30}
		\begin{tabular*}{\textwidth}{@{\extracolsep\fill}lp{.7\textwidth}}
			\toprule%
			\multicolumn{2}{@{}c@{}}{Last $ 15 $ Historiograph Documents Included}
			\\\cmidrule{1-2}
			Local/Global\textsuperscript{1} Citation & Document Reference\textsuperscript{2}/DOI\textsuperscript{3} \\
			\midrule
			$ 65 $/$ 156 $ & \href{https://doi.org/10.1016/j.eswa.2009.10.012}{Kim MJ, 2010, EXPERT SYST APPL} \\
			&	\footnotesize ENSEMBLE WITH NEURAL NETWORKS FOR BANKRUPTCY PREDICTION \\
			
			$ 65 $/$ 468 $ & \href{https://doi.org/10.1016/j.ejor.2015.05.030}{Lessmann S, 2015, EUR J OPER RES} \\
			&	\footnotesize BENCHMARKING STATE-OF-THE-ART CLASSIFICATION ALGORITHMS FOR CREDIT SCORING: AN UPDATE OF RESEARCH \\
			
			$ 62 $/$ 183 $ & \href{https://doi.org/10.2307/3665934}{Coats PK, 1993, FINANC MANAGE} \\
			&	\footnotesize RECOGNIZING FINANCIAL DISTRESS PATTERNS USING A NEURAL-NETWORK TOOL \\
			
			$ 61 $/$ 147 $ & \href{https://doi.org/10.1016/0167-9236(96)00018-8}{Lee KC, 1996, DECIS SUPPORT SYST} \\
			&	\footnotesize HYBRID NEURAL NETWORK MODELS FOR BANKRUPTCY PREDICTIONS \\
			
			$ 60 $/$ 194 $ & \href{https://doi.org/10.1016/S0957-4174(02)00045-3}{Park CS, 2002, EXPERT SYST APPL} \\
			&	\footnotesize A CASE-BASED REASONING WITH THE FEATURE WEIGHTS DERIVED BY ANALYTIC HIERARCHY PROCESS FOR BANKRUPTCY PREDICTION \\
			
			$ 56 $/$ 145 $ & \href{https://doi.org/10.1016/S0148-2963(97)00242-7}{Yang ZR, 1999, J BUS RES} \\
			&	\footnotesize PROBABILISTIC NEURAL NETWORKS IN BANKRUPTCY PREDICTION \\
			
			$ 55 $/$ 134 $ & \href{https://doi.org/10.1016/j.ejor.2016.01.012}{Liang D, 2016, EUR J OPER RES} \\
			&	\footnotesize FINANCIAL RATIOS AND CORPORATE GOVERNANCE INDICATORS IN BANKRUPTCY PREDICTION: A COMPREHENSIVE STUDY \\
			
			$ 53 $/$ 257 $ & \href{https://doi.org/10.1016/j.eswa.2005.12.008}{Wu CH, 2007, EXPERT SYST APPL} \\
			&	\footnotesize A REAL-VALUED GENETIC ALGORITHM TO OPTIMIZE THE PARAMETERS OF SUPPORT VECTOR MACHINE FOR PREDICTING BANKRUPTCY \\
			
			$ 52 $/$ 140 $ & \href{https://doi.org/10.1016/0167-9236(95)00033-X}{Serranocinca C, 1996, DECIS SUPPORT SYST} \\
			&	\footnotesize SELF ORGANIZING NEURAL NETWORKS FOR FINANCIAL DIAGNOSIS \\
			
			$ 52 $/$ 101 $ & \href{https://doi.org/10.1016/S0957-4174(96)00056-5}{Jo H, 1996, EXPERT SYST APPL} \\
			&	\footnotesize INTEGRATION OF CASE-BASED FORECASTING, NEURAL NETWORK, AND DISCRIMINANT ANALYSIS FOR BANKRUPTCY PREDICTION \\
			
			$ 52 $/$ 133 $ & \href{https://doi.org/10.1016/j.eswa.2008.03.020}{Chen WS, 2009, EXPERT SYST APPL} \\
			&	\footnotesize USING NEURAL NETWORKS AND DATA MINING TECHNIQUES FOR THE FINANCIAL DISTRESS PREDICTION MODEL \\
			
			$ 52 $/$ 103 $ & \href{https://doi.org/10.1016/j.ejor.2018.10.024}{Mai F, 2019, EUR J OPER RES} \\
			&	\footnotesize DEEP LEARNING MODELS FOR BANKRUPTCY PREDICTION USING TEXTUAL DISCLOSURES \\
			
			$ 51 $/$ 283 $ & \href{https://doi.org/10.1016/j.eswa.2006.02.016}{Kirkos E, 2007, EXPERT SYST APPL} \\
			&	\footnotesize DATA MINING TECHNIQUES FOR THE DETECTION OF FRAUDULENT FINANCIAL STATEMENTS \\
			
			$ 51 $/$ 174 $ & \href{https://doi.org/10.1016/j.dss.2011.10.007}{Olson DL, 2012, DECIS SUPPORT SYST} \\
			&	\footnotesize COMPARATIVE ANALYSIS OF DATA MINING METHODS FOR BANKRUPTCY PREDICTION \\
			
			$ 51 $/$ 177 $ & \href{https://doi.org/10.1016/j.eswa.2016.04.001}{Zieba M, 2016, EXPERT SYST APPL} \\
			&	\footnotesize ENSEMBLE BOOSTED TREES WITH SYNTHETIC FEATURES GENERATION IN APPLICATION TO BANKRUPTCY PREDICTION \\
			\botrule
		\end{tabular*}
		\footnotetext{Sorted according to local citation in descending order. For the reason of the substantial amount of data, this analysis has two parts, with the first part being in Table \ref{tab:MHCHD15}.}	
		\footnotetext[1]{Local citations are those from the research data, while global citations are those in the entire Web of Science database (included in the exported data from Web of Science)}
		\footnotetext[2]{Abbreviated document metadata and title}
		\footnotetext[3]{Link of a Digital Object Identifier}	
	\end{table}

\end{appendices}

\clearpage 

\bibliography{sn-bibliography}

\end{document}